\documentclass[a4paper,fleqn]{cas-sc}

\usepackage[numbers]{natbib}
\usepackage{amssymb}
\usepackage{algorithm}%
\usepackage{algorithmic}
\usepackage{graphicx}%
\usepackage{multirow}%
\usepackage{amsmath,amssymb,amsfonts}%
\usepackage{placeins}   
\usepackage{amsthm}%
\usepackage{mathrsfs}%
\usepackage[title]{appendix}%
\usepackage{xcolor}%
\usepackage{textcomp}%
\usepackage{manyfoot}%
\usepackage{booktabs}%
\usepackage{listings}%
\usepackage{subcaption}

\usepackage{amsmath}
\usepackage{import}%

\usepackage{xurl}
\usepackage{titletoc}

\def\tsc#1{\csdef{#1}{\textsc{\lowercase{#1}}\xspace}}
\tsc{WGM}
\tsc{QE}
\tsc{EP}
\tsc{PMS}
\tsc{BEC}
\tsc{DE}

\begin{document}
\let\WriteBookmarks\relax
\def\floatpagepagefraction{1}
\def\textpagefraction{.001}

\shorttitle{}    

\shortauthors{S.S. Murali Krishnan and D.F. Hougen}  

\title[mode = title] {SR4-Fit: A unified interpretable rule-based machine learning framework for informative and trustworthy decision-making}  



\author[1]{Shyam Sundar Murali Krishnan}[orcid=0000-0001-9239-4390]
\author[1]{Dean Frederick Hougen}[orcid=0000-0001-5393-1480]

\affiliation[1]{organization={School of Computer Science, University of Oklahoma},city={Norman},state={Oklahoma},country={USA}}

\begin{abstract}
In many high-stakes applications, machine learning is dominated by black-box models that require post hoc explanations to justify their predictions. These explanations are often unreliable because they do not reflect the model's actual computations, limiting accountability and trust. A natural alternative is to use models that are interpretable by design. However, existing rule-based approaches, such as RuleFit and decision trees, while transparent, often lack stability and predictive strength, reinforcing a perceived trade-off between traditional performance measures and model understandability. To address this, we propose Sparse Relaxed Regularized Regression Rule-Fit (SR4-Fit), an intrinsically interpretable algorithm for both classification and regression that produces compact and stable rule sets without sacrificing performance. Using demographic data from the U.S. Census Bureau’s American Community Survey, SR4-Fit predicts U.S. House election outcomes with high accuracy and interpretability while uncovering demographic interactions missed by black-box models. We further validate SR4-Fit across fourteen benchmark datasets (six classification and eight regression), where it outperforms existing rule-based methods, including RuleFit and decision trees in terms of accuracy, stability, and compactness while remaining competitive with black-box models in predictivity. These results demonstrate that interpretability and predictive reliability need not be mutually exclusive, offering a practical and transparent alternative for high-stakes decision-making.
\end{abstract}

\begin{highlights}
\item Proposed SR4-Fit to resolve the predictivity-interpretability trade-off for both classification and regression via sparse rule sets.
\item Introduced Rule Understandability Score and Pareto analysis show SR4-Fit dominates rule-based baselines.
\item Enhanced Interpretability Score incorporates rule complexity for fairer comparison.
\item SR4-Fit interpreted demographic interactions in House of Representatives elections via human-readable rules.
\item Validated across 14 benchmark datasets, confirming generalizability and robustness.
\end{highlights}


\maketitle

\section{Introduction}\label{intro}
Machine learning (ML) models such as random forests~\cite{breiman2001random}, Support Vector Machines (SVMs)~\cite{hearst1998support}, XGBoost\cite{friedman2001greedy}, and neural networks~\cite{mcculloch1943logical} have transformed how data-driven problems are approached across science and society. Despite their success, these models are often criticized as \emph{black boxes} that capture complex, nonlinear patterns but provide little clarity on how inputs influence predictions. In many domains, this opacity limits the ability to verify, explain, or trust automated decisions, especially when those decisions have real-world consequences~\cite{caruana2015intelligible,nesvijevskaia2021accuracy}. These concerns have motivated a growing interest in \emph{interpretable machine learning} which aims to design models that not only perform well but can also be understood and validated by humans~\cite{molnar2020interpretable}.

Interpretability becomes particularly important in areas where transparency and accountability are essential, such as the study of U.S. House of Representatives elections. Understanding how demographic and structural factors shape political outcomes is central to evaluating fair representation in a democracy. Electoral behavior depends not only on individual choices but also on how districts are drawn, which can amplify or diminish the influence of certain groups. Demographic characteristics such as race, education, income, and age strongly correlate with voting patterns and party preference~\cite{arrington2010affirmative}. Modeling these relationships helps estimate \emph{district security}---the likelihood that a political party will retain control in future elections---and can also provide insight into practices such as \emph{gerrymandering}---the deliberate manipulation of electoral district boundaries to grant a political party, incumbent, or group an unfair advantage~\cite{friedman2009rising}. In this setting, models that balance predictivity with interpretability can offer more than just forecasts; they can help explain why elections turn out the way they do, providing models that are \emph{informative} regarding the domain to which they are applied~\cite{richardson2020districts}.

Black-box models have proven capable of predicting congressional election results with high accuracy~\cite{richardson2020districts}. Yet the inner workings of these models often remain obscure, making it difficult to identify which demographic patterns drive the predictions. This reflects a well-known tension between predictive accuracy and interpretability~\cite{murdoch2019definitions}: complex models tend to predict well but are hard to understand, whereas simpler models may be easier to interpret but often lose predictive strength. Finding methods that achieve both remains an active and important challenge.

One common strategy has been to use post-hoc explanation tools such as SHapley Additive exPlanations (SHAP)~\cite{lundberg2017unified} and Local Interpretable Model-agnostic Explanations (LIME)~\cite{ribeiro2016should}. These approaches provide approximate insights into a model’s behavior, such as feature importance or local decision boundaries. While valuable for interpretive analysis, post-hoc explanations can diverge from what the model is actually computing, particularly for non-linear systems. As a result, they can aid understanding but do not fully solve the transparency problem.

An alternative direction focuses on intrinsically interpretable model algorithms whose structure makes their reasoning transparent by design~\cite{rudin2019stop}. These include decision trees~\cite{quinlan1986induction}, sparse linear models~\cite{vidaurre2013survey}, and rule-based systems~\cite{furnkranz2012foundations}. Because the logic behind each prediction is directly encoded in the model, such approaches make it easier to verify results and reason about causal patterns. While intrinsic interpretability does not automatically guarantee higher accuracy, it offers a solid basis for building models that are both reliable and understandable. In contexts like electoral forecasting, this combination supports both empirical rigor and public accountability.

Among intrinsically interpretable methods, rule-based models are particularly appealing because they express decisions as if–then statements that can be read and understood directly. These models align closely with human reasoning and lend themselves naturally to policy and social analysis. However, existing rule-based techniques vary widely in their design and scope. Some focus only on binary classification~\cite{obregon2023rulecosi+,10.1214/15-AOAS848,angelino2018learning}, while others prioritize simplicity to the point of losing predictive depth or are tailored to a single task type, either classification or regression ~\cite{benard2021interpretable}. These limitations reinforce the perception that interpretability often comes at the expense of performance.

The RuleFit algorithm~\cite{friedman2008predictive} represented a key attempt to balance this trade-off by combining rule extraction from tree ensembles with linear modeling. While RuleFit offers a useful middle ground, it can struggle with stability, consistency, and predictivity, especially when applied to high-dimensional or noisy data~\cite{yang2017scalable}. The rules it generates can overlap or vary across runs, complicating interpretation and limiting trust in the results. Further refinements are therefore needed to make rule-based modeling more robust and versatile.

To address these issues, we introduce \emph{Sparse Relaxed Regularized Regression Rule Fit} (SR4-Fit), an algorithm for generating intrinsically interpretable rulesets that extends the RuleFit framework using the sparse optimization principles of Sparse Relaxed Regularized Regression (SR3)~\cite{zheng2018unified}. SR4-Fit is designed to produce compact and stable rule sets while maintaining competitive predictive accuracy. Its formulation allows fine-grained control over sparsity, offering a single framework that works for both classification (binary and multi-class) and regression tasks.


\begin{figure}
\centering
\scriptsize

\begin{minipage}[t]{\linewidth}
    \centering
    \includegraphics[width=0.5\linewidth]{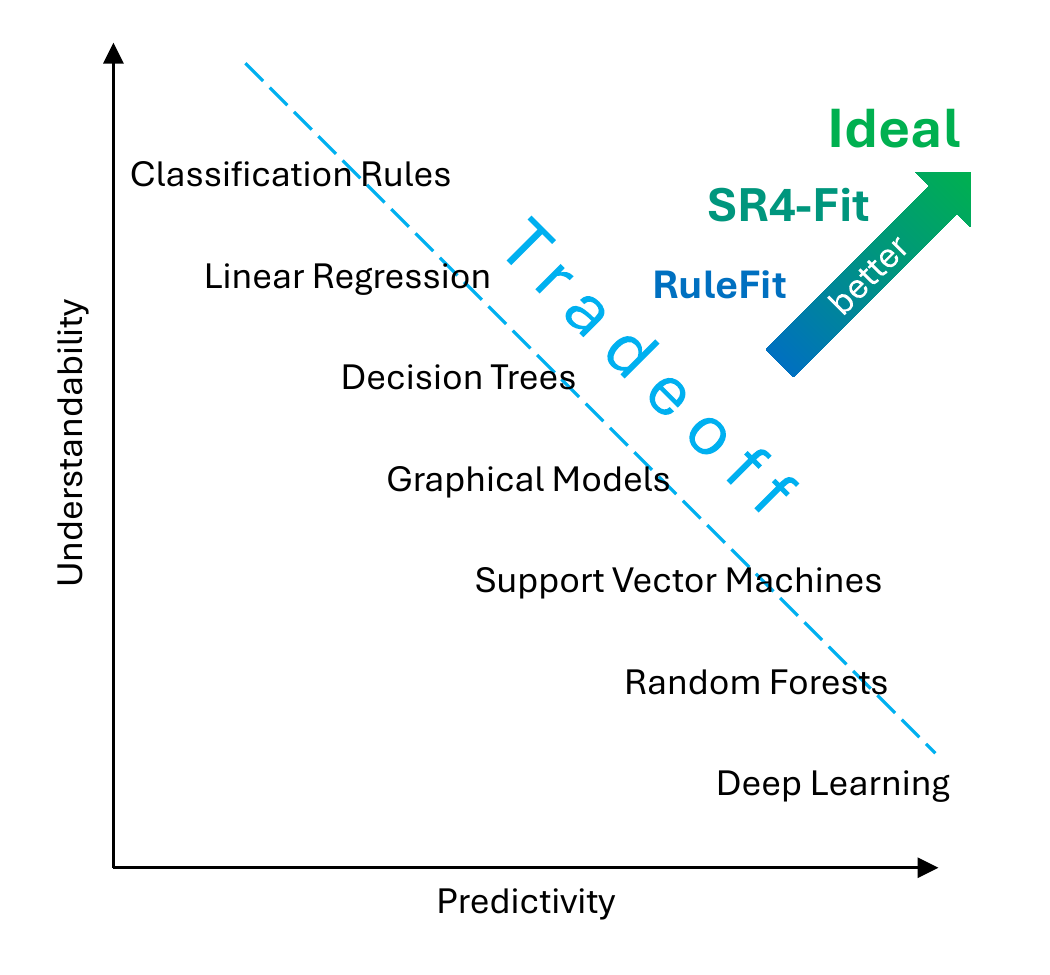}
\end{minipage}

\caption{Conceptual illustration of the predictivity--understandability trade-off across common machine learning models. SR4-Fit is positioned closer to the ideal region of high predictivity and high understandability, reflecting its aim to balance both criteria more effectively than existing rule-based approaches.}

\label{fig:tradeoff}
\end{figure}

Over the past decade, the literature on understandable AI has become replete with notional diagrams similar to Figure~\ref{fig:tradeoff}, showing greater understandability (or a related term such as explainability, interpretability, etc.) on one axis and greater predictivity (accuracy, 
learning performance, etc.) on the other, along with a sampling of ML types scattered across the plot to indicate a tradeoff between understandability and predictivity, with complex ML types such as deep learning and random forests shown as featuring high predictivity but low understandability; simple types such as decision rules and linear regression shown as featuring high understandability but low predictivity; and approaches of intermediate complexity such as decision trees, graphical models, and support vector 
machines shown as more moderate in both understandability and predictivity. Such diagrams have their origins in DARPA’s Explainable Artificial Intelligence Program~\cite{gunning2017explainable, gunning2019darpa}, and have been included in papers on explainable AI itself~\cite{angelov2021explainable} as well as in applications across a wide array of fields, including modeling human decision-making~\cite{guo2019interpretable}, wireless communications~\cite{morocho2019machine}, fraud detection~\cite{nesvijevskaia2021accuracy}, natural language processing~\cite{kumar2021explainable}, healthcare~\cite{abdullah2021review,nazir2023survey,viswan2024explainable}, ecology~\cite{pichler2023machine}, energy~\cite{chen2023interpretable,liu2023explainable,gugliermetti2024future}, and finance~\cite{sailer2026explainable}. 

Figure~\ref{fig:tradeoff} also illustrates the intuition that RuleFit~\cite{friedman2008predictive}, which first produces highly predictive random forests and then uses those to generate understandable decision rules, sacrifices less predictivity and less understandability than the intermediate complexity methods, and that SR4-Fit, by improving on the rule creation process of RuleFit, takes us further toward the ideal of high predictivity and high understandability.

Using demographic data from the U.S. Census Bureau’s American Community Survey, SR4-Fit predicts U.S. House election outcomes with strong accuracy while generating interpretable rules that highlight meaningful demographic interactions such as the influence of education, racial composition, and age distribution on party alignment. In the regression setting, we further examine the model’s ability to capture continuous electoral variation by using district-level Democratic and Republican vote percentages as target variables. This allows SR4-Fit to quantify how demographic gradients correspond to shifts in partisan support, offering a more detailed view of electoral behavior. Beyond elections, the model is tested on a suite of benchmark datasets, six for classification and eight for regression, where it performs competitively with both black-box and existing interpretable methods. Together, these results suggest that intrinsic interpretability and predictive reliability can coexist for tree-based algorithms, offering a practical path toward models that are transparent, stable, and empirically strong across different problem settings. Beyond the quantitative analysis of this data, we provide extensive discussion of the rules inferred by SR4-Fit.

\section{Contributions}\label{contrib}

The main contributions of this paper are as follows:

\begin{enumerate}
    \item We propose SR4-Fit, an intrinsically interpretable rule-based algorithm that directly addresses the traditional trade-off between interpretability and predictivity by producing compact, stable, and human-readable rule sets  without sacrificing performance.

    \item SR4-Fit provides a unified framework for both classification and regression tasks, extending the capabilities of existing rule-based methods such as RuleFit,  which are limited in predictive performance, and decision trees, which are limited in both stability and performance.

    \item We introduce the Rule Understandability Score (RUS), a three-component metric combining rule stability, compactness, and complexity that deliberately excludes predictive performance, enabling an independent assessment of interpretability. We employ RUS in a Pareto analysis against predictivity, providing a model-selection perspective that does not depend on a particular understandability-predictivity weighting scheme.
    
    \item We enhance the Interpretability Score (IPS) proposed by Margot and Luta~\cite{margot2021new} by incorporating rule complexity as an additional component alongside predictive accuracy, rule stability, and model compactness. Our enhanced interpretability score (E-IPS) provides a more comprehensive and principled basis for comparing rule-based models. SR4-Fit achieves consistently high E-IPS scores across both classification and regression tasks, demonstrating that it produces stable, compact, and interpretable rule sets while maintaining competitive predictive performance.

    \item We validate SR4-Fit on U.S. House election demographic data, where interpretable rules uncover meaningful demographic interactions that black-box models obscure, demonstrating the practical value of intrinsic interpretability in high-stakes decision-making.

    \item We further validate SR4-Fit across fourteen publicly available benchmark datasets (six classification and eight regression), confirming its generalizability and robustness across diverse domains.
\end{enumerate}

\section{Methodology}\label{method}

\begin{figure}
\centering
\scriptsize
\begin{minipage}{\linewidth}
    \centering
    \includegraphics[width=\linewidth]{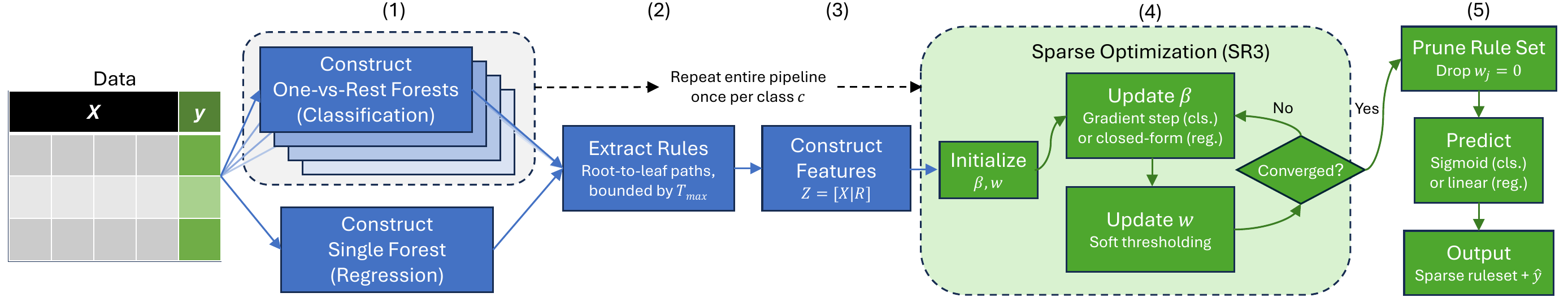}
\end{minipage}
\caption{
Overview of the SR4-Fit pipeline (including classification and regression).
}
\label{fig:methodology}
\end{figure}

SR4-Fit is a rule-based learning algorithm that combines the interpretability approach of RuleFit with the sparse optimization capabilities of SR3 to produce compact, stable, and human-readable predictive models for both classification and regression tasks. Like RuleFit, SR4-Fit extracts decision rules from an ensemble of decision trees to capture non-linear patterns in the data and then fits a sparse linear model over both the extracted rules and the original input features. To support classification, the SR3 optimization objective is modified by replacing the squared loss with logistic loss (see Algorithm~\ref{alg:classification}), allowing the model to output class probabilities rather than continuous values~\cite{11471514}. For regression, the squared-error loss is retained within the same optimization framework (see Algorithm~\ref{alg:regression}), enabling direct prediction of continuous target variables. In both settings, SR3's structured regularization balances predictive accuracy with sparsity, yielding interpretable models whose predictions can be traced through a compact set of human-readable rules. The algorithm proceeds in five stages: (1) random forest generation, (2) rule extraction, (3) feature construction, (4) sparse optimization, and (5) model pruning, as illustrated in Figure~\ref{fig:methodology}. 

\subsection{Random Forest Generation and Rule Extraction}
\label{Rule Extraction}

SR4-Fit begins by extracting interpretable decision rules from an ensemble of decision trees. In the classification setting, a separate forest is trained for each class using a one-versus-rest formulation \cite{rifkin2004defense}, where each forest is trained to distinguish the class of interest from all others. In the regression setting, a single forest is trained to predict the continuous target variable directly, enabling it to capture nonlinear dependencies between input features and the output. To control model complexity and ensure interpretability, the maximum total number of rules extracted is bounded by the hyperparameter $r_{\text{max}}$.

From each tree in the forest, rules are extracted by traversing every path from the root to a terminal node. Each path defines a rule as a conjunction of logical conditions on the input features. These rules are encoded as binary indicators: a rule evaluates to 1 if all its conditions are satisfied for a given input, and to 0 otherwise. For example, the rule
\begin{equation*}
X_1 \leq 2.7 \ \text{and} \ X_3 > 1.4
\end{equation*}
evaluates to 1 if and only if $X_1 \leq 2.7$ and $X_3 > 1.4$ hold 
simultaneously and to 0 otherwise.  Here, $X$ are the features. Because rules are drawn from every tree across the ensemble, the resulting rule set captures diverse and complementary patterns across the input space, reflecting the ensemble's collective coverage of the data.


\subsection{Feature Construction and Optimization Objective}
\label{Feature Construction and Optimization Objective}

Following rule extraction, SR4-Fit builds an extended feature matrix by combining the original input features with the binary rule indicators. For each data point, the rule evaluations are concatenated with the original feature values to form a comprehensive representation $Z = [X \;|\; R]$ where $X \in \mathbb{R}^{n \times d}$ where $n$ is the number of observations and $d$ is the number of features and $R \in \{0, 1\}^{n \times m}$. This expanded feature matrix allows the model to utilize both raw features and data-driven patterns captured by the trees.

To learn a sparse and accurate classifier, the algorithm minimizes the loss function $L(\beta, w)$ with parameters $\beta$, the model's predictive weight vector, and $w$, a vector enforcing sparsity, defined by:
\begin{equation}
L(\beta, w) = \sum_{i=1}^{n} \log(1 + \exp^{(-y_i Z_i^\top \beta)}) + \lambda \|w\|_1 + \frac{\kappa}{2} \|\beta - w\|_2^2
\end{equation}
where $y_i$ is the label.

For regression tasks, the same structure is retained, but the logistic loss is replaced by the squared-error loss:
\begin{equation}
L(\beta, w) = \frac{1}{2}\|y - Z\beta\|_2^2 + \lambda \|w\|_1 + \frac{\kappa}{2}\|\beta - w\|_2^2.
\label{eq:regression_loss_function}
\end{equation}

In these objective functions, $\lambda$ controls the sparsity and $\kappa$ controls how closely $\beta$ must follow $w$. The algorithm minimizes these objective functions by alternating between the two update steps. First, with $w$ fixed, $\beta$ is updated using gradient-based optimization on the logistic loss for classification and a closed-form ridge-type solution on the squared error loss for regression, both balancing fitting accuracy with quadratic regularization that enforces $\beta \approx w$. Second, with $\beta$ fixed, $w$ is updated via soft-thresholding, which is key for achieving sparsity \cite{tibshirani1996regression}, using
\begin{equation}
w_j = \operatorname{sign}(\beta_j) \cdot \max\left(|\beta_j| - \frac{\lambda}{\kappa},\, 0\right)
\label{eq:soft_threshold}
\end{equation}
where
\begin{equation*}
\operatorname{sign}(z) =
\begin{cases}
+1, & \text{if } z > 0 \\
\;\;0, & \text{if } z = 0 \\
-1, & \text{if } z < 0.
\end{cases}
\label{eq:sign}
\end{equation*}

This iterative process continues until the relative change in the objective function falls below a small tolerance $\varepsilon$ (typically $10^{-4}$). The result is a compact, interpretable model in which the nonzero components of $w$ identify the features and rules that contribute most strongly to prediction accuracy.

\begin{algorithm}
    \caption{SR4-Fit Classification Algorithm}
    \label{alg:classification}
    \begin{algorithmic}[1]
        \STATE \textbf{Input:} Dataset $(X, y)$, parameters $\lambda$, $\kappa$, $\varepsilon$ and $r_{\text{max}}$.
        \STATE \textbf{Output:} Predictive weights $\beta$.
        \FOR{each class $c = 0, 1, \ldots, C$}
            \STATE Train a random forest using one-vs-rest labeling for class $c$ \cite{breiman2001random}.
            \STATE Extract decision rules from each path in each tree until $r_{\text{max}}$ \cite{friedman2001greedy}.
            \STATE Encode rules as binary indicators $R \in \{0,1\}^{n \times m}$ \cite{cohen1995fast}.
            \STATE Form an extended feature matrix $Z = [X \;|\; R]$.
            \STATE Initialize $\beta$, $w$.
            \WHILE{not converged}
                \STATE Minimize $L(\beta, w) = \newline
                \hspace*{1em}\displaystyle\sum_{i=1}^n \log(1 + \exp^{(-y_i Z_i^\top \beta))} + \lambda \|w\|_1 + \frac{\kappa}{2}\|\beta - w\|_2^2$
                \STATE Update $w$,  $w_j = \operatorname{sign}(\beta_j) \cdot \max\left(|\beta_j| - \frac{\lambda}{\kappa},\, 0\right)$
            \ENDWHILE
            \STATE Prune rules where $w_j = 0$
        \ENDFOR
        \STATE \textbf{Prediction:}
        \STATE Compute $s(x) =  \displaystyle\sum_{j=1}^{d+m} \beta_j Z_j(x)$
        \STATE Compute predicted probability $\hat{y}(x) = \frac{1}{1 + e^{(-s(x))}}$
    \end{algorithmic}
\end{algorithm}

\begin{algorithm}
\caption{SR4-Fit Regression Algorithm}
\label{alg:regression}
\begin{algorithmic}[1]
    \STATE \textbf{Input:} Dataset $(X, y)$, parameters $\lambda$, $\kappa$, maximum rules $r_{\max}$, and tolerance $\varepsilon$.
    \STATE \textbf{Output:} Regression coefficients $\beta$.
    \STATE Train a random forest of depth $d$ on $(X, y)$.
    \STATE Extract decision rules from all root-to-leaf paths up to $r_{\max}$.
    \STATE Encode each rule as a binary feature; form an extended matrix $Z = [X \;|\; R]$.
    \STATE Initialize coefficients $\beta$, $w$.
    \WHILE{not converged}
        \STATE Update $\beta$ by solving
        \[
        \beta = \arg\min_\beta \frac{1}{2}\|y - Z\beta\|_2^2 + \frac{\kappa}{2}\|\beta - w\|_2^2
        \]
        \STATE Update $w$ using soft-thresholding:
        \[
        w_j = \text{sign}(\beta_j)\cdot \max(|\beta_j| - \tfrac{\lambda}{\kappa}, 0)
        \]
        \STATE Check convergence: $\frac{|L_{t-1} - L_t|}{L_{t-1}} < \varepsilon$, where
        \[
        L(\beta,w) = \tfrac{1}{2}\|y - Z\beta\|_2^2 + \lambda\|w\|_1 + \tfrac{\kappa}{2}\|\beta - w\|_2^2
        \]
    \ENDWHILE
    \STATE prune rules with $|w_j| = 0$.
    \STATE \textbf{Prediction:} For a new instance $x$, compute $\hat{y}(x) =  \sum_{j=1}^{d+m} \beta_j Z_j(x)$.
\end{algorithmic}
\end{algorithm}

\subsection{Model Pruning and Rule Selection}
\label{Model Pruning and Rule selection}

After optimization, SR4-Fit performs model pruning to retain only the most informative features and rules. The pruning step relies on the sparse auxiliary vector $w$, which reflects the importance of each feature and rule learned during training. Any component $w_j = 0$ indicates that the corresponding predictor contributes negligibly to the model and can be safely removed. The remaining nonzero elements of $w$ identify the subset of features and rules that form the final interpretable model.  

This separation of roles between $\beta$ and $w$, where $\beta$ captures predictive strength and $w$ enforces sparsity, provides a more stable and interpretable selection process than methods that apply sparsity constraints directly to the predictive coefficients \cite{zheng2018unified}. Because $w$ is updated through soft-thresholding, irrelevant or weakly correlated rules are automatically pruned, leading to a compact and transparent rule set.  

The pruning mechanism is identical for both classification and regression. By discarding inactive predictors and retaining only those associated with nonzero $w_j$ values, SR4-Fit yields a sparse model that is both easy to interpret and efficient to evaluate. This final rule set highlights the dominant feature interactions uncovered during training, offering clear and human-readable insights into how the model forms its predictions.

\subsection{Classification and Regression Prediction}
\label{Classification and Regression Prediction}

In the final stage, SR4-Fit uses the optimized coefficients $\beta$ to make predictions for both classification and regression tasks. For a new input instance $x$, the model evaluates the active rules and features and computes a linear score
\begin{equation*}
s(x) = \sum_{j=1}^{d+m} \beta_j Z_j(x),
\label{eq:prediction_function}
\end{equation*}
where $Z_j(x)$ represents the $j^{th}$ component of the extended feature vector that combines both raw features and rule indicators. The value $s(x)$ serves as the aggregated evidence derived from all selected predictors.

For classification tasks, SR4-Fit transforms this score into a probability through the sigmoid function:
\begin{equation*}
\hat{y}(x) = \sigma(s(x)) = \frac{1}{1 + e^{(-s(x))}},
\label{eq:sigmoid_output}
\end{equation*}
mapping the real-valued score to the interval $(0,1)$ and providing a probabilistic interpretation of the model’s confidence in the positive class. This probabilistic formulation not only supports threshold-based classification but also conveys uncertainty in predictions. For multiclass problems, SR4-Fit employs a one-vs-rest approach, training one binary classifier per class; the final class prediction corresponds to the one with the highest estimated probability. Because each classifier is constructed from sparse and interpretable rules, the decision path for every class prediction can be directly traced through the contributing rules and coefficients.

For regression tasks, SR4-Fit uses the same linear prediction structure without the sigmoid transformation, which provides a direct estimate of the target variable. Each nonzero coefficient $\beta_j$ quantifies the contribution of the corresponding rule or feature to the predicted outcome, allowing clear interpretation of how specific variables influence the continuous target.

Overall, SR4-Fit delivers a unified prediction framework that supports both probabilistic classification and continuous regression while preserving interpretability through its sparse, rule-based structure.

\section{Experimental Setup}\label{setup}
To thoroughly evaluate SR4-Fit quantitatively, we conducted extensive experiments on a total of 15 datasets (Section~\ref{ssec:datasets}) and introduce customized analysis methods appropriate for evaluation of interpretable ML (Section~\ref{ssec:analysis_methods}). Following the results and analysis (Section~\ref{results}), we provide detailed discussion of the rules discovered (Section~\ref{discussion}).

\subsection{Datasets}
\label{ssec:datasets}

Our evaluation features in-depth analysis on a dataset that explores connections between demographic features and election outcomes (Section~\ref{sssec:election_data}) considering both classification and regression problems using various configurations of this data, and a careful examination of the rules found, as well as broader numerical comparisons using 14 standard benchmark datasets (Section~\ref{sssec:benchmark_datasets}).

\subsubsection{Election Data}
\label{sssec:election_data}

The demographic data covers U.S. House of Representatives elections from 2006 to 2016, with the period 2006--2013 used for training and 2014--2016 for testing~\cite{census_s0601_2006_2016}. District-level election outcomes were obtained from the MIT Election Data and Science Lab's U.S. House 1976--2018 dataset \cite{medsl_us_house_1976_2018}. Following prior work~\cite{richardson2020districts}, four dataset configurations were constructed based on the demographic features included. The \emph{minimum dataset} consists of median age, male percentage, white percentage, and bachelor's degree attainment. The \emph{standard dataset} adds breakdowns of age, race, income, and education categories. The \emph{expanded dataset} further adds language spoken at home, marital status, and poverty indicators. The \emph{previous dataset} supplements the standard attributes with the party that won the district in the prior election.

For classification, experiments were conducted under two feature configurations: (i) including both Democratic (DEM) and Republican (REP) party voting percentages as features for each year in the training datasets, and (ii) excluding both party percentage variables to isolate model behavior under purely demographic inputs. For regression, two prediction targets were considered: DEM vote percentage and REP vote percentage. Each target was evaluated under two feature conditions, with and without the opposing party's vote percentage included as a feature, yielding four regression experiment configurations in total. These conditions were designed to assess model sensitivity to the availability of politically informative features, since party percentage variables can be highly predictive and can dominate model behavior when present.\footnote{While party voting percentages from previous elections are often highly predictive, they may also be misleading in some cases because they may reflect not only party loyalty, but also particulars of the candidates (or lack thereof) in a given election. For example, in Alabama District 5, the Democratic candidate won with greater than 98\% of the vote in 2006, and the Republican percentage that year was zero. This was because there was no Republican House candidate on the ballot that year in that district. In 2008, the vote was roughly 52\% Democratic to 48\% Republican, and in 2010 the vote was roughly 42\% Democratic to 58\% Republican. This shows that voting percentages can change rapidly from election to election.}
 
Previous research on U.S. demographic data~\cite{richardson2020districts} had shown that Random Forest and Support Vector Machine (SVM) models are strong predictors. Along with those models, XGBoost is included as an additional black-box baseline because it represents a current state-of-the-art in gradient boosting and consistently achieves top performance on structured tabular data \cite{friedman2001greedy}, making it a strong and relevant benchmark against which to assess the predictive competitiveness of SR4-Fit. Together, Random Forest, SVM, and XGBoost provide a rigorous and representative set of high-performing black-box baselines, enabling a direct comparison between black-box models and the interpretable SR4-Fit rule-based model. Additionally, SR4-Fit was compared with the traditional RuleFit algorithm, as SR4-Fit was designed with a similar interpretability objective, ensuring a fair evaluation setting. Along with RuleFit, decision trees and linear regression (logistic for classification problems and LASSO for regression problems) were also included to provide comparisons across interpretable and compact models. The hyperparameters ($r_{\text{max}}$, $\lambda$ , $\kappa$ and $\varepsilon$) were selected via grid search~\cite{pedregosa2011scikit} across a range of values. Each setting was evaluated over multiple trials, and the combination yielding higher accuracy was chosen.

All experiments on the U.S. demographic datasets were conducted over 30 independent trials for each model to provide statistically meaningful results. For classification tasks, predictive performance was assessed using standard classification metrics, including accuracy, precision, recall~\cite{swets1969effectiveness}, and F1-score~\cite{van1979information}, to provide a quantitative evaluation. For regression tasks, the predictive performance was assessed using Root Mean Squared Error (RMSE), Mean Absolute Error (MAE)~\cite{hodson2022root}, and Coefficient of Determination ($R^2$). The Dice-Sorensen index~\cite{chao2006abundance} was also employed for both classification and regression tasks to assess the structural stability of the rule sets across trials, measuring the overlap between rule sets produced in independent runs of the same model.

\subsubsection{Standard Benchmark Datasets}
\label{sssec:benchmark_datasets}

In addition, to demonstrate the broad applicability of the SR4-Fit algorithm in classification settings, six publicly available datasets were selected for experimentation. These include both binary and multiclass classification tasks, as well as imbalanced datasets, ensuring comprehensive evaluation. The Wisconsin Breast Cancer dataset~\cite{breast_cancer_wisconsin} provides diagnostic features from digitized images of breast masses to classify tumors as benign or malignant. The Ecoli dataset~\cite{ecoli_39} comprises protein localization sites across different cellular compartments. The Page Blocks dataset~\cite{page_blocks_classification_78} includes layout and geometric features of document blocks from scanned pages for classifying each block’s type. The Pima Indians Diabetes dataset~\cite{uci_pima_diabetes} contains medical and demographic attributes of female Pima Indian individuals to predict the onset of type 2 diabetes. The Vehicle Silhouette dataset~\cite{statlog_(vehicle_silhouettes)} involves shape descriptors extracted from vehicle silhouettes to classify them into categories based on geometric properties. Lastly, the Yeast dataset~\cite{yeast_110} includes sequence-based features of yeast proteins to predict their cellular localization. In this collection of datasets, the breast cancer and Pima Indians diabetes datasets are binary whereas the others are multi-class datasets.

Similarly, for regression settings, eight distinct benchmark datasets were used to demonstrate broad applicability. The Ozone dataset~\cite{friedman2001elements}  predicts daily average ozone measurements in Los Angeles in 1976. The MPG dataset~\cite{misc_auto_mpg_9} predicts fuel consumption for city cycles. The Machine dataset~\cite{misc_computer_hardware_29} predicts relative CPU performance. The Abalone dataset~\cite{misc_abalone_1} predicts ages of abalone. The Diabetes dataset~\cite{efron2004least} predicts how the disease will progress one year from the baseline. The bone mineral density of 261 adolescents in North America is predicted by the Bones dataset~\cite{misc_bone_density_friedman2001}. The Housing dataset~\cite{asuncion2003uci} predicts Boston home prices. Prostate-specific antigen levels in males slated for radical prostatectomy are predicted by the Prostate dataset~\cite{misc_prostate_friedman2001}.

\subsection{Analysis Methods}
\label{ssec:analysis_methods}

To assess whether SR4-Fit succeeds with regard to the standard predictivity-understandability trade-off conceptually visualized in Figure~\ref{fig:tradeoff}, we conduct a Pareto analysis that intentionally separates the predictivity and understandability axes through the introduction of a novel measure of understandability for rule-based ML systems (Section~\ref{sssec:parato_RUS}). This is complemented by an extensive numeric analysis using a single metric value that combines predictivity and understandability (Section~\ref{sssec:E-IPS}), extending a method used previously for interpretable ML~\cite{margot2021new}, as well as standard statistical analysis (Section~\ref{sssec:stat_analysis}) and visualizations (Section~\ref{sssec:viz}).

\subsubsection{Pareto Analysis and the Rule Understandability Score}
\label{sssec:parato_RUS}

To provide a clear perspective on the trade-off between predictive performance and rule understandability, Pareto analysis \cite{jin2008pareto} is employed. Pareto analysis examines whether one model simultaneously outperforms another along multiple independent axes, and our employment of it offers a model-selection perspective that does not depend on any particular predictivity-understandability weighting scheme.

For this analysis, we introduce the \emph{Rule Understandability Score} (RUS) as the measure of understandability. RUS is defined as a weighted combination of three components, each of which ranges from 0 to 1: (1)~Dice-Sørensen stability, (2)~normalized number of rules (inverted, so that fewer rules score higher), and (3)~normalized rule complexity (inverted, so that simpler rules score higher):
\begin{equation}
\text{RUS}(A) = {\gamma}_S S_n(A) + {\gamma}_{simp} (1-R_n^{\text{simp}}(A)) + {\gamma}_{comp} (1-R_n^{\text{comp}}(A))
\end{equation}
where $S_n(A)$ represents the stability score of model $A$, $R_n^{\text{simp}}(A)$ corresponds to the rule simplicity score of the model, and $R_n^{\text{comp}}(A)$ corresponds to the rule complexity of the model. The rule simplicity is computed through rule count, which is the average number of rules produced by the model per trial. The rule complexity is computed as the average number of conditions per rule, averaged across all rules a model produced in a trial. The weights ${\gamma}_P$, ${\gamma}_S$, ${\gamma}_{simp}$ and ${\gamma}_{comp}$ allow flexibility in emphasizing different aspects of understandability. In the present work, all three RUS components are equally weighted (${\gamma}_S = {\gamma}_{simp} = {\gamma}_{comp} = 1/3$). The components are normalized using MinMaxScaler in Python \citep{hao2019machine}, which maps each component to the range $[0, 1]$. Rule count and rule complexity are normalized because stability already fall on that scale, and keeping all the components on a common range ensures each contributes proportionally when combined into RUS.

Unlike the interpretability score (IPS, see Section~\ref{sssec:E-IPS}), which has been employed in prior work for measuring interpretability~\cite{margot2021new}, RUS deliberately excludes predictive performance when measuring model understandability so that the two axes used in the Pareto analysis remain independent. 

For the predictive performance axis in the Pareto analysis, we use mean F1 score as the measure for classification, while for regression tasks we use the mean $R^2$ score as the predictive performance measure. For classification tasks, the F1 score is used as the predictive performance axis because it balances precision and recall simultaneously, providing a more robust measure than accuracy on imbalanced datasets or precision and recall individually, which each capture only one dimension of predictive performance. For regression tasks, the coefficient of Determination ($R^2$) is used because it is bounded between 0 and 1, making it directly comparable across datasets with different scales and consistent with the classification setting where higher values always indicate better performance. This setup ensures that higher values indicate better performance on both axes.

\subsubsection{Extended Interpretability Score (E-IPS)}
\label{sssec:E-IPS}

To further quantify the interpretability of the model, and recognizing the desirability of having a single score for each model for comparative purposes, we introduce an \emph{Enhanced Interpretability Score} (E-IPS) as an extension of the original IPS proposed by Margot and Luta~\cite{margot2021new}. E-IPS is calculated based on a weighted combination of (1)~the F-1 score (for classification) or $R^2$ (for regression); (2)~the consistency or robustness of the model (Dice-Sørensen); (3)~the simplicity of the model (normalized number of rules); and (4)~the rule complexity (normalized number of components in a rule):
\begin{equation}
\text{E-IPS}(A) = {\gamma}_P P_n(A) + {\gamma}_S S_n(A) + {\gamma}_{simp} (1-R_n^{\text{simp}}(A)) + {\gamma}_{comp} (1-R_n^{\text{comp}}(A)).
\end{equation}
In the present work, each E-IPS component is equally weighted (${\gamma}_P={\gamma}_S = {\gamma}_{simp} = {\gamma}_{comp} = 0.25$). $R_n^{\text{simp}}(A)$ and $R_n^{\text{comp}}(A)$ are inverted to account for lower values, which would contribute to higher interpretability. 

E-IPS is computed and compared exclusively among rule-based models, specifically RuleFit, SR4-Fit, Random Forest, and Decision Tree, because interpretability, as captured by the E-IPS, is inherently a property of rule-based representations. Black-box models such as SVM and XGBoost and interpretable models such as logistic and LASSO regression do not produce rule sets. They therefore cannot be meaningfully evaluated on metrics that depend on rule structure, stability, and complexity. Restricting the E-IPS comparison to rule-based models ensures that the evaluation is both fair and methodologically consistent.

\subsubsection{Statistical Analysis}
\label{sssec:stat_analysis}

To assess whether the performance differences between SR4-Fit 
and RuleFit are statistically meaningful, the Wilcoxon signed-rank test \cite{noether1992introduction} is applied to compare the two models across all performance metrics. A significance threshold $p < 0.05$ is applied throughout. The comparison is conducted exclusively between SR4-Fit and RuleFit, as these are the two primary rule-based models sharing a similar interpretability objective, making a direct statistical comparison between them the most informative and methodologically appropriate. The full Wilcoxon test results, including W-statistics and p-values across all metrics and demographic dataset configurations, are provided in Supplemental Material Section~\ref{sup_sec_Wilcoxon}.

\subsubsection{Visualizations}
\label{sssec:viz}

To provide a high-level visual summary, aggregate Pareto figures are produced by averaging results across all dataset variants within each experimental condition, giving a big-picture view of how the models compare overall.
To facilitate more extensive visual comparative analysis, violin plots and line plots were also used~\cite{tukey1977exploratory}. These visualizations help to illustrate the variability, central tendency, and distribution of each model's performance in aggregate and across trials, making it easier to assess both consistency and predictive strength~\cite{suh2023metrics}.

\section{Results}\label{results}

SR4-Fit is evaluated across four experimental settings: (1)~U.S. House election classification, (2)~Democratic and Republican party voting percentage regression, (3)~a collection of six benchmark classification datasets, and (4)~a collection of eight benchmark regression datasets. In each setting, predictive performance is benchmarked against Random Forest, SVM, XGBoost, Decision Tree, RuleFit, and linear baselines (logistic regression for classification and LASSO regression for regression) (see Supplemental Material in the \hyperref[appendices]{Appendices}), while interpretability is assessed only for rule-based models, RuleFit, Decision Tree, and Random Forest, using the Rule Understandability Score (RUS) and the Enhanced Interpretability Score (E-IPS). All experiments are conducted over 30 independent trials. For the demographic datasets, experiments are conducted under both full-information settings, where party vote percentage features are included, and reduced-information settings, where these features are removed, enabling a systematic assessment of SR4-Fit's robustness and generalization under varying degrees of feature availability. We first present the aggregate Pareto analysis, which summarizes the predictivity--interpretability trade-off across all four experimental settings, followed by the detailed per-dataset E-IPS results in the subsections that follow.


\subsection{Pareto Analysis}
\label{sec:pareto_results}

\begin{figure}
\centering
\begin{minipage}{0.49\linewidth}
    \includegraphics[width=\linewidth]{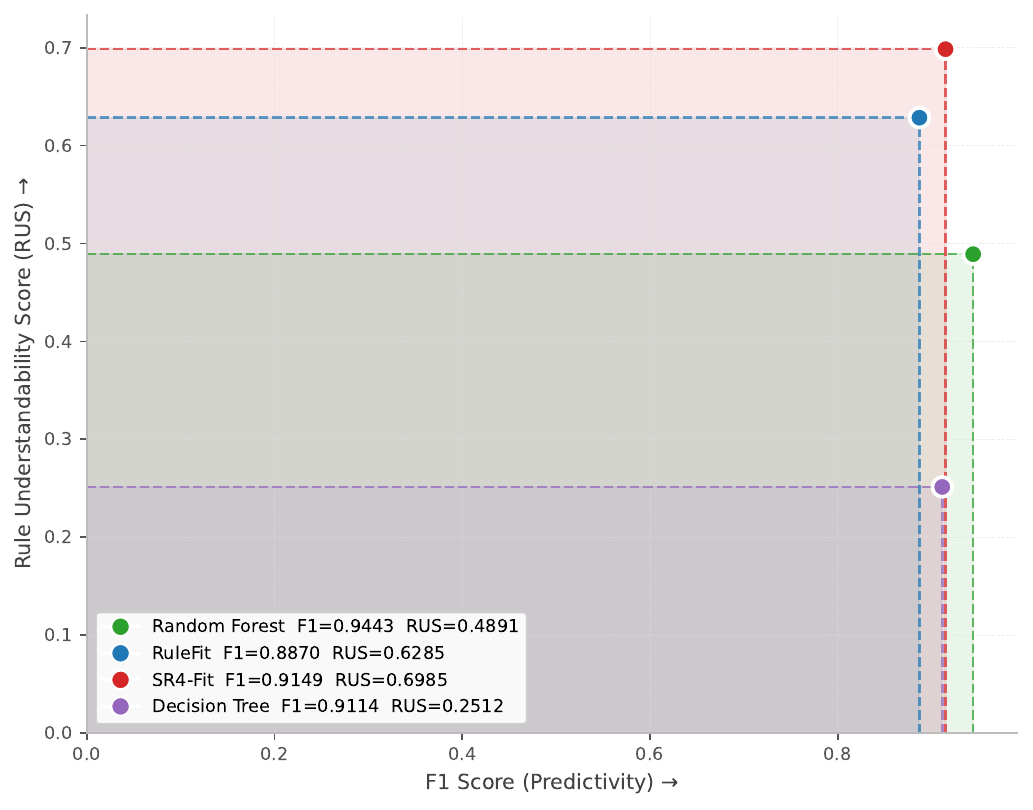}
\end{minipage}\hfill
\begin{minipage}{0.49\linewidth}
    \includegraphics[width=\linewidth]{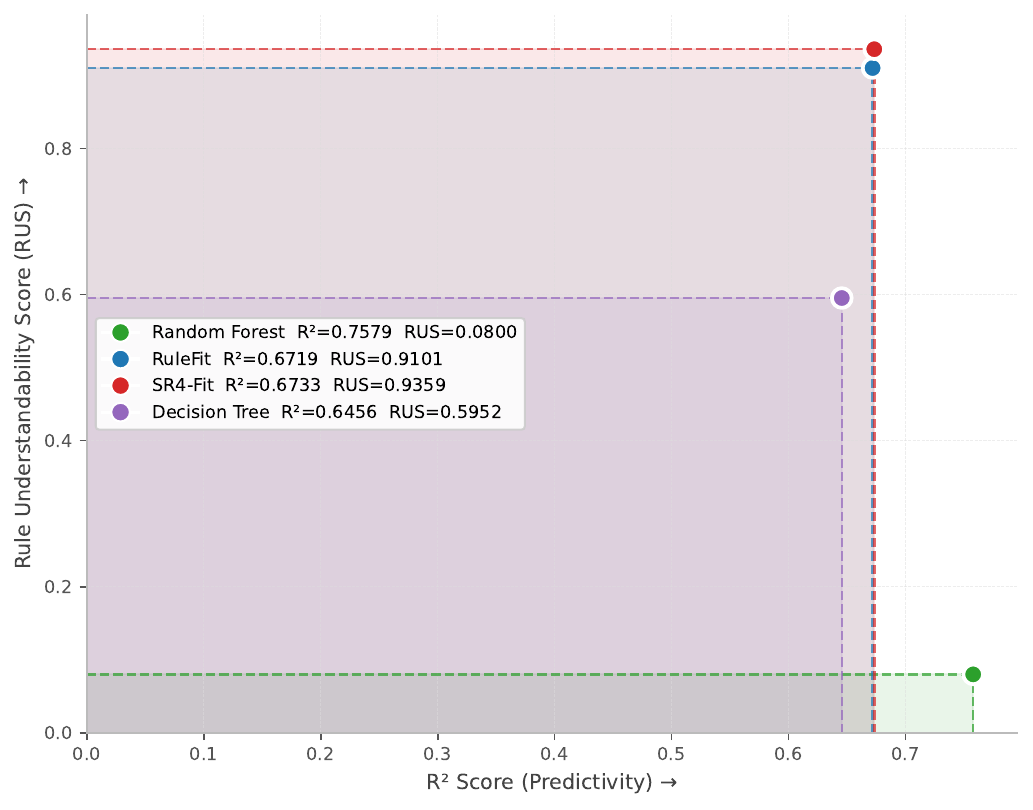}
\end{minipage}

\caption{
Aggregate Pareto plots for the U.S. House election classification and regression tasks, showing mean predictivity (F1 score for classification, left; $R^2$ score for regression, right) against Rule Understandability Score (RUS). The classification panel averages results across all four demographic datasets and both feature configurations (with and without party percentages). The regression panel averages results across all four datasets and all four experimental configurations (DEM as label with and without REP percentage; REP as label with and without DEM percentage).
}
\label{fig:pareto_voting}
\end{figure}

Figure~\ref{fig:pareto_voting} presents the aggregate Pareto plots for the U.S. House election classification and regression tasks. The left panel shows the classification results, averaged across all four demographic datasets and both feature configurations (with and without party percentages). SR4-Fit achieves the highest RUS (0.6985) among all models while maintaining a competitive mean F1 score (0.9149), Pareto-dominating both RuleFit (F1=0.8870, RUS=0.6285) and Decision Tree (F1=0.9114, RUS=0.2512), as SR4-Fit exceeds both models in predictivity and interpretability (RUS) simultaneously. Random Forest attains the highest mean F1 score (0.9443) but a substantially lower RUS (0.4891) than SR4-Fit, reflecting a predictivity-interpretability trade-off rather than outright dominance.

The right panel of Figure~\ref{fig:pareto_voting} shows the aggregate Pareto plot for the demographic regression tasks, averaged across all four datasets and all four experimental configurations (DEM as label with and without REP percentage; REP as label with and without DEM percentage). SR4-Fit Pareto-dominates both RuleFit ($R^2$=0.6719, RUS=0.9101) and Decision Tree ($R^2$=0.6456, RUS=0.5952), attaining a higher mean $R^2$ (0.6733) and higher RUS (0.9359) than either baseline. Random Forest attains the highest mean $R^2$ (0.7579) but the lowest RUS (0.0800) among all four models.

\begin{figure}
\centering
\begin{minipage}{0.49\linewidth}
    \includegraphics[width=\linewidth]{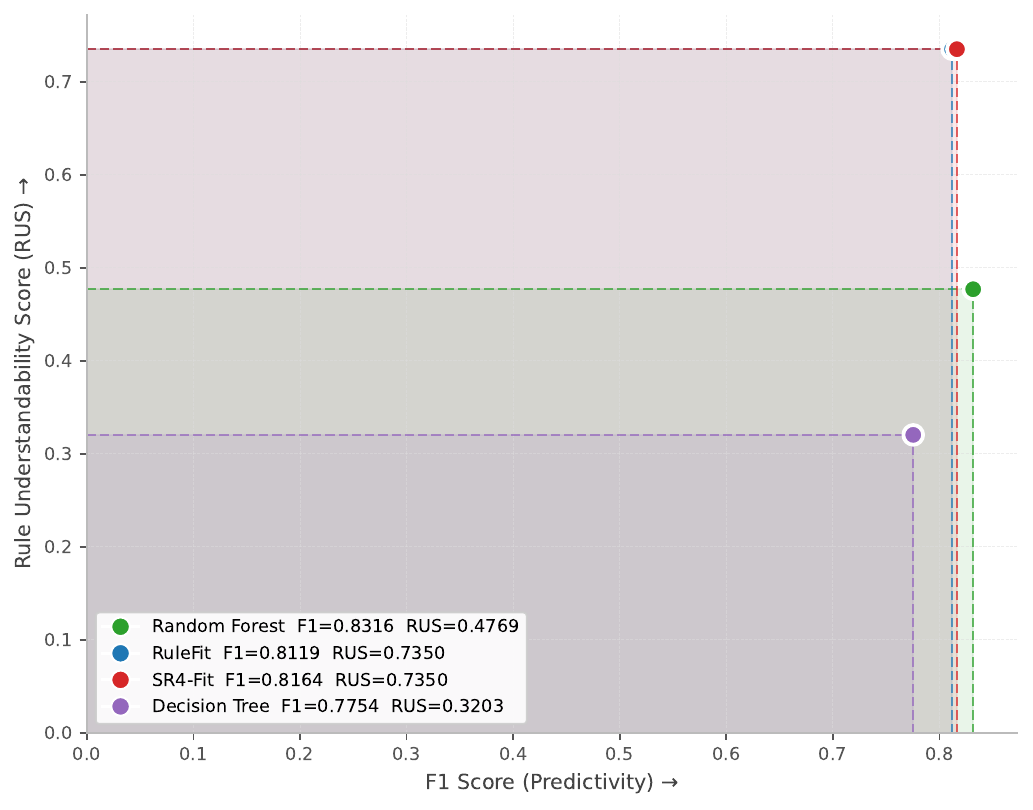}
\end{minipage}\hfill
\begin{minipage}{0.49\linewidth}
    \includegraphics[width=\linewidth]{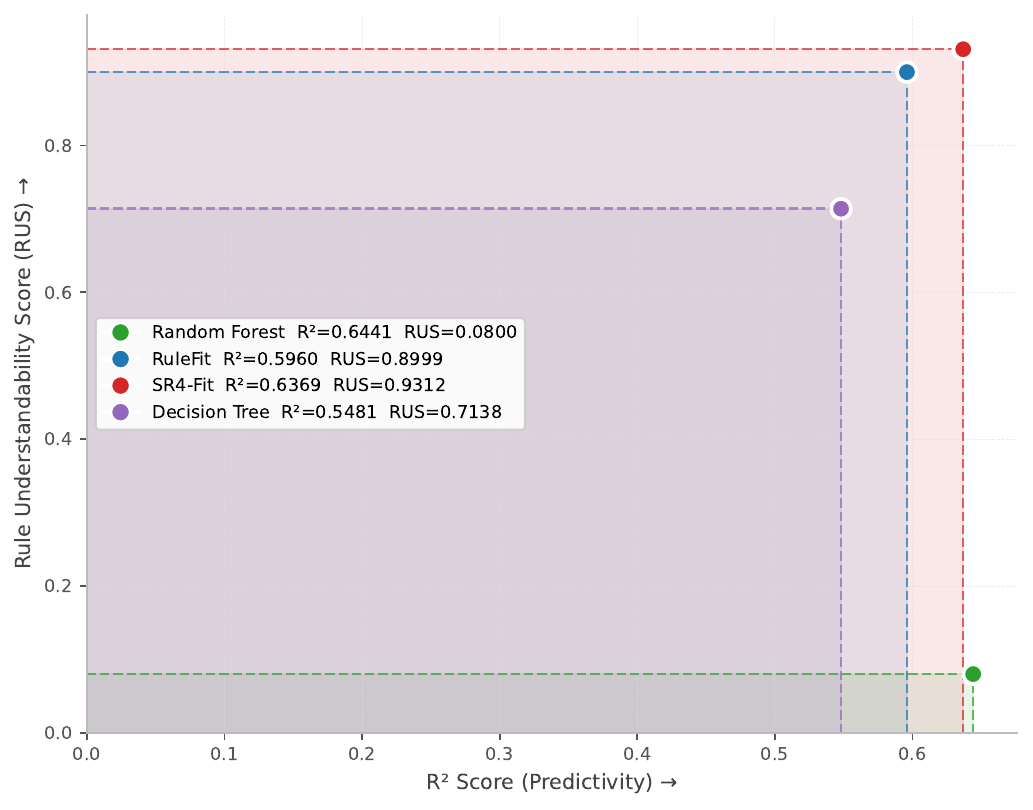}
\end{minipage}

\caption{
Aggregate Pareto plots for the public benchmark classification and regression tasks, showing mean predictivity (F1 score for classification, left; $R^2$ score for regression, right) against Rule Understandability Score (RUS). The classification panel averages results across the six public classification benchmark datasets. The regression panel averages results across the eight public regression benchmark datasets.
}
\label{fig:pareto_public}
\end{figure}

Figure~\ref{fig:pareto_public} presents the aggregate Pareto plots for the public benchmark classification and regression tasks. The left panel shows the results for the six public classification benchmark datasets. SR4-Fit and RuleFit achieve an identical RUS (0.7350), the highest among all four models, with SR4-Fit attaining a marginally higher mean F1 score (0.8164) than RuleFit (0.8119). SR4-Fit also Pareto-dominates Decision Tree (F1=0.7754, RUS=0.3203) on both axes. Random Forest attains the highest mean F1 score (0.8316) but a lower RUS (0.4769) than SR4-Fit, again reflecting a predictivity--interpretability trade-off.

The right panel of Figure~\ref{fig:pareto_public} shows the aggregate Pareto plot for the eight public regression benchmark datasets. SR4-Fit Pareto-dominates both RuleFit ($R^2$=0.5960, RUS=0.8999) and Decision Tree ($R^2$=0.5481, RUS=0.7138), attaining a higher mean $R^2$ (0.6369) and higher RUS (0.9312) than either baseline. Random Forest attains the highest mean $R^2$ (0.6441) but the lowest RUS (0.0800) among all four models, reflecting a predictivity--interpretability trade-off.

The RUS values for the rule-based interpretable models are consistently higher in the regression settings (approximately 0.90--0.94 for SR4-Fit and RuleFit) than in the classification settings (approximately 0.63--0.74), a pattern driven by Random Forest's markedly weaker performance on all three RUS components in regression relative to classification. In classification, Random Forest attains the lowest average rule complexity and comparatively lower rule count (when compared with regression data rule count) among all four models (approximately 1.9--2.0 conditions per rule; see Supplemental Material Section~\ref{sup_sec_HouseC} and Section~\ref{sup_sec_BenchC}) and near-zero stability (see Supplemental Section~\ref{sup_sec_BenchC}). In regression, by contrast, Random Forest shows no such offsetting strength: its rule complexity is inconsistent and, in several datasets, the highest among all four models (approximately 5--6 conditions per rule; see Supplemental Material Section~\ref{sup_sec_HouseRD}, Section~\ref{sup_sec_HouseRR}, and Section~\ref{sup_sec_BenchR}), while its rule counts remain on the order of 8000--9000, substantially larger than the rule counts produced by SR4-Fit and RuleFit (see Supplemental Material Section~\ref{sup_sec_HouseRD}, Section~\ref{sup_sec_HouseRR}, and Section~\ref{sup_sec_BenchR}), and its stability remains near zero. Because Random Forest underperforms simultaneously on rule count, rule complexity, and stability in regression, whereas its favorable rule complexity partially compensates for its other weaknesses in classification, the gap between Random Forest and the three rule-based models (RuleFit, SR4-Fit, and Decision Tree) is proportionally larger in regression than in classification. This widened gap is reflected directly in the higher RUS values attained by SR4-Fit and RuleFit in the regression settings.

Overall, across all four aggregate views, SR4-Fit consistently Pareto-dominates RuleFit and Decision Tree, achieving higher predictivity and higher RUS simultaneously in every setting. Relative to Random Forest, the same trade-off pattern holds throughout: Random Forest attains marginally higher predictivity, while SR4-Fit attains substantially higher RUS in each of the four settings. This consistent pattern reflects SR4-Fit's balance between predictive performance and interpretability across both election and public benchmark tasks.


\subsection{Classification Results}
\label{sec:cls_results}

\begin{figure}
\centering
\begin{minipage}{0.49\linewidth}
    \includegraphics[width=\linewidth]{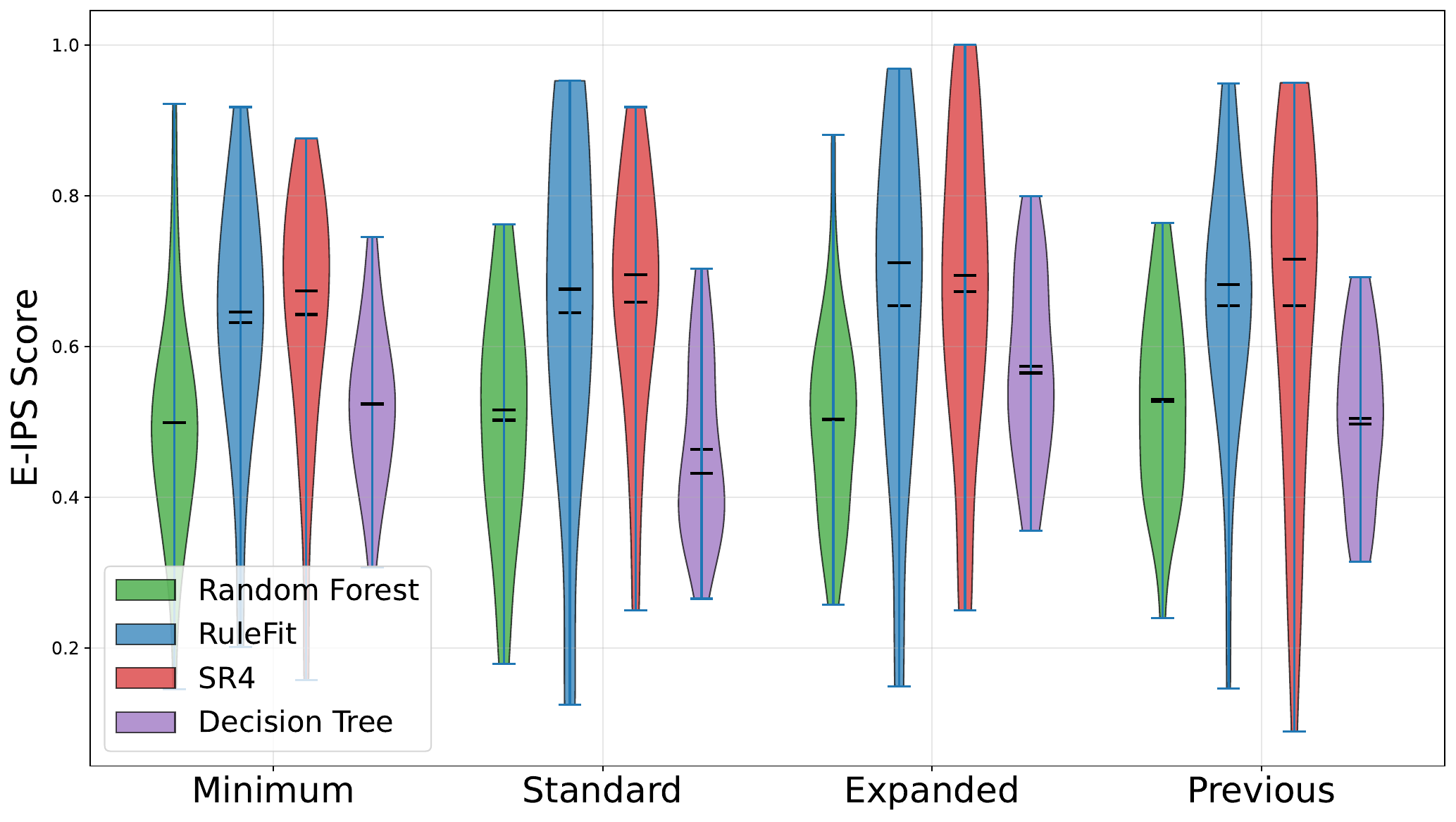}
\end{minipage}\hfill
\begin{minipage}{0.49\linewidth}
    \includegraphics[width=\linewidth]{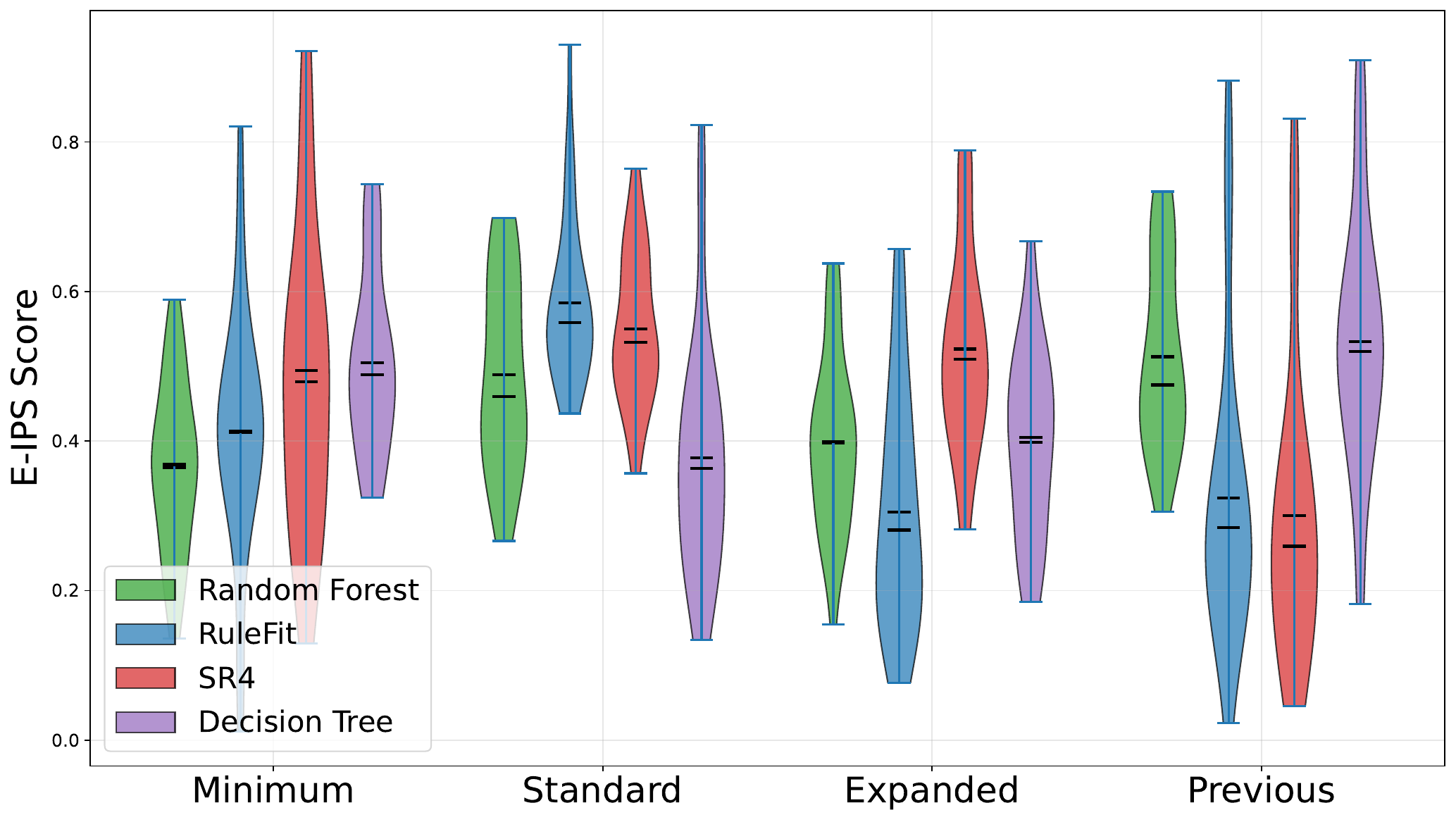}
\end{minipage}

\caption{Distribution of enhanced interpretability scores (E-IPS) across demographic classification datasets. The left panel reports results obtained using feature sets that include party percentage information, whereas the right panel shows results with party percentages removed.
}
\label{fig:E-IPS_violin_combined}
\end{figure}

As shown in Table~\ref{tab:E-IPS_cls_with_vs_without_party}, SR4-Fit achieves the highest E-IPS values in three of the four demographic datasets when party percentages are included as features, with scores of 0.6423, 0.6584, and 0.6728 for the Minimum, Standard, and Expanded datasets, respectively; in the Previous dataset, RuleFit attains a marginally higher E-IPS (0.6544) than SR4-Fit (0.6541). The violin plots in Figure~\ref{fig:E-IPS_violin_combined} confirm that SR4-Fit produces the tightest E-IPS distributions across trials, indicating greater interpretability stability than all competing rule-based models. SR4-Fit also achieves the highest predictive performance across all four datasets under this setting, outperforming Random Forest, SVM, RuleFit, logistic regression, decision tree, and XGBoost across accuracy, precision, recall, and F1 score. SR4-Fit achieves the highest Dice-Sorensen Index among rule-based models in three of the four datasets, with RuleFit leading in the standard dataset but closely followed by SR4-Fit. In contrast, Random Forest and Decision Tree produce near-zero stability scores across all datasets, indicating structurally inconsistent rule sets across trials. Regarding compactness, the decision tree produces the fewest rules when party percentages are included, followed closely by SR4-Fit and RuleFit, while Random Forest generates substantially larger rule sets. Full predictive line plots, stability tables, and rule compactness results are provided in Supplemental Material Section~\ref{sup_sec_BenchC}.

When party percentages are removed,  E-IPS values in Table~\ref{tab:E-IPS_cls_with_vs_without_party} decline across all models, reflecting the removal of highly informative features. Under this reduced-information setting, XGBoost and SVM maintain the highest predictive accuracy, while SR4-Fit produces smoother and more stable performance trajectories than RuleFit and Decision Tree. SR4-Fit achieves the highest E-IPS in the expanded dataset, while the decision tree leads in the minimum and previous datasets. However, this advantage comes at the cost of structural instability, the Decision Tree's Dice-Sørensen scores remain near zero across trials, indicating that its rule sets change substantially between runs and cannot be relied upon for consistent interpretation. SR4-Fit and RuleFit maintain more consistent rule structures under feature reduction. Detailed predictive line plots and stability analyses for the reduced-information setting are provided in Supplementary Material Section~\ref{sup_sec_BenchC}. Overall, SR4-Fit delivers the strongest and most consistent combination of predictive performance, interpretability, stability, and compactness when politically informative features are available and remains among the most stable rule-based models when they are not.


\begin{table}
\caption{Average enhanced interpretability score (E-IPS) $\pm$ standard deviation across 30 trials for demographic classification datasets. Bold indicates the best-performing method per dataset.
}
\label{tab:E-IPS_cls_with_vs_without_party}
\centering
\setlength{\tabcolsep}{4pt}
\renewcommand{\arraystretch}{1.15}

\begin{subtable}{\columnwidth}
\centering
\begin{tabular}{lcccc}
\toprule
\textbf{Dataset}
& \textbf{Random Forest} & \textbf{RuleFit} & \textbf{SR4-Fit} & \textbf{Decision Tree} \\
\midrule
Minimum  & 0.4991$\pm$0.1478 & 0.6313$\pm$0.1657 & \textbf{0.6423$\pm$0.1678} & 0.5246$\pm$0.1035 \\
Standard & 0.5023$\pm$0.1512 & 0.6447$\pm$0.2353 & \textbf{0.6584$\pm$0.1649} & 0.4635$\pm$0.1165 \\
Expanded & 0.5030$\pm$0.1294 & 0.6542$\pm$0.2184 & \textbf{0.6728$\pm$0.2064} & 0.5738$\pm$0.1231 \\
Previous & 0.5267$\pm$0.1249 & \textbf{0.6544$\pm$0.1711} & 0.6541$\pm$0.2264 & 0.5045$\pm$0.1062 \\
\bottomrule
\end{tabular}
\caption{\textbf{With party percentages.}}
\end{subtable}

\vspace{6pt}

\begin{subtable}{\columnwidth}
\centering
\begin{tabular}{lcccc}
\toprule
\textbf{Dataset}
& \textbf{Random Forest} & \textbf{RuleFit} & \textbf{SR4-Fit} & \textbf{Decision Tree} \\
\midrule
Minimum  & 0.3646$\pm$0.1111 & 0.4114$\pm$0.1635 & 0.4792$\pm$0.1996 & \textbf{0.5047$\pm$0.1214} \\
Standard & 0.4888$\pm$0.1261 & \textbf{0.5851$\pm$0.1095} & 0.5497$\pm$0.0981 & 0.3772$\pm$0.1584 \\
Expanded & 0.3978$\pm$0.1157 & 0.3050$\pm$0.1551 & \textbf{0.5230$\pm$0.1319} & 0.3982$\pm$0.1215 \\
Previous & 0.5128$\pm$0.1230 & 0.3240$\pm$0.1977 & 0.3000$\pm$0.1938 & \textbf{0.5329$\pm$0.1683} \\
\bottomrule
\end{tabular}
\caption{\textbf{Without party percentages.}}
\end{subtable}

\end{table}


\subsection{DEM Regression Results}
\label{sec:dem_results}

\begin{figure}
\centering
\begin{minipage}{0.49\linewidth}
    \includegraphics[width=\linewidth]{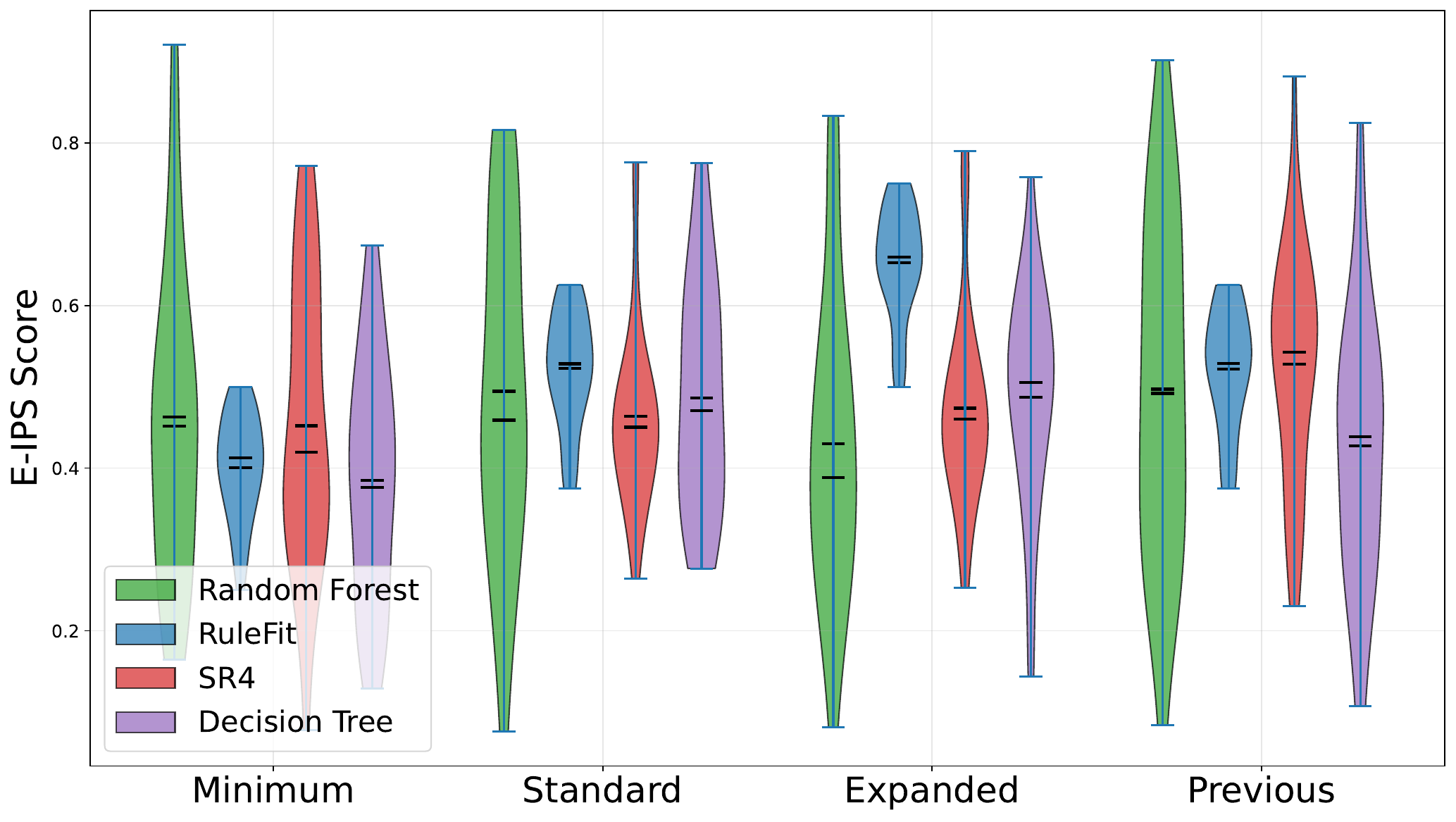}
\end{minipage}\hfill
\begin{minipage}{0.49\linewidth}
    \includegraphics[width=\linewidth]{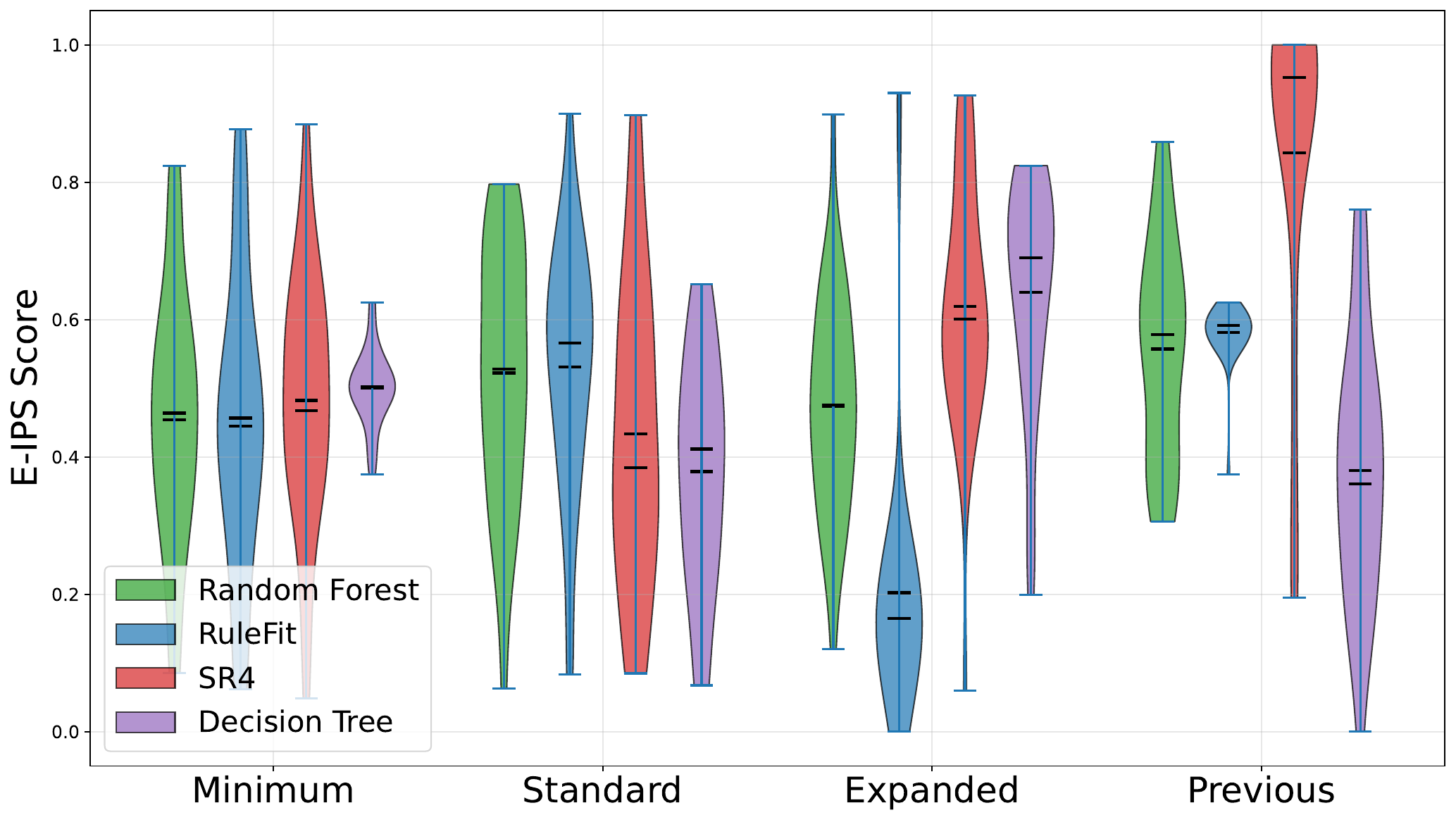}
\end{minipage}

\caption{
Distribution of enhanced interpretability scores (E-IPS) across demographic regression datasets when predicting Democratic vote percentage. Results are shown for models trained with all party-related features (left) and for models trained with Republican percentage features removed (right).
}
\label{fig:E-IPS_violin_combined_dem}
\end{figure}

As shown in Table~\ref{tab:E-IPS_dem_vs_dem_norep}, when REP percentage is included as a feature, RuleFit achieves the highest E-IPS in the standard and expanded datasets, while SR4-Fit leads in the minimum and previous datasets with scores of 0.4521 and 0.5278, respectively. RuleFit's higher E-IPS is accompanied by lower Dice-Sørensen stability and higher rule complexity than SR4-Fit. The violin plots in Figure~\ref{fig:E-IPS_violin_combined_dem} confirm that SR4-Fit maintains tighter and more stable E-IPS distributions than the baselines when REP percentage is included. In terms of predictive performance, XGBoost attains the lowest average error across most datasets, while SR4-Fit produces comparable RMSE, MAE, and $R^2$ values with notably smoother and more stable trajectories across trials. Full predictivity line plots, stability tables, and rule compactness results for this setting are provided in the Supplemental Material Section~\ref{sup_sec_HouseRD}).

When REP percentage is removed, SR4-Fit achieves the highest E-IPS in the previous dataset, with a score of 0.8427, outperforming RuleFit, Random Forest, and Decision Tree. In the minimum and expanded datasets, Decision Tree attains the highest E-IPS of 0.5006 and 0.6397, narrowly ahead of SR4-Fit 0.4822 and 0.6190, respectively.  The violin plots in Figure~\ref{fig:E-IPS_violin_combined_dem} confirm that SR4-Fit maintains tighter and more stable E-IPS distributions than competing models across both feature configurations, while all models show more concentrated distributions when REP percentage is excluded. Error values increase across all models under this reduced-information setting, with XGBoost still attaining the lowest average error, while SR4-Fit maintains comparatively smooth performance trends and more concentrated error distributions than the baselines. Detailed predictive line plots and stability analyses for the reduced-information setting are provided in Supplemental Material Section~\ref{sup_sec_HouseRD}.



\begin{table}
\caption{
Average enhanced interpretability score (E-IPS) $\pm$ standard deviation across 30 trials for demographic regression datasets. Results are reported for predicting Democratic vote percentage, with and without Republican percentage included as a feature. Bold indicates the best-performing method per dataset.
}
\label{tab:E-IPS_dem_vs_dem_norep}
\centering
\setlength{\tabcolsep}{4pt}
\renewcommand{\arraystretch}{1.15}

\begin{subtable}{\columnwidth}
\centering
\begin{tabular}{lcccc}
\toprule
\textbf{Dataset}
& \textbf{Random Forest} & \textbf{RuleFit} & \textbf{SR4-Fit} & \textbf{Decision Tree} \\
\midrule
Minimum  
& 0.4520$\pm$0.1848 & 0.4006$\pm$0.0643 & \textbf{0.4521$\pm$0.1708} & 0.3853$\pm$0.1440 \\
Standard 
& 0.4946$\pm$0.1977 & \textbf{0.5227$\pm$0.0697} & 0.4635$\pm$0.1110 & 0.4866$\pm$0.1393 \\
Expanded 
& 0.4297$\pm$0.1844 & \textbf{0.6524$\pm$0.0667} & 0.4737$\pm$0.1204 & 0.4874$\pm$0.1315 \\
Previous 
& 0.4971$\pm$0.2036 & 0.5219$\pm$0.0731 & \textbf{0.5278$\pm$0.1410} & 0.4274$\pm$0.1620 \\
\bottomrule
\end{tabular}
\caption{\textbf{With the Republican percentage as a feature.}}
\end{subtable}

\vspace{6pt}

\begin{subtable}{\columnwidth}
\centering
\begin{tabular}{lcccc}
\toprule
\textbf{Dataset}
& \textbf{Random Forest} & \textbf{RuleFit} & \textbf{SR4-Fit} & \textbf{Decision Tree} \\
\midrule
Minimum  
& 0.4541$\pm$0.1836 & 0.4448$\pm$0.2087 & 0.4822$\pm$0.1790 & \textbf{0.5006$\pm$0.0546} \\
Standard 
& 0.5223$\pm$0.1855 & \textbf{0.5309$\pm$0.1786} & 0.4338$\pm$0.2211 & 0.3786$\pm$0.1643 \\
Expanded 
& 0.4735$\pm$0.1587 & 0.2022$\pm$0.2001 & 0.6190$\pm$0.1787 & \textbf{0.6397$\pm$0.1637} \\
Previous 
& 0.5571$\pm$0.1550 & 0.5809$\pm$0.0433 & \textbf{0.8427$\pm$0.2512} & 0.3805$\pm$0.1844 \\
\bottomrule
\end{tabular}
\caption{\textbf{Without the Republican percentage as a feature.}}
\end{subtable}
\end{table}


\subsection{REP Regression Results}
\label{sec:rep_results}

\begin{figure}
\centering
\begin{minipage}{0.49\linewidth}
    \includegraphics[width=\linewidth]{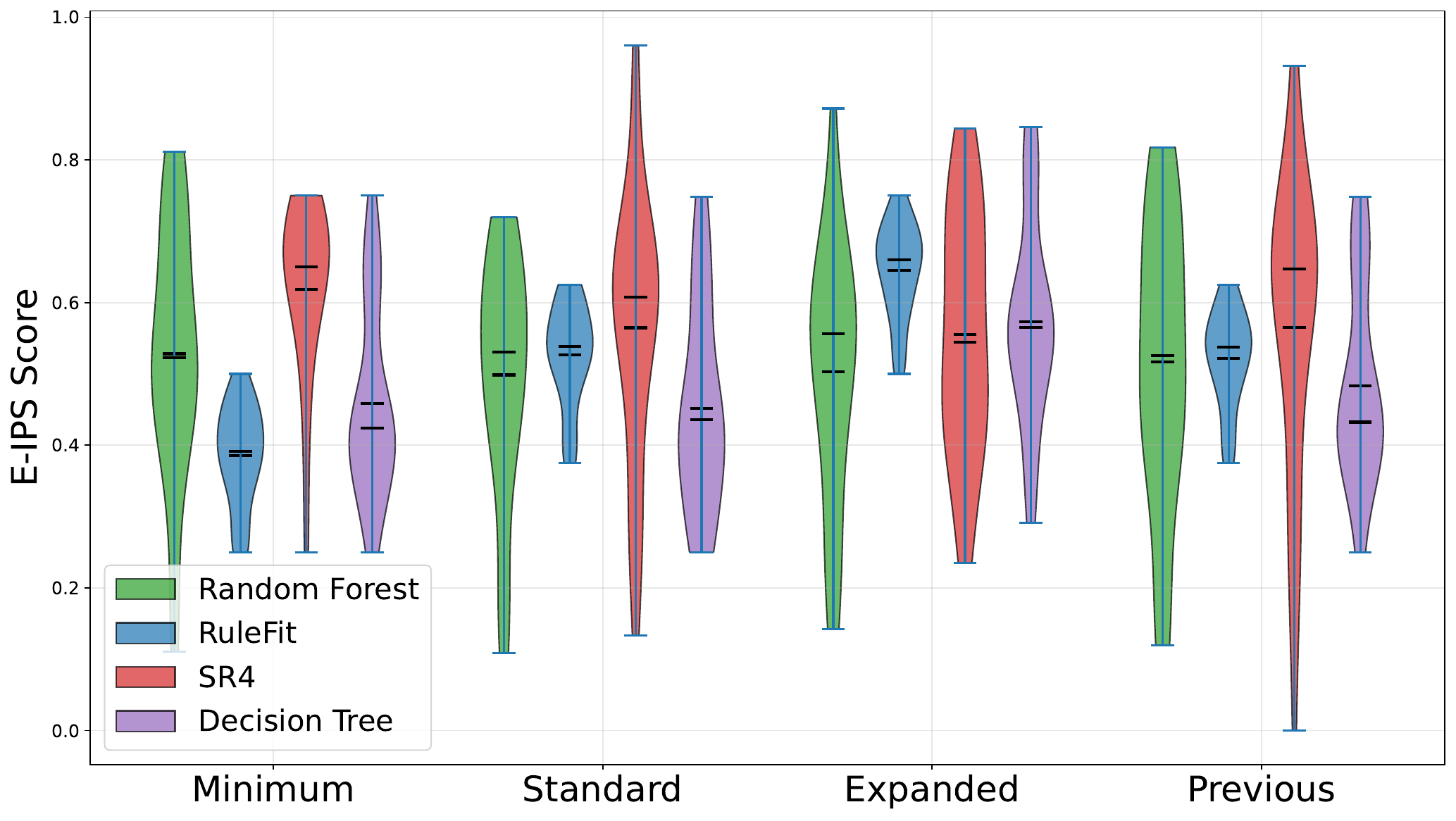}
\end{minipage}\hfill
\begin{minipage}{0.49\linewidth}
    \includegraphics[width=\linewidth]{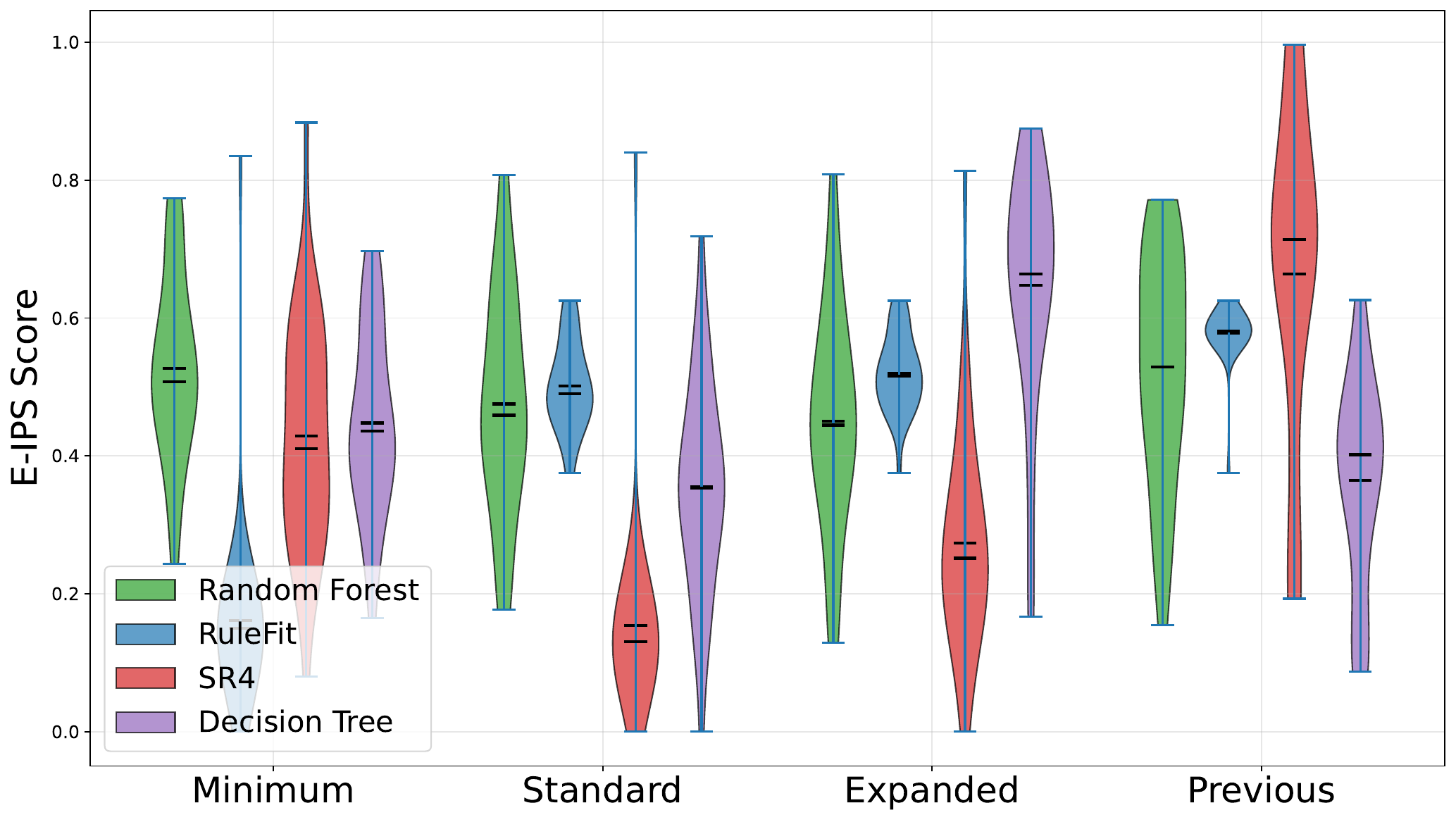}
\end{minipage}

\caption{
Distribution of enhanced interpretability scores (E-IPS) across demographic regression datasets when predicting Republican vote percentage. Results are shown for models trained with all party-related features (left) and for models trained with Democratic percentage features removed (right).
}
\label{fig:E-IPS_violin_combined_rep}
\end{figure}

As shown in Table~\ref{tab:E-IPS_rep_with_vs_no_dem}, SR4-Fit achieves the highest E-IPS in the Minimum, Standard, and Previous datasets when DEM percentage is included as a feature, with scores of 0.6182, 0.5654, and 0.5647, respectively, while RuleFit performs strongest in the Expanded dataset. The violin plots in Figure~\ref{fig:E-IPS_violin_combined_rep} confirm that SR4-Fit maintains tighter and more stable E-IPS distributions than the baselines when DEM percentage is included. In terms of predictive performance, XGBoost attains the lowest average error across most datasets, while SR4-Fit produces comparable RMSE, MAE, and $R^2$ values with smoother and more stable trajectories across trials. Despite RuleFit's advantage in the expanded dataset, SR4-Fit achieves higher Dice-Sørensen stability and lower rule complexity than RuleFit across most dataset configurations, indicating that SR4-Fit's interpretability advantage is accompanied by greater structural consistency and more compact rule sets. Full line plots, stability tables, and rule compactness results for this setting are provided in Supplemental Material Section~\ref{sup_sec_HouseRR}.

When DEM percentage is removed, E-IPS values decline across all models, but SR4-Fit retains the highest interpretability in the previous dataset with a score of 0.6639. The violin plots in Figure~\ref{fig:E-IPS_violin_combined_rep} confirm that SR4-Fit maintains tighter and more stable E-IPS distributions than competing models across both feature configurations, whereas RuleFit exhibits widening variance and Random Forest produces near-zero stability scores throughout. Error values increase across all models under this reduced-information setting, with XGBoost still achieving the lowest average error, while SR4-Fit maintains comparatively smooth performance trends and more concentrated error distributions than the baselines. Detailed line plots and stability analyses for the reduced-information setting are provided in Supplemental Material Section~\ref{sup_sec_HouseRR}.



\begin{table}
\caption{
Average enhanced interpretability score (E-IPS) $\pm$ standard deviation across 30 trials for demographic regression datasets. Results are reported for predicting Republican vote percentage, with and without Democratic percentage included as a feature. Bold indicates the best-performing method per dataset.
}
\label{tab:E-IPS_rep_with_vs_no_dem}
\centering
\setlength{\tabcolsep}{3pt}
\renewcommand{\arraystretch}{1.15}

\begin{subtable}{\columnwidth}
\centering
\begin{tabular}{lcccc}
\toprule
\textbf{Dataset}
& \textbf{Random Forest} & \textbf{RuleFit} & \textbf{SR4-Fit} & \textbf{Decision Tree} \\
\midrule
Minimum  
& 0.5281$\pm$0.1753 & 0.3858$\pm$0.0685 & \textbf{0.6182$\pm$0.1214} & 0.4583$\pm$0.1298 \\
Standard 
& 0.5166$\pm$0.1958 & 0.5221$\pm$0.0665 & \textbf{0.5654$\pm$0.2094} & 0.4830$\pm$0.1370 \\
Expanded 
& 0.5026$\pm$0.1753 & \textbf{0.6448$\pm$0.0646} & 0.5555$\pm$0.1658 & 0.5733$\pm$0.1400 \\
Previous 
& 0.4986$\pm$0.1640 & 0.5270$\pm$0.0701 & \textbf{0.5647$\pm$0.1853} & 0.4516$\pm$0.1421 \\
\bottomrule
\end{tabular}
\caption{\textbf{With the Democratic percentage as a feature.}}
\end{subtable}

\vspace{6pt}

\begin{subtable}{\columnwidth}
\centering
\begin{tabular}{lcccc}
\toprule
\textbf{Dataset}
& \textbf{Random Forest} & \textbf{RuleFit} & \textbf{SR4-Fit} & \textbf{Decision Tree} \\
\midrule
Minimum  
& \textbf{0.5271$\pm$0.1315} & 0.1610$\pm$0.1393 & 0.4290$\pm$0.1635 & 0.4478$\pm$0.1320 \\
Standard 
& 0.4754$\pm$0.1578 & \textbf{0.5016$\pm$0.0639} & 0.1536$\pm$0.1413 & 0.3551$\pm$0.1499 \\
Expanded 
& 0.4504$\pm$0.1578 & 0.5199$\pm$0.0562 & 0.2738$\pm$0.1583 & \textbf{0.6477$\pm$0.1725} \\
Previous 
& 0.5291$\pm$0.1660 & 0.5780$\pm$0.0422 & \textbf{0.6639$\pm$0.2393} & 0.3645$\pm$0.1469 \\
\bottomrule
\end{tabular}
\caption{\textbf{Without the Democratic percentage as a feature.}}
\end{subtable}
\end{table}


\subsection{Results on Standard Public Classification Datasets}
\label{sec:public_class_results}

\begin{figure}
\centering
\scriptsize

\begin{minipage}[t]{\linewidth}
    \centering
    \includegraphics[width=0.65\linewidth]{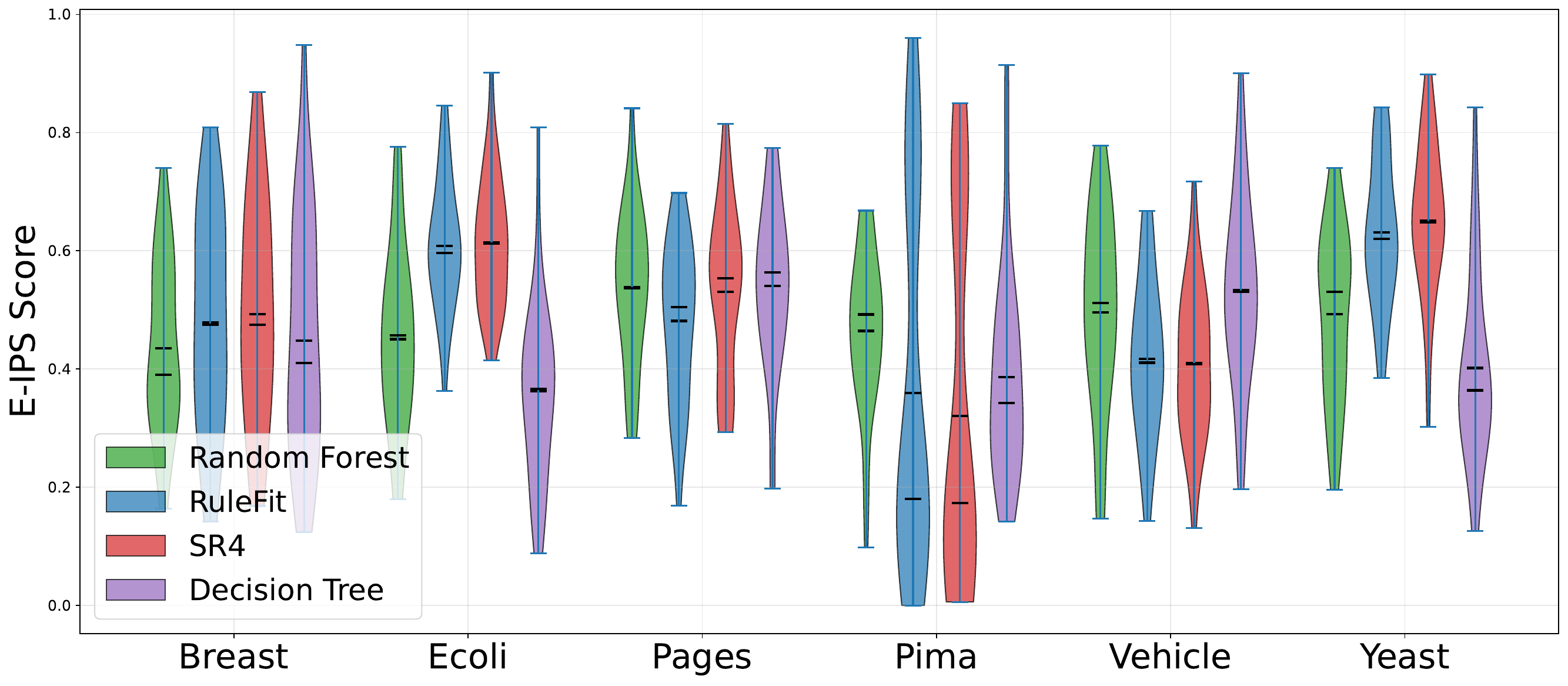}
\end{minipage}

\caption{
Violin plot comparison of enhanced interpretability scores for all rule-based models across standard public classification datasets.
}
\label{fig:E-IPS_violin_std}
\end{figure}

Across the six public classification benchmarks, SR4-Fit demonstrates a consistently strong balance between predictive performance, interpretability, and stability. As shown in Table~\ref{tab:E-IPS_public_class}, SR4-Fit achieves the highest E-IPS in three of the six datasets, including breast cancer (0.4931), E. coli (0.6141), and Yeast (0.6512), while remaining competitive in the remaining datasets. The violin plots in Figure~\ref{fig:E-IPS_violin_std} confirm that SR4-Fit's predictive distributions remain tightly clustered across all 30 trials, while RuleFit's distributions widen, and Random Forest exhibits the broadest and least consistent spread. The Dice–Sørensen stability results confirm that SR4-Fit and RuleFit achieve the highest stability across all six datasets, substantially outperforming Random Forest and Decision Tree, which produce near-zero scores. Although Random Forest occasionally achieves slightly higher accuracy on individual datasets, its trial-to-trial variability is substantially higher, and its interpretability remains near zero due to large, unstable rule sets. RuleFit performs competitively in accuracy but shows moderate instability and produces larger, more variable rule sets than SR4-Fit across most datasets.  Overall, the public classification experiments confirm that SR4-Fit reliably produces interpretable and stable rule sets while maintaining competitive predictive performance across diverse classification tasks. Full stability results, line plots, and violin plots for this setting are provided in Supplemental Material Section~\ref{sup_sec_BenchC}.


\begin{table}
\caption{
Average enhanced interpretability score (E-IPS) $\pm$ standard deviation across 30 trials for public classification datasets. Bold indicates the highest average E-IPS per dataset.
}
\label{tab:E-IPS_public_class}
\centering
\setlength{\tabcolsep}{3pt}
\renewcommand{\arraystretch}{1.15}
\begin{tabular}{lcccc}
\toprule
\textbf{Dataset}
& \textbf{Random Forest} & \textbf{RuleFit} & \textbf{SR4-Fit} & \textbf{Decision Tree} \\
\midrule
Breast Cancer
& 0.4350$\pm$0.1433 & 0.4793$\pm$0.1851 & \textbf{0.4931$\pm$0.1875} & 0.4477$\pm$0.2044 \\
E.~coli
& 0.4505$\pm$0.1475 & 0.6085$\pm$0.1087 & \textbf{0.6141$\pm$0.1093} & 0.3624$\pm$0.1494 \\
Page Blocks
& 0.5386$\pm$0.1296 & 0.4813$\pm$0.1322 & 0.5308$\pm$0.1381 & \textbf{0.5405$\pm$0.1391} \\
Pima Indians
& \textbf{0.4644$\pm$0.1330} & 0.3596$\pm$0.3061 & 0.3208$\pm$0.3011 & 0.3862$\pm$0.1781 \\
Vehicle
& 0.4958$\pm$0.1642 & 0.4175$\pm$0.1314 & 0.4099$\pm$0.1262 & \textbf{0.5338$\pm$0.1570} \\
Yeast
& 0.4927$\pm$0.1426 & 0.6313$\pm$0.1234 & \textbf{0.6512$\pm$0.1281} & 0.4017$\pm$0.1596 \\
\bottomrule
\end{tabular}

\end{table}


\subsection{Results on Standard Public Regression Datasets}
\label{sec:public_reg_results}

\begin{figure}
\centering
\scriptsize

\begin{minipage}[t]{\linewidth}
    \centering
    \includegraphics[width=0.75\linewidth]{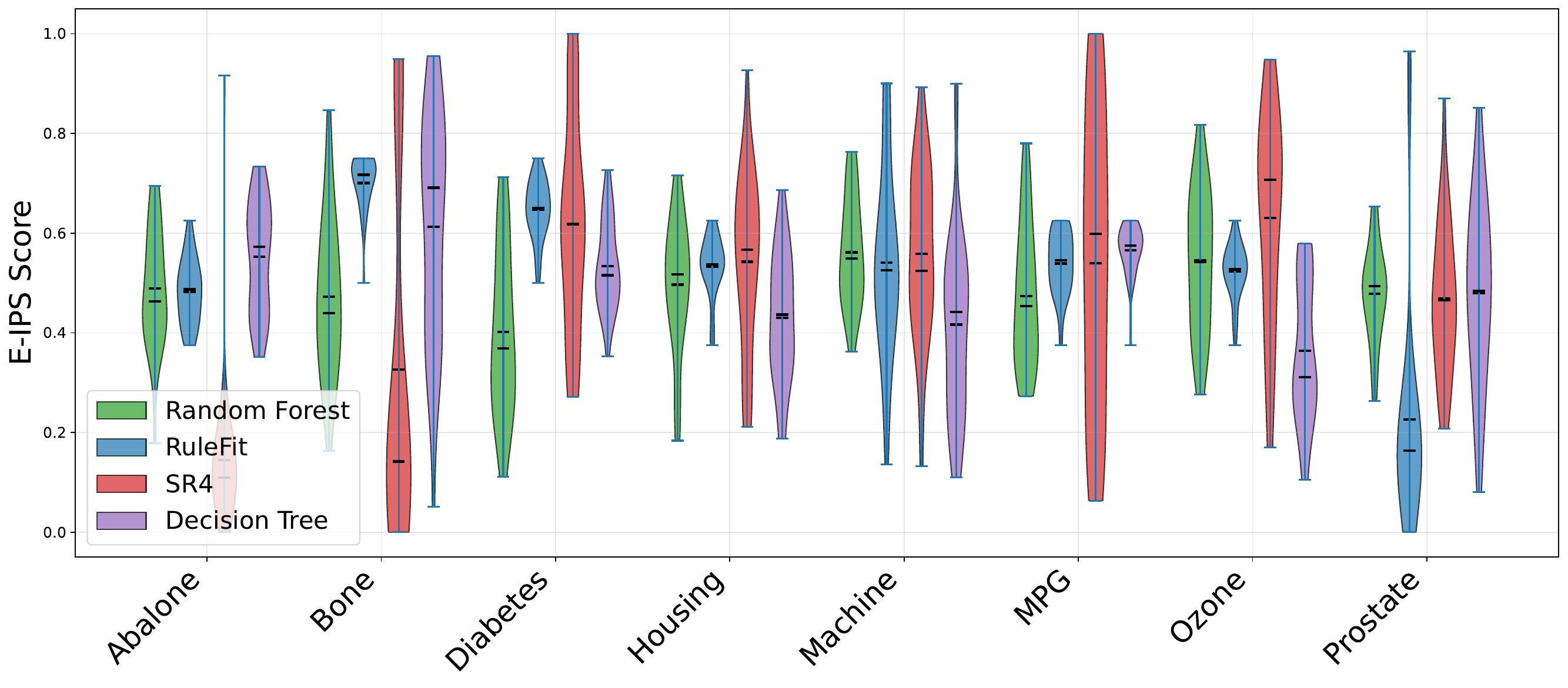}
\end{minipage}

\caption{
Violin plot comparison of enhanced interpretability scores for all rule-based models across standard public regression datasets.
}
\label{fig:E-IPS_violin_std_reg}
\end{figure}

Across the eight public regression benchmarks, SR4-Fit maintains competitive predictive accuracy while achieving stronger interpretability and rule stability than the baselines in most datasets. As shown in Table~\ref{tab:E-IPS_regression_datasets} and Figure~\ref{fig:E-IPS_violin_std_reg}, SR4-Fit attains the highest E-IPS in two of the eight datasets, Housing (0.5425) and Ozone (0.6303), while Decision Tree leads in four datasets: Abalone, Bone, MPG, and Prostate. RuleFit and Random Forest each lead in one dataset, Diabetes and Machine, respectively. The line plots in Supplemental Material Section~\ref{sup_sec_BenchC} show that SR4-Fit produces smooth RMSE, MAE, and $R^2$ curves with minimal variance across trials, in contrast to Random Forest's pronounced fluctuations and RuleFit's moderate instability. The Dice-Sørensen stability results confirm that SR4-Fit achieves the highest stability in six of the eight datasets, while Random Forest remains near zero and RuleFit performs inconsistently across datasets. Regarding compactness, SR4-Fit generates substantially fewer rules than Random Forest across all datasets and produces rule sets comparable in size to RuleFit but with markedly lower rule complexity in most cases. These findings demonstrate that SR4-Fit performs robustly across diverse regression benchmarks while providing more stable and concise rule representations than existing rule-based and ensemble baselines. Full line plots, stability results, and rule compactness results for this setting are provided in Supplemental Material Section~\ref{sup_sec_BenchR}. 


\begin{table}
\caption{
Average enhanced interpretability score (E-IPS) $\pm$ standard deviation 
across 30 trials for public regression datasets. Bold indicates 
the highest average E-IPS per dataset.
}
\label{tab:E-IPS_regression_datasets}
\centering
\setlength{\tabcolsep}{3pt}
\renewcommand{\arraystretch}{1.15}
\begin{tabular}{lcccc}
\toprule
\textbf{Dataset}
& \textbf{Random Forest} & \textbf{RuleFit} & \textbf{SR4-Fit} & \textbf{Decision Tree} \\
\midrule
Abalone
& 0.4890$\pm$0.1166 & 0.4824$\pm$0.0649 & 0.1444$\pm$0.1525 & \textbf{0.5524$\pm$0.1113} \\
Bone
& 0.4723$\pm$0.1565 & 0.5756$\pm$0.0549 & 0.3263$\pm$0.3564 & \textbf{0.6124$\pm$0.2261} \\
Diabetes
& 0.4019$\pm$0.1627 & \textbf{0.6474$\pm$0.0588} & 0.6176$\pm$0.2149 & 0.5338$\pm$0.0878 \\
Housing
& 0.4965$\pm$0.1320 & 0.5330$\pm$0.0613 & \textbf{0.5425$\pm$0.1730} & 0.4301$\pm$0.1264 \\
Machine
& \textbf{0.5615$\pm$0.1073} & 0.5410$\pm$0.1868 & 0.5588$\pm$0.1688 & 0.4165$\pm$0.1849 \\
MPG
& 0.4738$\pm$0.1402 & 0.5391$\pm$0.0615 & 0.5393$\pm$0.2905 & \textbf{0.5655$\pm$0.0507} \\
Ozone
& 0.5450$\pm$0.1397 & 0.5230$\pm$0.0548 & \textbf{0.6303$\pm$0.2125} & 0.3638$\pm$0.1400 \\
Prostate
& 0.4787$\pm$0.0951 & 0.2265$\pm$0.2401 & 0.4687$\pm$0.1499 & \textbf{0.4791$\pm$0.1833} \\
\bottomrule
\end{tabular}

\end{table}


\section{Discussion}\label{discussion}
The following analysis presents representative rules extracted from each dataset and interprets the associated demographic and structural characteristics of the regions these rules classify. In addition to 
classification-oriented rules, we also examine rules derived from regression-based demographic modeling, where SR4-Fit identifies structured combinations of predictors associated with continuous outcomes. In addition to this, detailed interpretations for the public classification and regression datasets are provided as well.

\subsection{Including Prior Party Voting Percentages (Classification)}
\label{ss:with_party_class}

In the minimum dataset, SR4-Fit builds a concise, interpretable model primarily driven by the Democratic and Republican voting percentages in prior elections. The most consistently important rule indicates that districts where the prior Democratic voting percentage is greater than about 48.7\% while the prior Republican voting percentage is at or below about 45.5\% are likely to elect Democrats. This pattern is not surprising, since districts where votes for Democratic candidates previously outnumbered votes for Republican candidates are likely to elect Democratic candidates in the future.

Similarly, districts where the prior Democratic voting percentage is at or below about 48.7\% and the prior Republican voting percentage is greater than about 45.5\% typically signal Republican elections. At first glance this may seem surprising because it could include districts in which Democrats slightly outnumber Republicans. However, several explanations are possible: In the predicted years (2014 and 2016), Republican turnout may be higher than Democratic turnout, independent voters may favor Republican candidates, or some Democrats may cross over to vote for Republican candidates in those districts and years.\footnote{Nationally, both higher Republican voter turnout and independents more frequently voting Republican were found using traditional post-election analyses of the 2014 and 2016 elections, whereas crossover voting was rare in both of these years \cite{pew_midterm_2014_ch1,Pew_2016_Chapter1,Barnes_2016,Roper_2016,WaPo_2016,CAP_2017,Pew_examination_2018}. This suggests that SR4-Fit rules find relevant patterns in the data.}

Interestingly, this latter pattern occasionally received negative coefficients in several trials, meaning it predicted Democratic outcomes in some cases. In those instances, the districts had a median age of 39.24 years compared to the overall average of 37.49 years, a White non-Hispanic population share of 69.46\% compared to 63.64\%, and a bachelor’s degree attainment rate of 19.06\% compared to 17.91\%. In other words, although districts without a clear Democratic majority often elect Republicans, districts that are older, whiter, and more highly educated sometimes elect Democrats. These demographic characteristics are widely known to influence voting patterns.

In the standard dataset, SR4-Fit builds an interpretable model shaped largely by age distribution, party support levels, and socioeconomic indicators. Age between 45 and 54 emerges as the most influential feature and consistently receives high positive coefficients, indicating Republican-leaning tendencies in regions where middle-aged populations are prominent. Similarly, the age groups 65–74 and 18–24 also show strong positive associations with Republican classification, highlighting the predictive importance of both older and younger demographic groups.

A more complex rule involving closely balanced prior Democratic and Republican voting percentages combined with relatively low median income (around \$23,717 or lower) has only a small positive influence, suggesting that lower income levels may slightly shift predictions toward Republican classification in closely contested regions. Another narrow-range rule involving modest Democratic support, moderate Republican support, and bachelor’s degree attainment above about 21.7\% also receives a small positive coefficient. This suggests that relatively high educational attainment may sometimes push moderately divided districts toward Republican classification.

In the expanded dataset, SR4-Fit captures more nuanced voting behavior through interpretable patterns primarily driven by prior Democratic and Republican support percentages along with demographic factors. One pattern involving moderately split Democratic and Republican support tends to produce Democratic classifications. Districts satisfying this pattern tend to have a slightly older population, with a median age of about 38.97 years compared to the overall average of 37.49 years. These districts also show higher Asian population shares (8.48\% versus 4.86\%) and higher multiracial population shares (3.95\% versus 2.68\%). In addition, they have somewhat higher White non-Hispanic population shares, higher bachelor’s degree attainment rates (19.38\% compared to 17.91\%), higher median incomes (about \$30,764 compared to \$26,873), and lower poverty levels (11.22\% compared to 14.63\%). Together, these characteristics suggest that moderate diversity combined with higher socioeconomic status tends to correspond with Democratic outcomes in these districts.

In the previous dataset, SR4-Fit identifies patterns where both Democratic and Republican support levels are relatively high but Republican support is slightly stronger, resulting in Republican classification. Districts following this pattern tend to have slightly older populations, with a median age of about 37.70 years compared to the overall average of 37.49 years. These districts are also less diverse, with higher White non-Hispanic population shares (71.58\% versus 63.64\%) and lower representation of Black, Asian, Hispanic, and multiracial populations. Educational attainment is also somewhat higher in these districts, with about 19.37 percent holding bachelor’s degrees and 11.68 percent holding graduate or professional degrees compared to 17.91\% and 10.65\% overall.

\subsection{Without Prior Party Voting Percentages (Classification)}
\label{ss:no_party_class}

For the minimum dataset, several demographic patterns emerge that help explain the predicted electoral outcomes. Districts with White population between approximately 26.8\% and 42.25\% tend to predict Democratic outcomes, suggesting that districts with greater racial diversity are more likely to support Democratic candidates. Similarly, districts with moderate White population levels between approximately 41.55\% and 54.65\%, combined with bachelor’s degree attainment above 12.95\%, also tend to predict Democratic outcomes, indicating that moderate educational attainment, together with demographic diversity, may contribute to Democratic support. In contrast, districts characterized by very high White population shares exceeding approximately 77.15\%, male population proportions above about 48.35\%, and bachelor’s degree attainment greater than approximately 17.35\%  more likely to predict Republican outcomes. Age structure also appears to influence predictions: districts with lower White population shares ($\leq$ 42.7) combined with older median ages above about 41.7 years, as well as districts with lower White population shares ($\leq$ 41.55\%) and younger median ages below approximately 40.05 years, tend to predict Republican outcomes. Additionally, temporal factors appear to play a role, as districts with White population shares greater than about 53.85\%, male population proportions above roughly 48.55\%, and median ages above approximately 35.55 years were associated with Democratic outcomes.

Across the extracted rules in the standard dataset, districts with lower White population shares (generally $\leq$42.7\%) combined with moderate working-age populations aged 25–44 (approximately 23.65\%–24.55\%) and younger median ages ($\leq$41.7 years) frequently predict Republican outcomes, indicating that districts with younger working-age demographics may lean Republican in the rule set. Several rules also associate Republican predictions with districts where middle-aged populations (Age 45–54: $\ge$12.65\%) or pre-retirement populations (Age 55–64: $\ge$10.45\%) are present, while older populations such as Age 65–74 remain relatively small ($\leq$8–9\%), suggesting that certain age structures influence voting behavior. In addition, Republican predictions appear in districts with low minority population shares, including Black population $\leq$1.95\% and Asian population $\leq$0.85\%, as well as moderate Hispanic population shares $\leq$38.2\%. Some rules further associate Republican outcomes with family-structured populations containing both youth (Age 5–17 $\ge$13.45\%) and elderly populations (Age 75+ $\ge$8.2\%), suggesting the presence of multi-generational demographic structures. In contrast, Democratic predictions are observed in districts with lower White population shares ($\leq$42.25\%) combined with higher median income levels ($\ge$ \$30,298.5), indicating that economically stronger and demographically diverse districts may favor Democratic candidates. Additional Democratic outcomes appear in districts with higher proportions of young adults (Age 18–24 $\ge$11\%) and moderate educational attainment levels, such as some college education exceeding 27.45\% and high school graduation rates above 30.3\%, suggesting that education and younger populations may contribute to Democratic voting patterns.

Across the rules extracted from the expanded dataset, many rules involve districts where the proportion of individuals who have never been married is $\leq$32.95\% and the proportion of married individuals is $\leq$54.05–54.45\%, suggesting that marital status composition is a common structural feature in the model. Democratic predictions frequently occur in districts with greater demographic diversity, including Black population shares between approximately 20.75\% and 29.0\% or above 4.85\%, Hispanic population shares greater than about 1.85\%, and Asian population shares above roughly 1.45\%, combined with higher educational attainment such as graduate or professional degree levels exceeding about 4.85–10.3\% and median income levels above roughly \$19,394. These districts may also contain younger adult populations (Age 18–24 above approximately 8.65–11.7\%), indicating the influence of demographic diversity, education, and younger age groups in Democratic predictions. In contrast, Republican predictions appear in districts with moderate educational attainment, including bachelor’s degree levels above about 10.9\% and relatively low proportions of individuals without a high school diploma ($\leq$12.15\%), along with lower Black population shares ($\leq$8.9\%), moderate language diversity where households speaking languages other than English remain $\leq$34.15\%, and moderate income groups in the \$65,000–\$74,999 range exceeding about 2.25\%.

Across the extracted rules in the previous dataset, Republican predictions frequently occur in districts with moderate educational attainment, such as populations with some college or associate degrees exceeding approximately 31.65\%, combined with low Asian population shares ($\leq$1.15\%) and moderate representation of other racial groups ($\ge$1.95\%). Additional Republican outcomes appear in districts with large working-age populations (Age 25–44 above roughly 24.45–24.55\%) and higher White population shares exceeding approximately 54.75\%, particularly when family-age populations are present, such as youth populations (Age 5–17 $\ge$18.15\%). Republican predictions also appear in districts with moderate racial diversity, including Asian population shares greater than about 2.05\% and multiracial populations above approximately 0.95\%, as well as districts where Black population shares remain below roughly 38.45\%. In contrast, Democratic predictions tend to occur in districts with very high Hispanic population shares ($\ge$67.2\%), as well as districts where lower educational attainment levels are present, such as less than high school completion exceeding approximately 19.15\%, combined with greater representation of other racial groups ($\ge$2.30\%).

\subsection{Including REP Percentage and DEM Percentage as Label (Regression)}
\label{ss:dem_reg}

In the minimum dataset, the rules primarily highlight the strong influence of prior Republican vote share (REP percentage) on predicting Democratic vote percentage. When Republican support is extremely low (REP percentage$\leq$$-2.0186$), the model increases the predicted Democratic percentage by 23.72, indicating that districts with very weak Republican presence strongly favor Democratic outcomes. When Republican support remains low but not extremely low ($-2.0186$$<$REP Percentage$\leq$$-0.7020$) and the share of the white non-Hispanic population is relatively smaller ($\leq$$-0.4248$), the prediction increases by 18.12, suggesting that demographic composition further reinforces Democratic vote share in these districts. Additionally, in districts with very low Republican support (REP percentage$\leq$$-2.0338$) and moderate levels of bachelor’s degree attainment ($\leq$0.2639) increase the predicted Democratic percentage by 7.56.

In the standard dataset, the rules again emphasize the strong role of prior Republican vote share (REP percentage) in predicting Democratic vote percentage. When Republican support is extremely low (REP percentage$\leq$$-2.0308$), the model increases the predicted Democratic vote share by 16.96, indicating that districts with very weak Republican presence strongly favor Democratic outcomes. Additionally, when Republican support is low (REP percentage$\leq$$-0.7970$) and a relatively high Black population share (Black$>$1.6788), the predicted Democratic percentage increases by 15.74.

In the expanded dataset, when Republican support is relatively low (REP percentage$\leq$$-0.8104$) and the share of the Black or African American population is high ($>$1.5405), the model increases the predicted Democratic vote percentage by 17.12. This suggests that districts with weaker Republican presence and a higher proportion of Black residents tend to exhibit substantially stronger Democratic electoral support.

In the previous dataset, the rules highlight the influence of prior Republican vote share, demographic characteristics, and historical voting patterns in predicting Democratic vote percentage. When Republican support falls within a moderately low range ($-0.7020$$<$REP percentage$\leq$$-0.3607$) and the share of the population aged 25–44 years is relatively higher ($>$$-0.4926$), the predicted Democratic percentage increases by 9.19, suggesting that districts with larger working-age populations tend to favor Democratic outcomes when Republican support is moderate. When prior Republican vote share remains moderately low ($-0.8065$$<$REP percentage$\leq$$-0.1539$) but the 25–44 age population is smaller, combined with higher proportions of residents with less than a high school education ($>$$-0.4735$), the prediction increases by 6.84. Finally, when Republican support is very low (REP percentage$\leq$ $-0.8128$), the model increases the predicted Democratic vote percentage by 3.91, even when the Black population share remains below the specified threshold.

\subsection{Without REP Percentage and DEM Percentage as Label (Regression)}
\label{ss:dem_norep_reg}

In the minimum dataset without including prior Republican vote share as a feature, the rules highlight the influence of racial composition, education, and demographic characteristics in predicting Democratic vote percentage. When the share of the White non-Hispanic population is relatively low ($\leq$$-1.0747$), the predicted Democratic percentage increases by 17.82. A similar pattern appears when the White non-Hispanic population share remains low ($\leq$$-0.7819$) combined with specific ranges of male population share, increasing the prediction by 18.73, indicating strong Democratic support in districts with a lower White population share. Additionally, districts with higher educational attainment (0.4004$<$Bachelor’s degree$\leq$1.3966) and higher White population share ($>$0.5837) still increase the predicted Democratic percentage by 14.04, suggesting that education plays an important role in shaping voting behavior.

In the standard dataset without including prior Republican vote share as a feature, the rules emphasize the importance of racial composition, educational attainment, and age distribution in predicting Democratic vote percentage. Districts with low White population share (White$\leq$$-1.0973$) and very high Black population share (Black$>$2.9733) produce the strongest increase in Democratic vote share, contributing 19.53 to the prediction. Similarly, districts with higher Asian population share ($>$0.0735) and lower proportions of younger residents increase the predicted Democratic percentage by 18.08. Other rules highlight the role of age distribution, particularly larger shares of the 25–44 age group ($>$0.6508), which increases predictions by 15.06 to 15.21 when combined with lower White population share and specific education levels. Additionally, districts with higher educational attainment (graduate degree$>$0.0250) or higher proportions of high school graduates ($>$0.8662) also contribute positively to Democratic vote predictions.

In the expanded dataset without including prior Republican vote share as a feature, the rules highlight the influence of marital status, racial composition, education, income, and age distribution in predicting Democratic vote percentage. Districts with lower shares of the White non-Hispanic population and higher shares of Black residents show strong increases in Democratic vote share, contributing 17.14 to the prediction. Similarly, districts characterized by high proportions of high school graduates combined with lower median income produce the strongest effect, increasing predicted Democratic vote share by 27.95. Other rules highlight the importance of marital status, particularly districts with higher shares of individuals who have never married, which increase Democratic vote predictions by 13.17 to 15.14 when combined with specific age distributions and demographic characteristics. Additionally, racial diversity and language composition further refine predictions, contributing 12.27 to 13.49 to the predicted Democratic percentage.

In the previous dataset without including prior Republican vote share as a feature, when the previous party indicator reflects Democratic alignment and the share of the White non-Hispanic population is relatively higher ($>$$-0.7791$), the predicted Democratic vote percentage increases by 17.10. Similarly, when previous electoral alignment remains Democratic and the White non-Hispanic population share is relatively lower ($\leq$$-0.7882$), the model increases the predicted Democratic percentage by 14.84.

\subsection{Including DEM Percentage and REP Percentage as Label (Regression)}
\label{ss:rep_reg}

In the minimum dataset where Republican vote percentage is the target variable and prior Democratic vote percentage is included as a feature, the rule highlights the inverse relationship between the two party vote shares. Specifically, when prior Democratic vote share is relatively low ($-2.1595$$<$DEM percentage$\leq$$-0.5436$), the model increases the predicted Republican vote percentage by 6.47. This indicates that districts with weaker Democratic support tend to exhibit higher Republican vote percentages, reflecting the expected competitive relationship between the two major parties in electoral outcomes.

In the standard dataset where Republican vote percentage is the label and prior Democratic vote percentage is included as a feature, when Democratic vote share is extremely low (DEM percentage$\leq$$-2.1753$), the predicted Republican vote percentage increases substantially by up to 15.11 units, representing the strongest contribution among the rules. When Democratic support remains relatively low ($-2.1629$$<$DEM percentage$\leq$$-0.5604$), the predicted Republican vote share increases by 8.00 to 10.16, particularly in districts with higher White population share or lower educational attainment. Additional rules show that moderately low prior Democratic vote share ($-0.5604$$<$DEM percentage$\leq$$-0.1930$) contributes smaller increases ranging from 5.43 to 5.73, depending on demographic and socioeconomic characteristics such as lower Hispanic population share and lower Black population share and higher median income.

In the expanded dataset where Republican vote percentage is the label and prior Democratic vote percentage is included as a feature, when prior Democratic vote share is extremely low (DEM percentage$\leq$$-2.1629$), the predicted Republican vote percentage increases substantially by 16.18, particularly in districts where the income group earning \$35,000–\$49,999 remains below the specified threshold. When Democratic support remains relatively low ($-2.1629$$<$DEM percentage$\leq$$-0.5604$) and the share of married individuals is relatively higher, the model increases the predicted Republican vote share by 9.95. Finally, when prior Democratic vote share falls within a moderate low range ($-0.5604$$<$DEM percentage$\leq$$-0.2231$) and the share of individuals who have never married remains below the specified threshold, the predicted Republican percentage increases by 6.24.

In the previous dataset where Republican vote percentage is the label and prior Democratic vote percentage is included as a feature, when prior Democratic vote share is relatively low ($-2.1323$$<$DEM percentage$\leq$$-0.5254$), the predicted Republican vote percentage increases significantly by up to 12.92, especially in districts with lower Black population share. Similarly, districts with low Democratic support ($-2.1851$$<$DEM percentage$\leq$$-0.4622$) and specific educational characteristics contribute 11.97 to Republican vote share. When prior Democratic vote share falls within moderate low ranges (approximately $-0.57$ to $-0.21$), the predicted Republican percentage increases more modestly by 6.05 to 6.85, depending on demographic factors such as higher White population share, lower Hispanic population share, and racial composition.

\subsection{Without DEM Percentage and REP Percentage as Label (Regression)}
\label{ss:rep_nodem_reg}

In the minimum dataset where Republican vote percentage is the target variable and prior Democratic vote share is not included as a feature, the rules highlight the importance of demographic composition, education, age structure, and geographic context in explaining Republican vote share. Districts with higher shares of White non-Hispanic population ($>$$-0.9646$) and higher male population share tend to increase predicted Republican vote percentages. The strongest contribution occurs when the proportion of individuals with bachelor’s degrees is relatively low ($\leq$$-0.3805$) and the median population age is lower ($\leq$$-0.1370$), increasing the predicted Republican vote share by 20.05. Other rules show that districts with higher educational attainment ($>$ 0.2185 bachelor’s degree share) it contribute 12.90, while similar demographic conditions contribute 7.72.

In the standard dataset where Republican vote percentage is the label and prior Democratic vote share is not included as a feature, the rules highlight several demographic factors associated with higher Republican vote share. Districts with higher White population share ($>$$-0.9312$) and lower proportions of graduate-level educational attainment ($\leq$0.3989) increase the predicted Republican vote percentage by 9.21. Additionally, individual demographic features such as White population share ($+10.20$) and population aged 65–74 ($+11.57$) contribute strongly to Republican vote predictions, suggesting that districts with larger White and older populations tend to exhibit higher Republican vote percentages. Educational attainment and ethnic composition also contribute smaller but meaningful increases, with bachelor’s degree share adding 6.47 and Hispanic or Latino population share contributing 5.41.

In the expanded dataset where Republican vote percentage is the label and prior Democratic vote share is not included as a feature, the rules highlight the influence of marital status, age distribution, education, income, and economic status on Republican vote share. Districts with lower proportions of individuals who have never married ($\leq$$-0.2858$) and higher shares of individuals with less than a high school education ($>$$-0.3833$) increase the predicted Republican vote percentage by 9.69, particularly during later election periods. Additional rules show that moderate levels of never-married individuals combined with higher median age groups such as 45–54 contribute 7.64 to Republican vote predictions. Demographic features such as population aged 65–74 ($+4.24$) and bachelor’s degree attainment ($+4.45$) also contribute positively, while Hispanic or Latino population share adds 9.04 to the prediction. Socioeconomic characteristics including median income levels and poverty thresholds further influence predictions, contributing 4.74 to 6.80 depending on the combination of conditions.

In the previous political dataset where Republican vote percentage is the label and prior Democratic vote share is not included as a feature, the rules highlight the importance of historical electoral alignment, demographic composition, age distribution, and educational attainment in explaining Republican vote share. Districts where the previous party indicator suggests prior Republican alignment and the share of individuals with graduate or professional degrees exceeds $-0.2243$ increase the predicted Republican vote percentage by 14.06, representing the strongest contribution among the rules. Additional demographic factors also contribute positively, including population aged 65–74 ($+6.40$) and White non-Hispanic population share ($+5.74$), suggesting that districts with larger older and White populations tend to favor Republican electoral outcomes. Educational attainment further contributes to the prediction, with bachelor’s degree share adding 3.84.

\subsection{Interpretation SR4-Fit Rules for Public Classification Datasets}

Across the extracted rules for the breast cancer dataset, SR4-Fit identifies several cellular morphology patterns that distinguish malignant from benign tumors. Malignant classifications frequently occur when features indicating cellular irregularity and abnormal nuclei are elevated. Malignant outcomes appear when clump thickness exceeds approximately 0.7301 and bare nuclei values are greater than about $-0.4242$, even when uniformity of cell size remains at or below 0.1140. Malignancy is also associated with high variability in cell size (uniformity cell size $>$ 0.4405) combined with marginal adhesion values greater than about $-0.4647$, indicating irregular cell growth and stronger cell clustering. Additional malignant rules involve nuclear structure features such as bland chromatin values above approximately $-0.3861$ and bare nuclei values greater than about $-0.0123$, as well as rules where uniformity of cell shape remains below about 0.0954 with normal nucleoli values below approximately 0.2066 and mitoses below roughly 0.5179, highlighting the importance of chromatin structure and nuclear activity in identifying malignant cells. In contrast, benign classifications typically occur when cellular structures remain more regular. For instance, benign outcomes appear when uniformity of cell shape is relatively low ($\leq$$-0.2395$), bare nuclei values remain within moderate ranges (between about $-0.0123$ and 0.5370), and uniformity of cell size remains above approximately $-0.2125$. Additional benign patterns occur when clump thickness remains small ($\leq$0.3753) and single epithelial cell size remains above about $-0.3305$, indicating thinner and more stable cell clusters. Other benign rules involve lower marginal adhesion values ($\leq$$-0.4647$), even when uniformity of cell size exceeds 0.4405, suggesting that larger cell sizes without strong adhesion may still correspond to non-invasive tissue behavior.

Across the extracted rules for the E. coli dataset, most rules predicting the cp (cytoplasmic) protein class involve moderate or low values of alm1 ($\leq$0.6955) combined with strongly negative mcg values ($\leq$$-1.6469$ or $\leq$$-1.7241$) and moderate ranges of alm2 and aac, suggesting weaker signal characteristics associated with cytoplasmic proteins. In contrast, rules predicting the im (inner membrane) protein class consistently involve higher feature values, including alm1 greater than approximately 0.3473 to 0.6258, mcg greater than about 0.5914, and alm2 exceeding approximately 1.5317, indicating stronger signal patterns associated with inner membrane protein localization.

Across the extracted rules for the page blocks dataset, SR4-Fit identifies patterns primarily driven by geometric properties of document blocks, including height, eccentricity, block area, length, pixel composition, and white-to-black transitions. Most rules predict class 1, which reflects the dominant category in the dataset and typically corresponds to standard document regions such as text blocks or simple rectangular areas. These rules commonly involve moderate or small block heights ($\leq$0.8981), low eccentricity values ($\leq$$-0.4422$), and moderate pixel composition measures, indicating regular block shapes and consistent document textures. Some rules rely solely on geometric attributes such as area or length, suggesting that the physical dimensions of blocks are strong predictors of common document structures. In contrast, class 2 predictions occur when blocks become larger or taller (height $>$ 0.8453) while maintaining specific eccentricity values, indicating structural elements that differ from typical text regions.

Across the extracted rules for the Pima Indians Diabetes dataset, diabetic predictions generally occur when glucose levels exceed approximately 0.96 to 1.15, often combined with moderate or high BMI values ($>$ $-1.1286$) or additional factors such as pregnancy count, insulin levels, or blood pressure. These conditions reflect well-known clinical indicators of diabetes risk, particularly elevated blood glucose and body mass index. In contrast, non-diabetic classifications occur when glucose levels remain relatively low ($\leq$ 0.3945 or $\leq$ $-0.2001$), often accompanied by lower BMI values and younger age groups, suggesting lower metabolic risk. Some rules also indicate that genetic risk factors, represented by the Diabetes Pedigree Function, may play a secondary role when glucose levels remain low.

Across the extracted rules for the vehicle dataset, SR4-Fit identifies interpretable patterns based on vehicle shape and geometric descriptors, including aspect ratios, elongatedness, circularity, rectangularity, variance, and scatter measures derived from vehicle silhouettes. Bus classifications occur when vehicle shapes exhibit larger elongated structures or compact rectangular silhouettes, often associated with moderate or low aspect ratios and specific variance ranges. Car classifications tend to appear when vehicle shapes are more compact, balanced, and circular, often characterized by moderate circularity and reduced scatter or hollow ratios. In contrast, van classifications are typically associated with box-like and rectangular silhouettes, reflected by moderate variance values, lower circularity, and specific rectangularity thresholds.

Across the extracted rules for the yeast dataset, the rules that predict the CYT (cytosolic) class indicate that proteins belonging to this class exhibit distinctive combinations of signal intensities across these attributes. Several rules involve very low vac values ($\leq$ $-2.3346$) combined with moderate or low mcg and gvh values, suggesting that weak localization signals for certain cellular compartments correspond to cytosolic proteins. Other rules predict CYT when mcg and gvh values are relatively high (mcg $>$ 1.7113 or gvh $>$ 2.1397), indicating alternative signal patterns associated with cytosolic localization. Additional rules rely on very low alm values ($\leq$ $-1.3277$) or low mitochondrial indicators (mit $\leq$ 0.4656) combined with moderate nuclear signal values, suggesting that the absence of strong compartment-specific signals can also indicate cytosolic proteins.

\subsection{Interpretation SR4-Fit Rules for Public Regression Datasets}

In the Abalone dataset, the regression rules primarily emphasize the role of shell weight in determining the predicted outcome. When the shell weight is relatively high (shell weight$>$0.8448) and the shucked weight remains moderate (shucked weight$\leq$0.8860), the model adds 0.56 to the prediction, indicating the strongest positive contribution. Abalones with moderate shell weight (0.1150$<$shell weight$\leq$0.8448) also increase the prediction by 0.48, suggesting that mid-range shell weight still has a strong positive relationship with the target variable. In contrast, when shell weight is lower ($-1.3939$$<$shell weight$\leq$$-0.6147$) and the encoded sex value is less than 0.6568, the model contributes only 0.12, indicating a comparatively weaker influence.

In the Bone dataset, the regression rules primarily highlight the influence of age on the predicted outcome. When age is relatively low (age$\leq$$-0.4755$) and the encoded gender value is greater than $-0.0568$, the model contributes 0.01 to the prediction, indicating a small positive effect. Similarly, when age falls within the range $-1.0279$$<$age$\leq$$-0.5504$, the prediction increases by 0.01, suggesting that individuals in this younger age interval also contribute positively to the outcome.

In the Diabetes dataset, when BMI is very high (BMI$>$1.5267) and S2 is relatively low ($\leq$0.5068), the prediction increases substantially by 63.8, while a similar condition with S5$>$$-0.0210$, BMI$>$1.5237, and S2$\leq$0.5107 results in an even stronger contribution of 68.7. When BMI remains high but S2 becomes larger, the contribution drops to 16.76 or 21.97, suggesting that S2 moderates the impact of high BMI. For moderate BMI (0.0745$<$ BMI$\leq$1.5267), elevated S6 ($>$0.6737) increases the prediction by 31.04, while lower S6 reduces the effect to 4.56. Additional rules show that high S5 combined with elevated blood pressure (BP$>$1.3623) contributes 55.17, indicating strong metabolic interactions. Even when BMI is relatively low, combinations involving S3$\leq$$-0.7763$, S4$>$1.9298, or S1$\leq$0.9309 still produce contributions around 24, demonstrating that biomarker interactions beyond BMI can also significantly influence predicted diabetes progression.

In the Housing dataset, the rules highlight the strong influence of the average number of rooms (RM) on predicted housing prices. When room counts are very high (RM$>$1.6312 or RM$>$1.7026) and favorable neighborhood characteristics such as low pupil–teacher ratios (PTRATIO$\leq$0.6006 or $\leq$$-0.0796$), moderate tax rates (TAX$\leq$0.7703), or reasonable environmental conditions (NOX$\leq$0.8502) are present, the predicted housing value increases substantially by 12.12 to 15.15. For moderately high room counts (0.9091$<$RM$\leq$1.6312) combined with low lower-status population percentages (LSTAT$\leq$0.4702) and reasonable distances to employment centers (DIS$>$$-0.9158$), the prediction increases more modestly by 3.46. Even when room counts are smaller, combinations involving controlled crime rates, housing age, and socio-economic factors still contribute small positive effects of 0.58 to 2.04.

In the Machine dataset, the rules emphasize the importance of hardware capacity and baseline performance metrics in predicting machine performance. When PRP values are extremely high (PRP$>$3.3872 or PRP$>$3.9534) and large memory capacity (MMAX), the model increases the predicted performance by 20.19 to 27.50, while combinations of high PRP and specific model categories can contribute as much as 47.75. The strongest contribution occurs when both memory capacity (MMAX$>$2.9077) and maximum channel capacity (CHMAX$>$1.2501) are high, resulting in an increase of 49.42. Additional rules show that model and vendor identifiers also influence predictions, indicating that certain machine configurations or manufacturers correspond to higher predicted performance.

In the MPG dataset, the rules highlight the importance of engine size, horsepower, vehicle weight, and model year in determining fuel efficiency. Vehicles with very small engine displacement ($\leq$$-0.9550$), low horsepower ($\leq$$-0.7266$), and newer model years (model year$>$0.3881) increase the predicted MPG by 5.32, while low-cylinder engines with extremely low horsepower ($\leq$$-1.2563$) in newer vehicles produces an even stronger effect of 7.45. Lightweight vehicles also contribute positively, as seen when weight$\leq$$-0.8996$ combined with small displacement and newer model years increases MPG by 4.05. Additional rules show that moderate horsepower combined with low displacement and light vehicle weight adds 3.39 to the prediction. In contrast, vehicles with larger engines and higher horsepower, particularly from older model years, produce smaller effects such as 1.16, indicating weaker contributions to fuel efficiency.

In the Ozone dataset, the rules emphasize the importance of temperature, atmospheric pressure conditions, humidity, and visibility in determining ozone levels. When temperatures at Sandburg and El Monte are very high (Temp Sandburg$>$1.2484 or Temp El Monte$>$1.3626) and IBT values are elevated, the predicted ozone concentration increases substantially by 6.21 to 7.07. The strongest increase occurs when Sandburg temperature exceeds 0.7464, IBT$>$0.9268, and visibility is extremely low ($\leq$$-0.9002$), resulting in an increase of 8.23, indicating that hot and stagnant atmospheric conditions strongly promote ozone formation. Additional rules show that low visibility combined with high temperatures and low wind speeds contributes between 5.04 and 5.71, reflecting the accumulation of pollutants when atmospheric dispersion is limited. Even when IBT values are lower, combinations of moderate temperature, reduced visibility, and higher humidity still increase ozone predictions by 2.36.

In the prostate dataset, the rules emphasize the role of tumor volume (lcavol), seminal vesicle invasion (svi), prostate weight (lweight), and other clinical markers in predicting PSA levels. When tumor volume is very high (lcavol$>$1.2430) and benign hyperplasia measurements are elevated (lbph$>$$-0.4522$), the predicted outcome increases by 0.48, indicating a strong association between tumor size and PSA levels. Similarly, the presence of seminal vesicle invasion (svi$>$0.6291) combined with high tumor volume (lcavol$>$1.2786) increases the prediction by 0.43, highlighting the influence of tumor spread. Additional rules show that high tumor volume combined with capsular penetration (lcp$>$2.1708) or moderate tumor volume with controlled Gleason scores (pgg45$\leq$2.1094) contributes smaller increases ranging from 0.17 to 0.23. Even when tumor volume is lower, combinations involving larger prostate weight and older age still increase predictions by 0.19.

\section{Conclusions}\label{conclusions}
This work presented SR4-Fit, an interpretable machine learning framework designed to generate compact, stable, and rule-based predictive models while maintaining competitive predictive performance. The primary application examined in this study is the U.S. House of Representatives election datasets, where the objective is to explain variations in Democratic and Republican vote percentages using demographic and socioeconomic characteristics derived from the American Community Survey (ACS). By extracting interpretable decision rules, SR4-Fit enables transparent analysis of how demographic patterns across congressional districts relate to electoral outcomes.

The rule sets generated for the U.S. House election data reveal interpretable relationships between demographic and socioeconomic variables, such as race, age distribution, education levels, income, marital status, and historical voting behavior. These rule-based explanations provide insight into how combinations of district-level characteristics influence Democratic and Republican vote shares, illustrating the usefulness of interpretable models for political and policy analysis, where transparency is critical.

To support and validate the proposed approach beyond the primary electoral application, SR4-Fit was also evaluated on several public regression and classification benchmark datasets. These additional experiments demonstrate that the method performs consistently across diverse domains while producing interpretable rule structures.

Interpretability in this study is evaluated using our newly introduced rule understandability score (RUS) which combines combines key aspects of interpretable modeling: rule stability, number of extracted rules, and rule complexity. RUS is utilized in 
a Pareto analysis placing RUS on one axis and predictive accuracy (measured using the F1 score for classification and the coefficient of determination, $R^2$, for regression) on the other.
We further introduce and utilize an enhanced interpretability score (E-IPS) that combines understandability and predictivity into a single numeric value.
In both RUS and E-IPS, stability is assessed using the Dice-Sørensen similarity index across repeated trials, reflecting the consistency of rule structures. The number of rules measures model compactness, while rule complexity captures the number of conditions required to represent each rule. 

The experimental results show that SR4-Fit achieves a strong balance between predictive performance and interpretability. Compared with existing rule-based baselines, the framework consistently produces more stable rule sets with lower rule complexity while maintaining competitive F1 scores in classification and competitive $R^2$ values in regression, Pareto-dominating RuleFit and Decision Tree across all four experimental settings: election classification, election regression, public classification, and public regression. Relative to Random Forest, SR4-Fit consistently trades a modest amount of predictivity for substantially greater rule understandability in every setting, reflecting a single, coherent predictivity-interpretability trade-off rather than divergent behavior across tasks.

Overall, the results demonstrate that SR4-Fit provides a practical and reliable approach for interpretable machine learning in complex real-world settings. By generating stable, compact, and human-readable rule sets, the framework enables transparent understanding of predictive relationships without sacrificing F1 or $R^2$ performance relative to existing rule-based methods. The analysis of the U.S. House election dataset illustrates how interpretable rules can reveal meaningful connections between demographic characteristics and electoral outcomes, highlighting the broader potential of interpretable models for social and policy-related research. At the same time, strong F1 and $R^2$ performance across multiple public benchmark datasets confirm the general applicability of the proposed method. These findings suggest that SR4-Fit can serve as an effective tool for domains where understandability, reliability, and trust in model outputs are essential, contributing to the growing need for interpretable machine learning approaches in data-driven decision-making.

\section{Acknowledgements}

This work was supported in part by funding from the Institute for Community and Society Transformation (ICAST) at the University of Oklahoma, which is an interdisciplinary initiative that supports collaborative research and fosters partnerships across disciplines to address complex societal challenges. The authors gratefully acknowledge ICAST for its support of this project.

\bibliographystyle{elsarticle-num}
\bibliography{refs,ElectionOutcomesArticles}
\vspace{0.5in}
\appendix
\startcontents[appendices]

\begin{center}
    {\Large\bfseries Appendices (Supplemental Material)}
\end{center}
\vspace{0.5em}

\begingroup
\large
\setcounter{tocdepth}{2} 
\printcontents[appendices]{}{1}{}
\endgroup

\section{Statistical Significance Test (Wilcoxon Test)}\label{sup_sec_Wilcoxon}
\label{appendices}

To assess whether the differences in predictive performance between SR4-Fit and RuleFit are statistically meaningful, one-sided Wilcoxon signed-rank tests were conducted across all performance metrics for each dataset and feature configuration, with a significance threshold of $p < 0.05$.

Table~\ref{tab:wilcoxon_cls} reports the results for the classification task. When prior party voting percentages are included as features, SR4-Fit's advantage over RuleFit is statistically significant across all four metrics (accuracy, precision, recall, and F1 score) in the Minimum, Standard, and Expanded datasets. In the previous dataset, none of the four metrics reach significance, consistent with the closely matched E-IPS and predictive performance observed between the two models on this dataset. When prior party voting percentages are removed, the differences become significant across all four metrics in every dataset, with most $p$-values falling below $0.0001$, indicating that SR4-Fit's advantage over RuleFit is most pronounced and most consistent in the reduced-information setting.

Table~\ref{tab:wilcoxon_dem_regression} reports the results for the DEM regression task. The pattern here is more mixed than in classification. When the prior Republican voting percentage is included as a feature, significant differences favoring SR4-Fit appear for RMSE and $R^2$ in the Standard and Expanded datasets and for MAE in the Minimum dataset, while no metric reaches significance in the Previous dataset. When the prior Republican voting percentage is removed, significant differences emerge for RMSE and $R^2$ in the expanded and previous datasets, while the minimum and standard datasets show no significant differences. Across both feature configurations, MAE rarely reaches significance, whereas RMSE and $R^2$ tend to move together, reaching significance in the same dataset when either does.

Table~\ref{tab:wilcoxon_rep_regression} reports the results for the REP regression task. When prior Democratic voting percentage is included as a feature, MAE is the only metric that reaches significance, doing so consistently across all four datasets, while RMSE and $R^2$ do not reach significance in any dataset under this configuration. When the prior Democratic voting percentage is removed, the pattern shifts: RMSE and $R^2$ become significant in the Standard, Expanded, and Previous datasets, while MAE does not reach significance in any dataset in this configuration.

Across all three tables, no comparison yields a statistically significant result in RuleFit's favor, consistent with the one-sided test formulation used throughout. Where significant differences are observed, they consistently favor SR4-Fit, reinforcing the pattern described in the main text: SR4-Fit's predictive advantage over RuleFit is strongest and most consistent in the classification task, particularly under the without-prior-party-voting-percentage configuration, and more selectively significant in the regression tasks, where the specific metric reaching significance (RMSE/$R^2$ versus MAE) varies by dataset and feature configuration.


\begin{table}
\caption{Wilcoxon test W-statistic and p-value comparison between RuleFit and SR4-Fit over 30 trials across demographic classification datasets. Statistically significant differences ($p < 0.05$) are shown in \textbf{bold}.}

\setlength{\tabcolsep}{1.5pt}
\centering
\resizebox{\linewidth}{!}{
\begin{subtable}{\linewidth}
\centering
\begin{tabular}{lcccccccc}
\toprule
\multirow{2}{*}{\textbf{Dataset}} & \multicolumn{2}{c}{\textbf{Accuracy}} & \multicolumn{2}{c}{\textbf{Precision}} & \multicolumn{2}{c}{\textbf{Recall}} & \multicolumn{2}{c}{\textbf{F1 Score}} \\
\cmidrule(lr){2-3} \cmidrule(lr){4-5} \cmidrule(lr){6-7} \cmidrule(lr){8-9}
& \textbf{W-stat} & \textbf{p-value} & \textbf{W-stat} & \textbf{p-value} & \textbf{W-stat} & \textbf{p-value} & \textbf{W-stat} & \textbf{p-value} \\
\midrule
Minimum  & \textbf{74.5}  & \textbf{0.0024}   & \textbf{33.0}  & \textbf{0.0178}   & \textbf{19.0}   & \textbf{0.0374}   & \textbf{73.5}  & \textbf{0.0034}   \\
Standard & \textbf{128.0} & \textbf{0.0009}   & \textbf{47.0}  & \textbf{0.0234}   & \textbf{110.0}  & \textbf{0.0023}   & \textbf{159.0} & \textbf{0.0007}   \\
Expanded & \textbf{341.0} & \textbf{0.0001}   & \textbf{192.0} & \textbf{0.0006}   & \textbf{192.5}  & \textbf{0.0037}   & \textbf{367.0} & \textbf{0.0001}   \\
Previous & 72.5           & 0.1037            & 48.0           & 0.8495            & 148.5           & 0.0522            & 174.0          & 0.0615            \\
\bottomrule
\end{tabular}
\caption{\textbf{With prior party voting percentages.}}
\end{subtable}
}

\vspace{6pt}

\resizebox{\linewidth}{!}{
\begin{subtable}{\linewidth}
\centering
\begin{tabular}{lcccccccc}
\toprule
\multirow{2}{*}{\textbf{Dataset}} & \multicolumn{2}{c}{\textbf{Accuracy}} & \multicolumn{2}{c}{\textbf{Precision}} & \multicolumn{2}{c}{\textbf{Recall}} & \multicolumn{2}{c}{\textbf{F1 Score}} \\
\cmidrule(lr){2-3} \cmidrule(lr){4-5} \cmidrule(lr){6-7} \cmidrule(lr){8-9}
& \textbf{W-stat} & \textbf{p-value} & \textbf{W-stat} & \textbf{p-value} & \textbf{W-stat} & \textbf{p-value} & \textbf{W-stat} & \textbf{p-value} \\
\midrule
Minimum  & \textbf{418.0} & \textbf{$<$0.0001} & \textbf{337.0} & \textbf{0.0158}    & \textbf{397.0} & \textbf{$<$0.0001} & \textbf{439.0} & \textbf{$<$0.0001} \\
Standard & \textbf{465.0} & \textbf{$<$0.0001} & \textbf{465.0} & \textbf{$<$0.0001} & \textbf{464.0} & \textbf{$<$0.0001} & \textbf{465.0} & \textbf{$<$0.0001} \\
Expanded & \textbf{465.0} & \textbf{$<$0.0001} & \textbf{465.0} & \textbf{$<$0.0001} & \textbf{416.0} & \textbf{0.0001}    & \textbf{465.0} & \textbf{$<$0.0001} \\
Previous & \textbf{465.0} & \textbf{$<$0.0001} & \textbf{465.0} & \textbf{$<$0.0001} & \textbf{465.0} & \textbf{$<$0.0001} & \textbf{465.0} & \textbf{$<$0.0001} \\
\bottomrule
\end{tabular}
\caption{\textbf{Without prior party voting percentages.}}
\end{subtable}
}

\label{tab:wilcoxon_cls}
\end{table}

\begin{table}
\caption{Wilcoxon signed-rank test W-statistic and p-value comparison between RuleFit and SR4-Fit over 30 trials across demographic datasets for Democratic percentage label regression. Statistically significant differences ($p < 0.05$) are shown in \textbf{bold}.}

\setlength{\tabcolsep}{1.5pt}
\centering
\resizebox{\linewidth}{!}{
\begin{subtable}{\linewidth}
\centering
\begin{tabular}{lcccccc}
\toprule
\multirow{2}{*}{\textbf{Dataset}} & \multicolumn{2}{c}{\textbf{RMSE}} & \multicolumn{2}{c}{\textbf{MAE}} & \multicolumn{2}{c}{\textbf{R\textsuperscript{2}}} \\
\cmidrule(lr){2-3} \cmidrule(lr){4-5} \cmidrule(lr){6-7}
& \textbf{W-stat} & \textbf{p-value} & \textbf{W-stat} & \textbf{p-value} & \textbf{W-stat} & \textbf{p-value} \\
\midrule
Minimum  & 193.0          & 0.2083            & \textbf{13.0}  & \textbf{$<$0.0001} & 274.0          & 0.1967            \\
Standard & \textbf{52.0}  & \textbf{0.0001}   & 375.0          & 0.9983            & \textbf{415.0} & \textbf{0.0001}   \\
Expanded & \textbf{127.0} & \textbf{0.0150}   & 408.0          & 0.9998            & \textbf{333.0} & \textbf{0.0194}   \\
Previous & 425.0          & 1.0000            & 221.0          & 0.4065            & 37.0           & 1.0000            \\
\bottomrule
\end{tabular}
\caption{\textbf{With prior Republican voting percentage as feature.}}
\end{subtable}
}

\vspace{6pt}

\resizebox{\linewidth}{!}{
\begin{subtable}{\linewidth}
\centering
\begin{tabular}{lcccccc}
\toprule
\multirow{2}{*}{\textbf{Dataset}} & \multicolumn{2}{c}{\textbf{RMSE}} & \multicolumn{2}{c}{\textbf{MAE}} & \multicolumn{2}{c}{\textbf{R\textsuperscript{2}}} \\
\cmidrule(lr){2-3} \cmidrule(lr){4-5} \cmidrule(lr){6-7}
& \textbf{W-stat} & \textbf{p-value} & \textbf{W-stat} & \textbf{p-value} & \textbf{W-stat} & \textbf{p-value} \\
\midrule
Minimum  & 325.0          & 0.9715            & 379.0          & 0.9987            & 139.0          & 0.9728            \\
Standard & 189.0          & 0.1855            & 398.0          & 0.9997            & 280.0          & 0.1643            \\
Expanded & \textbf{126.0} & \textbf{0.0142}   & 394.0          & 0.9996            & \textbf{335.0} & \textbf{0.0175}   \\
Previous & \textbf{28.0}  & \textbf{$<$0.0001} & 325.0         & 0.9715            & \textbf{439.0} & \textbf{$<$0.0001} \\
\bottomrule
\end{tabular}
\caption{\textbf{Without prior Republican voting percentage as feature.}}
\end{subtable}
}

\label{tab:wilcoxon_dem_regression}
\end{table}

\begin{table}
\caption{Wilcoxon signed-rank test W-statistic and p-value comparison between RuleFit and SR4-Fit over 30 trials across demographic datasets for Republican percentage label regression. Statistically significant differences ($p < 0.05$) are shown in \textbf{bold}.}

\setlength{\tabcolsep}{1.5pt}
\centering
\resizebox{\linewidth}{!}{
\begin{subtable}{\linewidth}
\centering
\begin{tabular}{lcccccc}
\toprule
\multirow{2}{*}{\textbf{Dataset}} & \multicolumn{2}{c}{\textbf{RMSE}} & \multicolumn{2}{c}{\textbf{MAE}} & \multicolumn{2}{c}{\textbf{R\textsuperscript{2}}} \\
\cmidrule(lr){2-3} \cmidrule(lr){4-5} \cmidrule(lr){6-7}
& \textbf{W-stat} & \textbf{p-value} & \textbf{W-stat} & \textbf{p-value} & \textbf{W-stat} & \textbf{p-value} \\
\midrule
Minimum  & 252.0          & 0.6558            & \textbf{145.0} & \textbf{0.0360}   & 213.0          & 0.6558            \\
Standard & 307.0          & 0.9373            & \textbf{70.0}  & \textbf{0.0004}   & 157.0          & 0.9398            \\
Expanded & 185.0          & 0.1643            & \textbf{89.0}  & \textbf{0.0016}   & 281.0          & 0.1592            \\
Previous & 165.0          & 0.0825            & \textbf{144.0} & \textbf{0.0344}   & 299.0          & 0.0857            \\
\bottomrule
\end{tabular}
\caption{\textbf{With prior Democratic voting percentage as feature.}}
\end{subtable}
}
\vspace{6pt}
\resizebox{\linewidth}{!}{
\begin{subtable}{\linewidth}
\centering
\begin{tabular}{lcccccc}
\toprule
\multirow{2}{*}{\textbf{Dataset}} & \multicolumn{2}{c}{\textbf{RMSE}} & \multicolumn{2}{c}{\textbf{MAE}} & \multicolumn{2}{c}{\textbf{R\textsuperscript{2}}} \\
\cmidrule(lr){2-3} \cmidrule(lr){4-5} \cmidrule(lr){6-7}
& \textbf{W-stat} & \textbf{p-value} & \textbf{W-stat} & \textbf{p-value} & \textbf{W-stat} & \textbf{p-value} \\
\midrule
Minimum  & 242.0          & 0.5775            & 247.0          & 0.6172            & 224.0          & 0.5694            \\
Standard & \textbf{34.0}  & \textbf{$<$0.0001} & 274.0         & 0.8033            & \textbf{432.0} & \textbf{$<$0.0001} \\
Expanded & \textbf{56.0}  & \textbf{0.0001}   & 251.0          & 0.6482            & \textbf{409.0} & \textbf{0.0001}   \\
Previous & \textbf{137.0} & \textbf{0.0247}   & 212.0          & 0.3366            & \textbf{327.0} & \textbf{0.0260}   \\
\bottomrule
\end{tabular}
\caption{\textbf{Without prior Democratic voting percentage as feature.}}
\end{subtable}
}
\label{tab:wilcoxon_rep_regression}
\end{table}

\FloatBarrier

\section{Results of U.S House of Representative Elections Classification}\label{sup_sec_HouseC}

\begin{figure}
\centering
\begin{minipage}{0.49\linewidth}
    \centering
    \includegraphics[width=\linewidth]{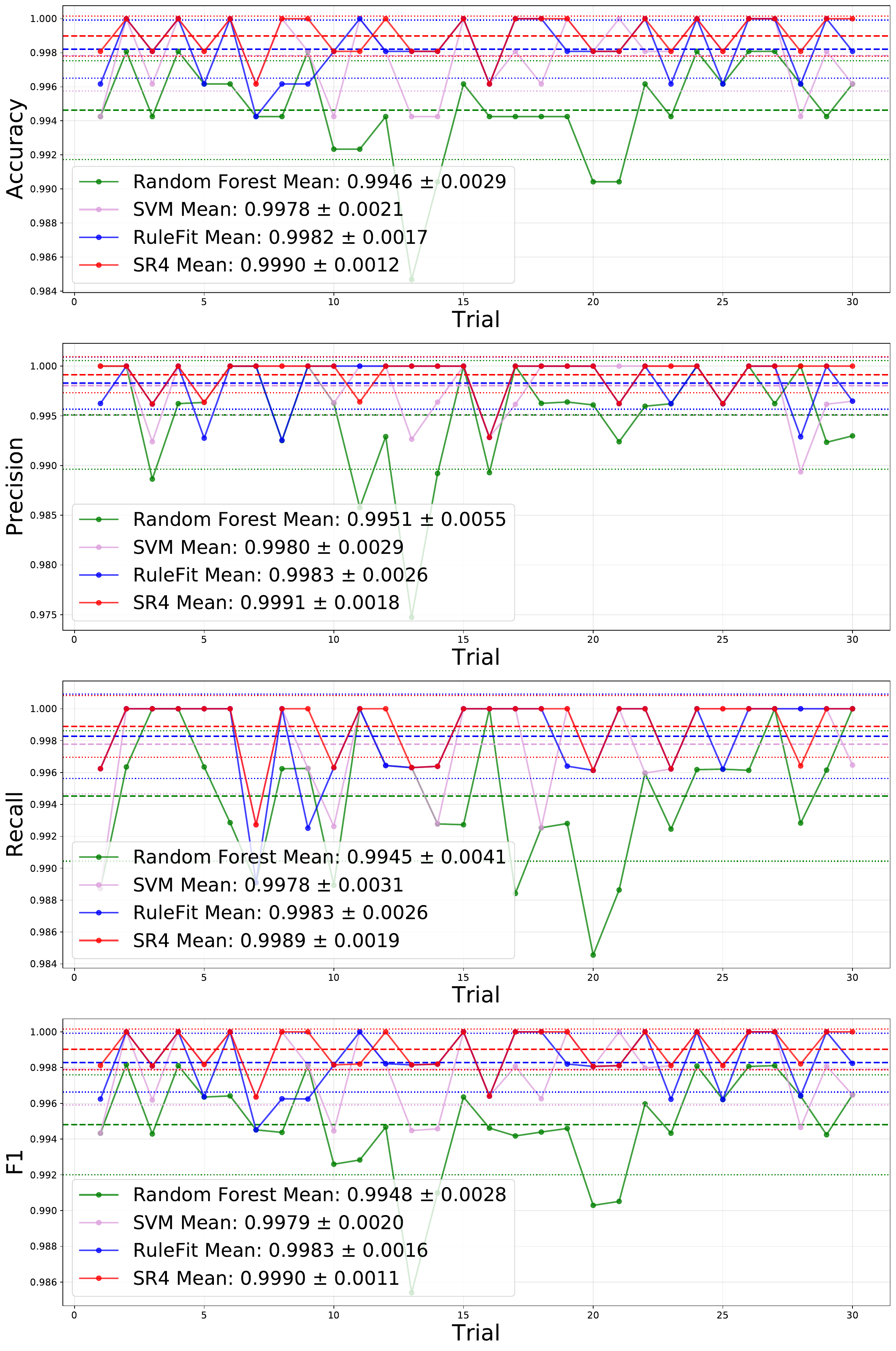}
\end{minipage}\hfill
\begin{minipage}{0.49\linewidth}
    \centering
    \includegraphics[width=\linewidth]{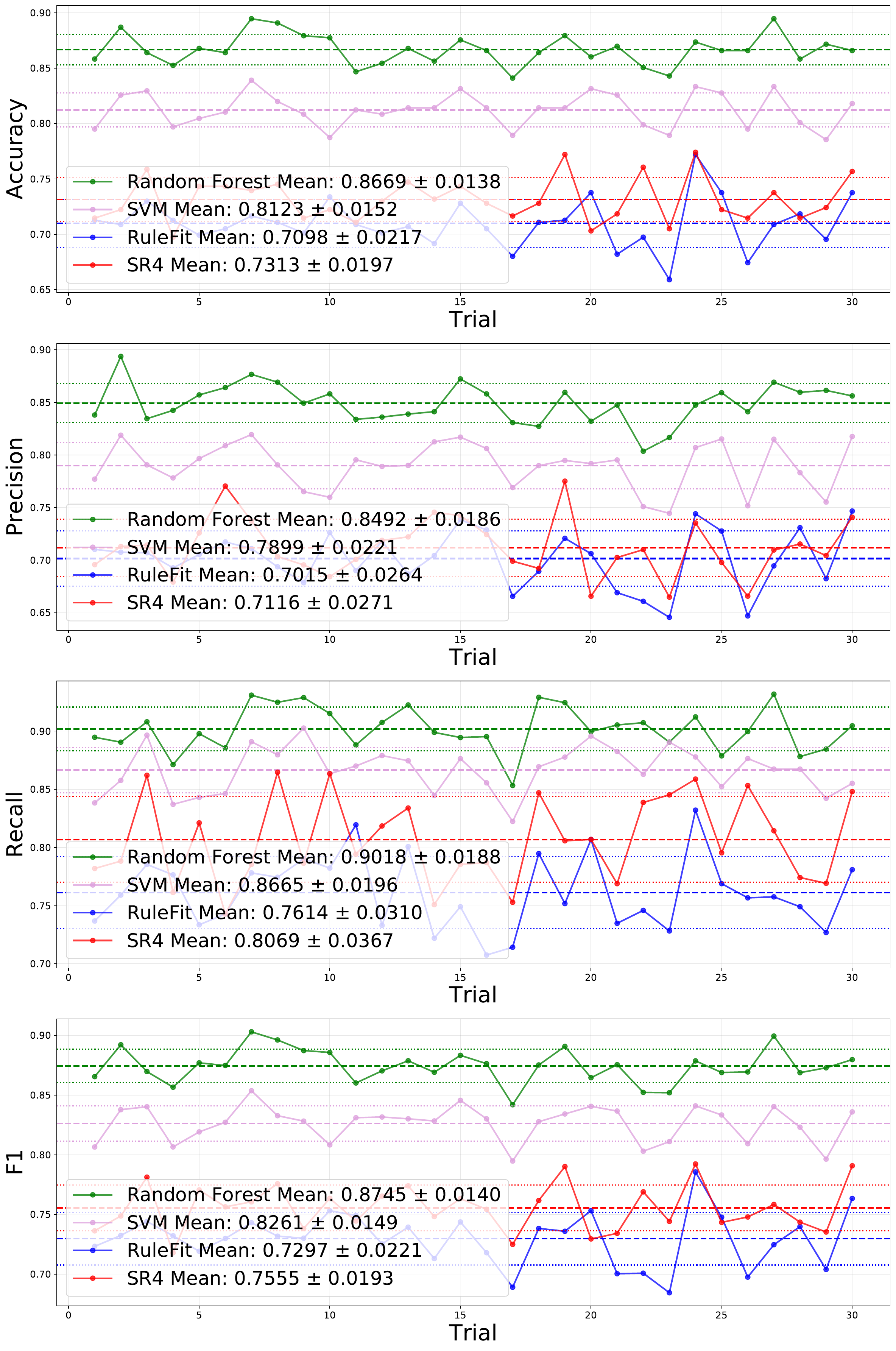}
\end{minipage}

\caption{
Line plot comparison of model performance metrics for minimum data across 30 trials.
The left panel reports results obtained using feature sets that include prior party voting percentage information,
whereas the right panel shows results with prior party voting percentages removed.
}
\label{fig:lineplot_grid_min}
\end{figure}

\begin{figure}
\centering
\begin{minipage}{0.49\linewidth}
    \centering
    \includegraphics[width=\linewidth]{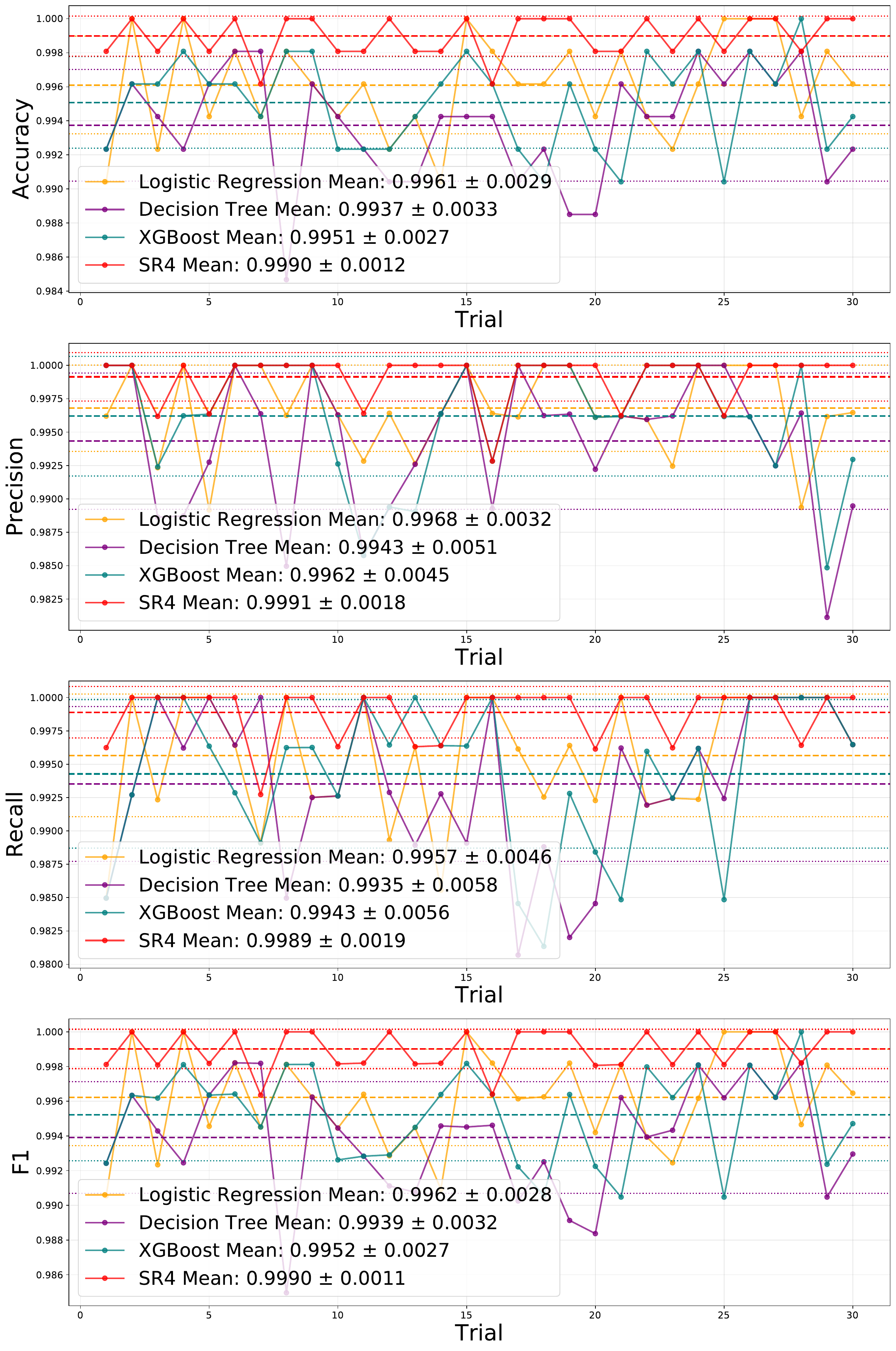}
\end{minipage}\hfill
\begin{minipage}{0.49\linewidth}
    \centering
    \includegraphics[width=\linewidth]{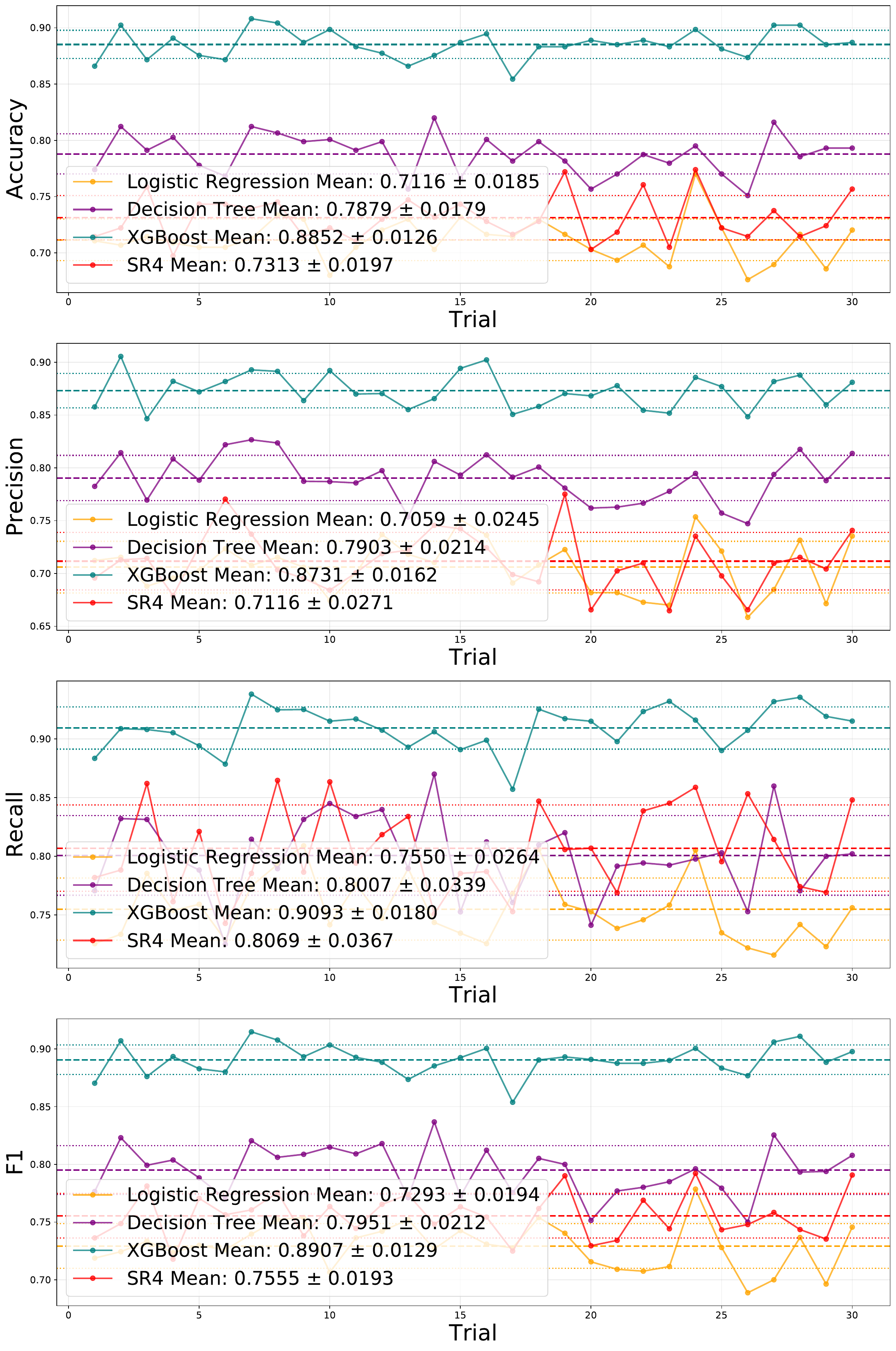}
\end{minipage}

\caption{
Line plot comparison of model (logistic regression, Decision Tree, XG boost) performance metrics for minimum data across 30 trials. The left panel reports results obtained using feature sets that include prior party voting percentage information, whereas the right panel shows results with prior party voting percentages removed.
}
\label{fig:lineplot_grid_min_other}
\end{figure}

\begin{figure}
\centering
\begin{minipage}{0.49\linewidth}
    \centering
    \includegraphics[width=\linewidth]{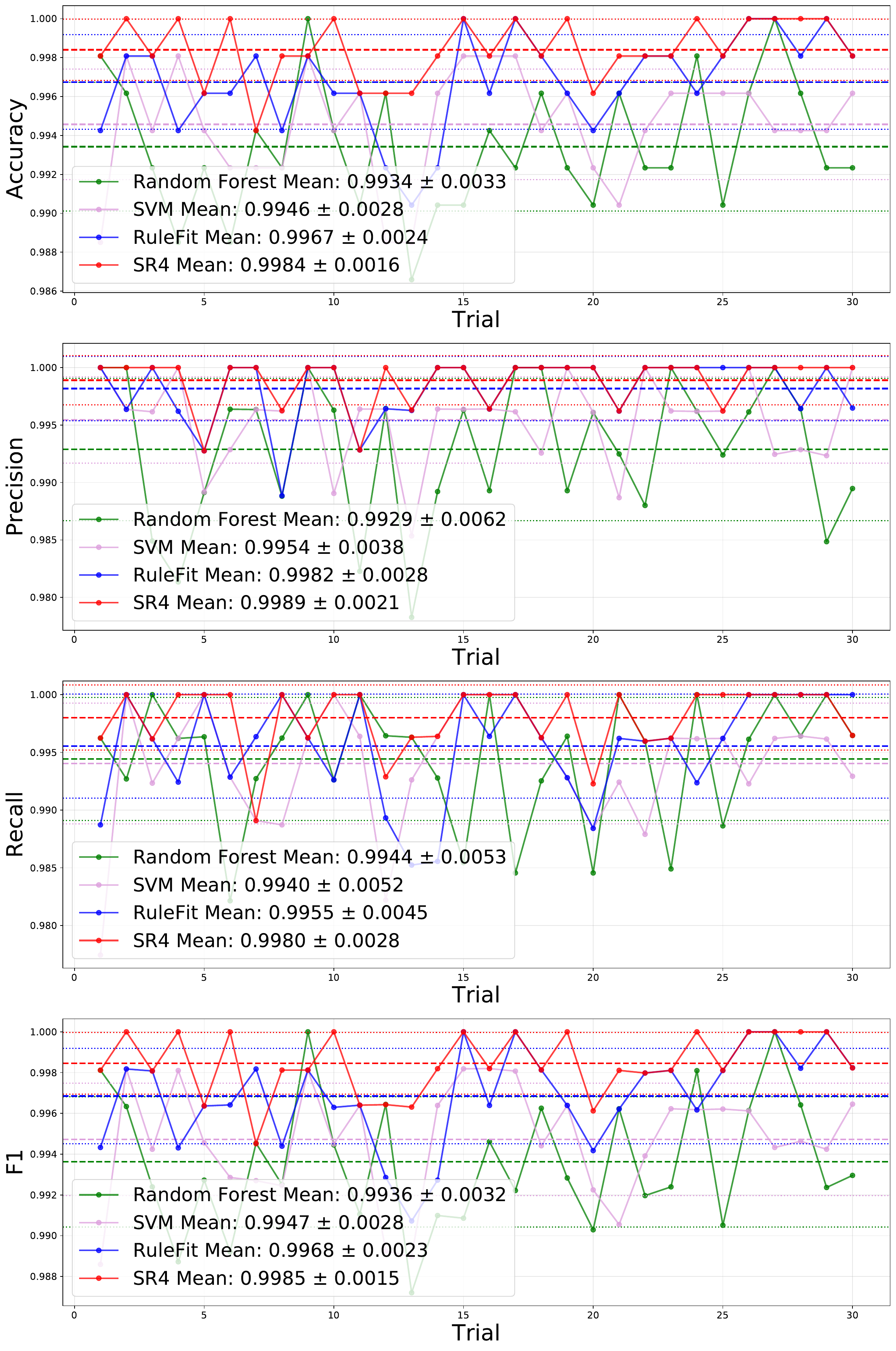}
\end{minipage}\hfill
\begin{minipage}{0.49\linewidth}
    \centering
    \includegraphics[width=\linewidth]{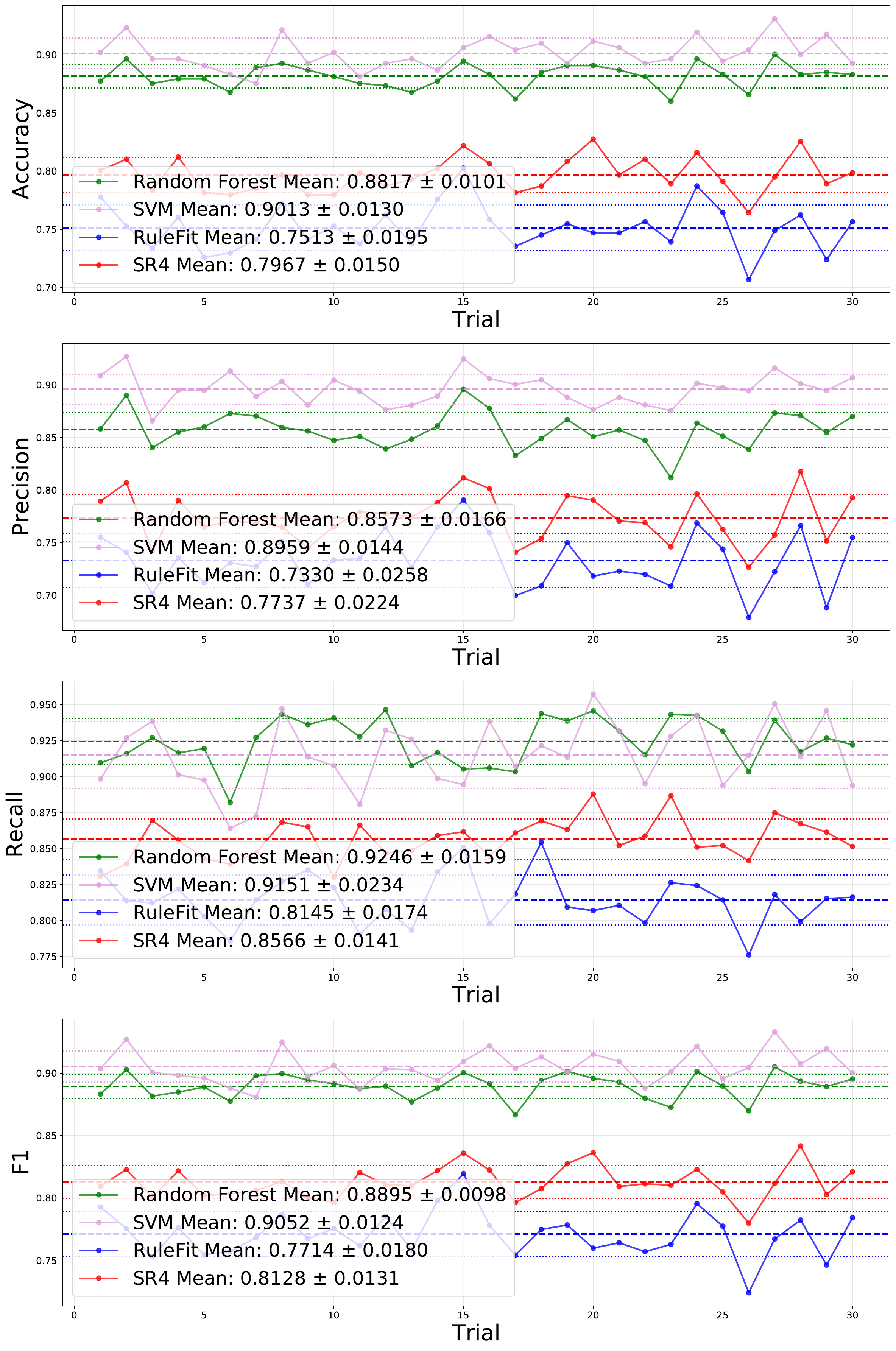}
\end{minipage}

\caption{
Line plot comparison of model performance metrics for standard data across 30 trials.
The left panel reports results obtained using feature sets that include prior party voting percentage information,
whereas the right panel shows results with prior party voting percentages removed.
}
\label{fig:lineplot_grid_std}
\end{figure}

\begin{figure}
\centering
\begin{minipage}{0.49\linewidth}
    \centering
    \includegraphics[width=\linewidth]{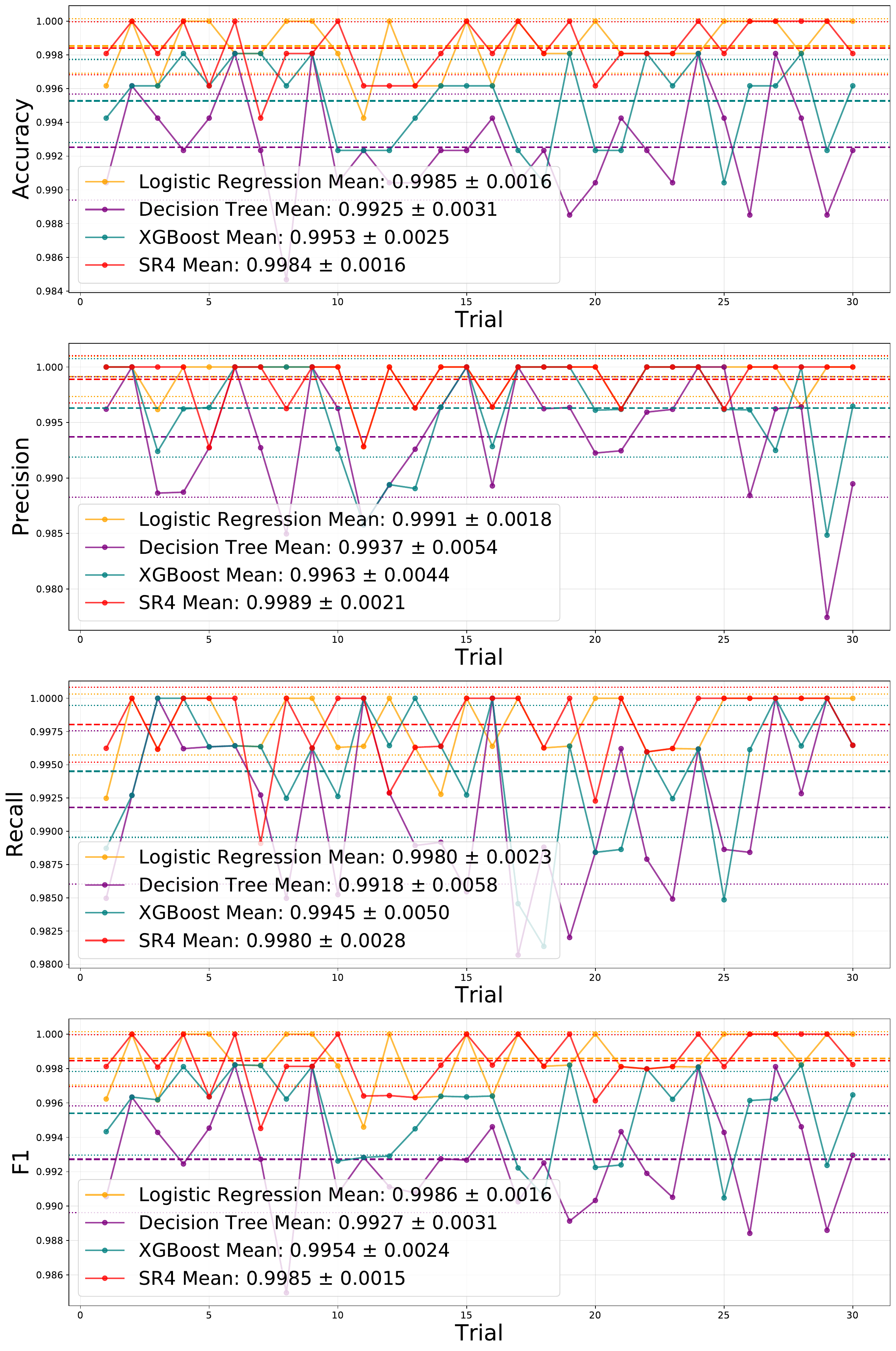}
\end{minipage}\hfill
\begin{minipage}{0.49\linewidth}
    \centering
    \includegraphics[width=\linewidth]{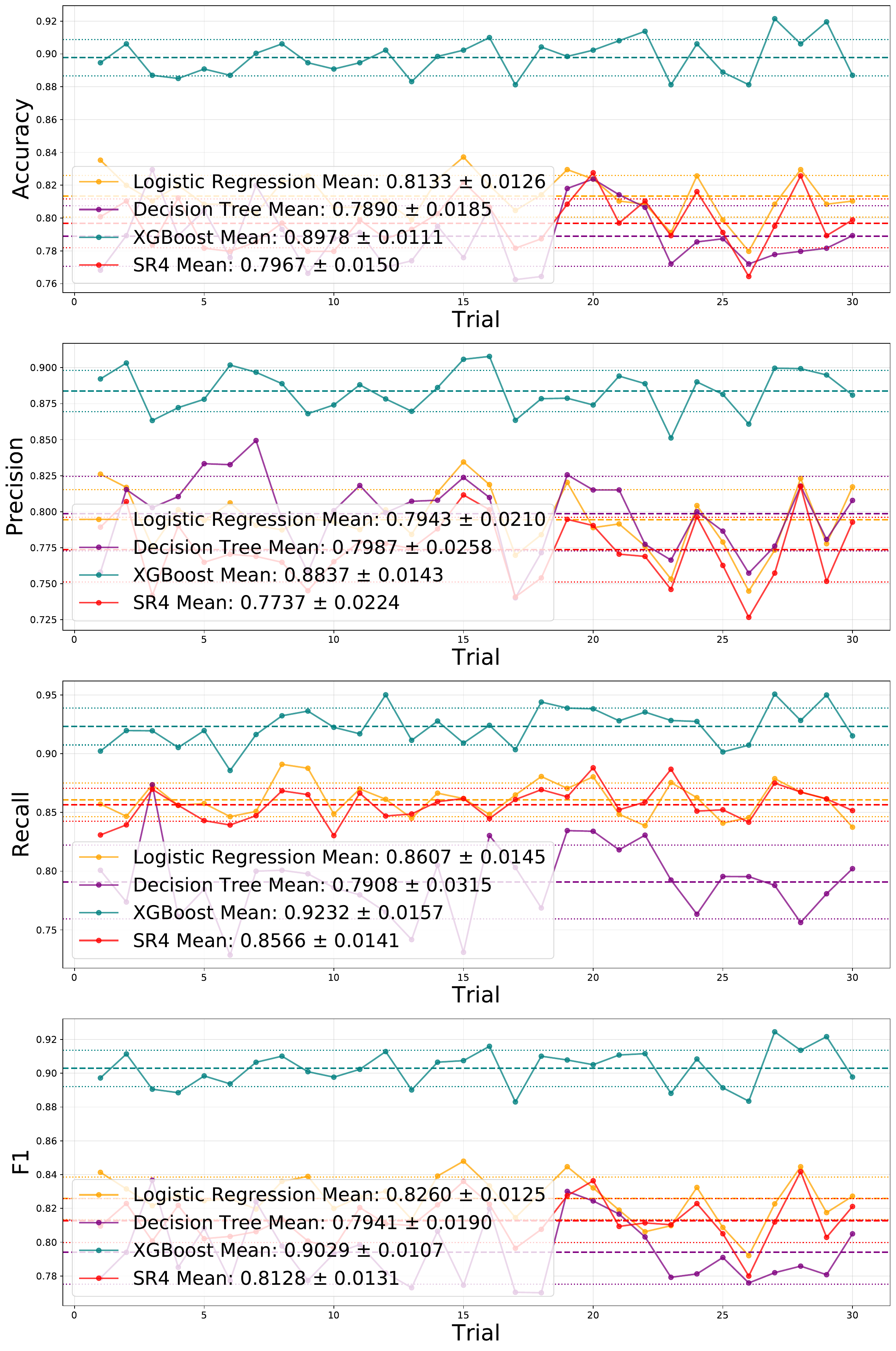}
\end{minipage}

\caption{
Line plot comparison of model (logistic regression, Decision Tree, XG boost) performance metrics for standard data across 30 trials. The left panel reports results obtained using feature sets that include prior party voting percentage information, whereas the right panel shows results with prior party voting percentages removed.
}
\label{fig:lineplot_grid_std_other}
\end{figure}

\begin{figure}
\centering
\begin{minipage}{0.49\linewidth}
    \centering
    \includegraphics[width=\linewidth]{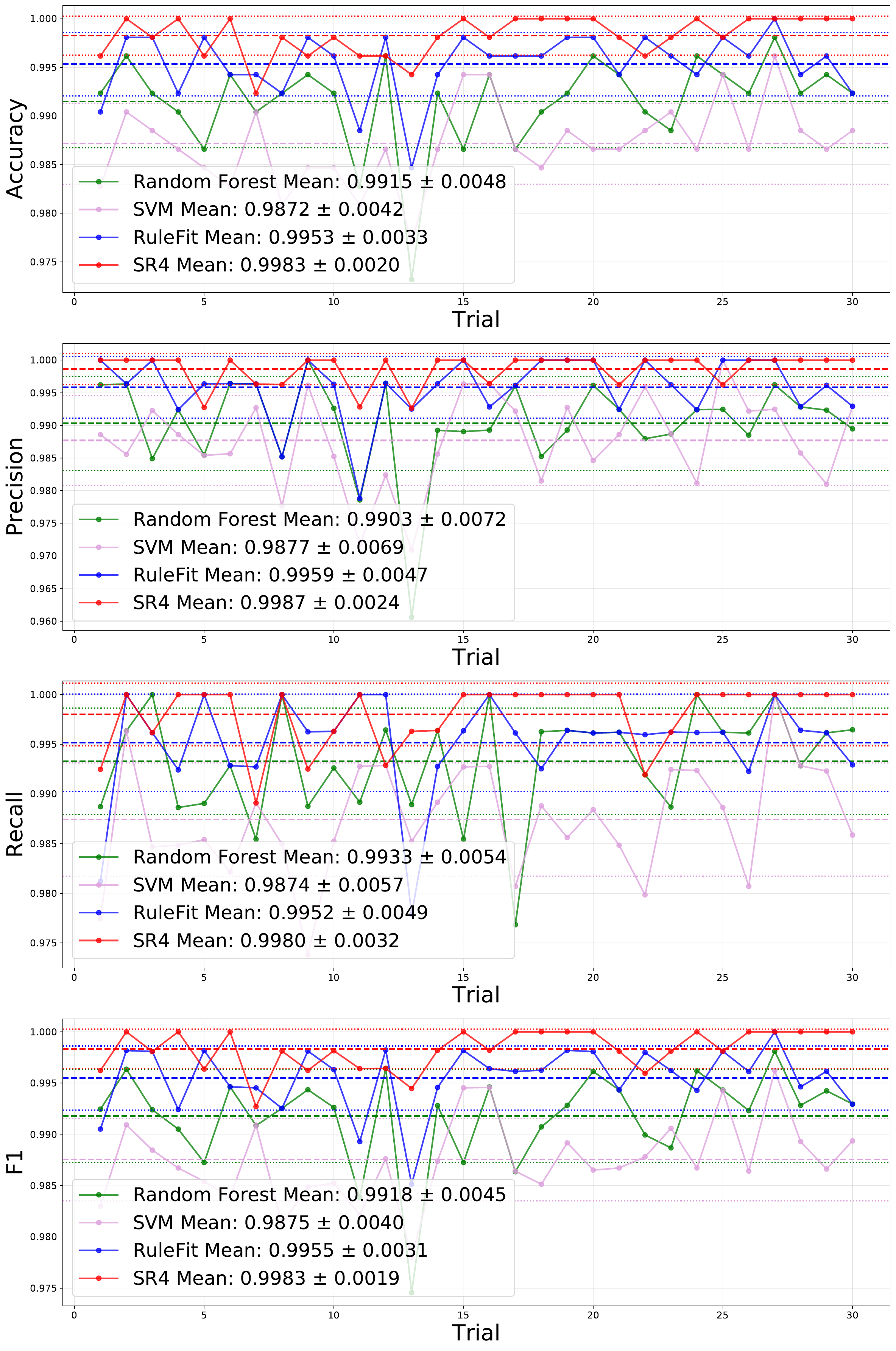}
\end{minipage}\hfill
\begin{minipage}{0.49\linewidth}
    \centering
    \includegraphics[width=\linewidth]{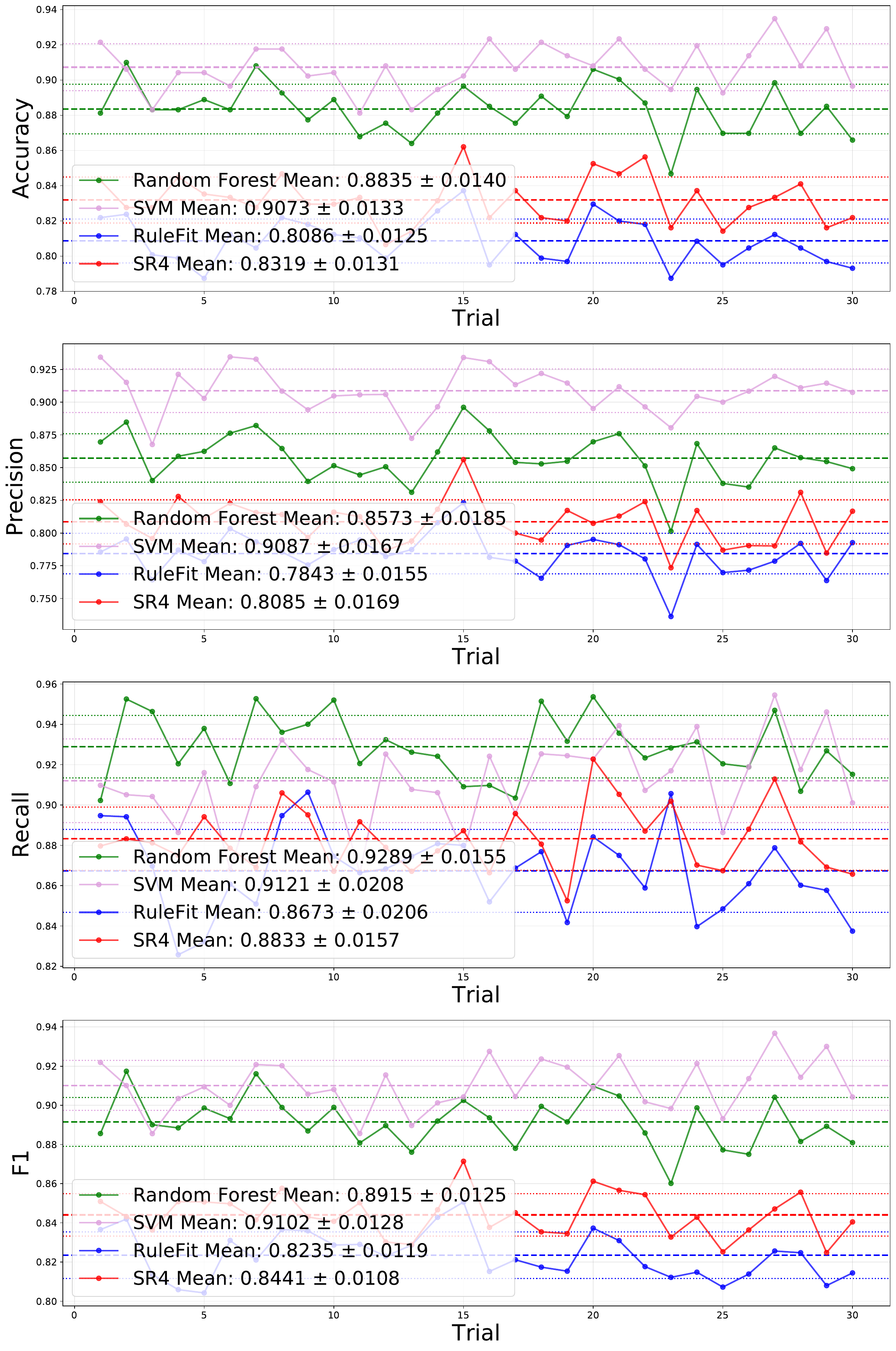}
\end{minipage}

\caption{
Line plot comparison of model performance metrics for expanded data across 30 trials. The left panel reports results obtained using feature sets that include prior party voting percentage information,
whereas the right panel shows results with prior party voting percentages removed.
}
\label{fig:lineplot_grid_exp}
\end{figure}

\begin{figure}
\centering
\begin{minipage}{0.49\linewidth}
    \centering
    \includegraphics[width=\linewidth]{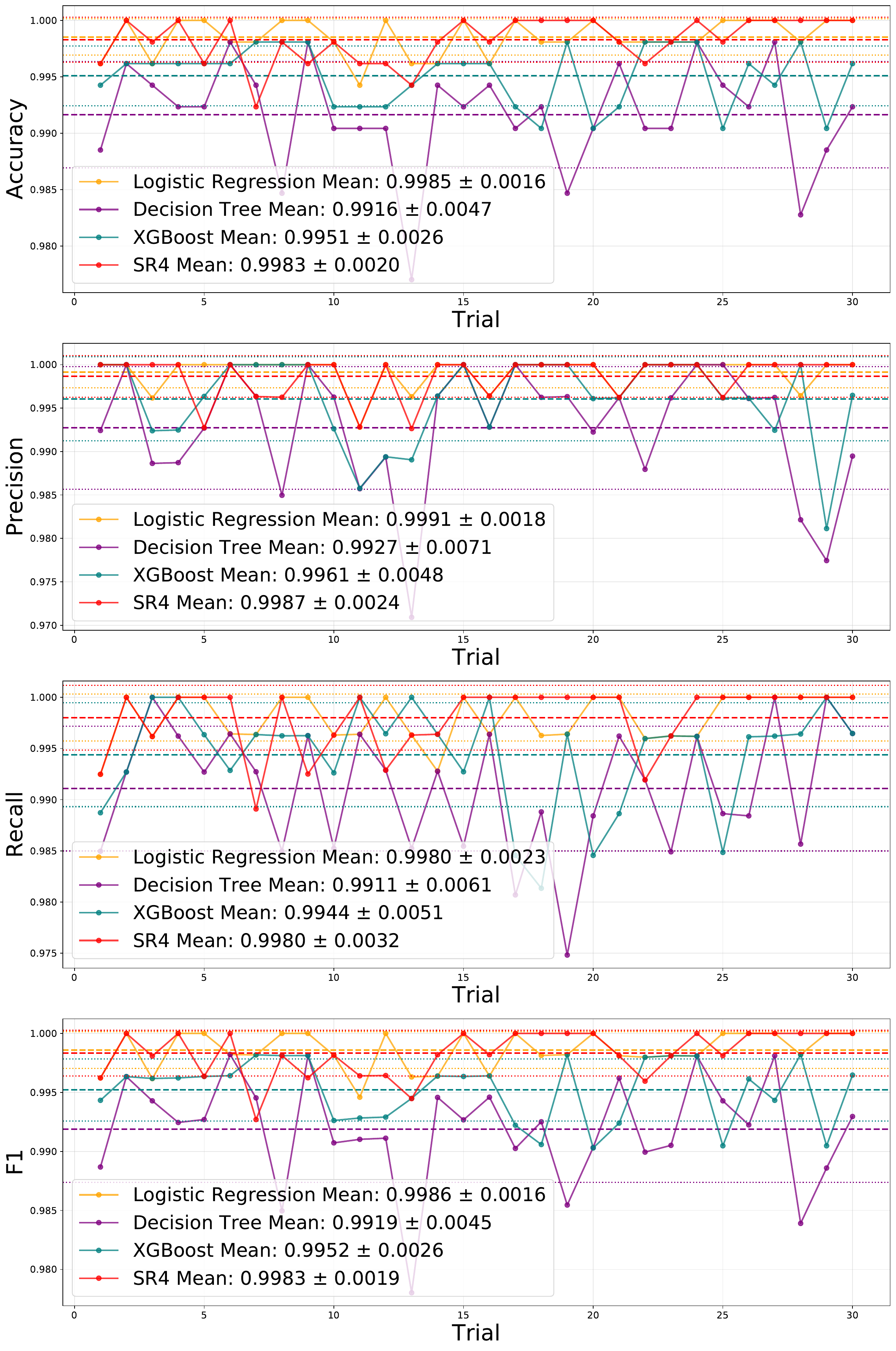}
\end{minipage}\hfill
\begin{minipage}{0.49\linewidth}
    \centering
    \includegraphics[width=\linewidth]{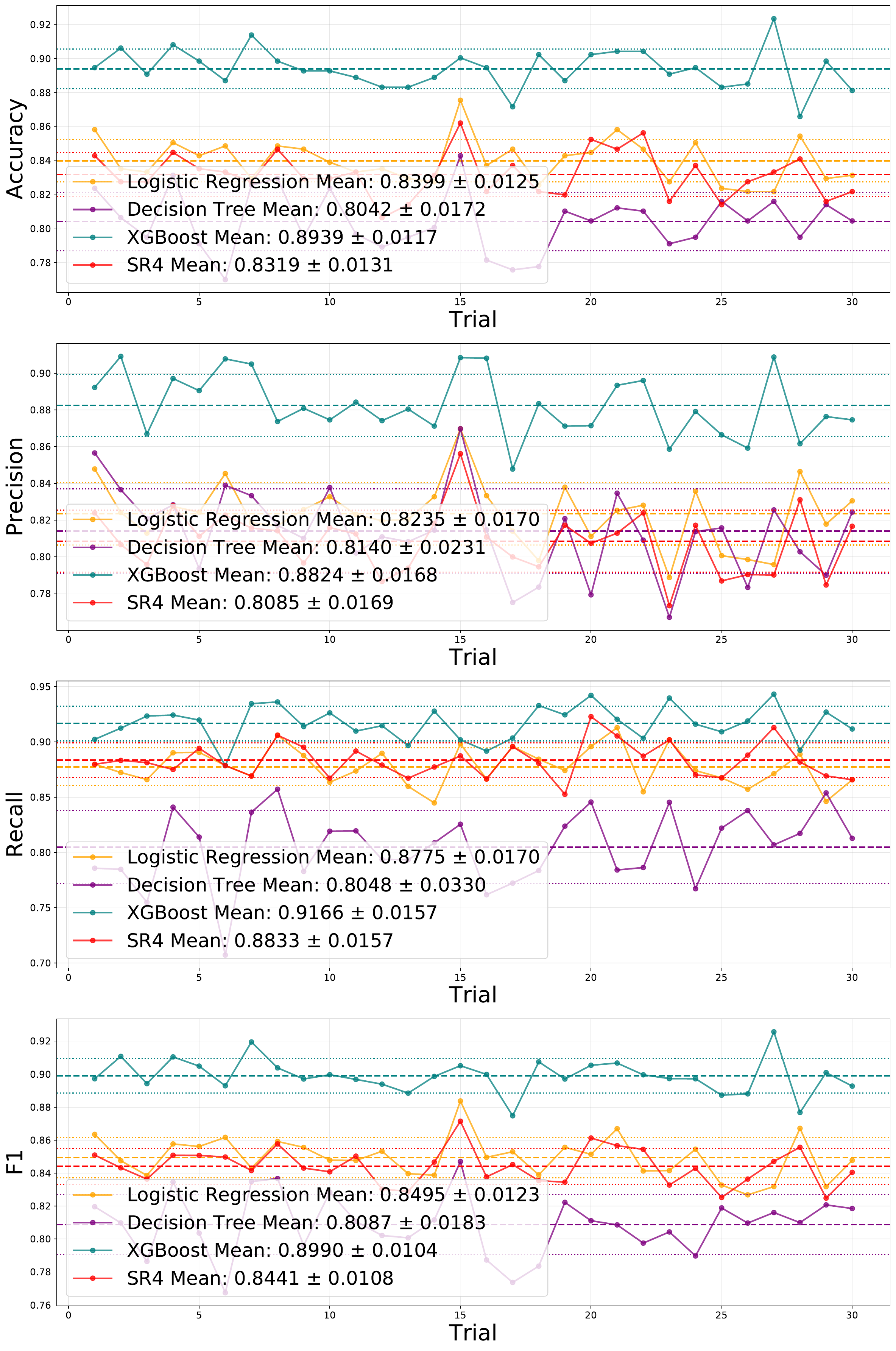}
\end{minipage}

\caption{
Line plot comparison of model (logistic regression, Decision Tree, XG boost) performance metrics for expanded data across 30 trials. The left panel reports results obtained using feature sets that include prior party voting percentage information, whereas the right panel shows results with prior party voting percentages removed.
}
\label{fig:lineplot_grid_exp_other}
\end{figure}

\begin{figure}
\centering
\begin{minipage}{0.49\linewidth}
    \centering
    \includegraphics[width=\linewidth]{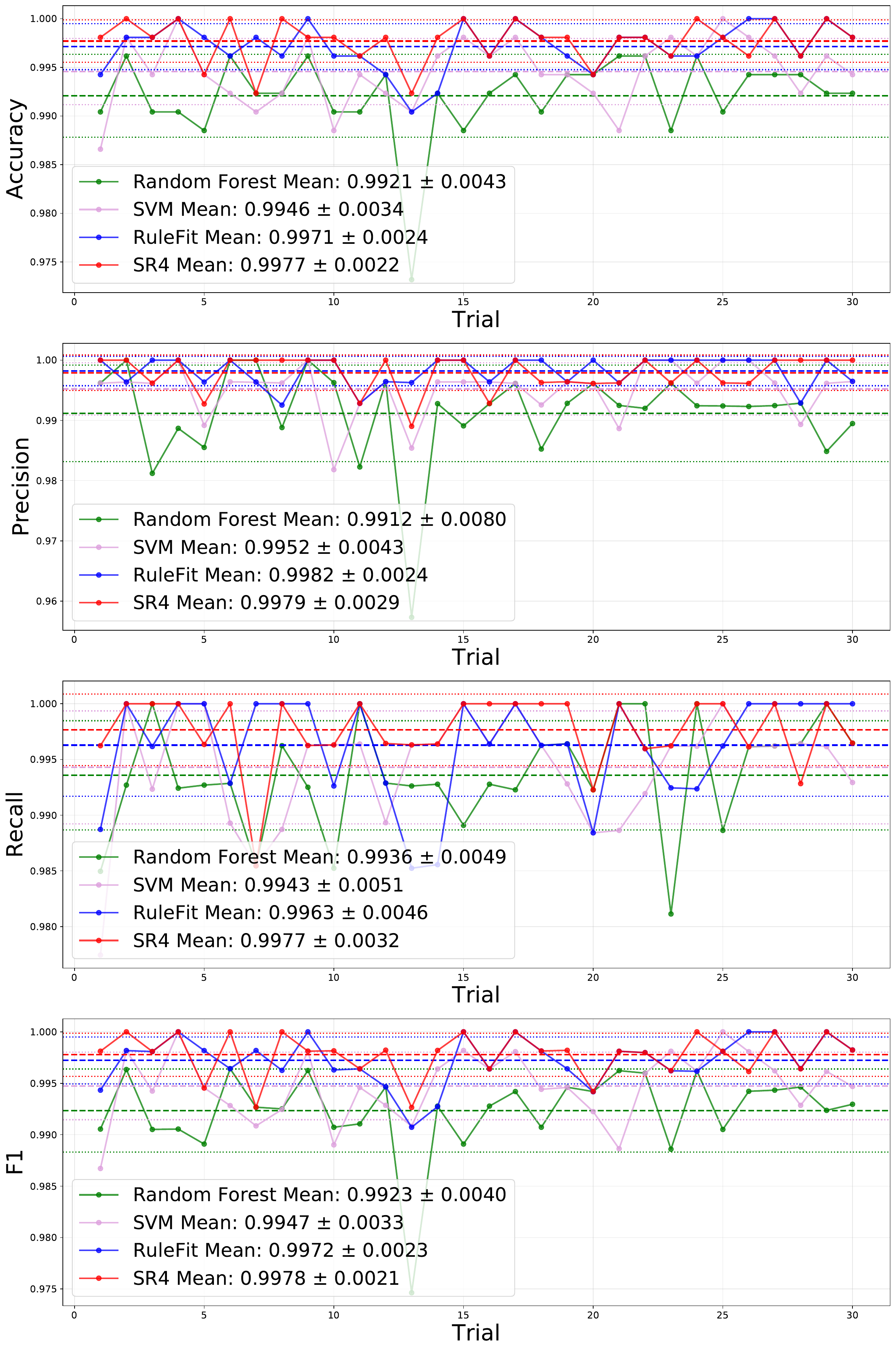}
\end{minipage}\hfill
\begin{minipage}{0.49\linewidth}
    \centering
    \includegraphics[width=\linewidth]{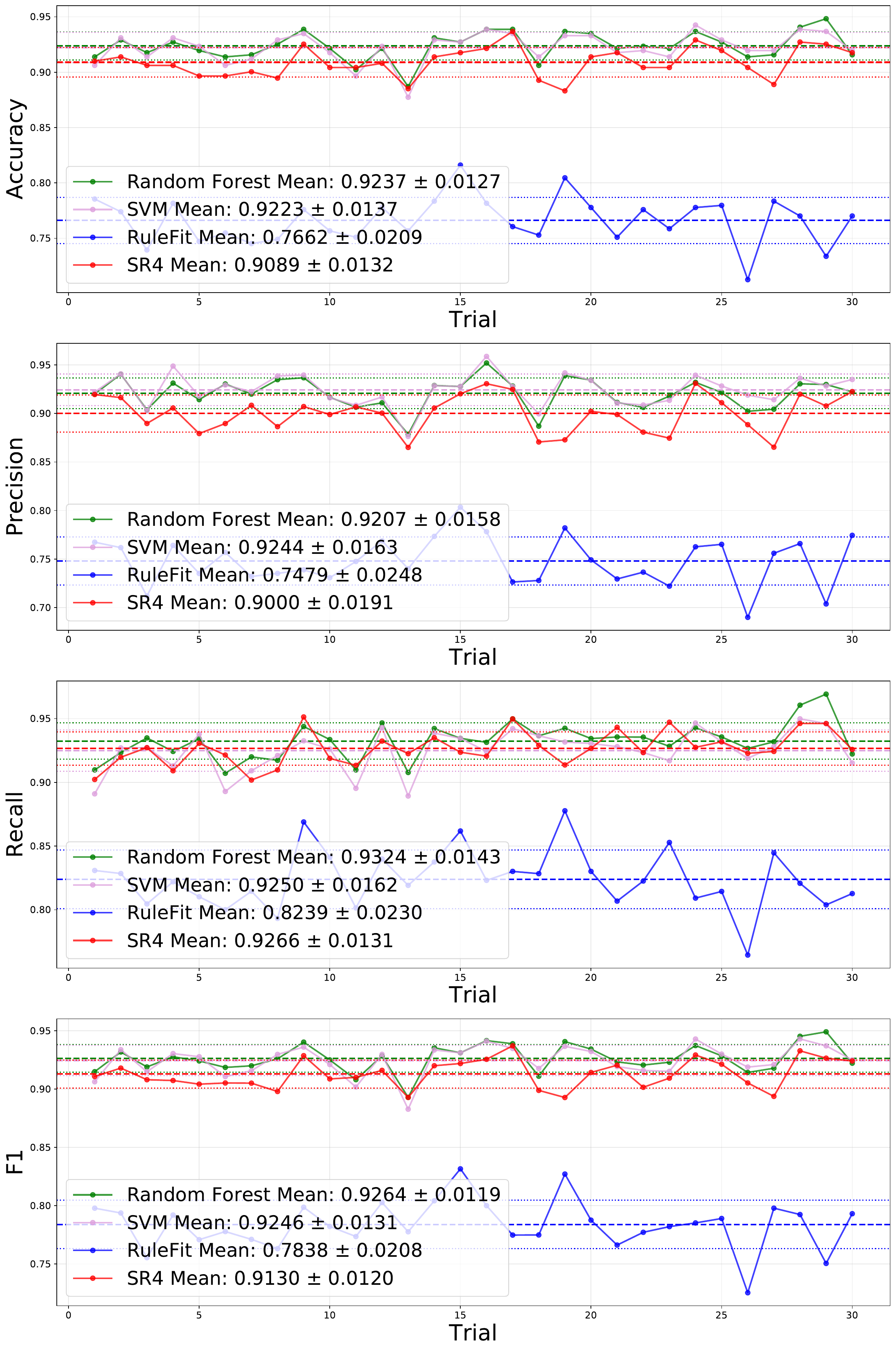}
\end{minipage}

\caption{
Line plot comparison of model performance metrics for previous data across 30 trials. The left panel reports results obtained using feature sets that include prior party voting percentage information,
whereas the right panel shows results with prior party voting percentages removed.
}
\label{fig:lineplot_grid_pre}
\end{figure}

\begin{figure}
\centering
\begin{minipage}{0.49\linewidth}
    \centering
    \includegraphics[width=\linewidth]{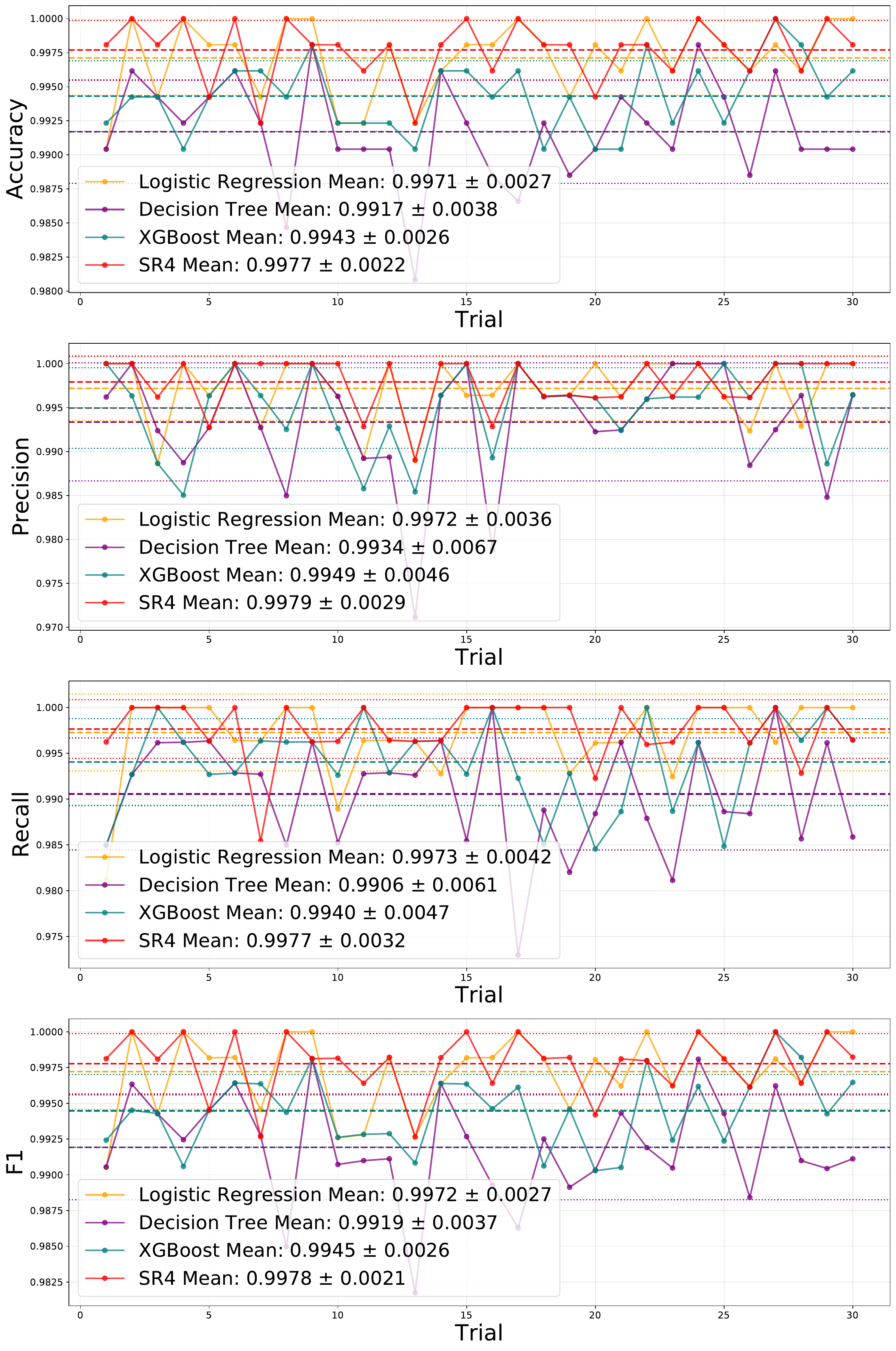}
\end{minipage}\hfill
\begin{minipage}{0.49\linewidth}
    \centering
    \includegraphics[width=\linewidth]{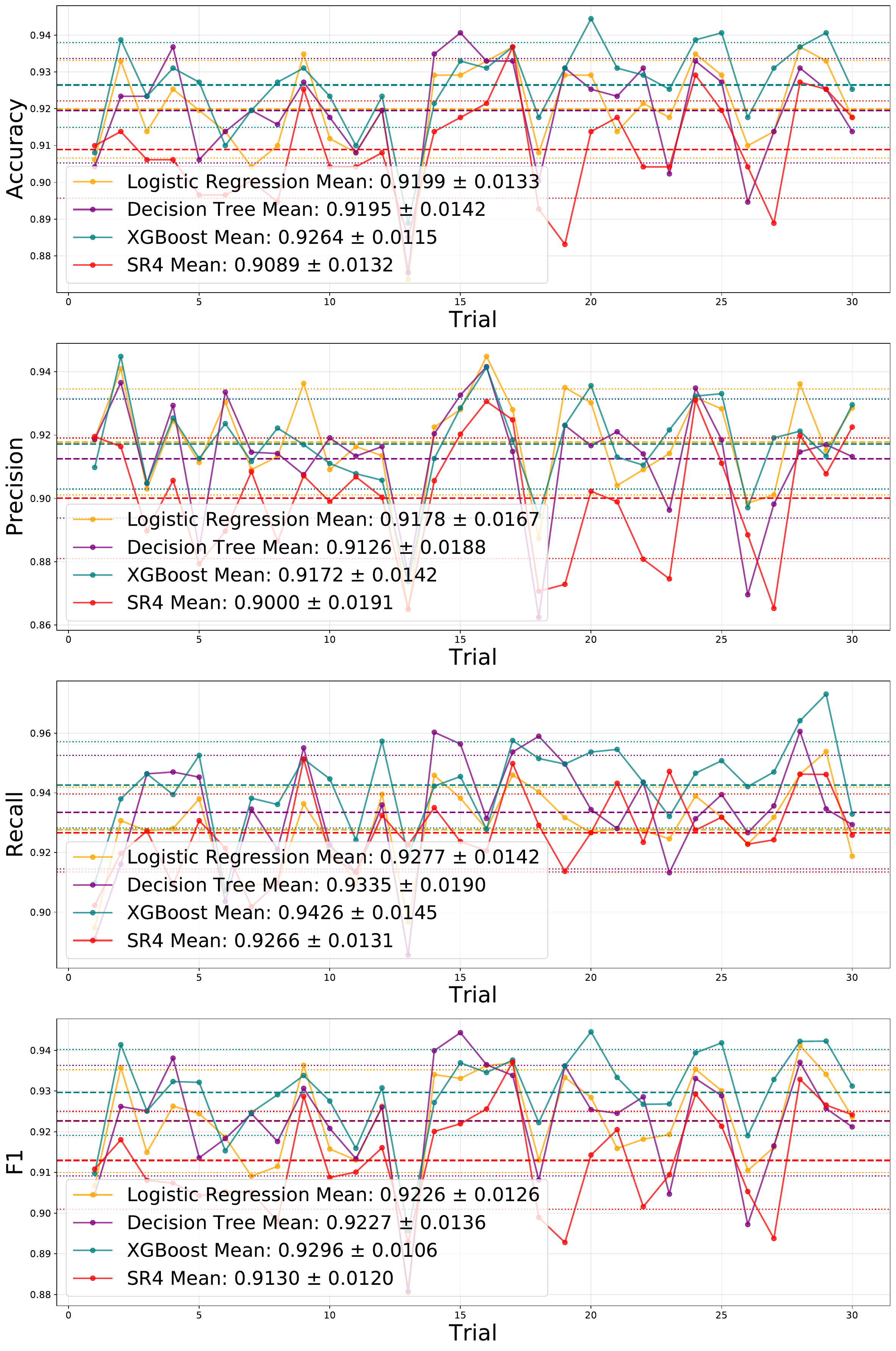}
\end{minipage}

\caption{
Line plot comparison of model (logistic regression, Decision Tree, XG boost) performance metrics for previous data across 30 trials. The left panel reports results obtained using feature sets that include prior party voting percentage information, whereas the right panel shows results with prior party voting percentages removed.
}
\label{fig:lineplot_grid_pre_other}
\end{figure}


\begin{figure}
\centering
\begin{minipage}{0.48\linewidth}
    \centering
    \includegraphics[width=\linewidth]{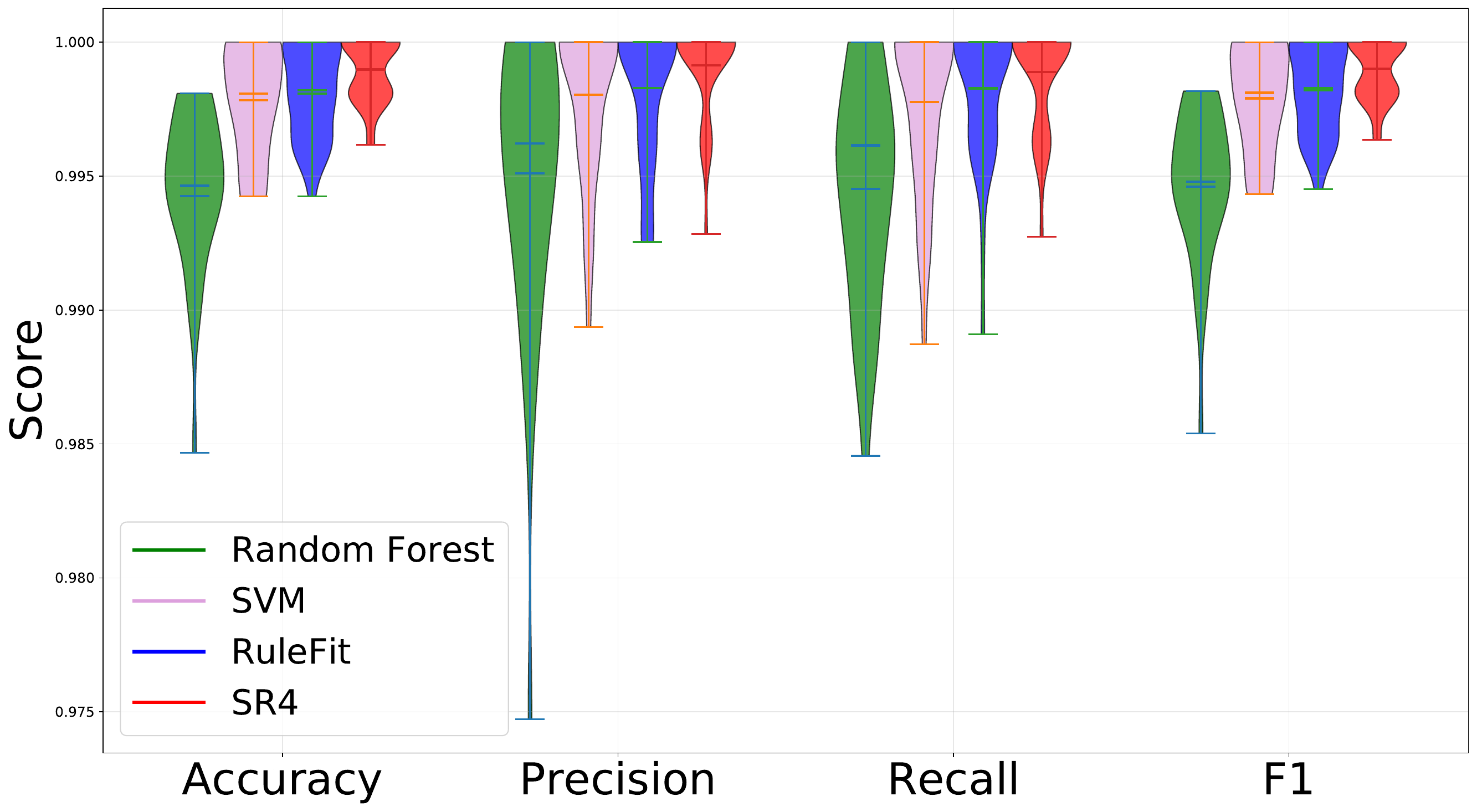}
    \subcaption{Minimum}
\end{minipage}\hfill
\begin{minipage}{0.48\linewidth}
    \centering
    \includegraphics[width=\linewidth]{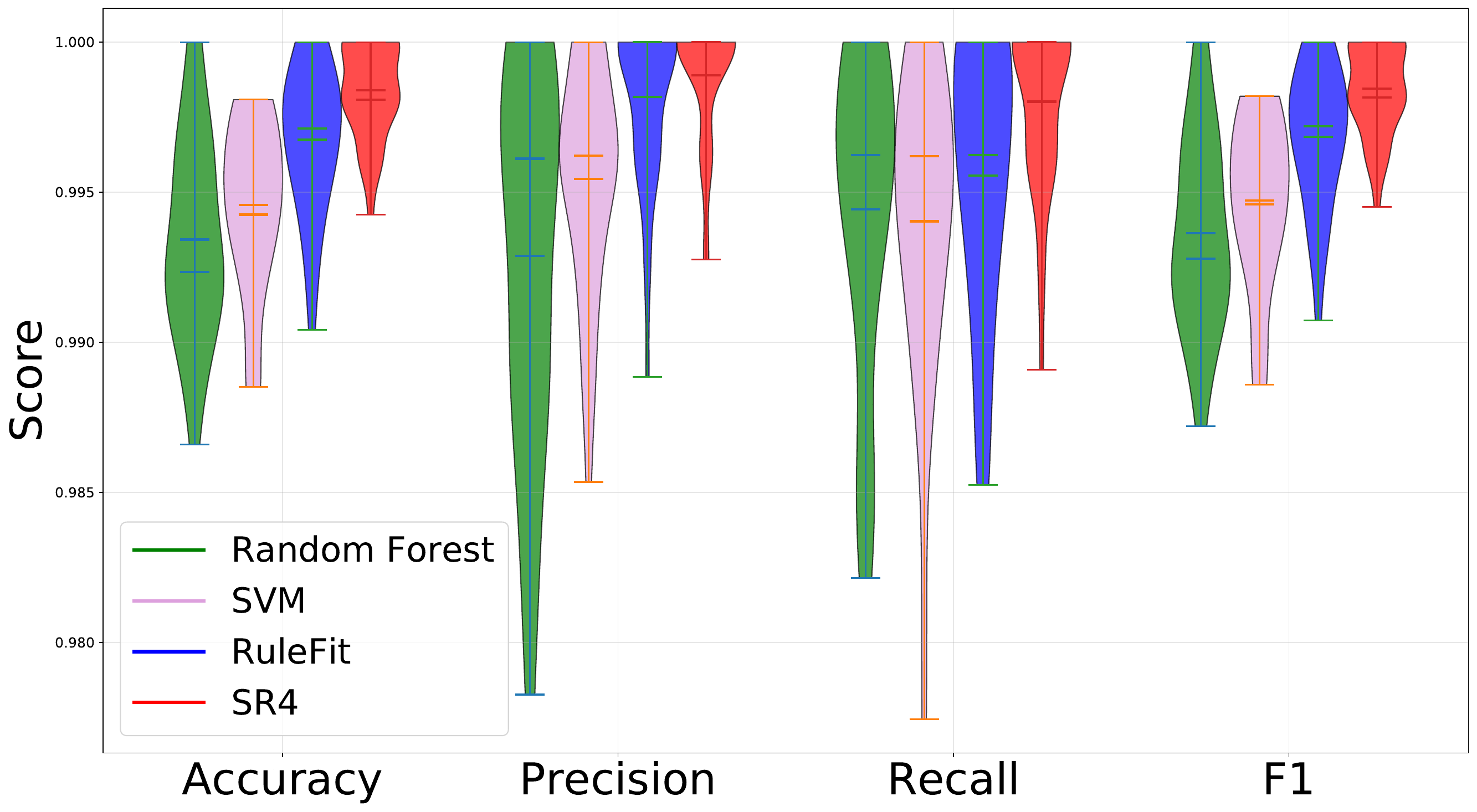}
    \subcaption{Standard}
\end{minipage}

\vspace{0.4cm}

\begin{minipage}{0.48\linewidth}
    \centering
    \includegraphics[width=\linewidth]{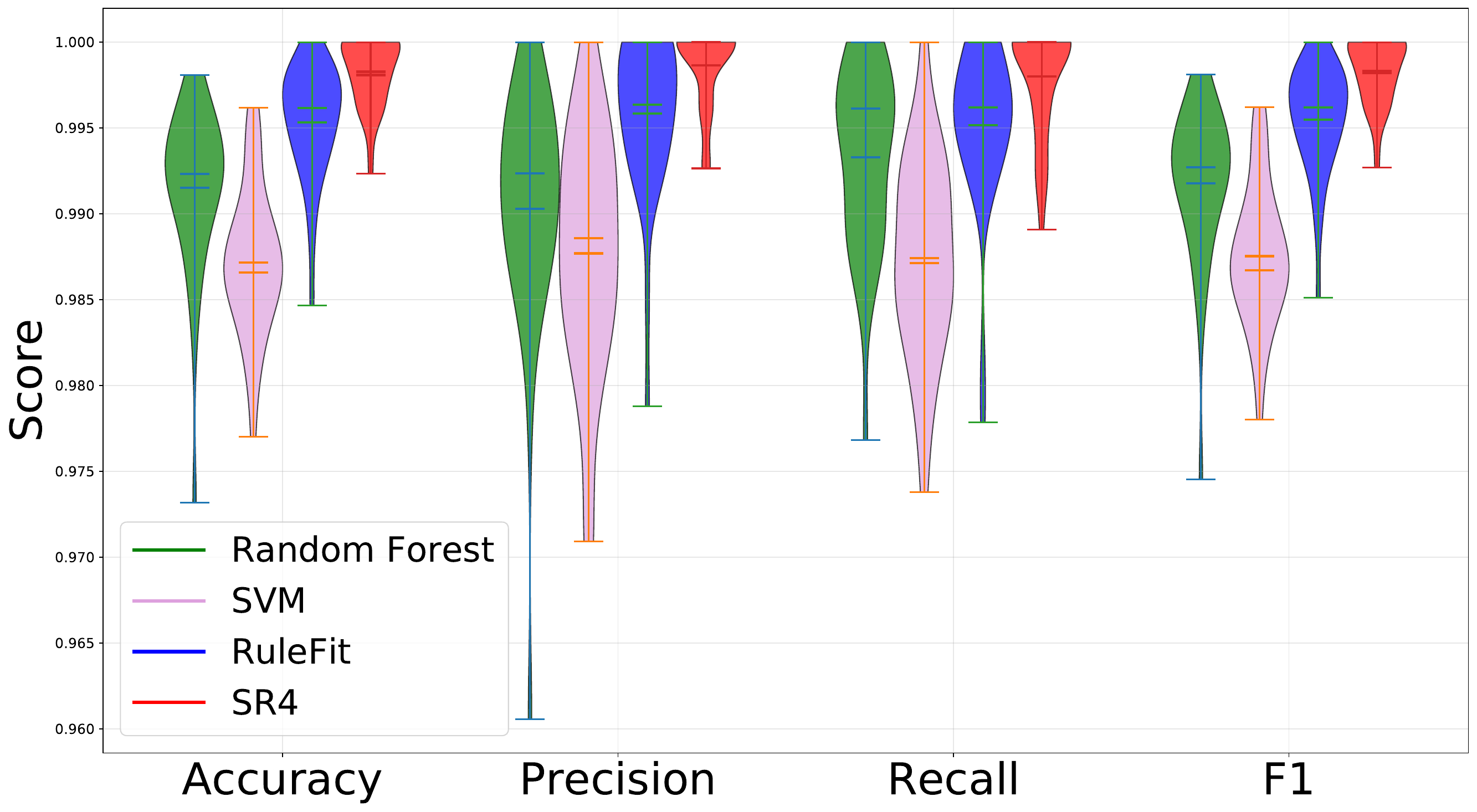}
    \subcaption{Expanded}
\end{minipage}\hfill
\begin{minipage}{0.48\linewidth}
    \centering
    \includegraphics[width=\linewidth]
    {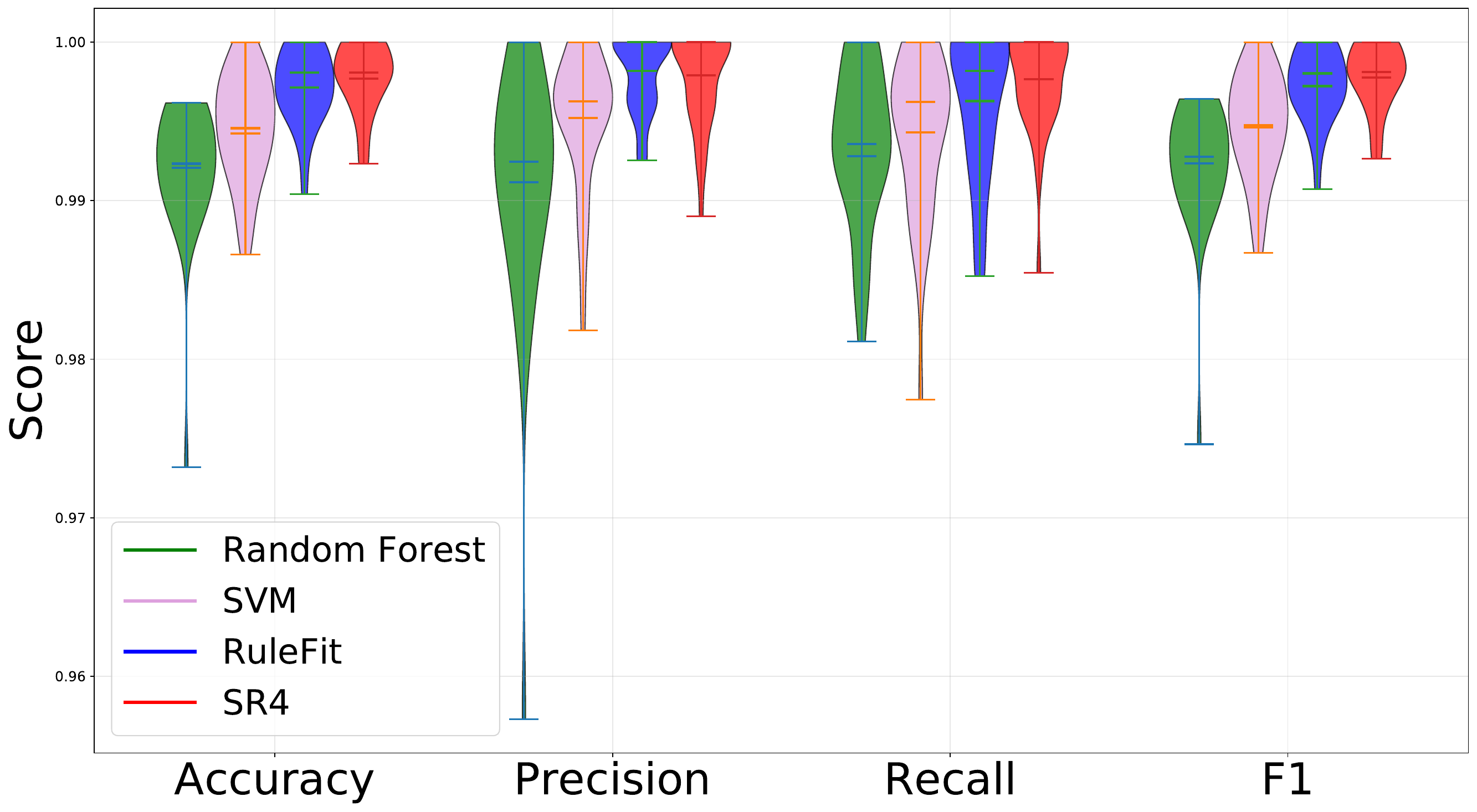}
    \subcaption{Previous}
\end{minipage}

\caption{
Violin plot comparison of model performance metrics across multiple datasets for the election classification task that includes prior party voting percentage information.}
\label{fig:violin_all_models}
\end{figure}

\begin{figure}
\centering
\begin{minipage}{0.48\linewidth}
    \centering
    \includegraphics[width=\linewidth]{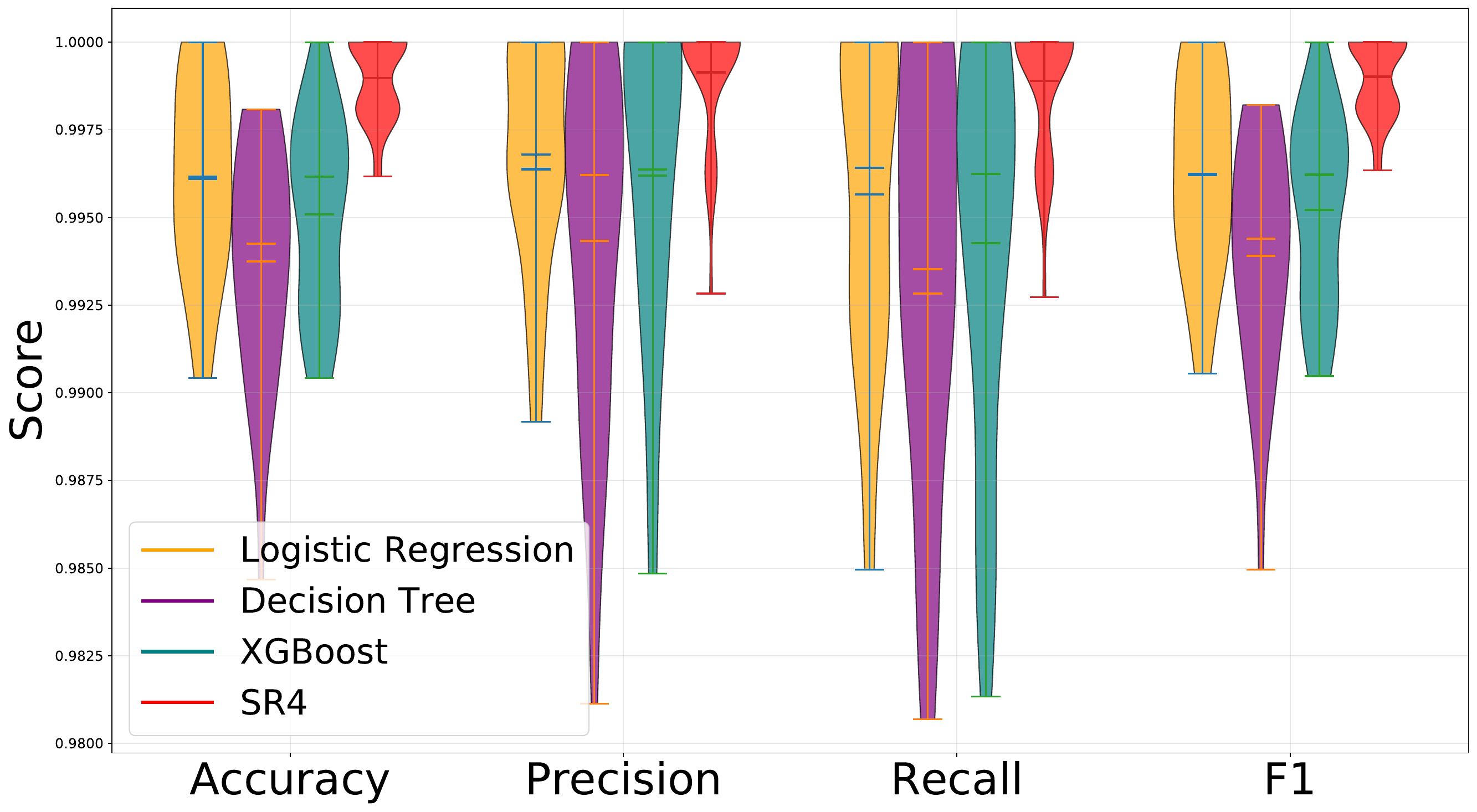}
    \subcaption{Minimum}
\end{minipage}\hfill
\begin{minipage}{0.48\linewidth}
    \centering
    \includegraphics[width=\linewidth]{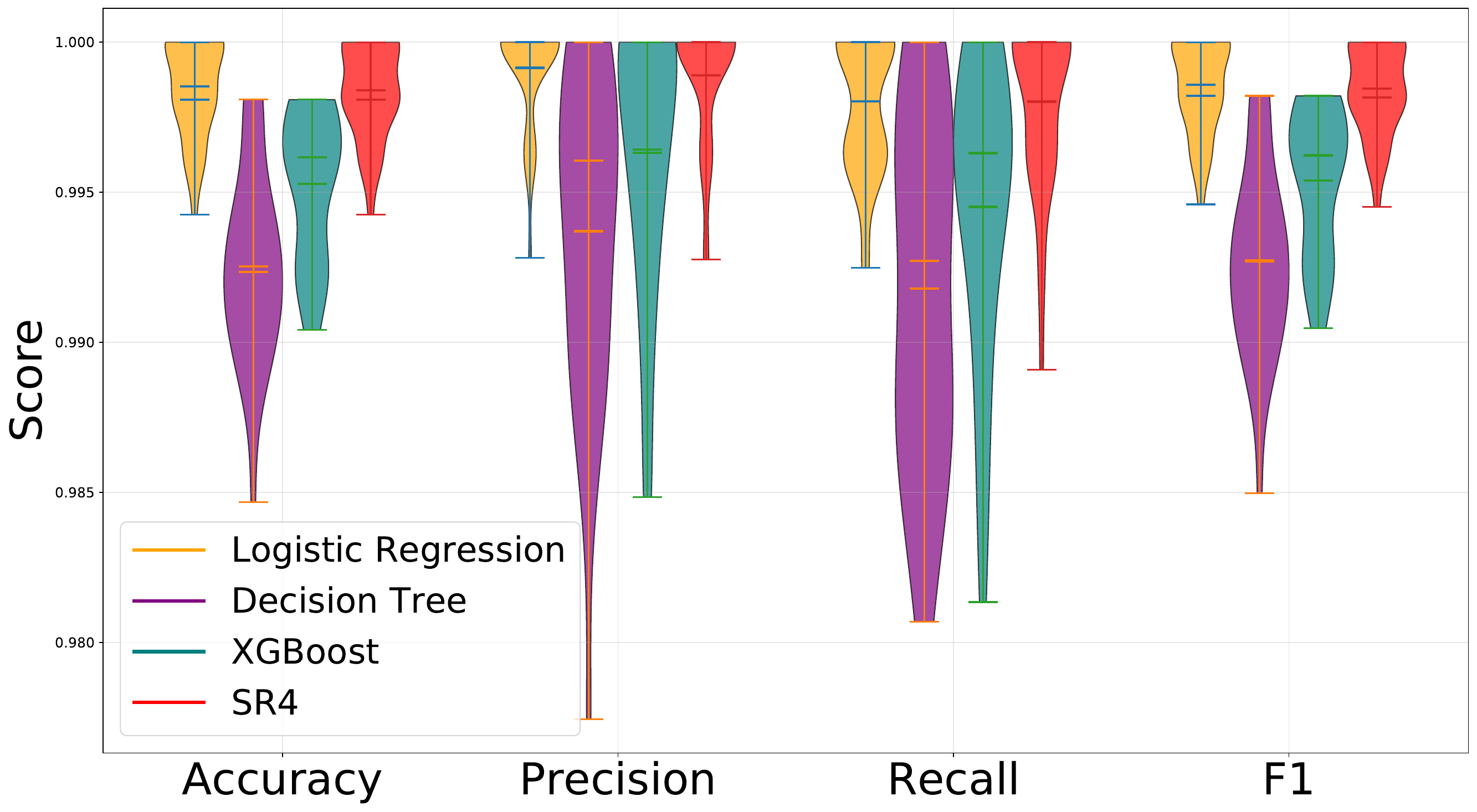}
    \subcaption{Standard}
\end{minipage}

\vspace{0.4cm}

\begin{minipage}{0.48\linewidth}
    \centering
    \includegraphics[width=\linewidth]{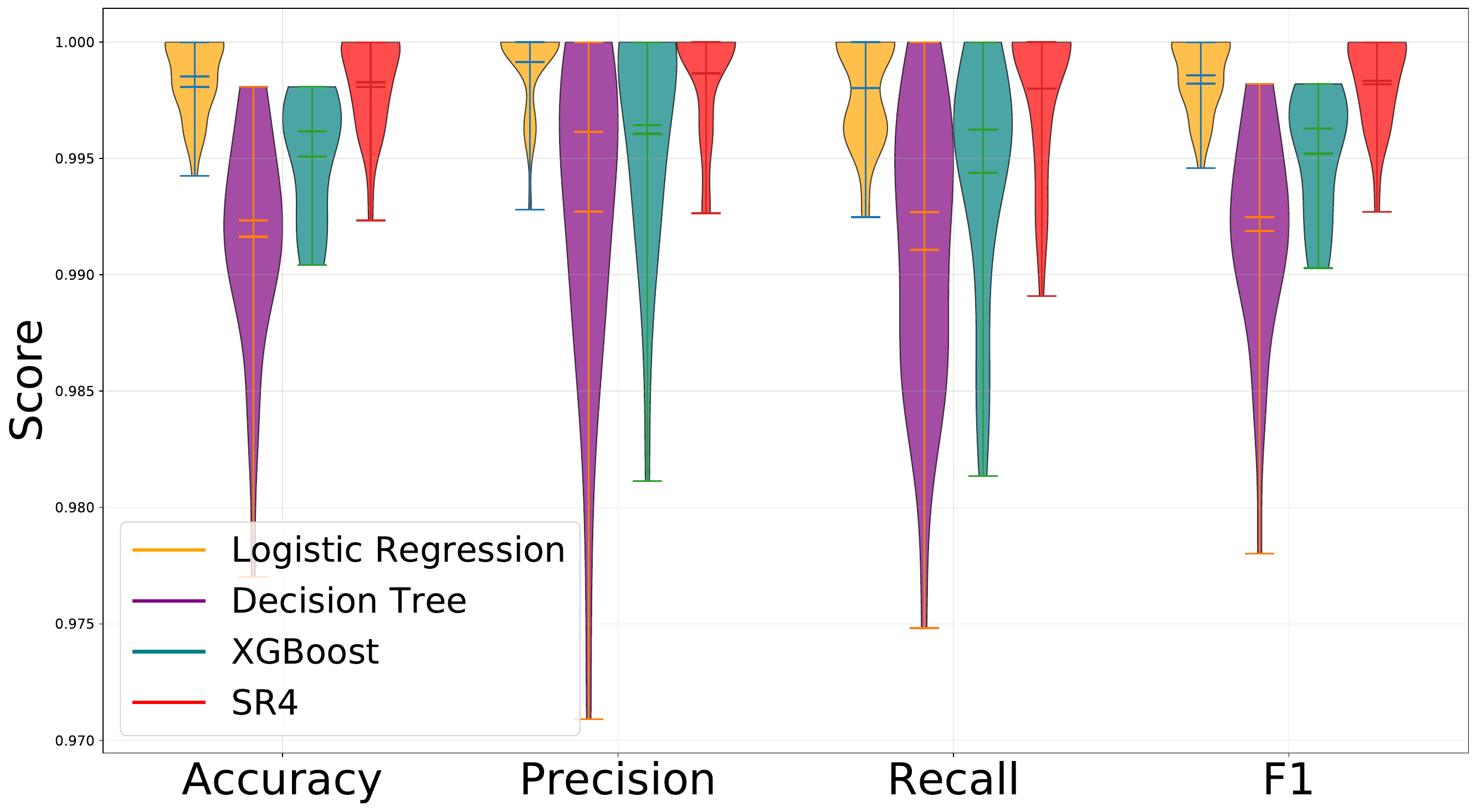}
    \subcaption{Expanded}
\end{minipage}\hfill
\begin{minipage}{0.48\linewidth}
    \centering
    \includegraphics[width=\linewidth]{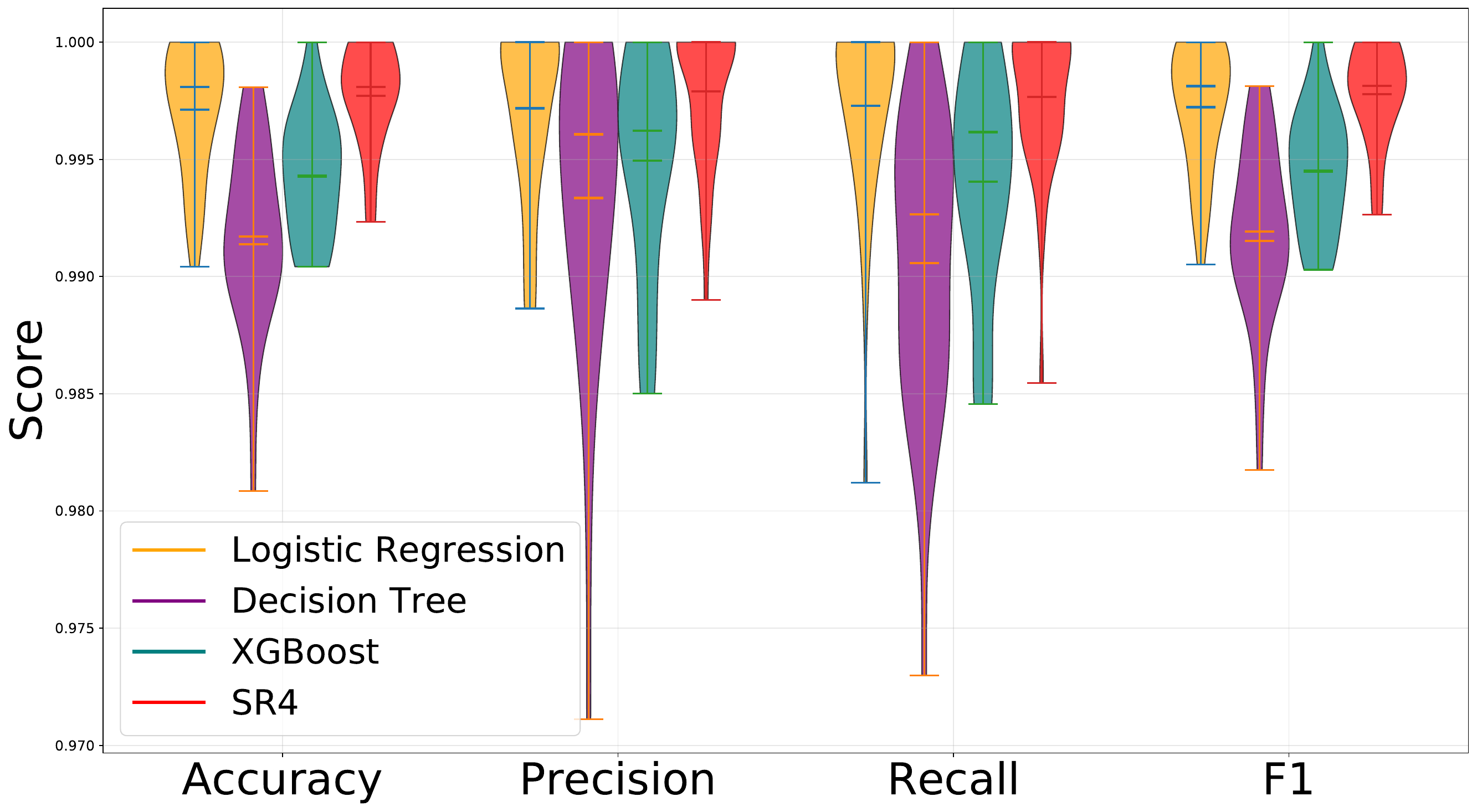}
    \subcaption{Previous}
\end{minipage}

\caption{
Violin plot comparison of model (logistic regression, Decision Tree, XG boost) performance metrics across multiple datasets for the election classification task that includes prior party voting percentage information.
}
\label{fig:violin_all_models_other}
\end{figure}


\begin{figure}
\centering
\begin{minipage}{0.48\linewidth}
    \centering
    \includegraphics[width=\linewidth]{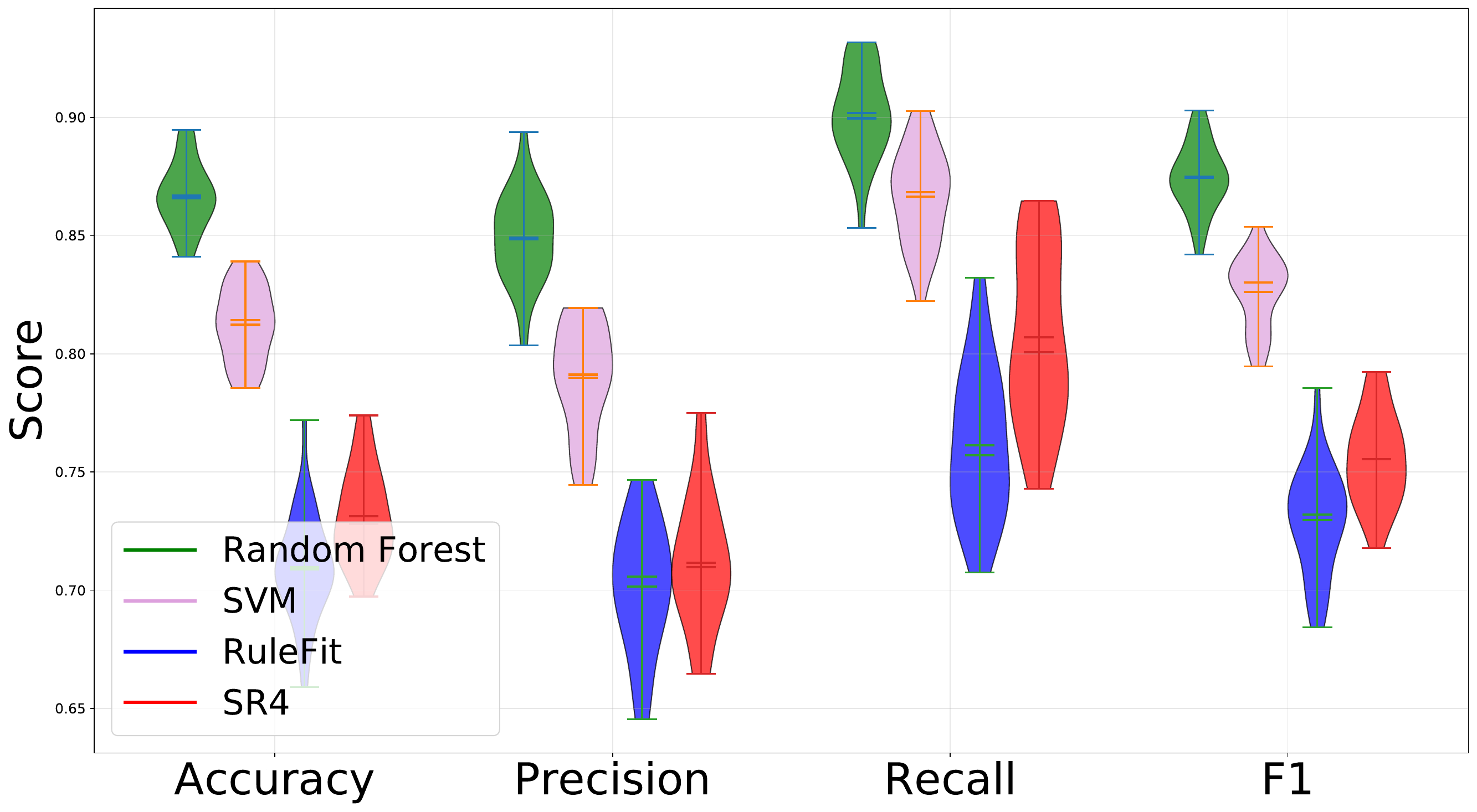}
    \subcaption{Minimum}
\end{minipage}\hfill
\begin{minipage}{0.48\linewidth}
    \centering
    \includegraphics[width=\linewidth]{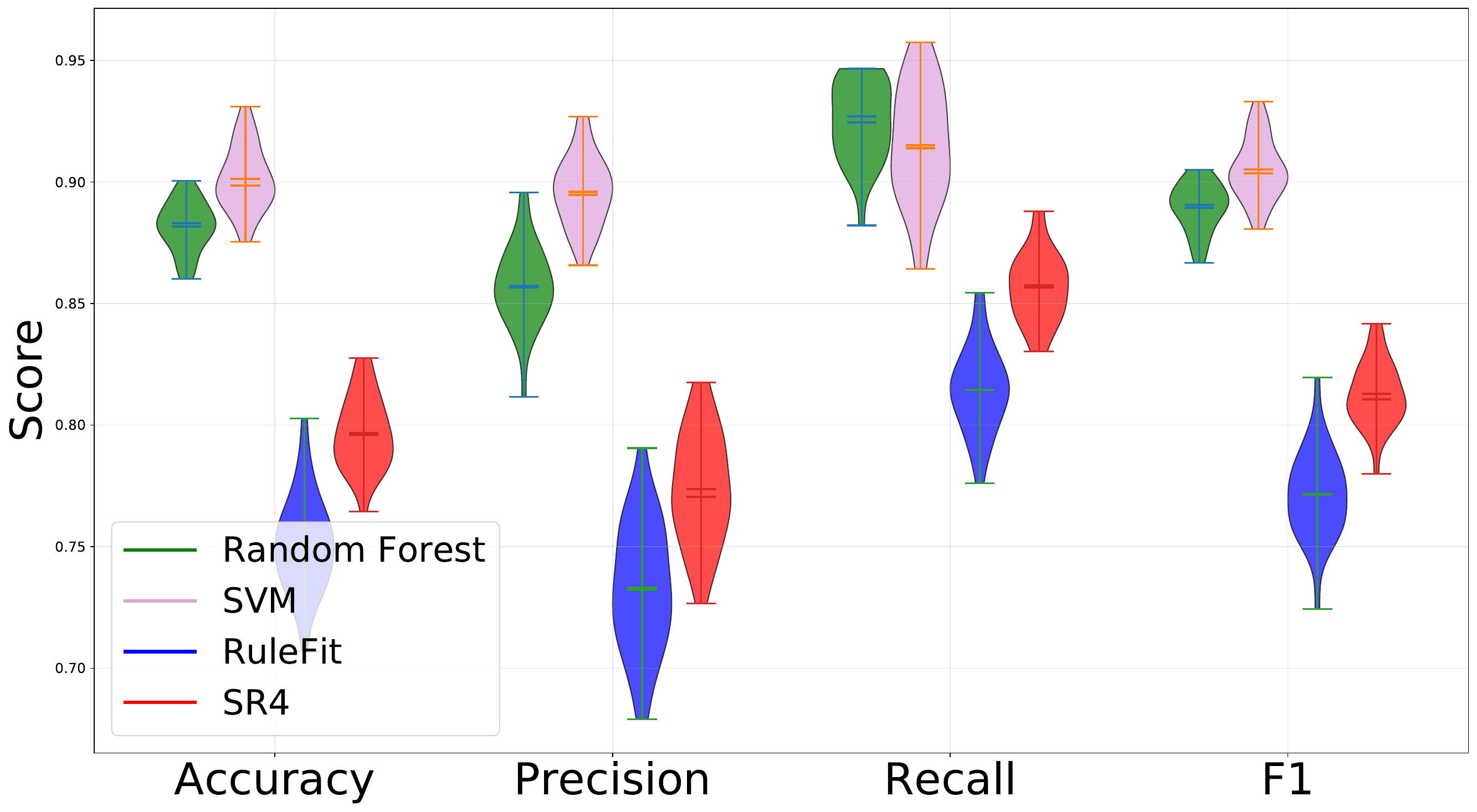}
    \subcaption{Standard}
\end{minipage}

\vspace{0.4cm}

\begin{minipage}{0.48\linewidth}
    \centering
    \includegraphics[width=\linewidth]{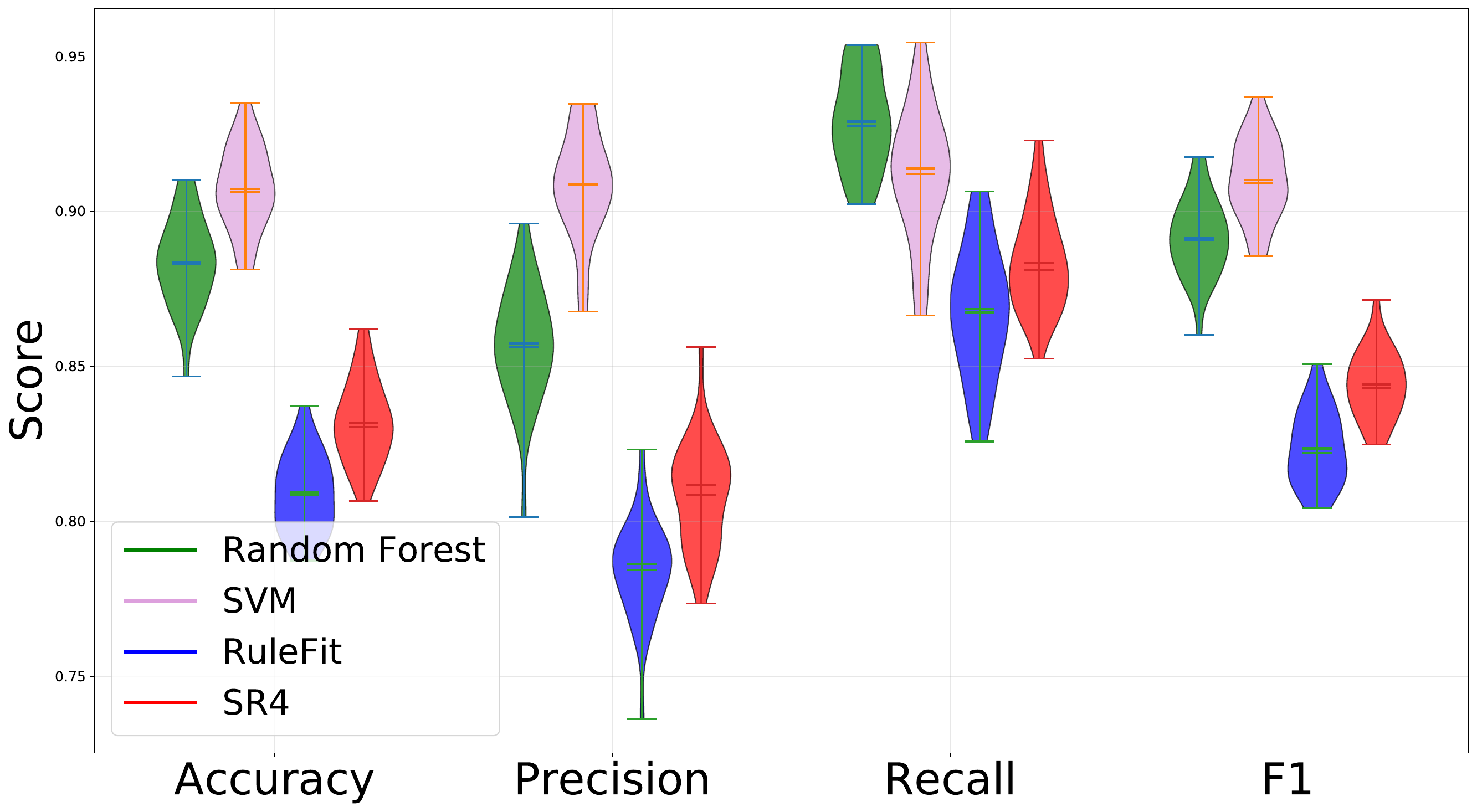}
    \subcaption{Expanded}
\end{minipage}\hfill
\begin{minipage}{0.48\linewidth}
    \centering
    \includegraphics[width=\linewidth]{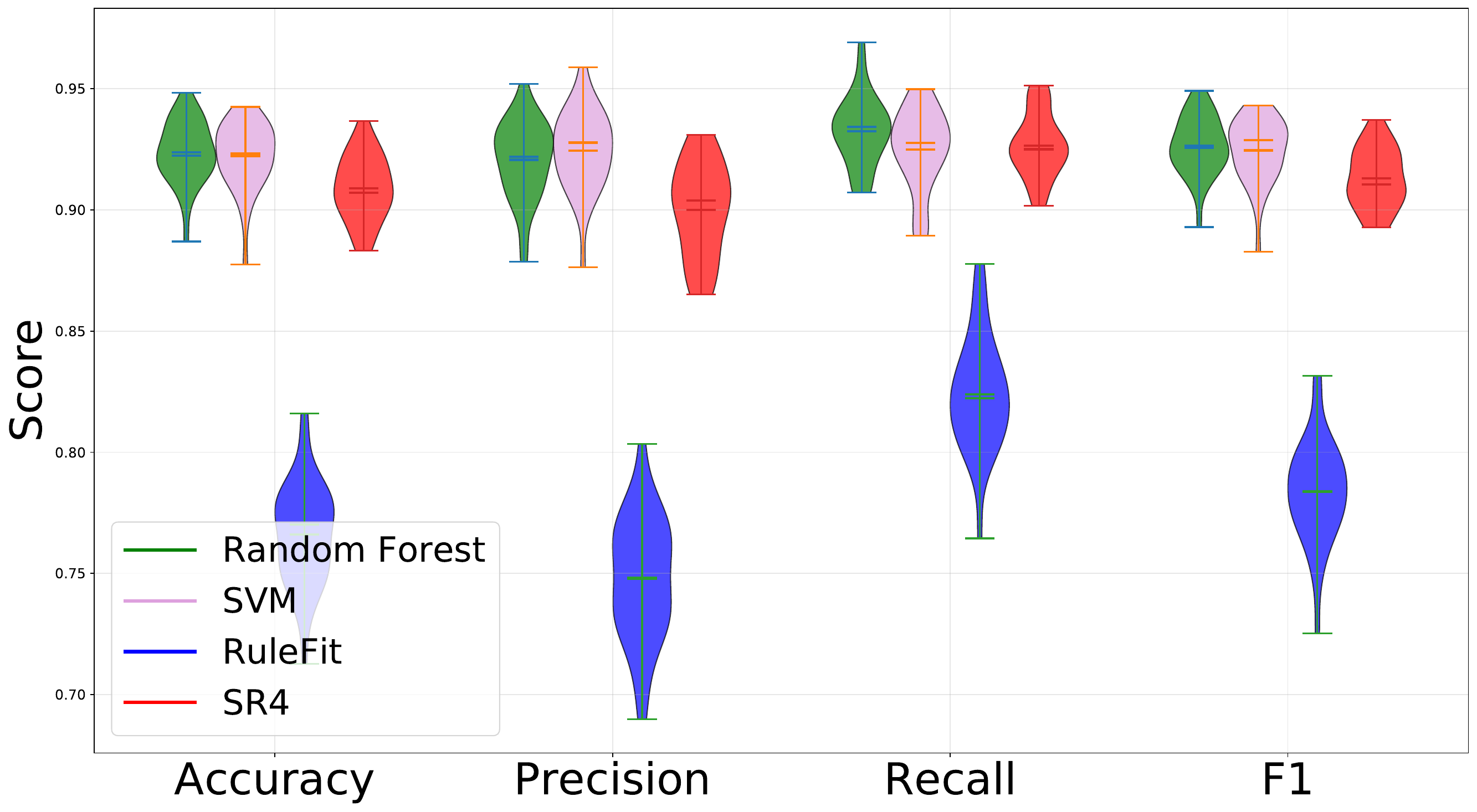}
    \subcaption{Previous}
\end{minipage}

\caption{
Violin plot comparison of model performance metrics across multiple datasets for the election classification task that does not include prior party voting percentage information.
}
\label{fig:violin_all_models_wpp}
\end{figure}

\begin{figure}
\centering
\begin{minipage}{0.48\linewidth}
    \centering
    \includegraphics[width=\linewidth]{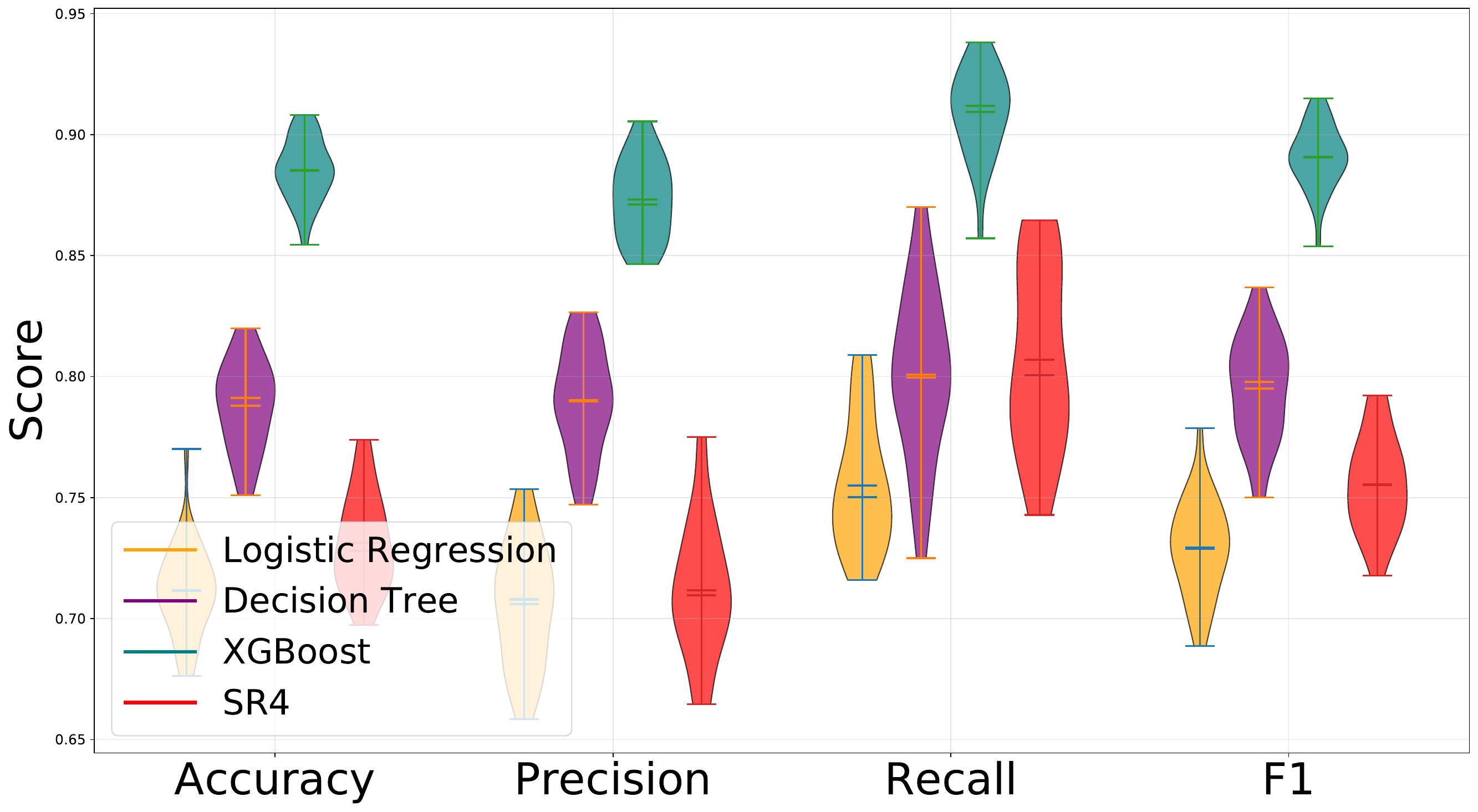}
    \subcaption{Minimum}
\end{minipage}\hfill
\begin{minipage}{0.48\linewidth}
    \centering
    \includegraphics[width=\linewidth]{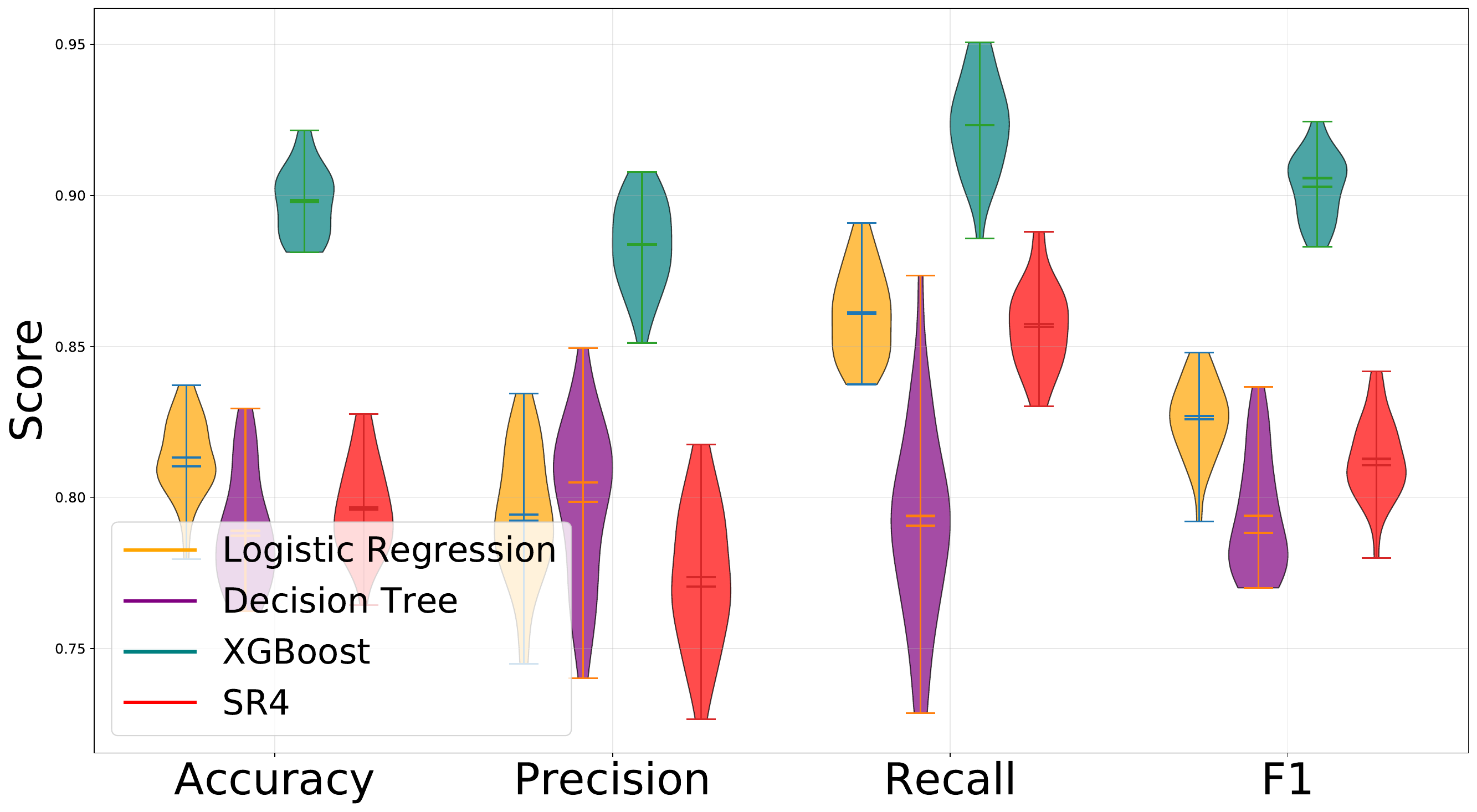}
    \subcaption{Standard}
\end{minipage}

\vspace{0.4cm}

\begin{minipage}{0.48\linewidth}
    \centering
    \includegraphics[width=\linewidth]{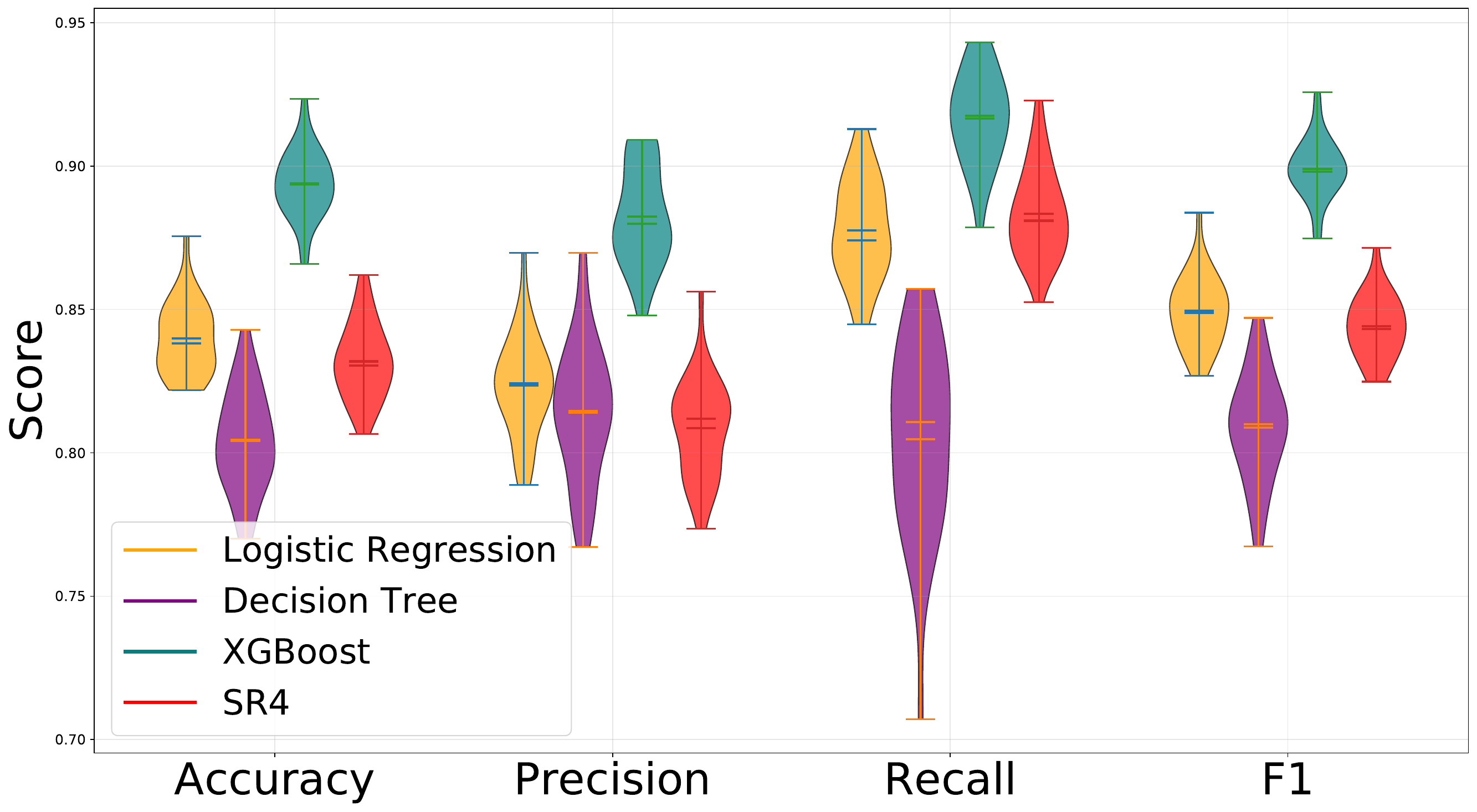}
    \subcaption{Expanded}
\end{minipage}\hfill
\begin{minipage}{0.48\linewidth}
    \centering
    \includegraphics[width=\linewidth]{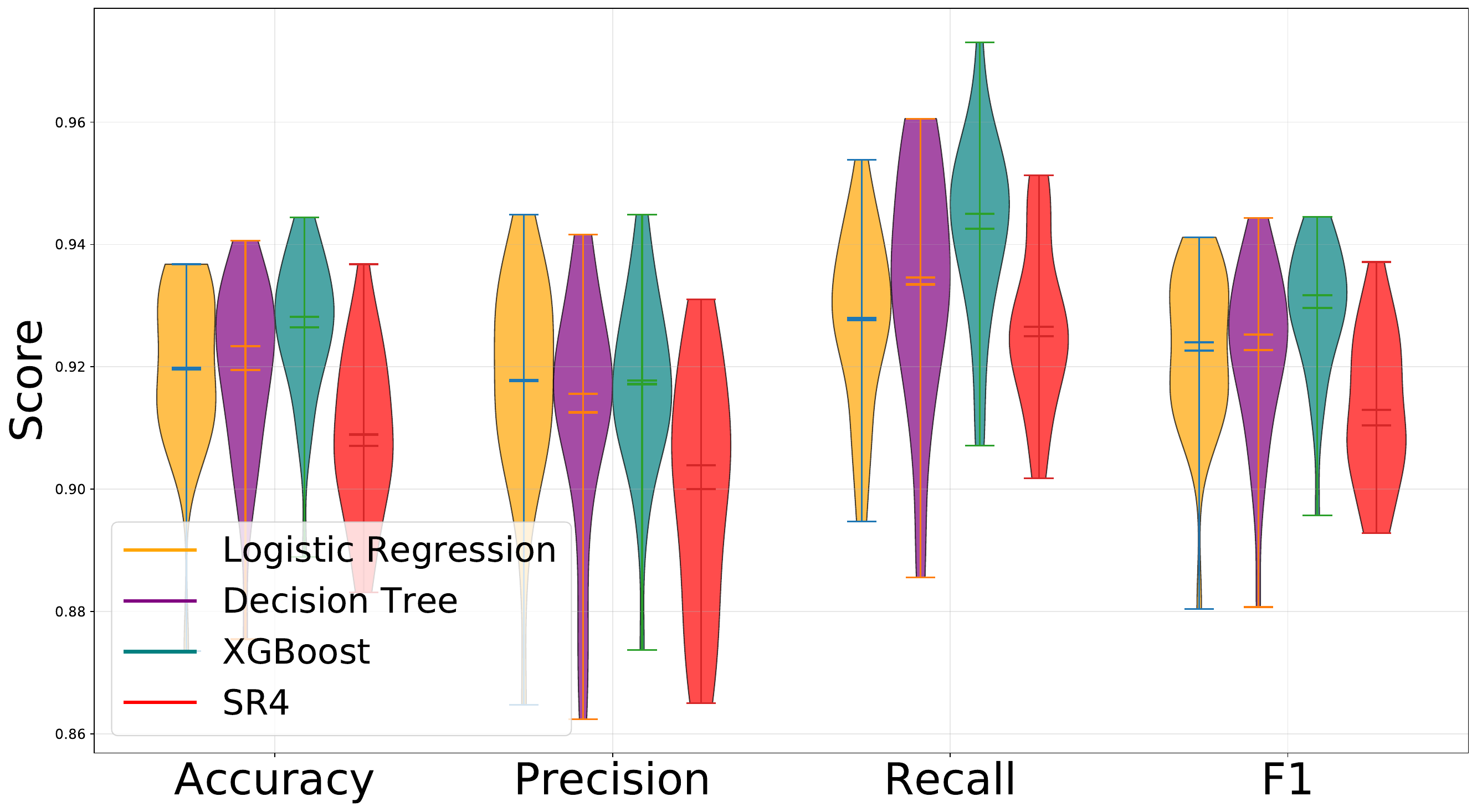}
    \subcaption{Previous}
\end{minipage}

\caption{
Violin plot comparison of model (logistic regression, Decision Tree, XG boost) performance metrics across multiple datasets for the election classification task that does not include prior party voting percentage information.
}
\label{fig:violin_all_models_wpp_other}
\end{figure}


\begin{table}
\caption{
Average Dice–Sørensen Index $\pm$ standard deviation across 30 trials for election classification datasets. Bold indicates the best-performing method per dataset.
}

\centering
\setlength{\tabcolsep}{3pt}
\renewcommand{\arraystretch}{1.15}

\begin{subtable}{\columnwidth}
\centering
\begin{tabular}{lcccc}
\toprule
\textbf{Dataset}
& \textbf{Random Forest} & \textbf{RuleFit} & \textbf{SR4-Fit} & \textbf{Decision Tree} \\
\midrule
Minimum  
& 0.0487$\pm$0.0107 & \textbf{0.4776$\pm$0.0162} & \textbf{0.4776$\pm$0.0162} & 0.0453$\pm$0.0337 \\
Standard 
& 0.0308$\pm$0.0087 & \textbf{0.7814$\pm$0.0111} & 0.7370$\pm$0.0110 & 0.0415$\pm$0.0293 \\
Expanded 
& 0.0378$\pm$0.0111 & \textbf{0.8540$\pm$0.0080} & \textbf{0.8540$\pm$0.0080} & 0.0411$\pm$0.0288 \\
Previous 
& 0.0587$\pm$0.0098 & 0.7437$\pm$0.0108 & \textbf{0.7873$\pm$0.0108} & 0.0471$\pm$0.0331 \\
\bottomrule
\end{tabular}
\caption{\textbf{With prior party voting percentage as a feature.}}
\end{subtable}

\vspace{6pt}

\begin{subtable}{\columnwidth}
\centering
\begin{tabular}{lcccc}
\toprule
\textbf{Dataset}
& \textbf{Random Forest} & \textbf{RuleFit} & \textbf{SR4-Fit} & \textbf{Decision Tree} \\
\midrule
Minimum  
& 0.0210$\pm$0.0073 & 0.0400$\pm$0.0021 & \textbf{0.2047$\pm$0.0065} & 0.0027$\pm$0.0017 \\
Standard 
& 0.0347$\pm$0.0145 & \textbf{0.3425$\pm$0.0006} & 0.2066$\pm$0.0004 & 0.0005$\pm$0.0005 \\
Expanded 
& 0.0094$\pm$0.0058 & 0.1815$\pm$0.0011 & \textbf{0.3056$\pm$0.0002} & 0.0009$\pm$0.0008 \\
Previous 
& 0.0618$\pm$0.0149 & \textbf{0.6459$\pm$0.0098} & \textbf{0.6459$\pm$0.0098} & 0.0314$\pm$0.0210 \\
\bottomrule
\end{tabular}
\caption{\textbf{Without prior party voting percentage as feature.}}
\end{subtable}

\label{tab:dice_with_vs_without_party}
\end{table}


\begin{table}
\caption{
Average number of rules $\pm$ standard deviation across 30 trials for election classification datasets. Bold indicates the most compact model.
}

\centering
\setlength{\tabcolsep}{3pt}
\renewcommand{\arraystretch}{1.15}

\begin{subtable}{\columnwidth}
\centering
\begin{tabular}{lcccc}
\toprule
\textbf{Dataset}
& \textbf{Random Forest} & \textbf{RuleFit} & \textbf{SR4-Fit} & \textbf{Decision Tree} \\
\midrule
Minimum  
& 86.43$\pm$8.70 & 19.33$\pm$1.06 & 19.33$\pm$1.06 & \textbf{9.97$\pm$0.76} \\
Standard 
& 68.00$\pm$5.19 & 36.13$\pm$0.94 & 38.30$\pm$0.99 & \textbf{10.03$\pm$0.85} \\
Expanded 
& 82.07$\pm$8.99 & 54.13$\pm$0.94 & 54.13$\pm$0.94 & \textbf{10.03$\pm$0.93} \\
Previous 
& 153.93$\pm$11.25 & 39.30$\pm$0.99 & 37.13$\pm$0.94 & \textbf{8.77$\pm$1.01} \\
\bottomrule
\end{tabular}
\caption{\textbf{With prior party voting percentage as a feature.}}
\end{subtable}

\vspace{6pt}

\begin{subtable}{\columnwidth}
\centering
\begin{tabular}{lcccc}
\toprule
\textbf{Dataset}
& \textbf{Random Forest} & \textbf{RuleFit} & \textbf{SR4-Fit} & \textbf{Decision Tree} \\
\midrule
Minimum  
& 89.70$\pm$5.16 & 177.80$\pm$13.47 & \textbf{34.80$\pm$1.54} & 253.87$\pm$15.9 \\
Standard 
& \textbf{32.50$\pm$7.23} & 76.00$\pm$0.00 & 126.00$\pm$0.00 & 104.00$\pm$3.301 \\
Expanded 
& \textbf{23.27$\pm$6.69} & 242.73$\pm$2.98 & 144.00$\pm$0.00 & 90.56$\pm$5.67 \\
Previous 
& 30.03$\pm$6.71 & 42.87$\pm$0.35 & 42.87$\pm$0.35 & \textbf{22.03$\pm$1.62} \\
\bottomrule
\end{tabular}
\caption{\textbf{Without prior party voting percentage as a feature.}}
\end{subtable}

\label{tab:rules_with_vs_without_party}
\end{table}


\begin{table}
\caption{
Average rule complexity (number of conditions per rule) $\pm$ standard deviation for election classification datasets across 30 trials.
Bold indicates less complex rules.
}

\centering
\setlength{\tabcolsep}{3pt}
\renewcommand{\arraystretch}{1.15}

\begin{subtable}{\columnwidth}
\centering
\begin{tabular}{lcccc}
\toprule
\textbf{Dataset}
& \textbf{Random Forest} & \textbf{RuleFit} & \textbf{SR4-Fit} & \textbf{Decision Tree} \\
\midrule
Minimum  
& \textbf{1.90$\pm$0.03} & 3.00$\pm$0.15 & 3.00$\pm$0.15 & 4.41$\pm$0.20 \\
Standard 
& 1.91$\pm$0.03 & \textbf{1.84$\pm$0.11} & 2.21$\pm$0.11 & 4.42$\pm$0.23 \\
Expanded 
& 1.90$\pm$0.02 & \textbf{1.60$\pm$0.09} & \textbf{1.60$\pm$0.09} & 4.44$\pm$0.30 \\
Previous 
& 1.92$\pm$0.02 & 2.19$\pm$0.11 & \textbf{1.83$\pm$0.11} & 4.03$\pm$0.20 \\
\bottomrule
\end{tabular}
\caption{\textbf{With prior party voting percentage as feature.}}
\end{subtable}

\vspace{6pt}

\begin{subtable}{\columnwidth}
\centering
\begin{tabular}{lcccc}
\toprule
\textbf{Dataset}
& \textbf{Random Forest} & \textbf{RuleFit} & \textbf{SR4-Fit} & \textbf{Decision Tree} \\
\midrule
Minimum  
& \textbf{1.99$\pm$0.01} & 8.11$\pm$0.14 & 4.43$\pm$0.07 & 10.95$\pm$0.19 \\
Standard 
& \textbf{1.97$\pm$0.03} & 5.61$\pm$0.32 & 6.79$\pm$0.10 & 8.36$\pm$0.31 \\
Expanded 
& \textbf{1.98$\pm$0.03} & 8.05$\pm$0.21 & 7.41$\pm$0.32 & 7.55$\pm$0.14 \\
Previous 
& \textbf{1.97$\pm$0.03} & 2.62$\pm$0.03 & 2.62$\pm$0.03 & 4.64$\pm$0.08 \\
\bottomrule
\end{tabular}
\caption{\textbf{Without prior party voting percentage as feature.}}
\end{subtable}

\label{tab:complexity_with_vs_without_party}
\end{table}


Across all configurations of the U.S. House of Representatives election classification task, SR4-Fit consistently emerges as the best combination of a stable, accurate, and compact model, even when compared with all the baselines. The combined evidence from line plots, violin plots, stability tables, and rule-complexity analyses provides a comprehensive view of model behavior both with and without the party-percentage features.

In the minimum dataset, the line plots in Figure~\ref{fig:lineplot_grid_min} and Figure~\ref{fig:lineplot_grid_min_other} show that when prior party voting percentage is included, all models achieve high performance; however, SR4-Fit exhibits the smoothest trajectories across accuracy, precision, recall, and F1 with small variance, outperforming logistic regression, Decision Tree, and XGBoost. The distributional comparisons through violin plots in Figure~\ref{fig:violin_all_models} and Figure~\ref{fig:violin_all_models_other} confirm that SR4-Fit produces tighter metric distributions than other models, which display broader spreads and heavier fluctuations. When prior party voting percentage is removed, performance declines across models, with XGBoost maintaining the highest predictive scores, while SR4-Fit maintains comparatively smoother trends despite lower accuracy, which is further confirmed through the violin plots in Figure~\ref{fig:violin_all_models_wpp} and Figure~\ref{fig:violin_all_models_wpp_other} where the distributions are tighter for SR4-Fit despite lower accuracy. Stability analysis in Table~\ref{tab:dice_with_vs_without_party} shows that with and without prior party voting percentages, SR4-Fit achieves the highest Dice–Sørensen Index when compared with all rule-based models, showing its structural consistency and robustness. Table~\ref{tab:rules_with_vs_without_party} shows that in the case of prior party voting percentages as a feature, the Decision Tree has the most compact structure, which is preceded by both SR4-Fit and RuleFit, whereas without including prior party voting percentages, SR4-Fit has a more compact structure with respect to the number of rules. Table~\ref{tab:complexity_with_vs_without_party} shows that Random Forest has lower average rule complexity compared to all the other models.

In the standard dataset, the line plots in Figure~\ref{fig:lineplot_grid_std} and Figure~\ref{fig:lineplot_grid_std_other} show that when prior party voting percentage is included, SR4-Fit exhibits the smoothest trajectories with small variance, outperforming all models. The distributional comparisons through violin plots in Figure~\ref{fig:violin_all_models} and Figure~\ref{fig:violin_all_models_other} confirm that SR4-Fit produces tighter metric distributions than other models. When the prior party voting percentages are removed, performance declines across models, with SVM maintaining the highest predictive scores. Stability analysis in Table~\ref{tab:dice_with_vs_without_party} shows that with and without prior party voting percentages, RuleFit achieves the highest Dice–Sørensen Index when compared with all rule-based models, which is closely preceded by SR4-Fit. Table~\ref{tab:rules_with_vs_without_party} shows that in the case of prior party voting percentages as a feature, the Decision Tree has the most compact structure, whereas without including prior party voting percentages, Random Forest has a more compact structure with respect to the number of rules. Table~\ref{tab:complexity_with_vs_without_party} shows that Random Forest has lower average rule complexity when prior party voting percentages are not included, whereas RuleFit has lower average rule complexity when prior party voting percentages are included.

In the expanded dataset, the line plots in Figure~\ref{fig:lineplot_grid_exp} and Figure~\ref{fig:lineplot_grid_exp_other} show that when prior party voting percentage is included,  SR4-Fit exhibits the smoothest trajectories with small variance, outperforming all models. When the prior party voting percentages are removed, performance declines across models, with SVM maintaining the highest predictive scores. Stability analysis in Table~\ref{tab:dice_with_vs_without_party} shows that with and without prior party voting percentages, SR4-Fit achieves the highest Dice–Sørensen Index when compared with other rule-based models. Table~\ref{tab:rules_with_vs_without_party} shows that in the case of prior party voting percentages as a feature, the Decision Tree has the most compact structure, whereas without including prior party voting percentages, Random Forest has a more compact structure with respect to the number of rules. Table~\ref{tab:complexity_with_vs_without_party} shows that Random Forest has lower average rule complexity when prior party voting percentages are not included, whereas both SR4-Fit and RuleFit have lower average rule complexity when prior party voting percentages are included.

In the previous dataset, the line plots in Figure~\ref{fig:lineplot_grid_pre} and Figure~\ref{fig:lineplot_grid_pre_other} show that when prior party voting percentage is included,  SR4-Fit exhibits the smoothest trajectories with small variance, outperforming all models. When the prior party voting percentages are removed, performance declines across models, with Random Forest maintaining the highest predictive scores. Stability analysis in Table~\ref{tab:dice_with_vs_without_party} shows that with and without prior party voting percentages, SR4-Fit achieves the highest Dice–Sørensen Index when compared with other rule-based models. Table~\ref{tab:rules_with_vs_without_party} shows that in both cases, the Decision Tree has the most compact structure. Table~\ref{tab:complexity_with_vs_without_party} shows that Random Forest has lower average rule complexity when prior party voting percentages are not included, whereas SR4-Fit has lower average rule complexity when prior party voting percentages are included.

\FloatBarrier

\section{Results of U.S House of Representative Elections Regression (DEM)}\label{sup_sec_HouseRD}
\begin{figure}
\centering
\begin{minipage}{0.49\linewidth}
    \centering
    \includegraphics[width=\linewidth]{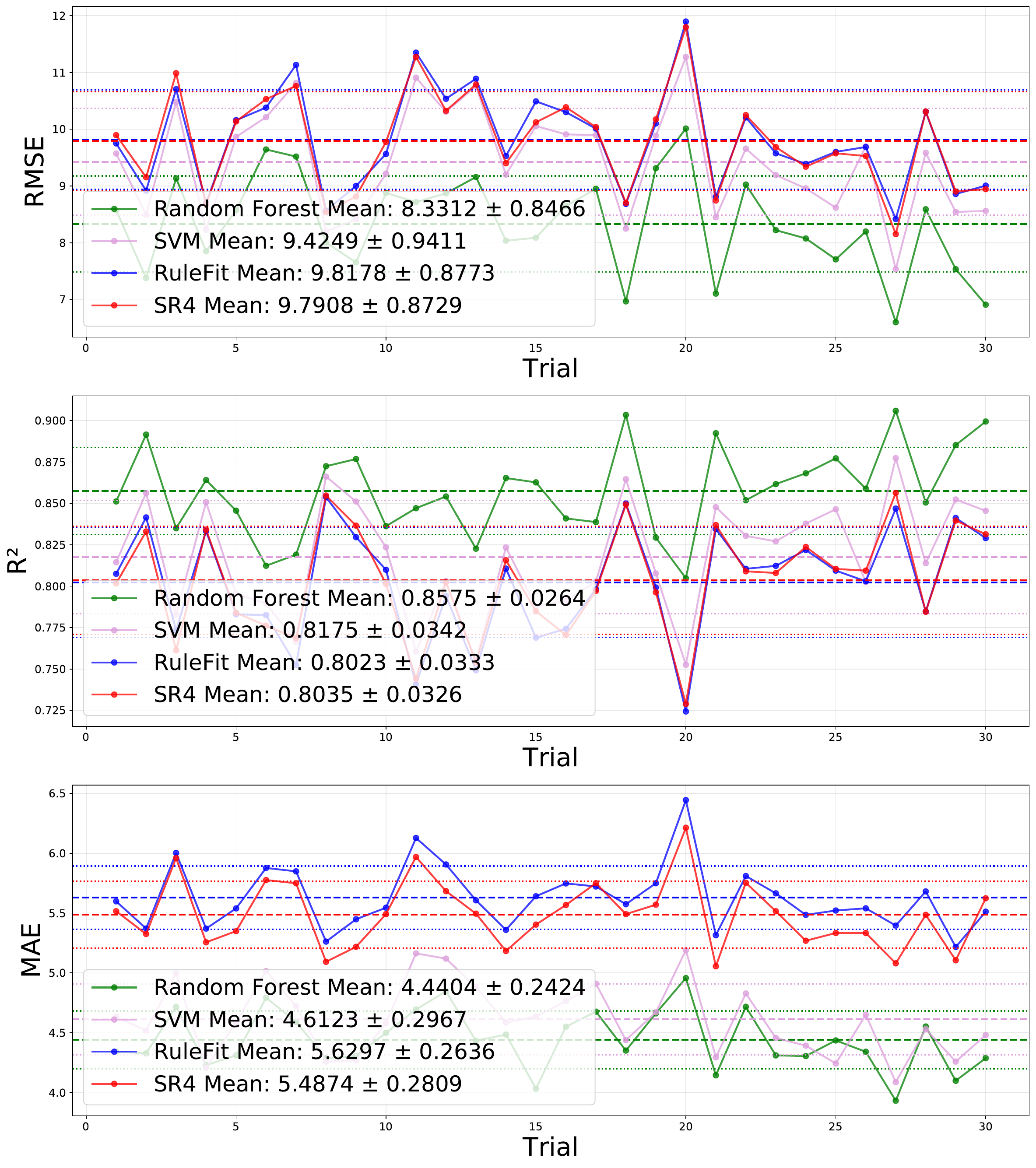}
\end{minipage}\hfill
\begin{minipage}{0.49\linewidth}
    \centering
    \includegraphics[width=\linewidth]{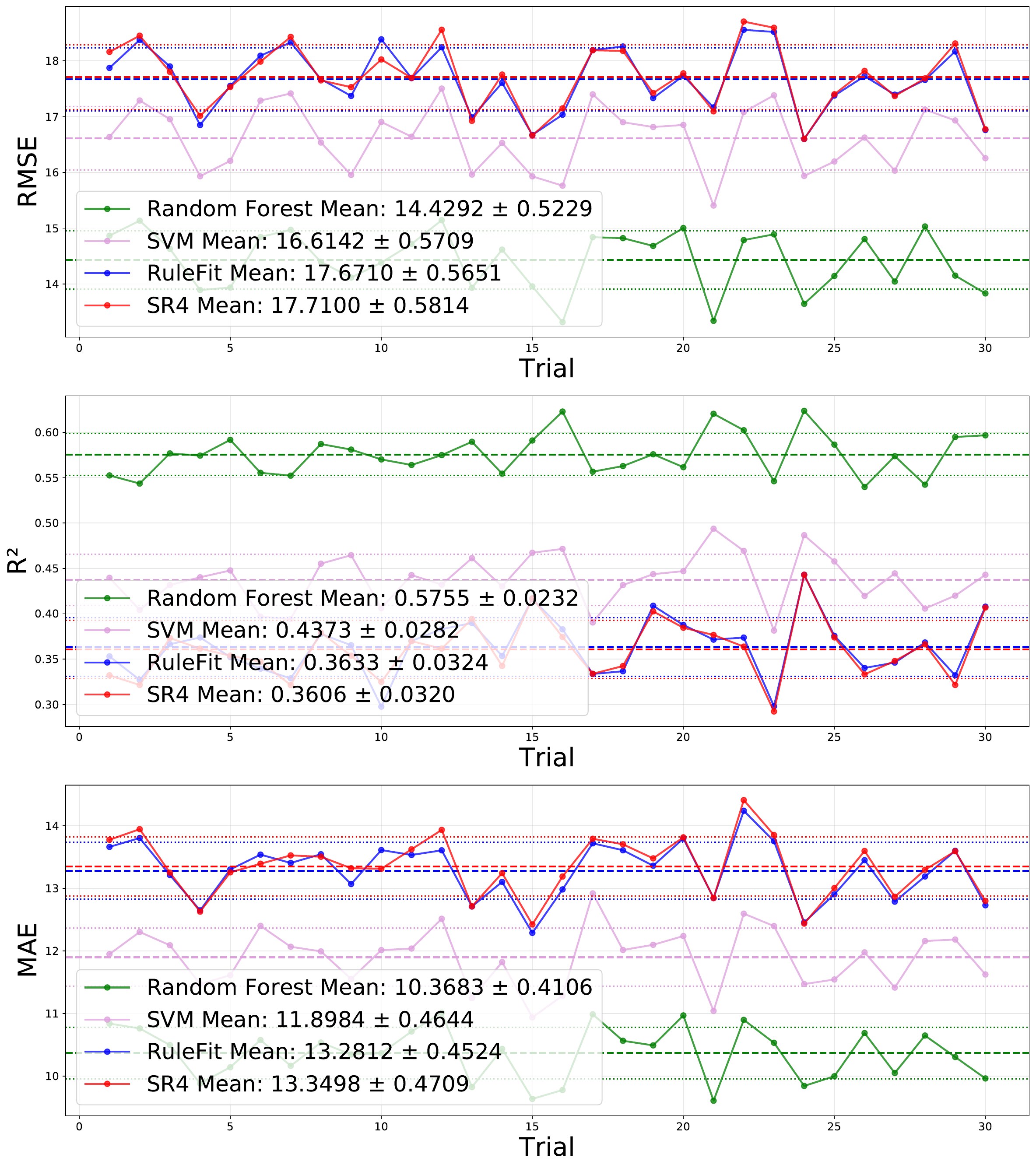}
\end{minipage}

\caption{
Line plot comparison of model performance metrics for minimum data across 30 trials with DEM percentage as label. The left panel reports results obtained with party percentage (REP) included, whereas the right panel shows results with party percentage (REP) not included.
}
\label{fig:lineplot_grid_min_dem}
\end{figure}

\begin{figure}
\centering
\begin{minipage}{0.49\linewidth}
    \centering
    \includegraphics[width=\linewidth]{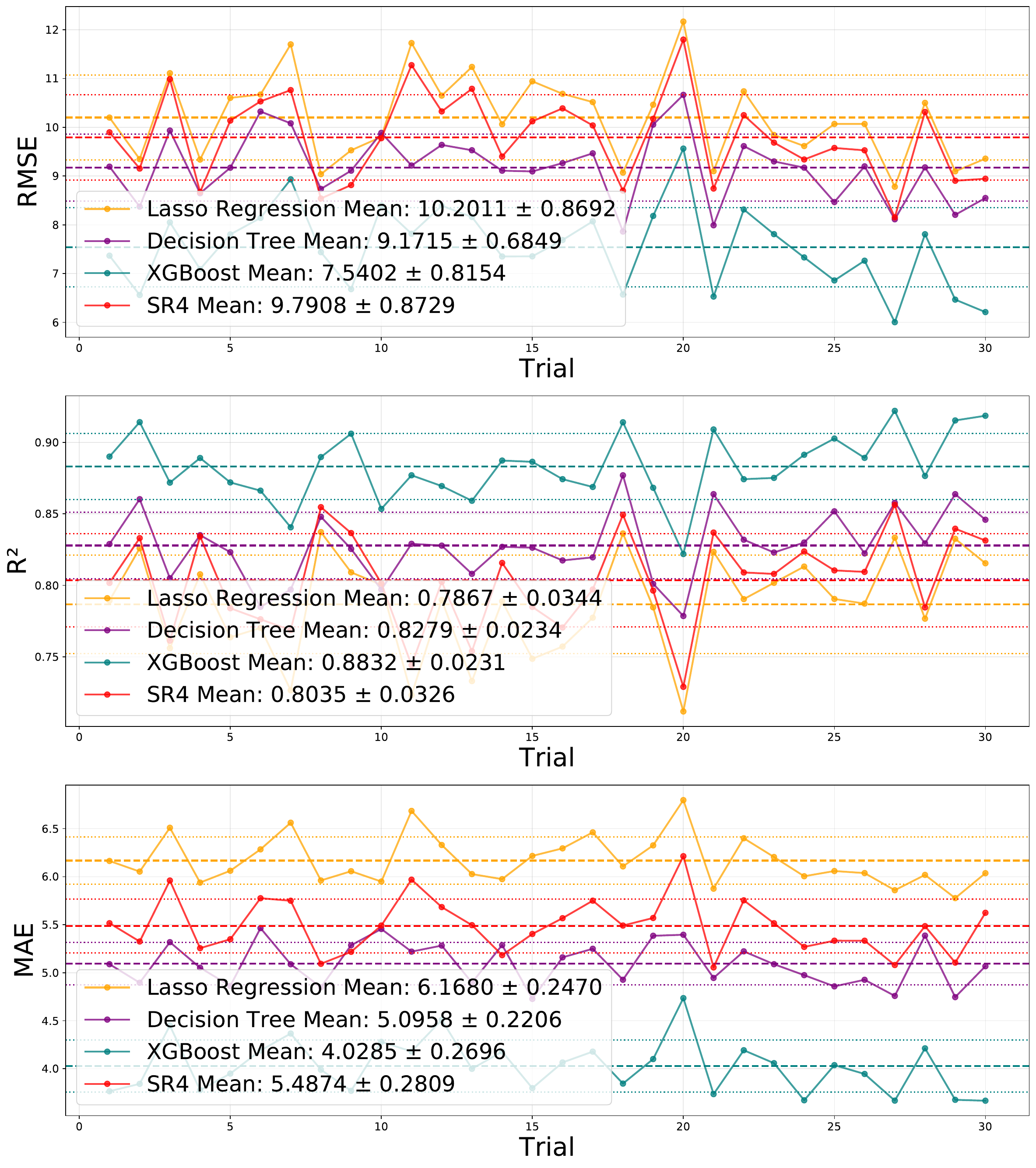}
\end{minipage}\hfill
\begin{minipage}{0.49\linewidth}
    \centering
    \includegraphics[width=\linewidth]{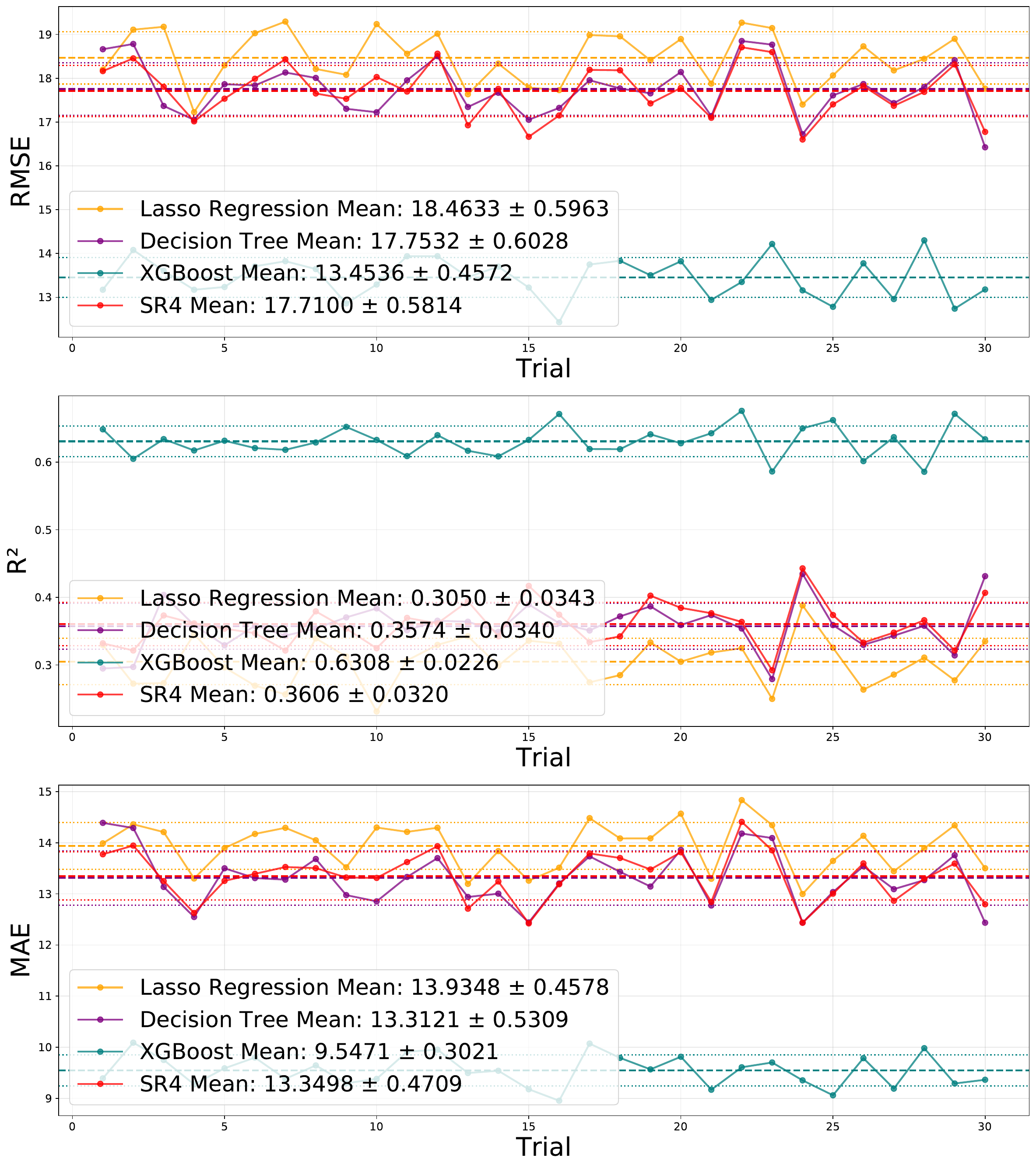}
\end{minipage}

\caption{
Line plot comparison of model (LASSO regression, Decision Tree, XG boost) performance metrics for minimum data across 30 trials. The left panel reports results obtained with party percentage (REP) included, whereas the right panel shows results with party percentage (REP) not included.
}
\label{fig:lineplot_grid_min_dem_other}
\end{figure}

\begin{figure}
\centering
\begin{minipage}{0.49\linewidth}
    \centering
    \includegraphics[width=\linewidth]{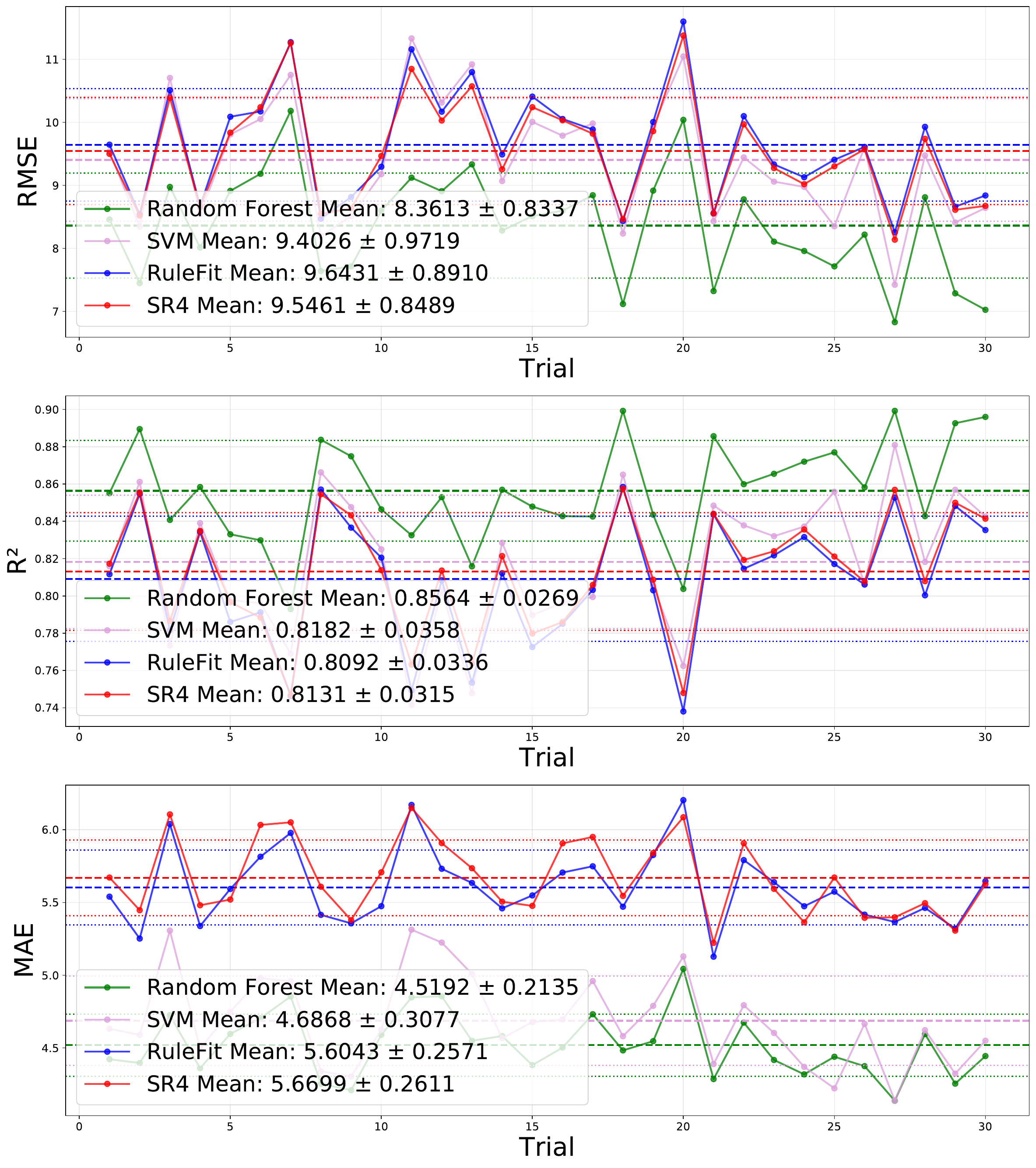}
\end{minipage}\hfill
\begin{minipage}{0.49\linewidth}
    \centering
    \includegraphics[width=\linewidth]{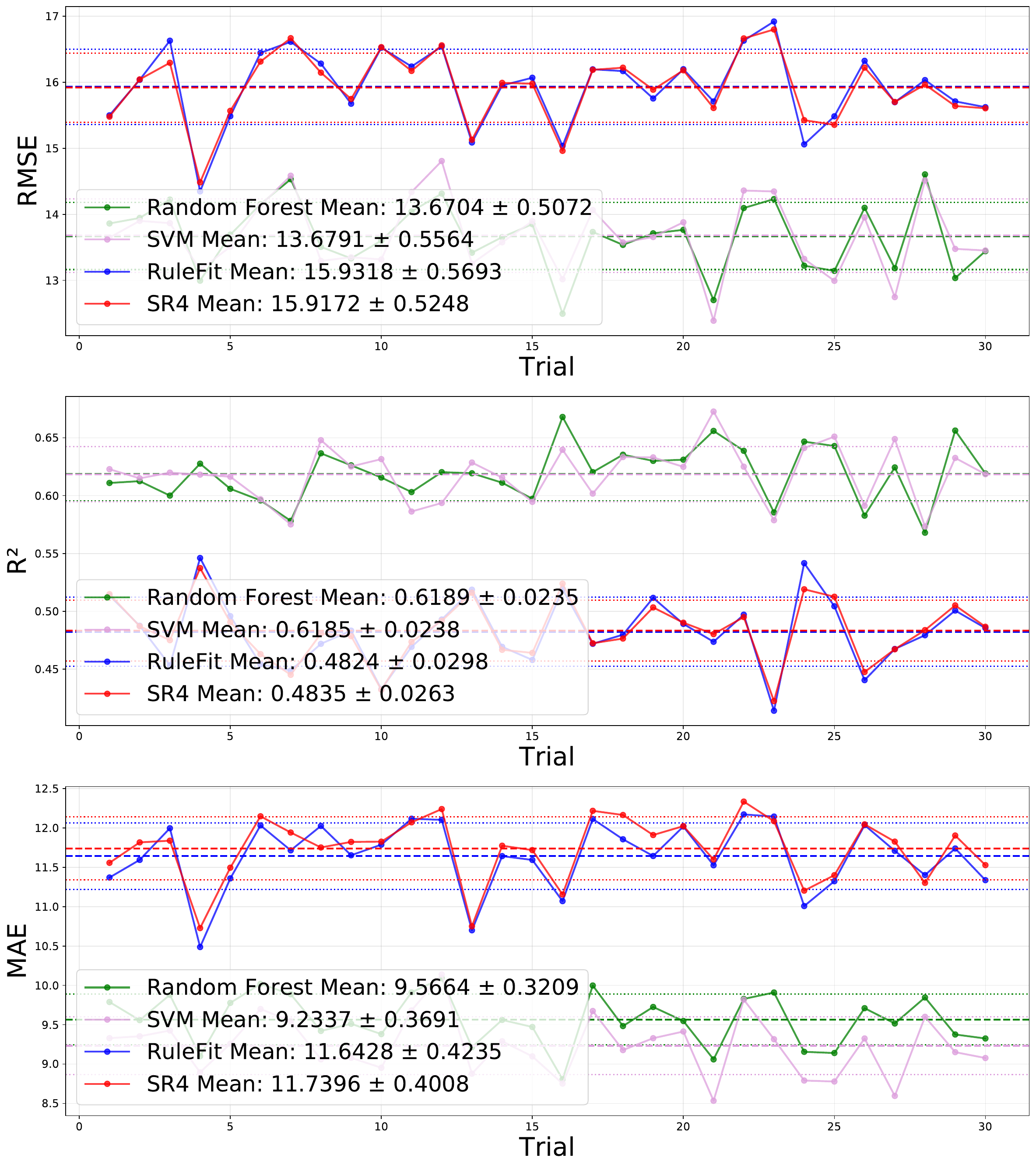}
\end{minipage}

\caption{
Line plot comparison of model performance metrics for standard data across 30 trials with DEM percentage as label. The left panel reports results obtained with party percentage (REP) included, whereas the right panel shows results with party percentage (REP) not included.
}
\label{fig:lineplot_grid_std_dem}
\end{figure}

\begin{figure}
\centering
\begin{minipage}{0.49\linewidth}
    \centering
    \includegraphics[width=\linewidth]{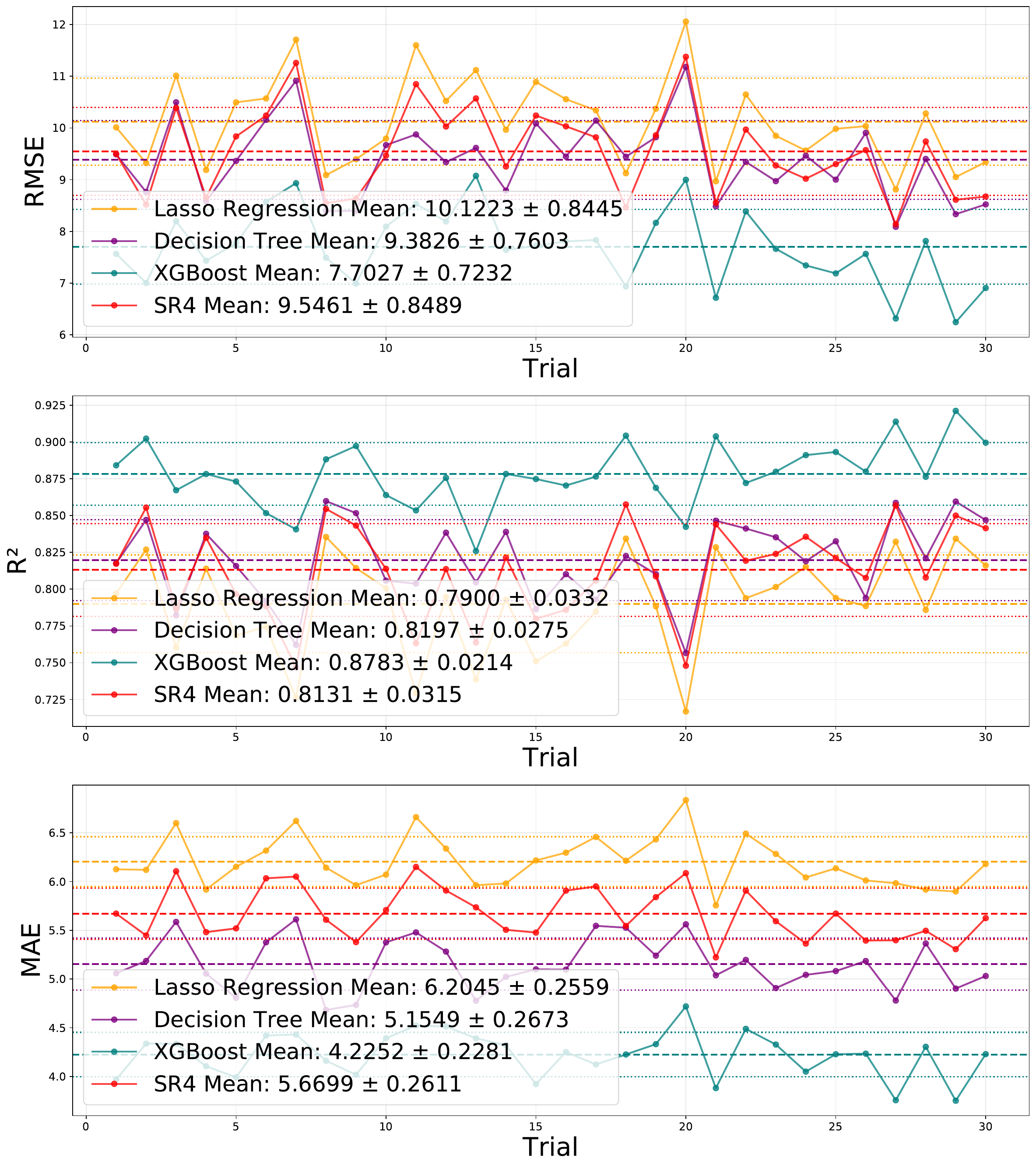}
\end{minipage}\hfill
\begin{minipage}{0.49\linewidth}
    \centering
    \includegraphics[width=\linewidth]{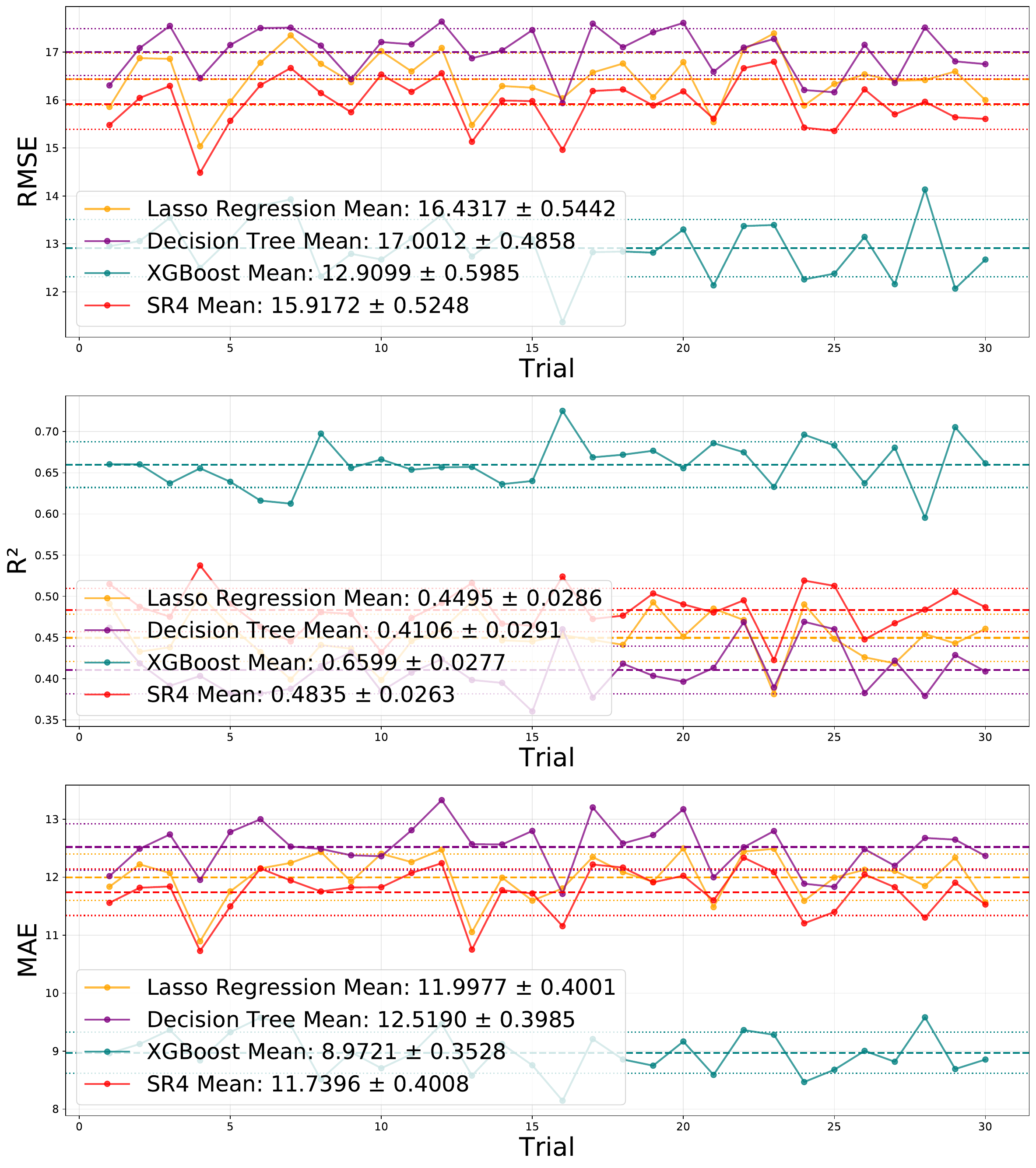}
\end{minipage}

\caption{
Line plot comparison of model (LASSO regression, Decision Tree, XG boost) performance metrics for standard data across 30 trials. The left panel reports results obtained with party percentage (REP) included, whereas the right panel shows results with party percentage (REP) not included.
}
\label{fig:lineplot_grid_std_dem_other}
\end{figure}

\begin{figure}
\centering
\begin{minipage}{0.49\linewidth}
    \centering
    \includegraphics[width=\linewidth]{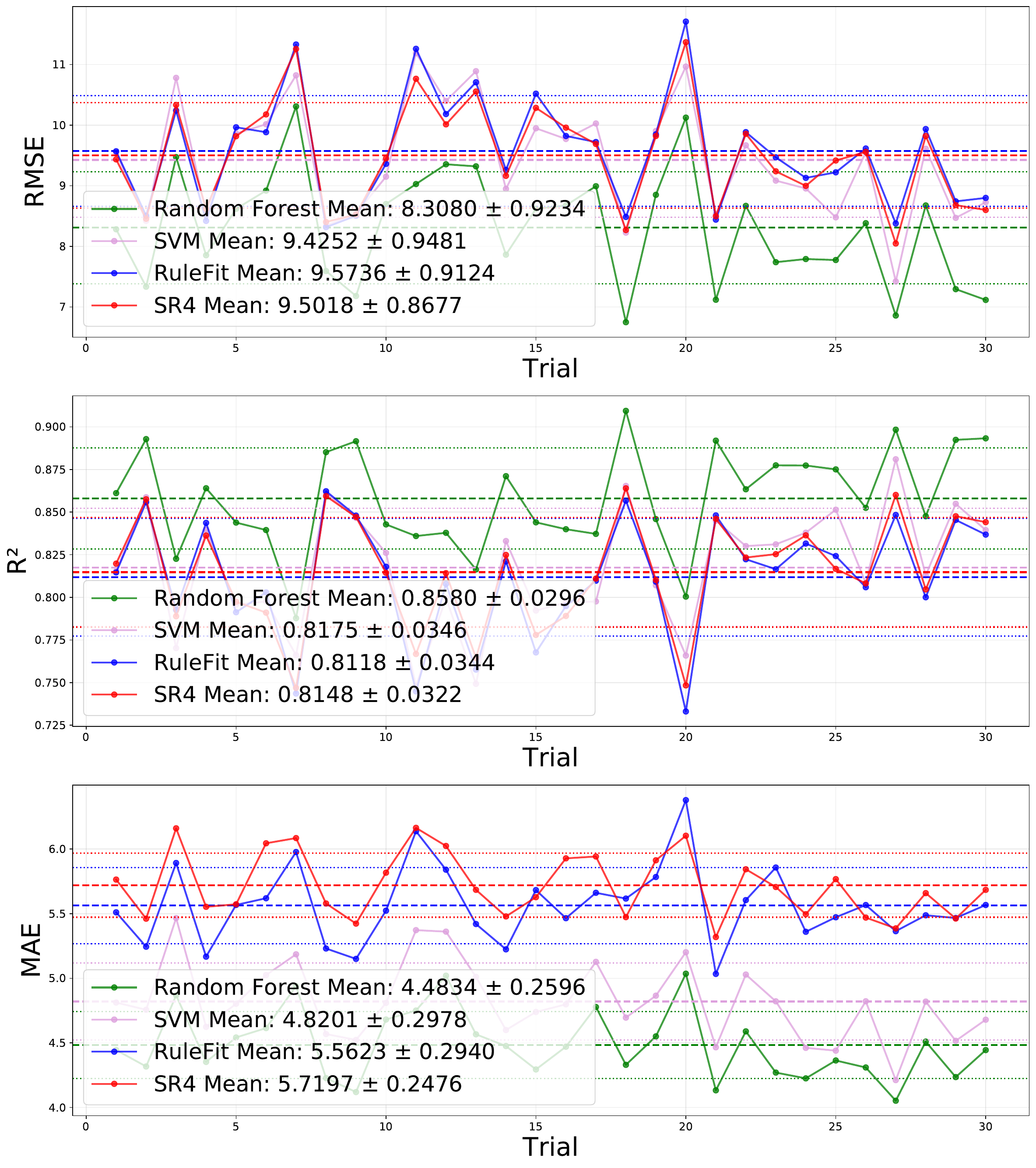}
\end{minipage}\hfill
\begin{minipage}{0.49\linewidth}
    \centering
    \includegraphics[width=\linewidth]{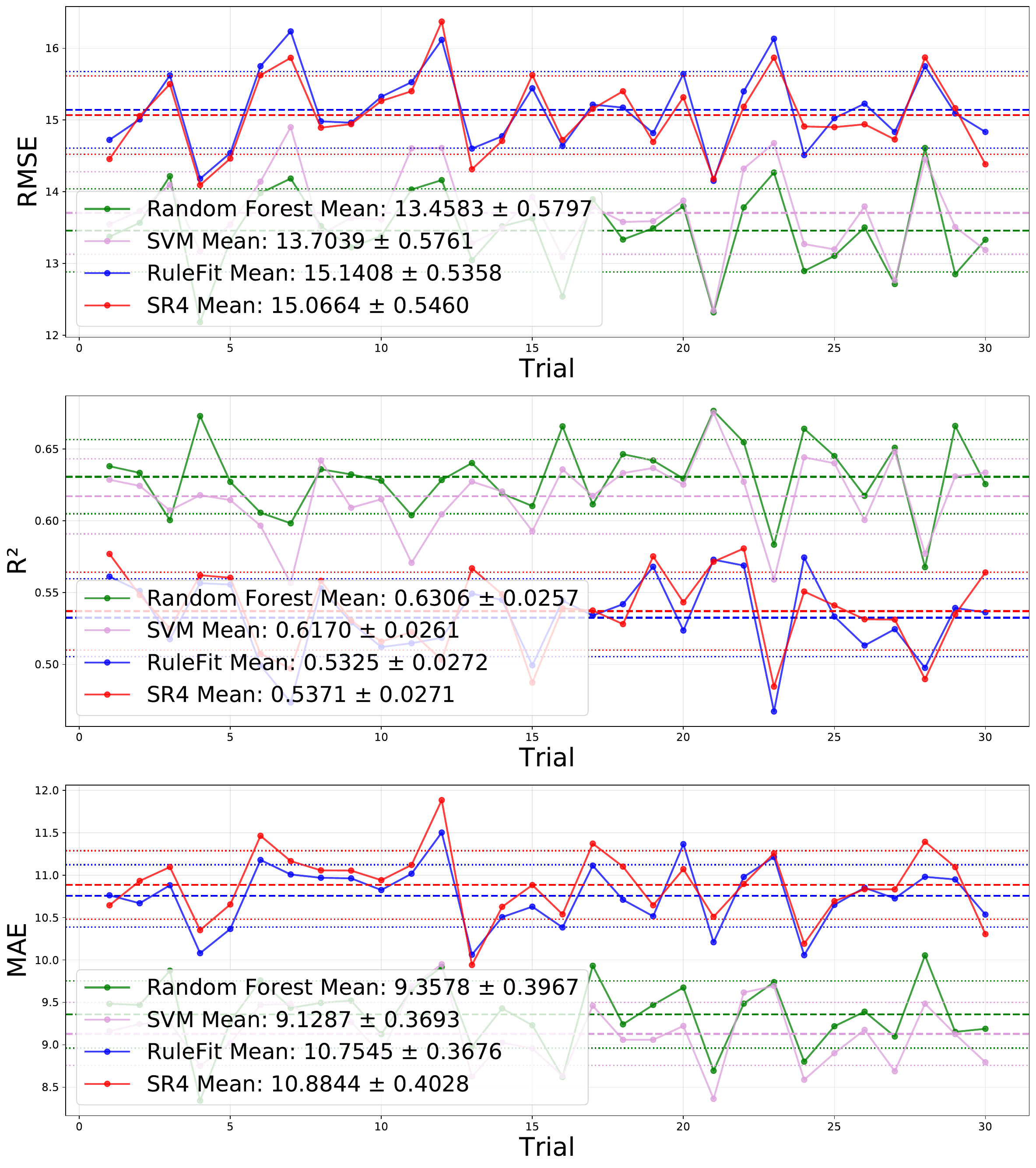}
\end{minipage}

\caption{
Line plot comparison of model performance metrics for expanded data across 30 trials with DEM percentage as label. The left panel reports results obtained with party percentage (REP) included, whereas the right panel shows results with party percentage (REP) not included.
}
\label{fig:lineplot_grid_exp_dem}
\end{figure}

\begin{figure}
\centering
\begin{minipage}{0.49\linewidth}
    \centering
    \includegraphics[width=\linewidth]{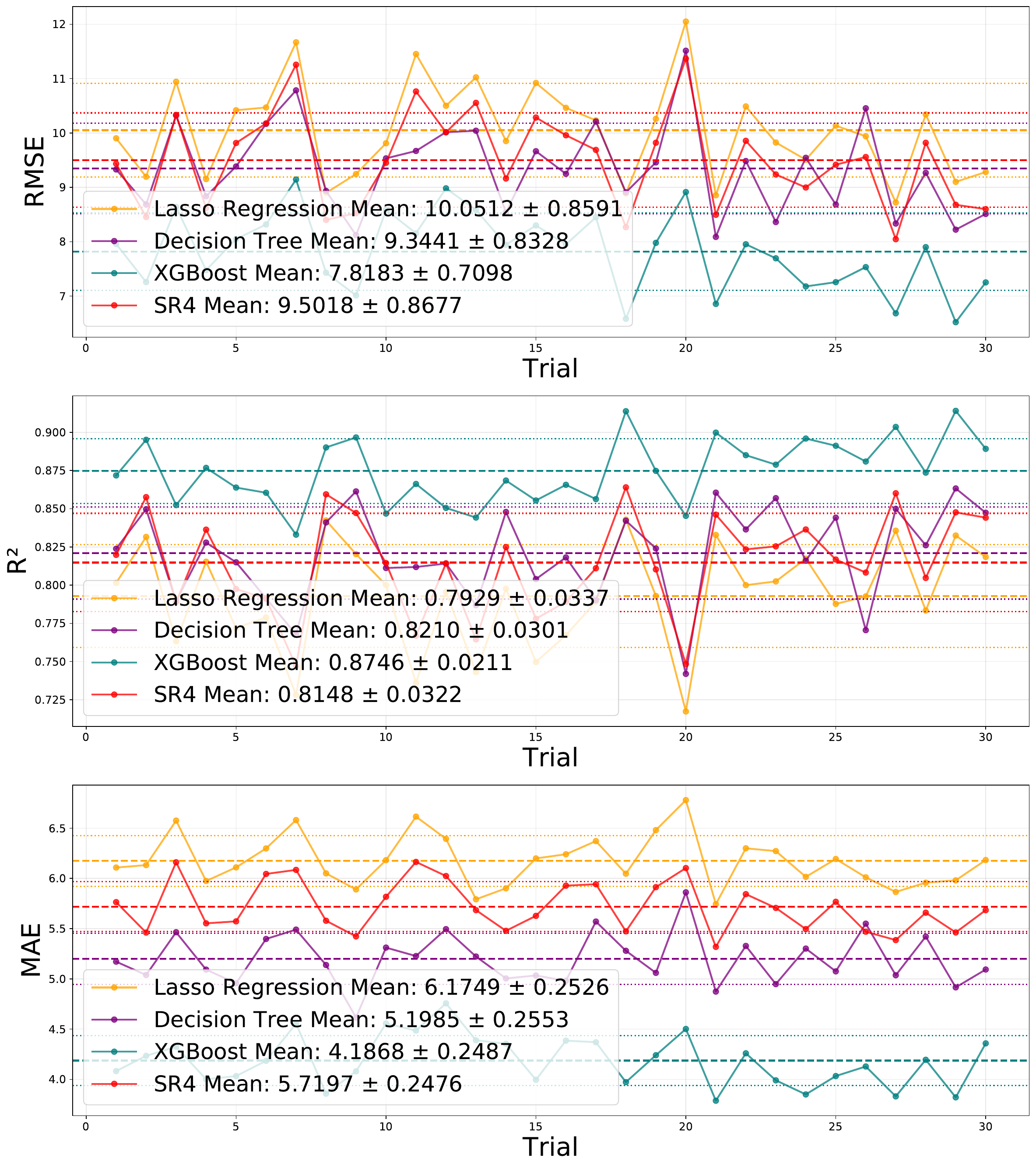}
\end{minipage}\hfill
\begin{minipage}{0.49\linewidth}
    \centering
    \includegraphics[width=\linewidth]{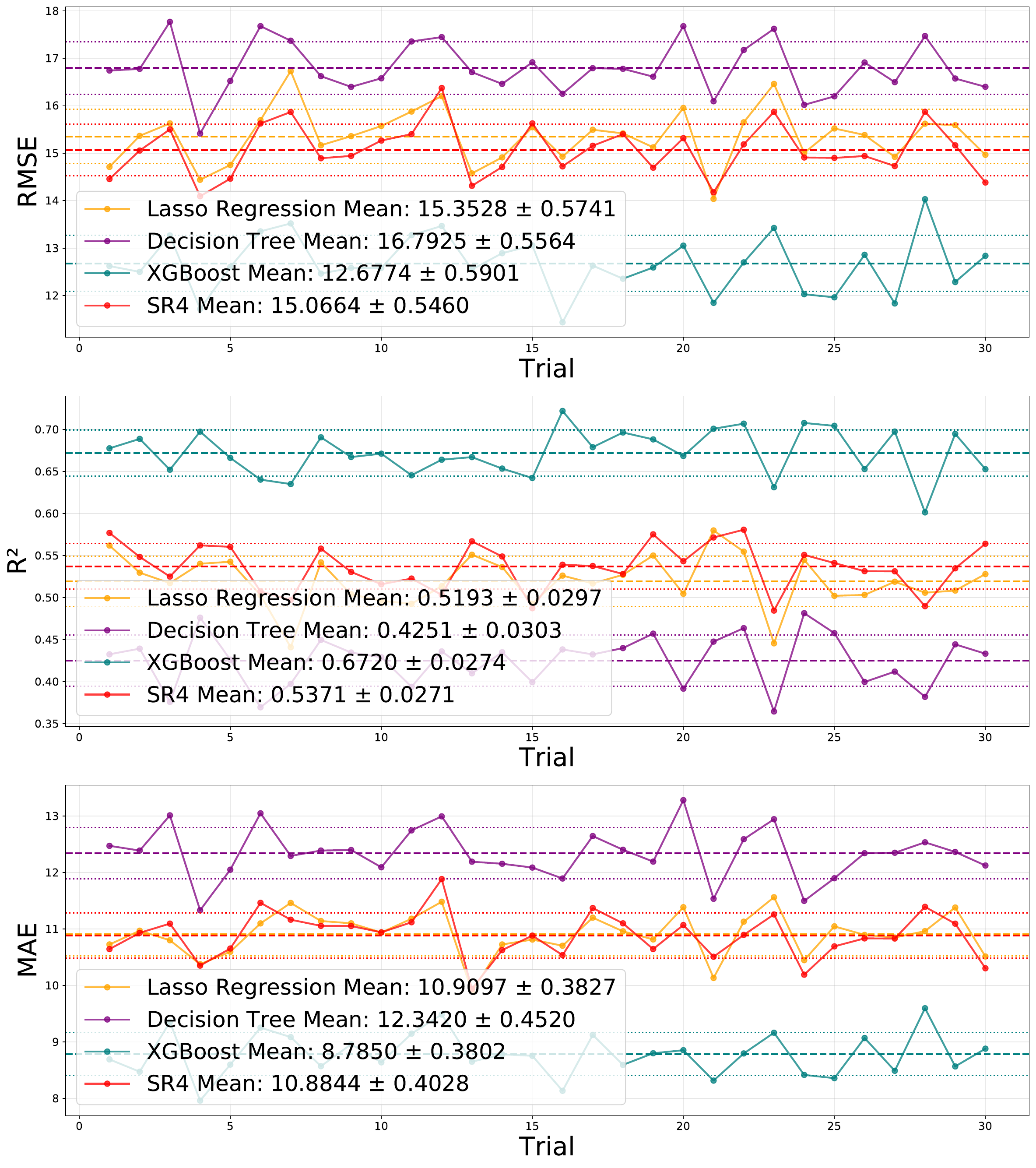}
\end{minipage}

\caption{
Line plot comparison of model (LASSO regression, Decision Tree, XG boost) performance metrics for expanded data across 30 trials. The left panel reports results obtained with party percentage (REP) included, whereas the right panel shows results with party percentage (REP) not included.
}
\label{fig:lineplot_grid_exp_dem_other}
\end{figure}

\begin{figure}
\centering
\begin{minipage}{0.49\linewidth}
    \centering
    \includegraphics[width=\linewidth]{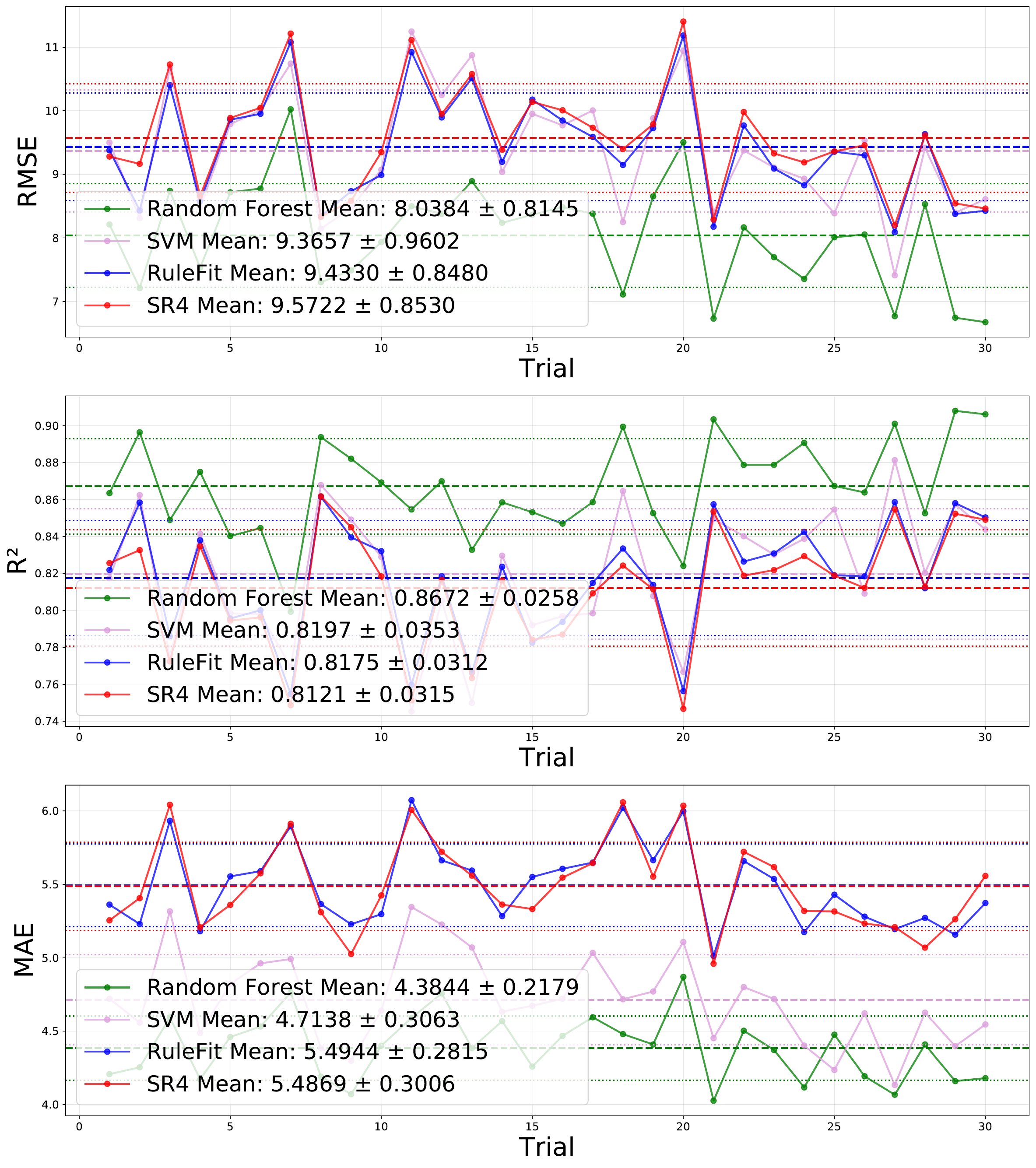}
\end{minipage}\hfill
\begin{minipage}{0.49\linewidth}
    \centering
    \includegraphics[width=\linewidth]{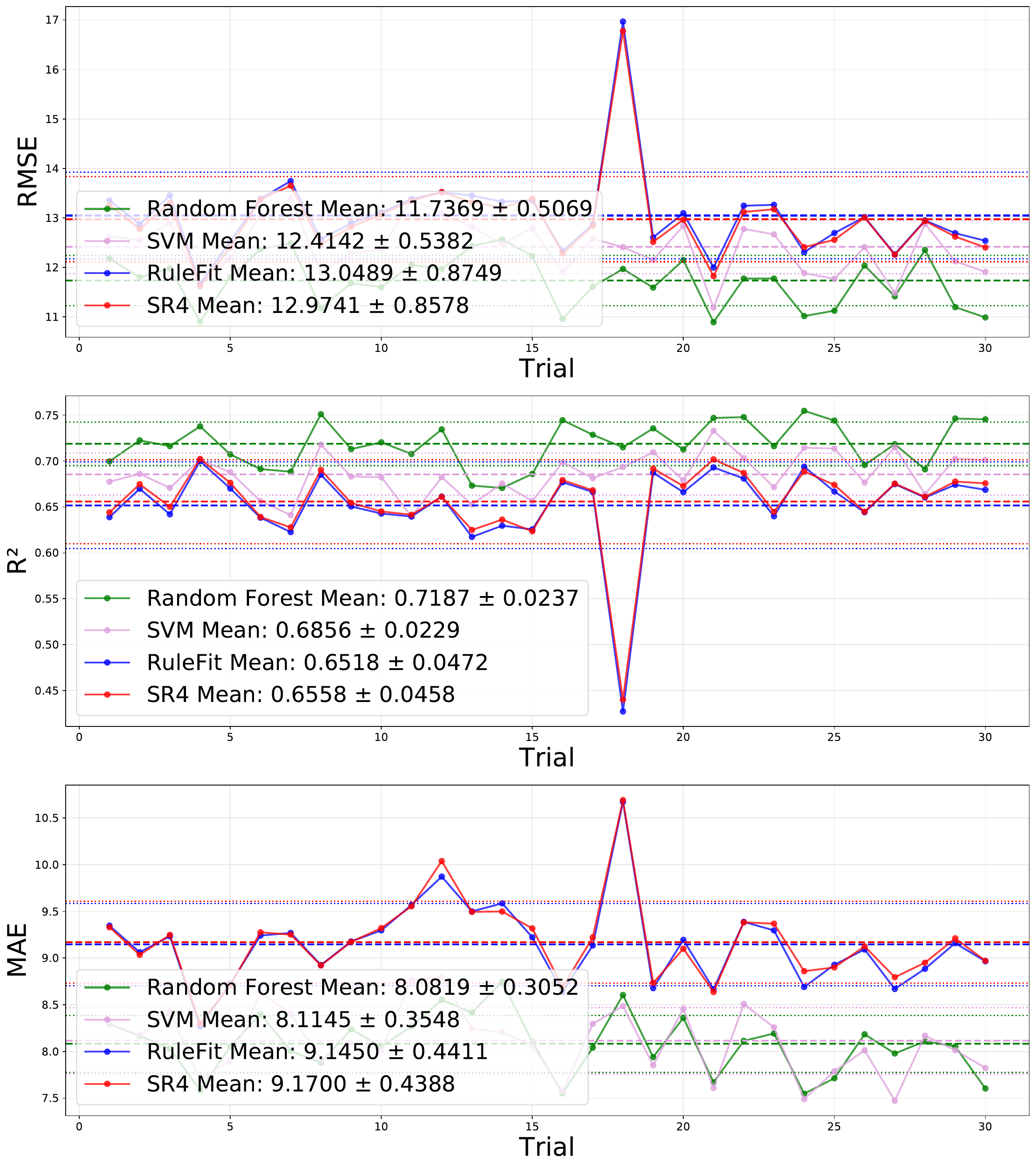}
\end{minipage}

\caption{
Line plot comparison of model performance metrics for previous data across 30 trials with DEM percentage as label. The left panel reports results obtained with party percentage (REP) included, whereas the right panel shows results with party percentage (REP) not included.
}
\label{fig:lineplot_grid_pre_dem}
\end{figure}

\begin{figure}
\centering
\begin{minipage}{0.49\linewidth}
    \centering
    \includegraphics[width=\linewidth]{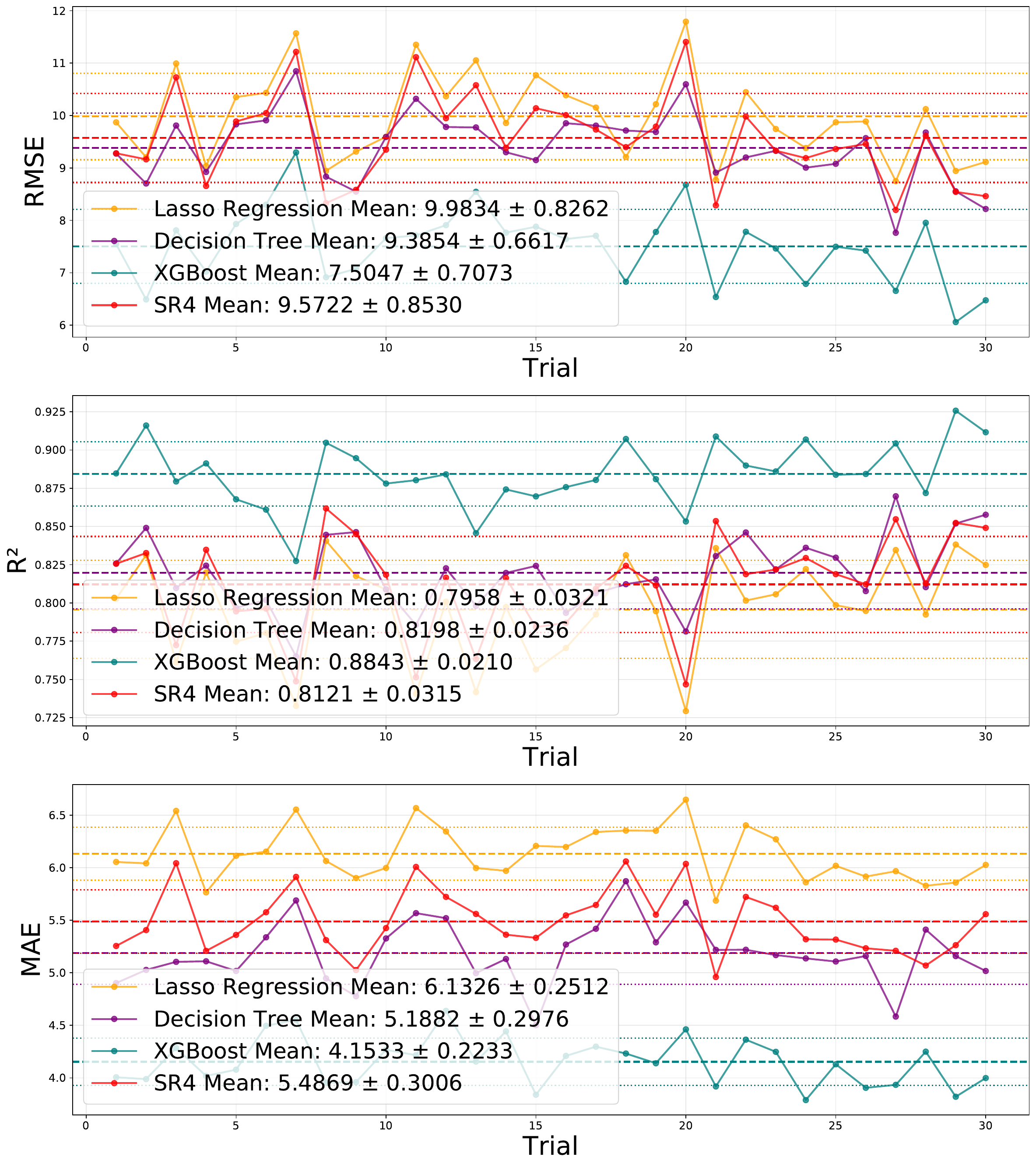}
\end{minipage}\hfill
\begin{minipage}{0.49\linewidth}
    \centering
    \includegraphics[width=\linewidth]{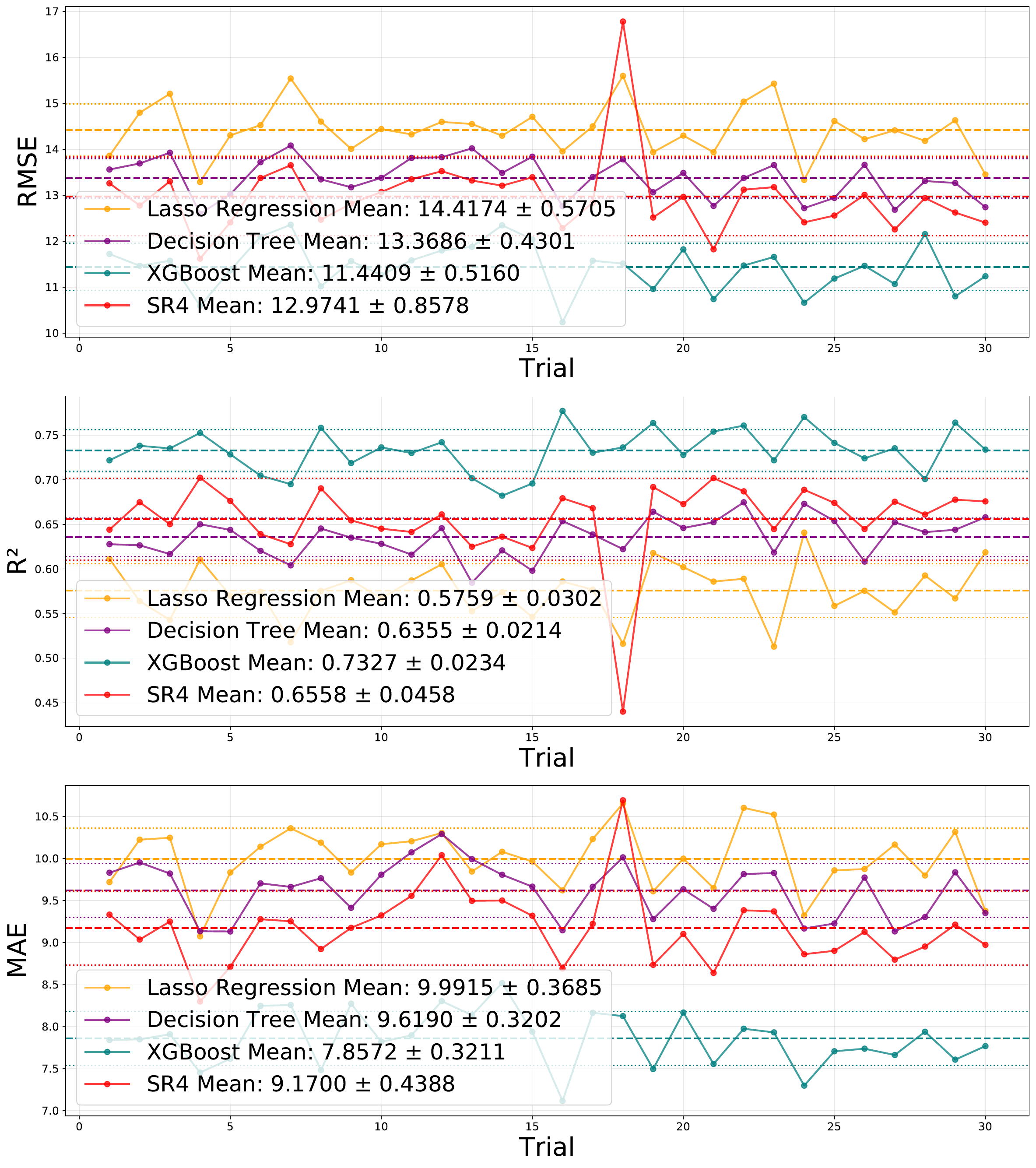}
\end{minipage}

\caption{
Line plot comparison of model (LASSO regression, Decision Tree, XG boost) performance metrics for previous data across 30 trials. The left panel reports results obtained with party percentage (REP) included, whereas the right panel shows results with party percentage (REP) not included.
}
\label{fig:lineplot_grid_pre_dem_other}
\end{figure}


\begin{figure}
\centering
\begin{minipage}{0.48\linewidth}
    \centering
    \includegraphics[width=\linewidth]{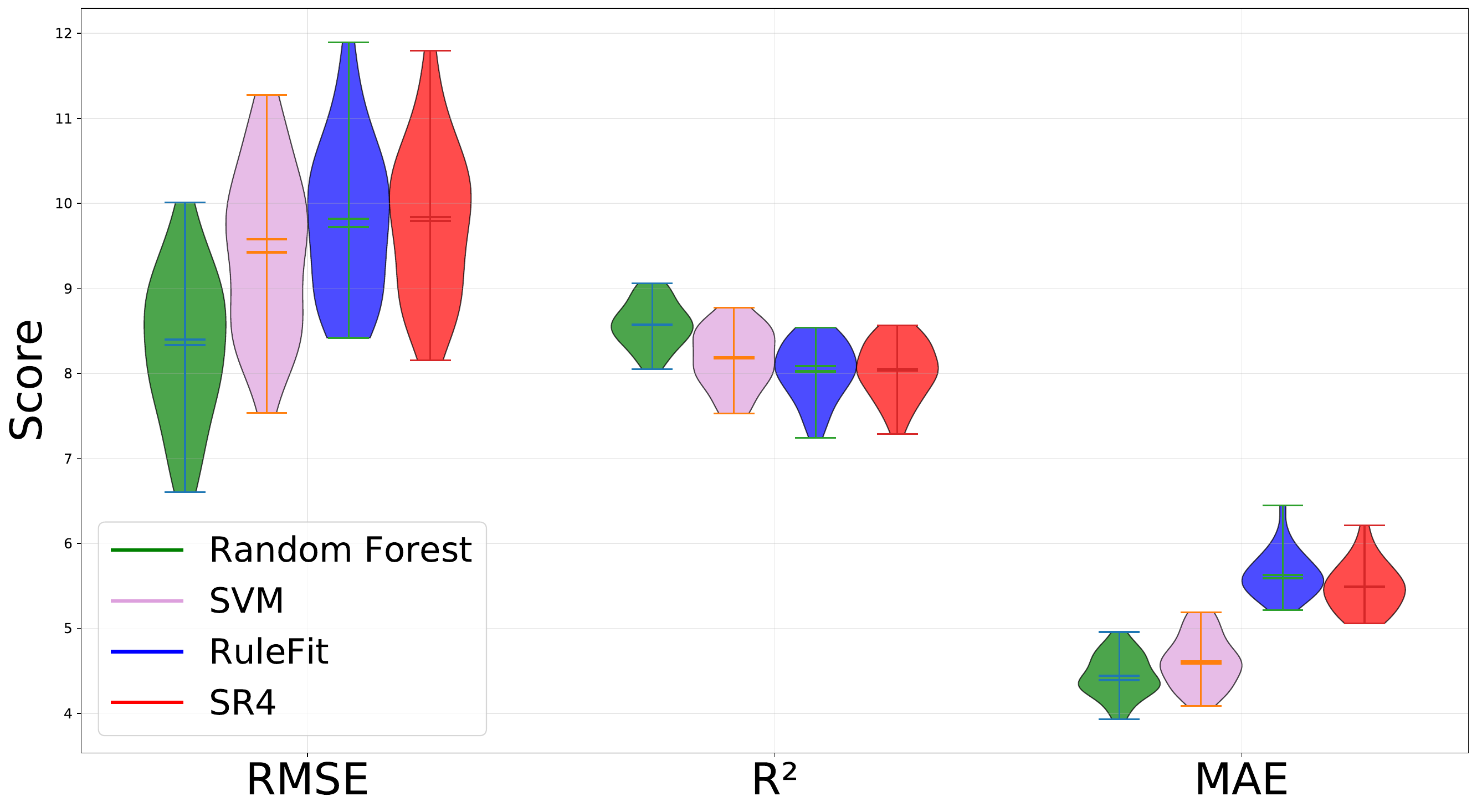}
    \subcaption{Minimum}
\end{minipage}\hfill
\begin{minipage}{0.48\linewidth}
    \centering
    \includegraphics[width=\linewidth]{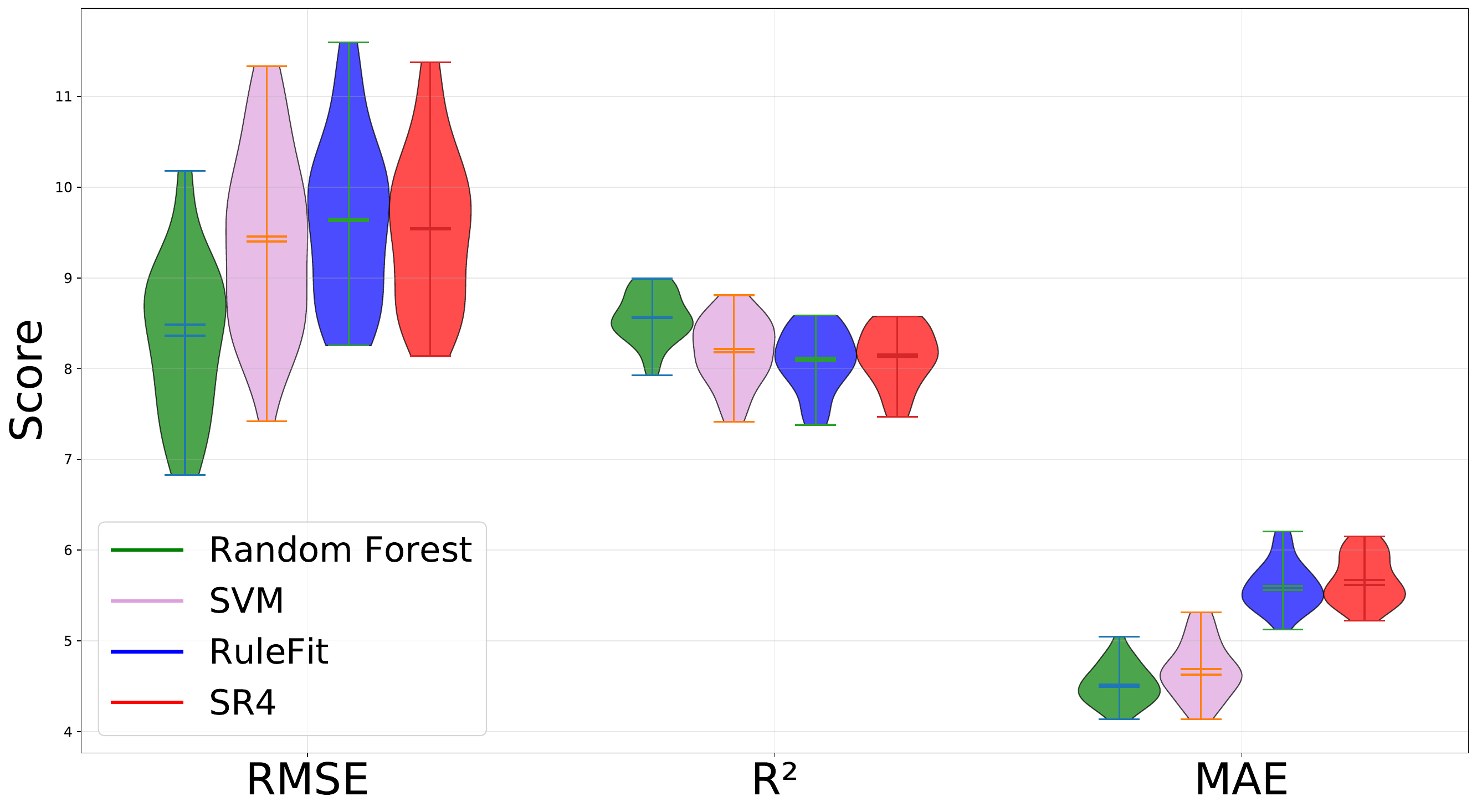}
    \subcaption{Standard}
\end{minipage}

\vspace{0.4cm}

\begin{minipage}{0.48\linewidth}
    \centering
    \includegraphics[width=\linewidth]{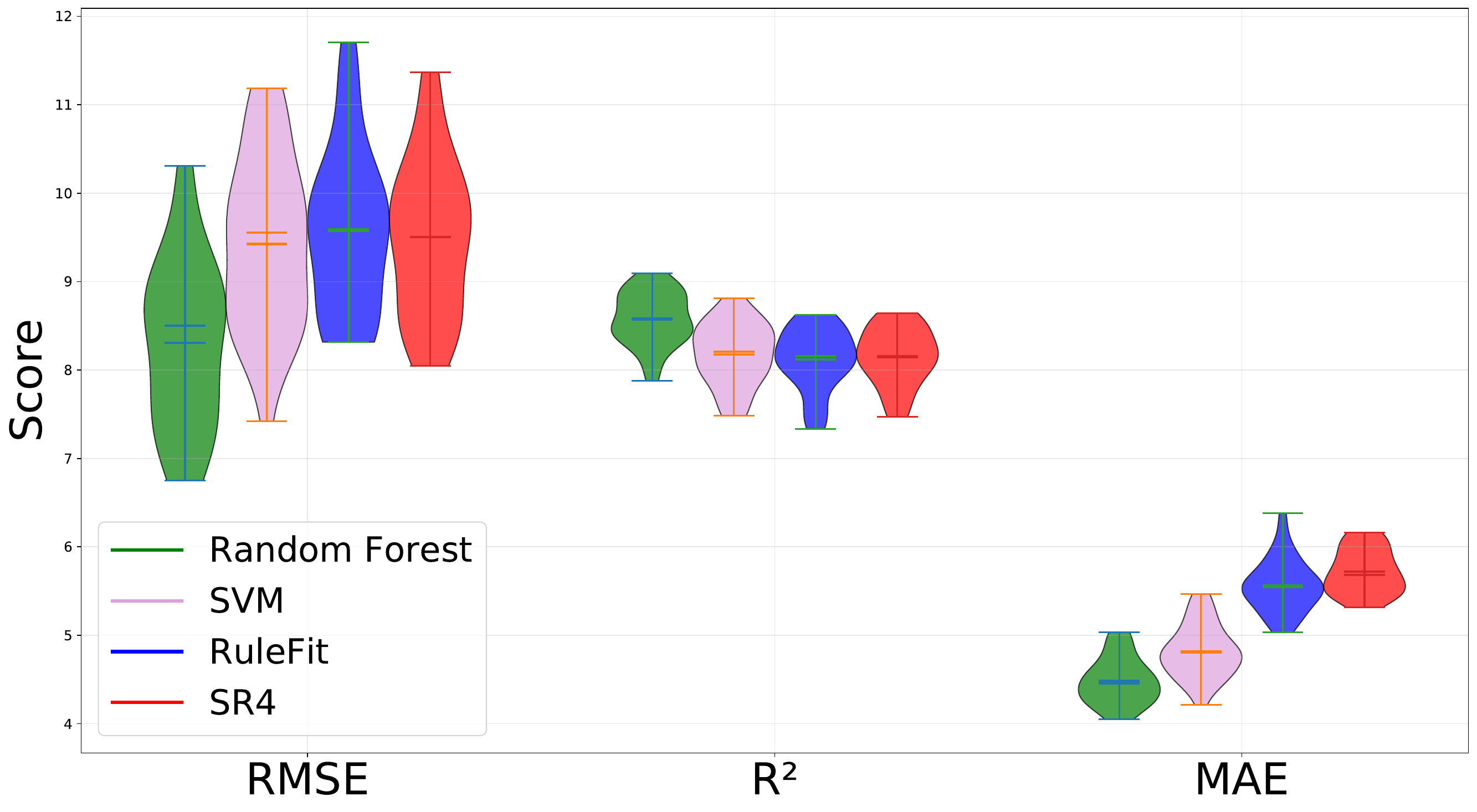}
    \subcaption{Expanded}
\end{minipage}\hfill
\begin{minipage}{0.48\linewidth}
    \centering
    \includegraphics[width=\linewidth]{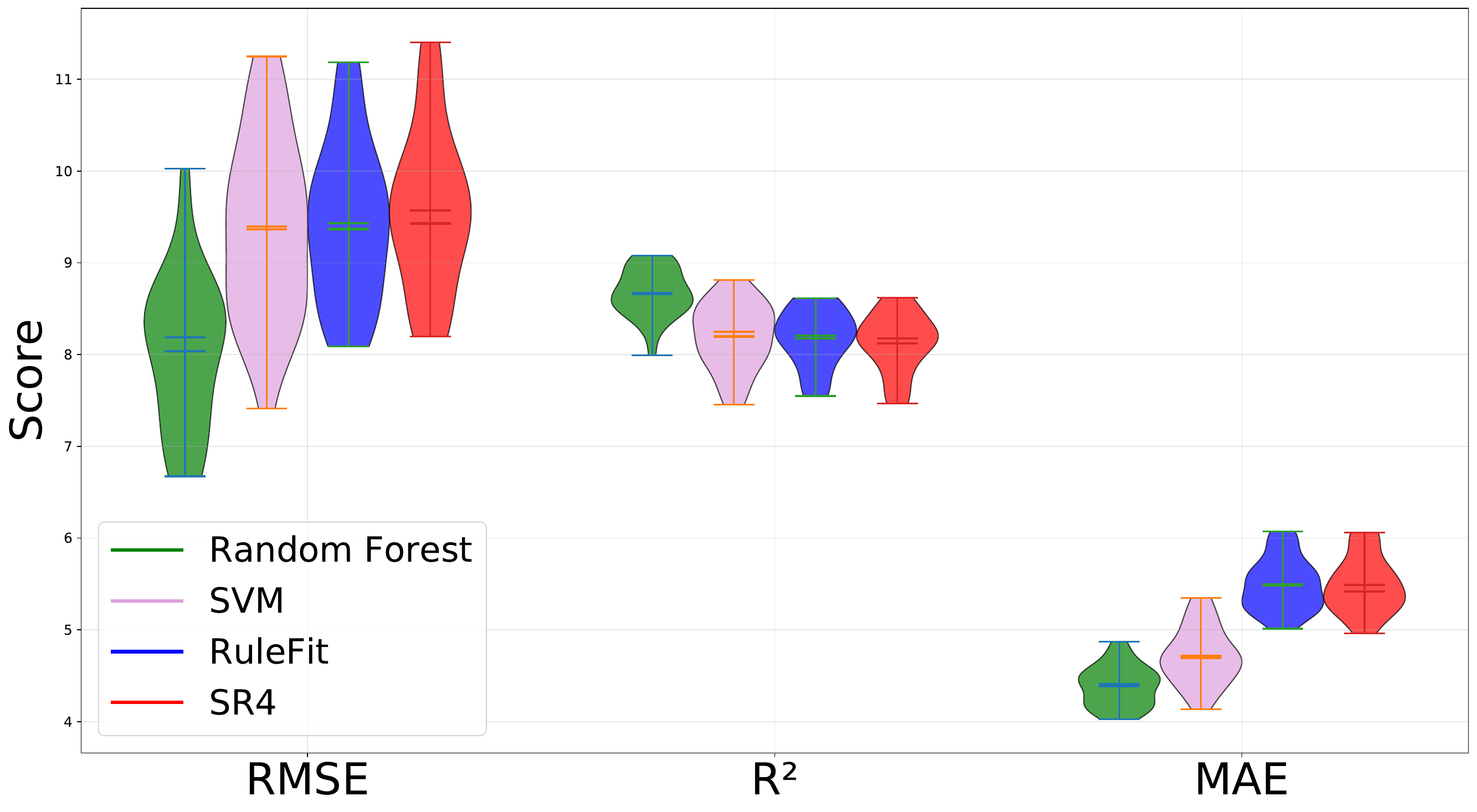}
    \subcaption{Previous}
\end{minipage}

\caption{
Violin plot comparison of model performance metrics across multiple datasets for the election regression task that has DEM percentage as a label and includes party percentage (REP).
}
\label{fig:violin_all_models_dem}
\end{figure}

\begin{figure}
\centering
\begin{minipage}{0.48\linewidth}
    \centering
    \includegraphics[width=\linewidth]{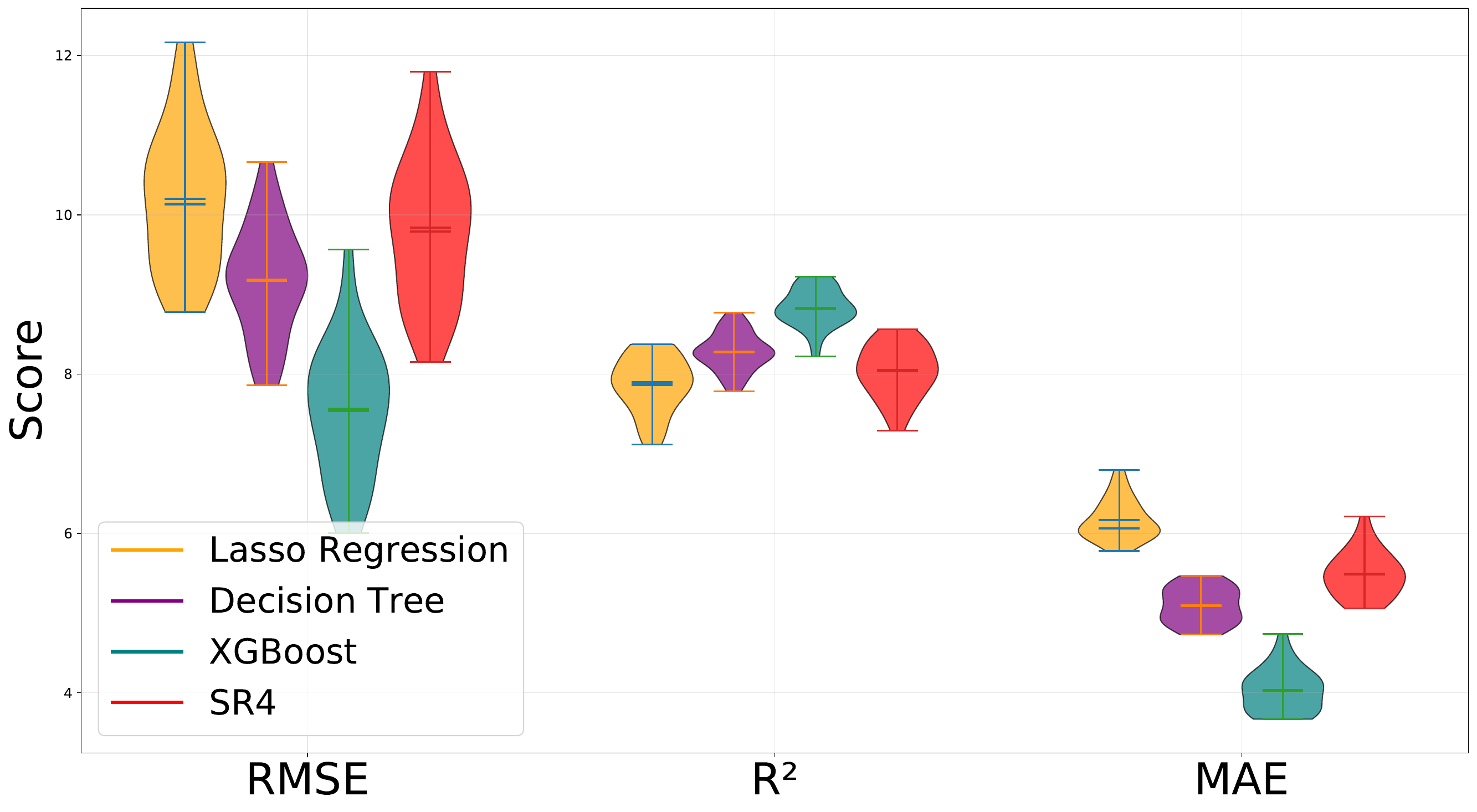}
    \subcaption{Minimum}
\end{minipage}\hfill
\begin{minipage}{0.48\linewidth}
    \centering
    \includegraphics[width=\linewidth]{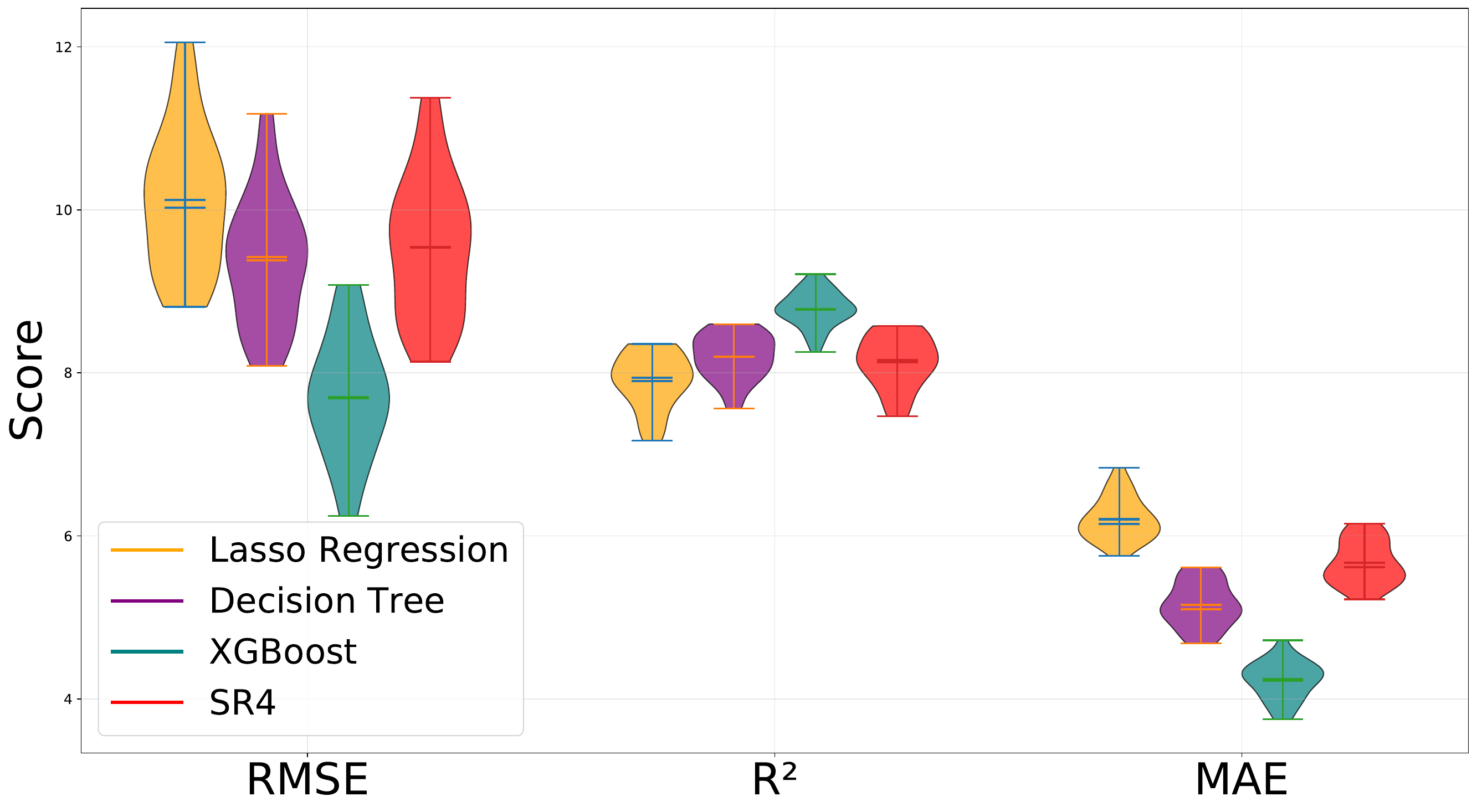}
    \subcaption{Standard}
\end{minipage}

\vspace{0.4cm}

\begin{minipage}{0.48\linewidth}
    \centering
    \includegraphics[width=\linewidth]{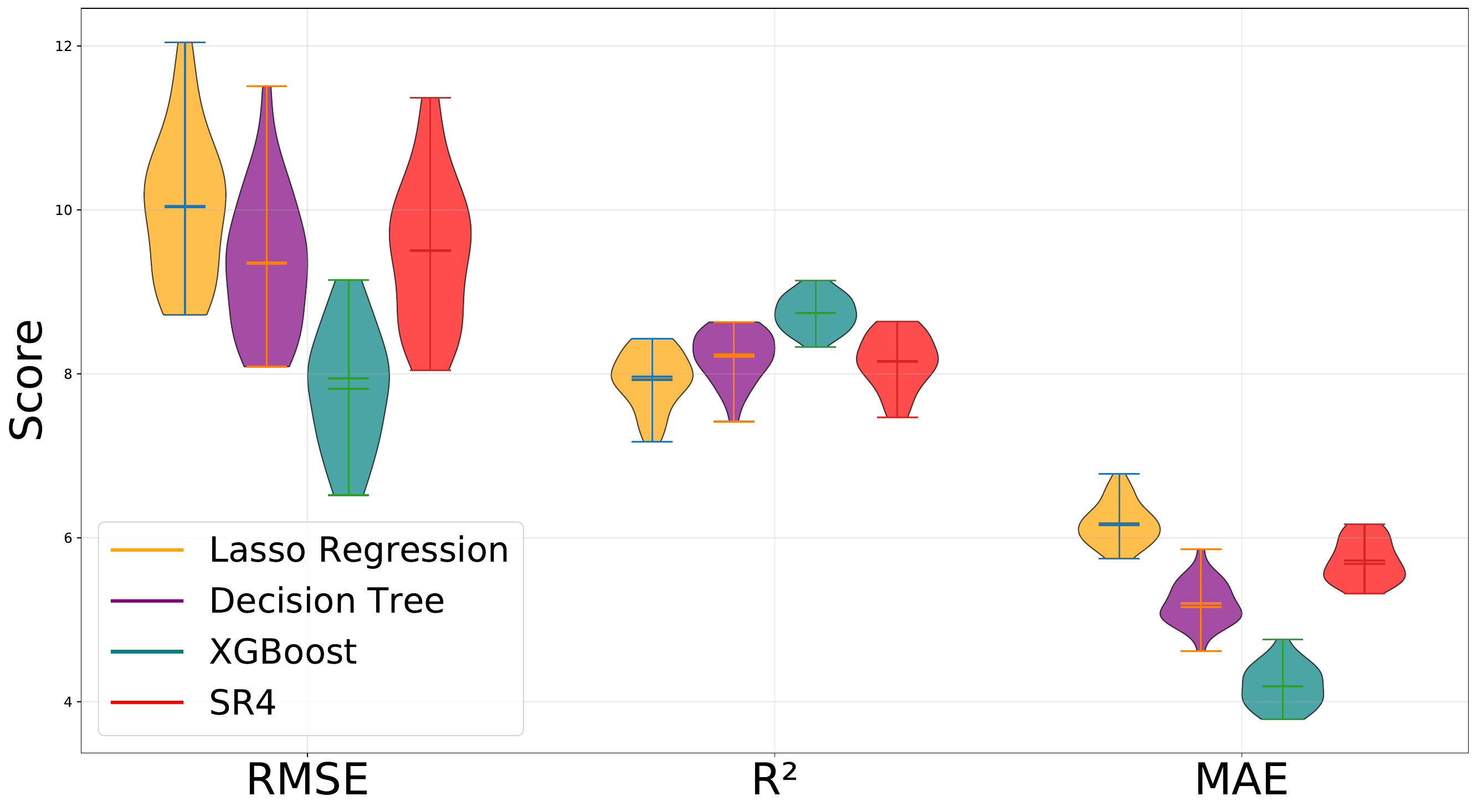}
    \subcaption{Expanded}
\end{minipage}\hfill
\begin{minipage}{0.48\linewidth}
    \centering
    \includegraphics[width=\linewidth]{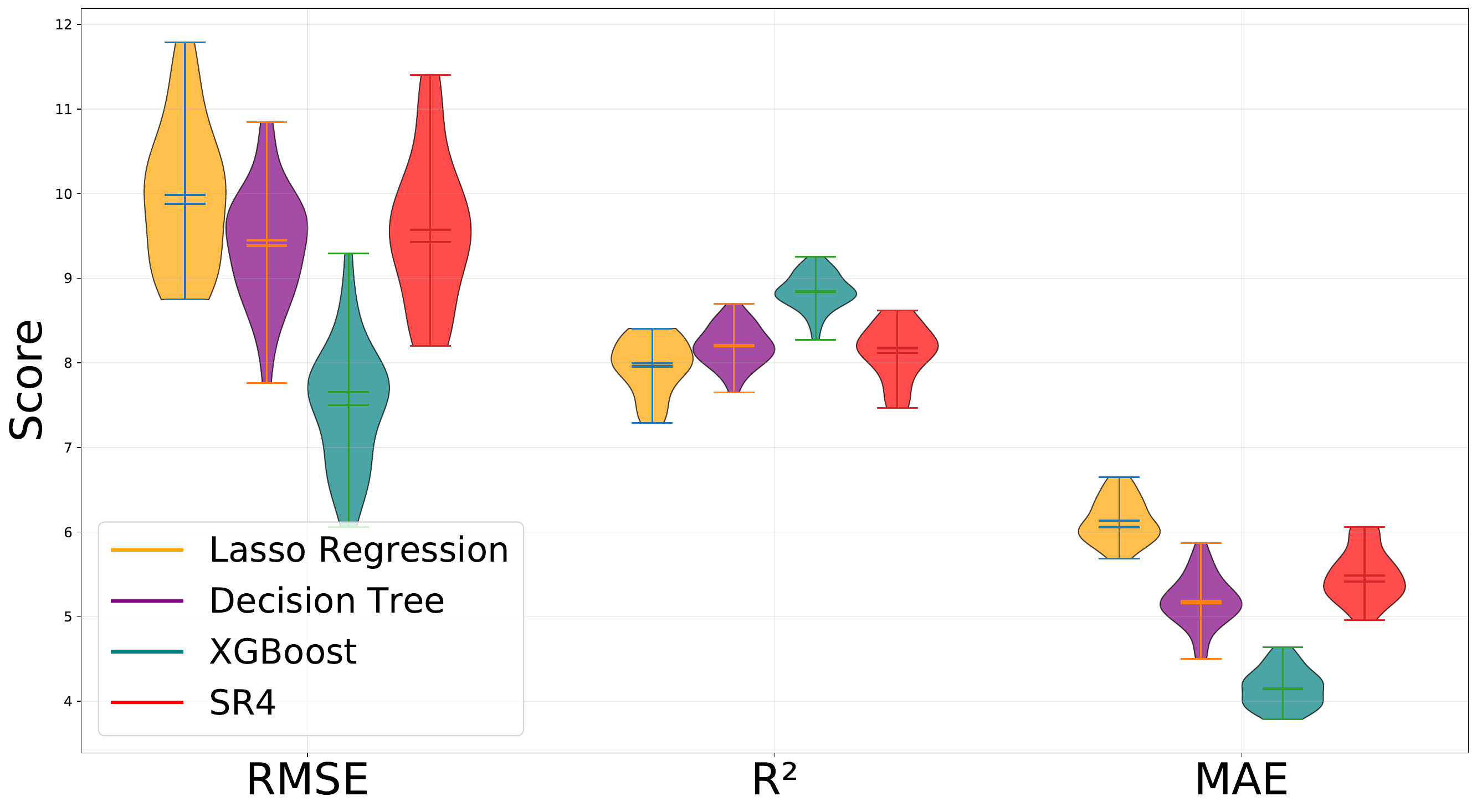}
    \subcaption{Previous}
\end{minipage}

\caption{
Violin plot comparison of model (LASSO regression, Decision Tree, XG boost) performance metrics across multiple datasets for the election regression task that has DEM percentage as a label and includes party percentage (REP).
}
\label{fig:violin_all_models_dem_other}
\end{figure}


\begin{figure}
\centering
\begin{minipage}{0.48\linewidth}
    \centering
    \includegraphics[width=\linewidth]{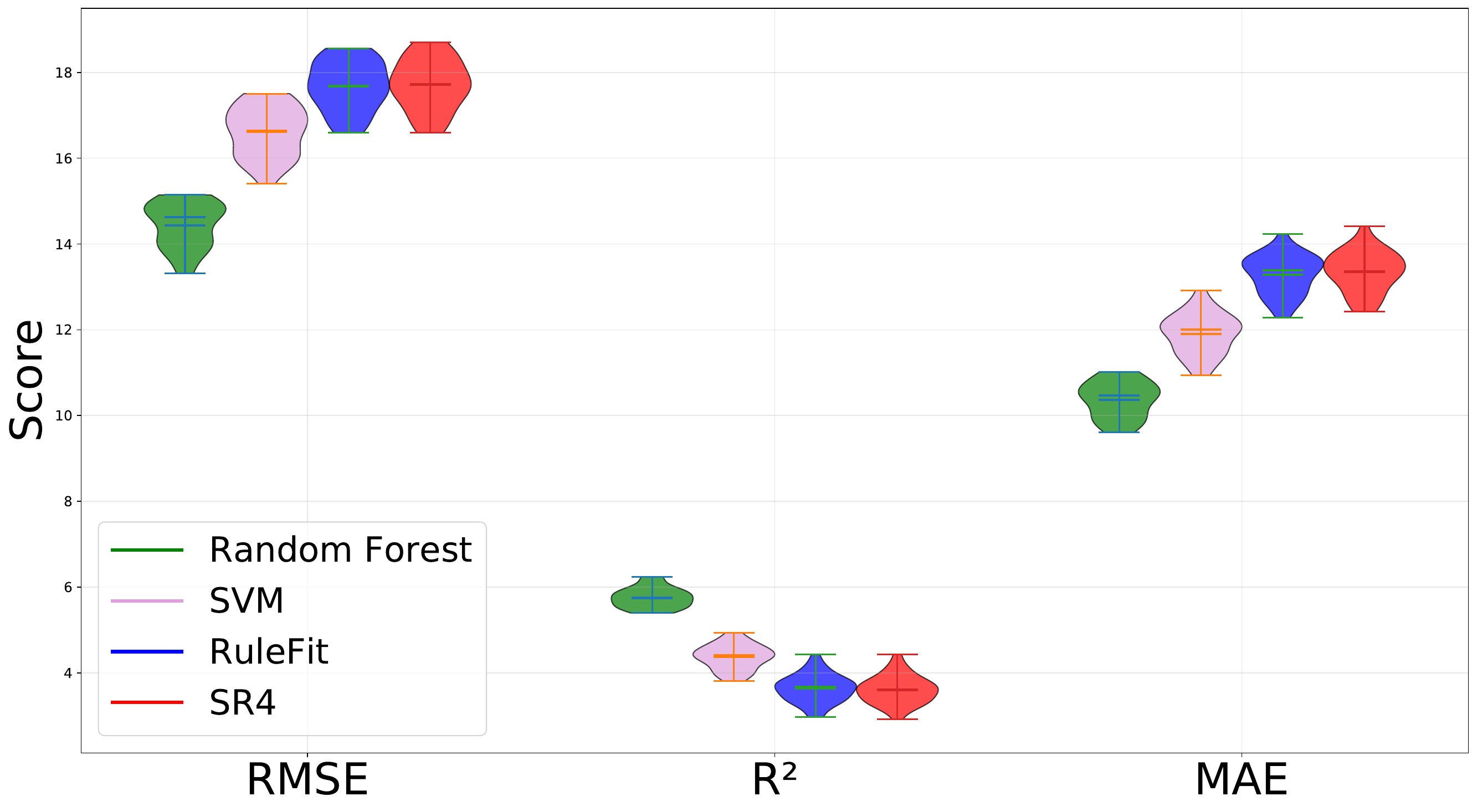}
    \subcaption{Minimum}
\end{minipage}\hfill
\begin{minipage}{0.48\linewidth}
    \centering
    \includegraphics[width=\linewidth]{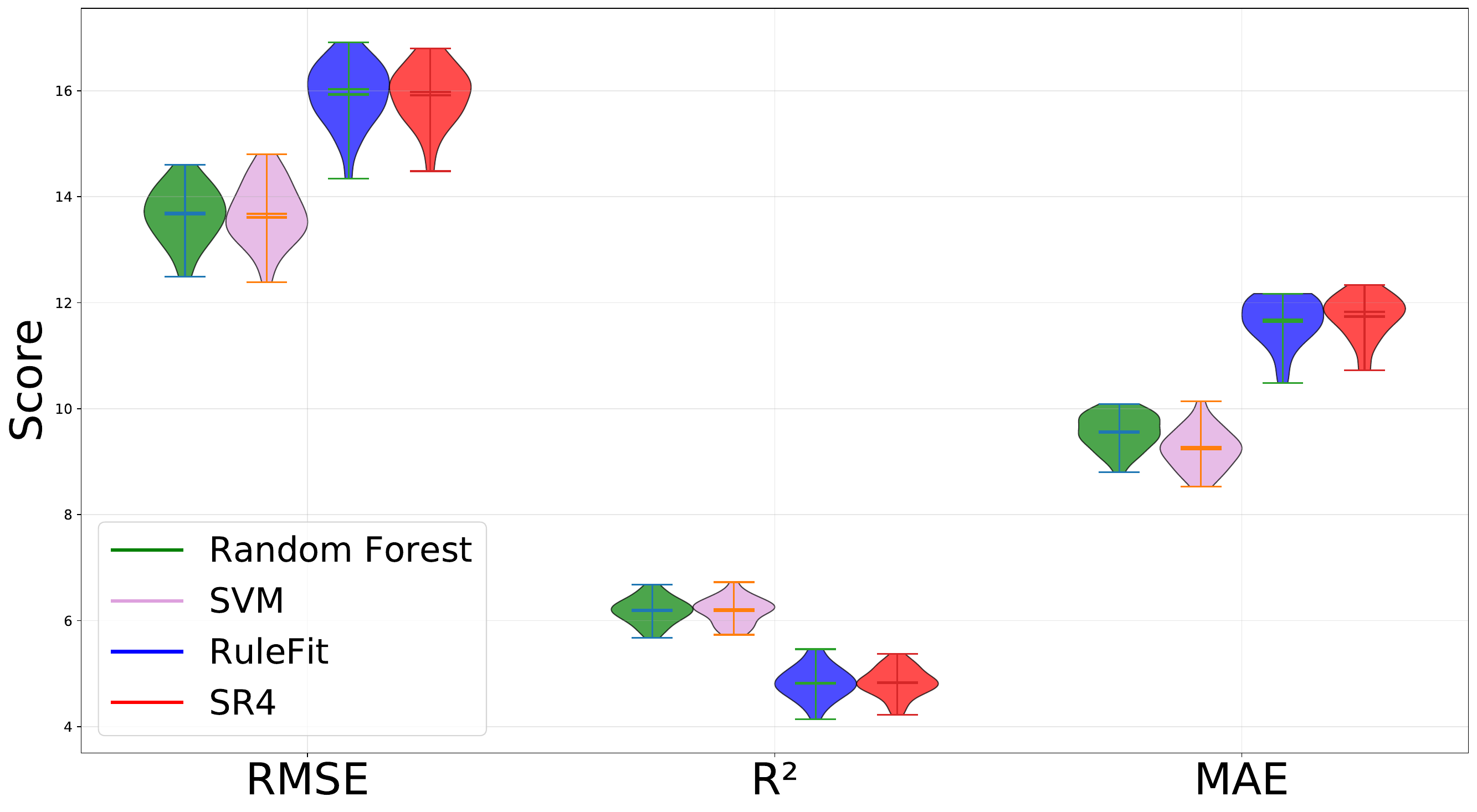}
    \subcaption{Standard}
\end{minipage}

\vspace{0.4cm}

\begin{minipage}{0.48\linewidth}
    \centering
    \includegraphics[width=\linewidth]{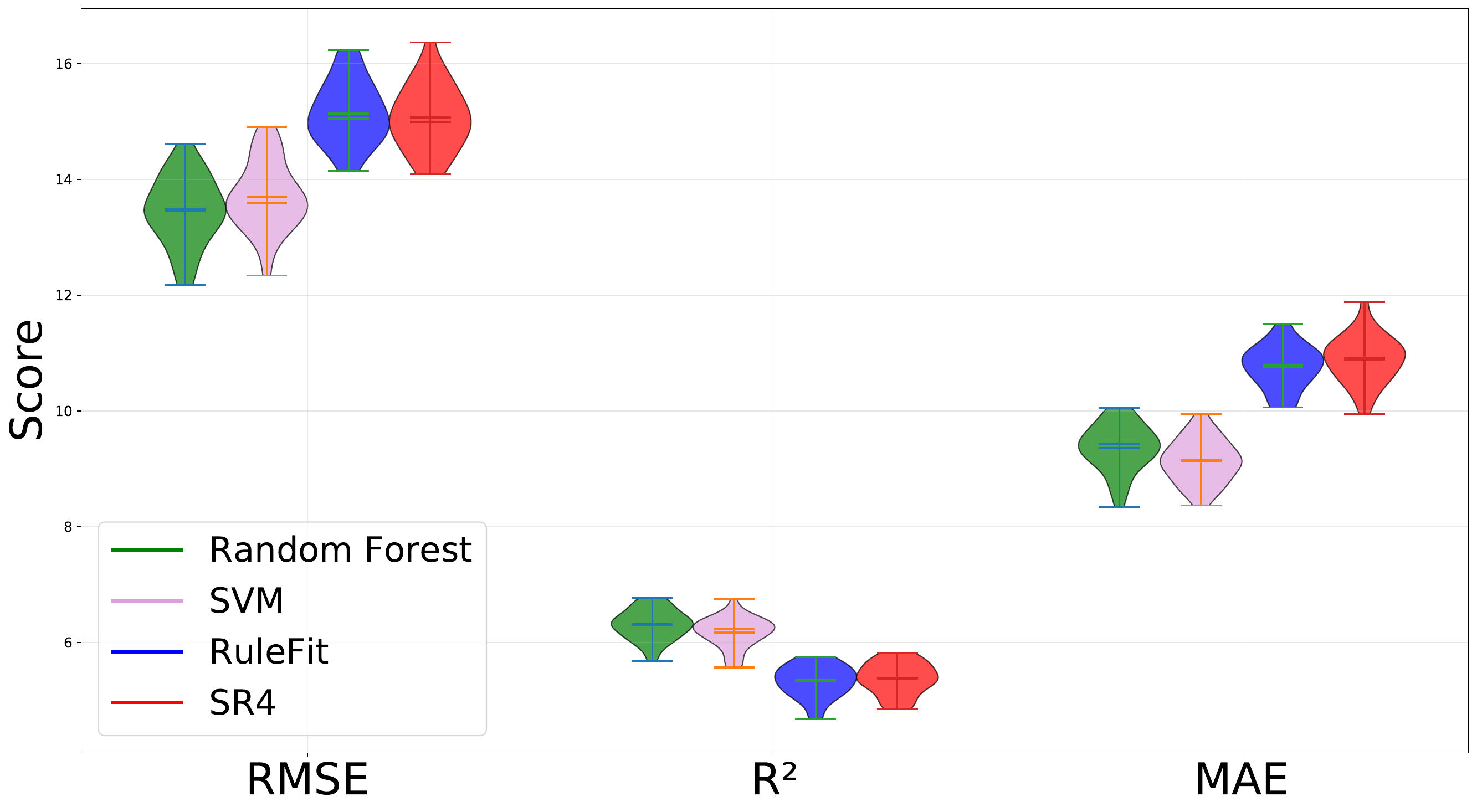}
    \subcaption{Expanded}
\end{minipage}\hfill
\begin{minipage}{0.48\linewidth}
    \centering
    \includegraphics[width=\linewidth]{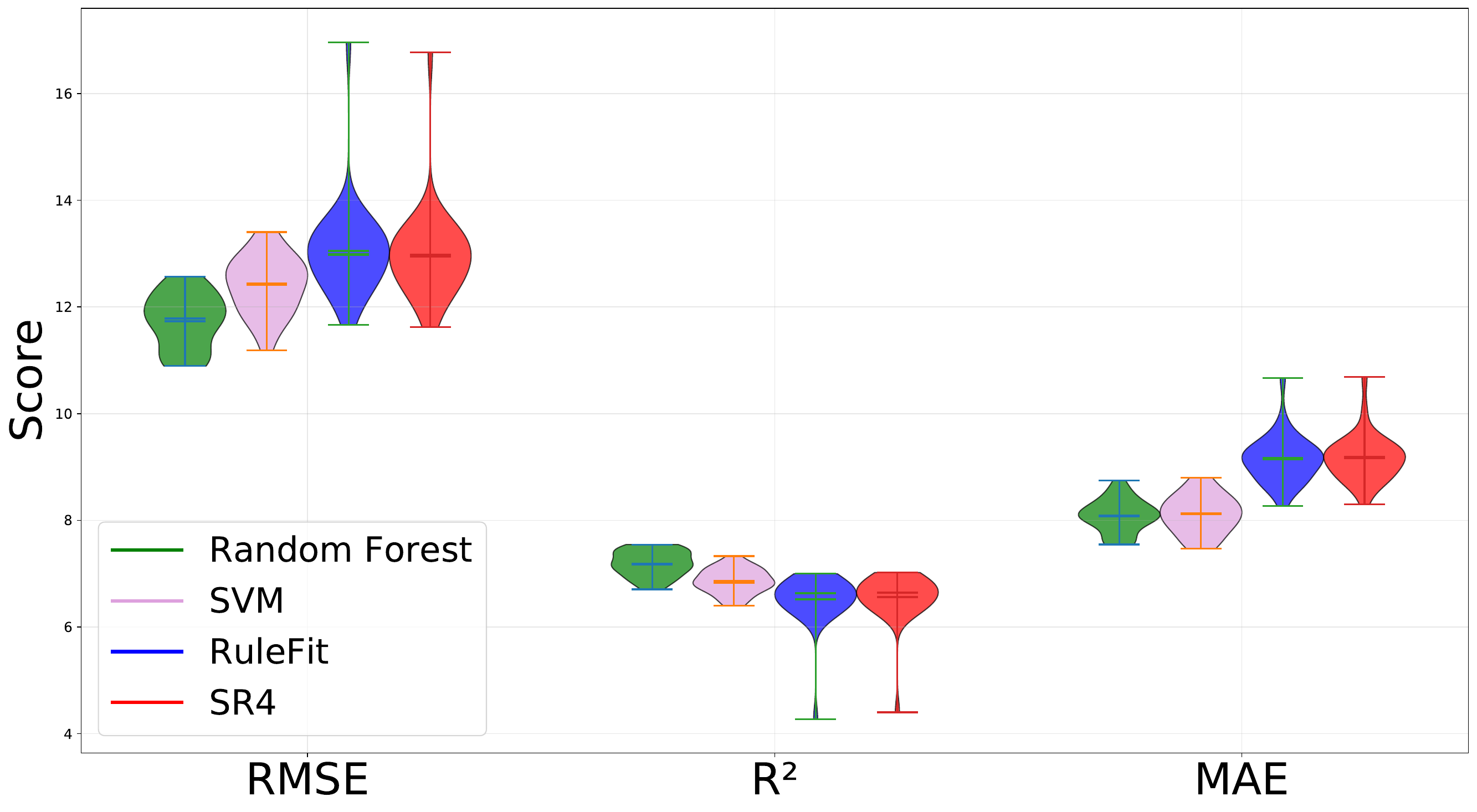}
    \subcaption{Previous}
\end{minipage}

\caption{
Violin plot comparison of model performance metrics across multiple datasets for the election regression task that has DEM percentage as a label and does not include party percentage (REP).
}
\label{fig:violin_all_models_dem_wpp}
\end{figure}

\begin{figure}
\centering
\begin{minipage}{0.48\linewidth}
    \centering
    \includegraphics[width=\linewidth]{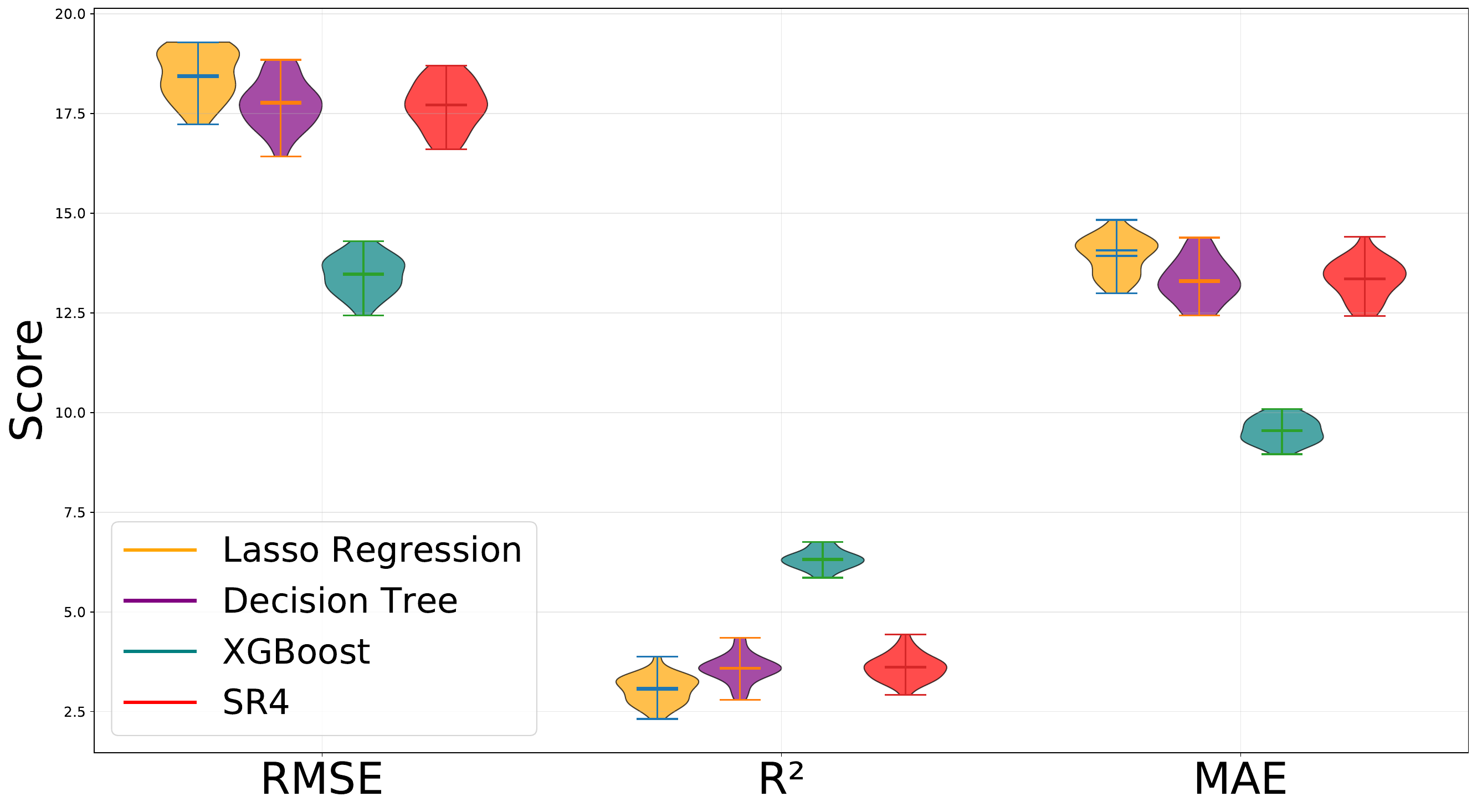}
    \subcaption{Minimum}
\end{minipage}\hfill
\begin{minipage}{0.48\linewidth}
    \centering
    \includegraphics[width=\linewidth]{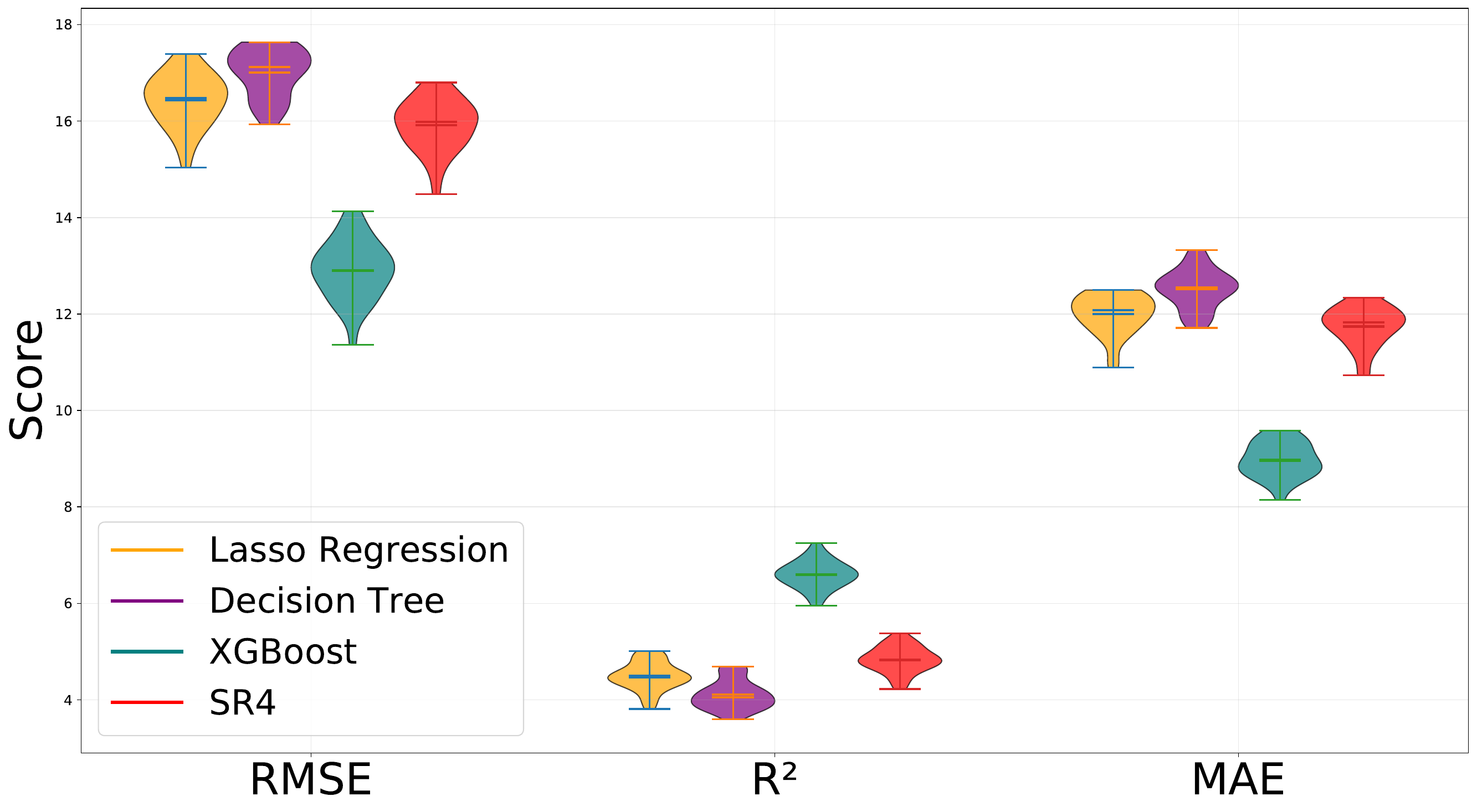}
    \subcaption{Standard}
\end{minipage}

\vspace{0.4cm}

\begin{minipage}{0.48\linewidth}
    \centering
    \includegraphics[width=\linewidth]{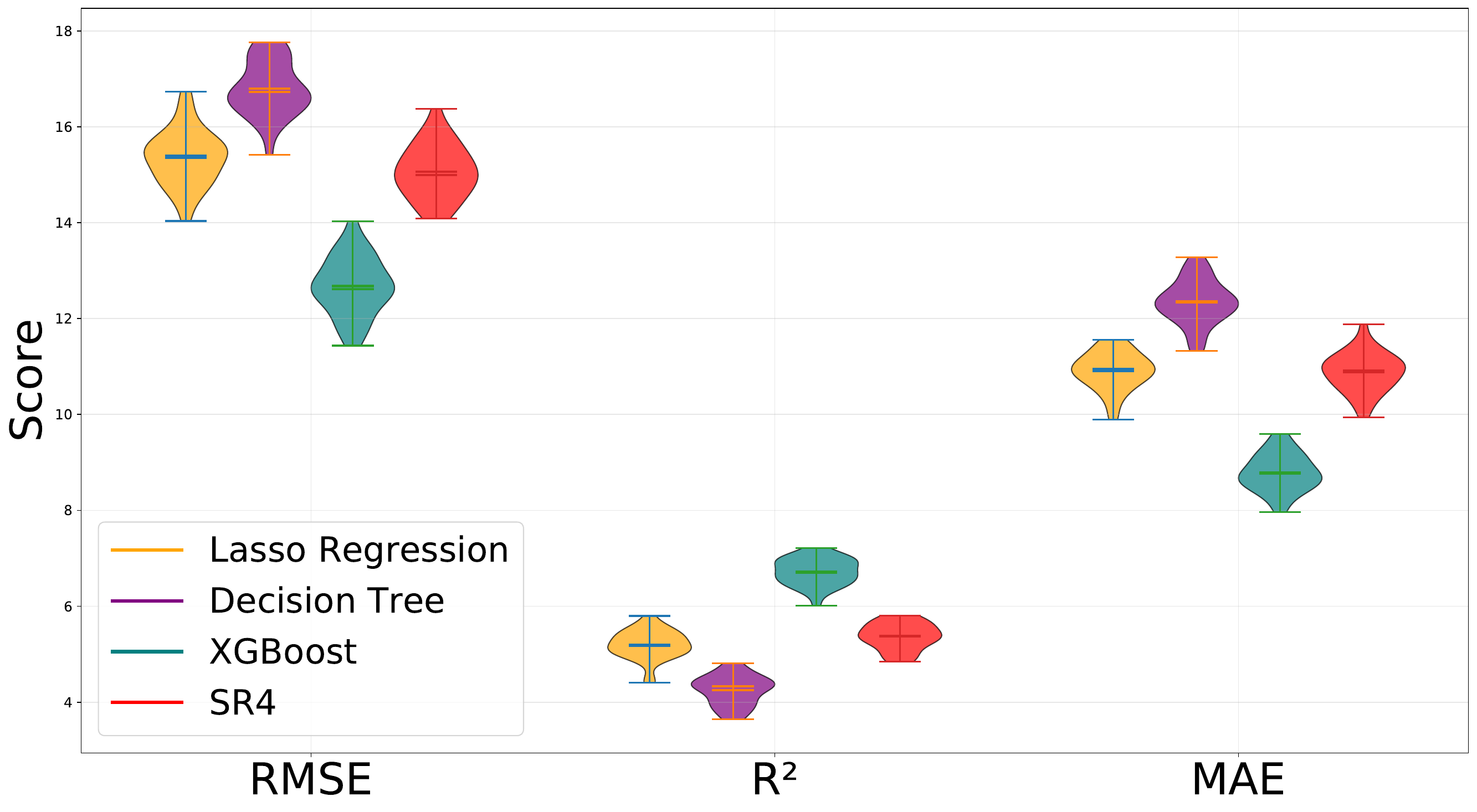}
    \subcaption{Expanded}
\end{minipage}\hfill
\begin{minipage}{0.48\linewidth}
    \centering
    \includegraphics[width=\linewidth]{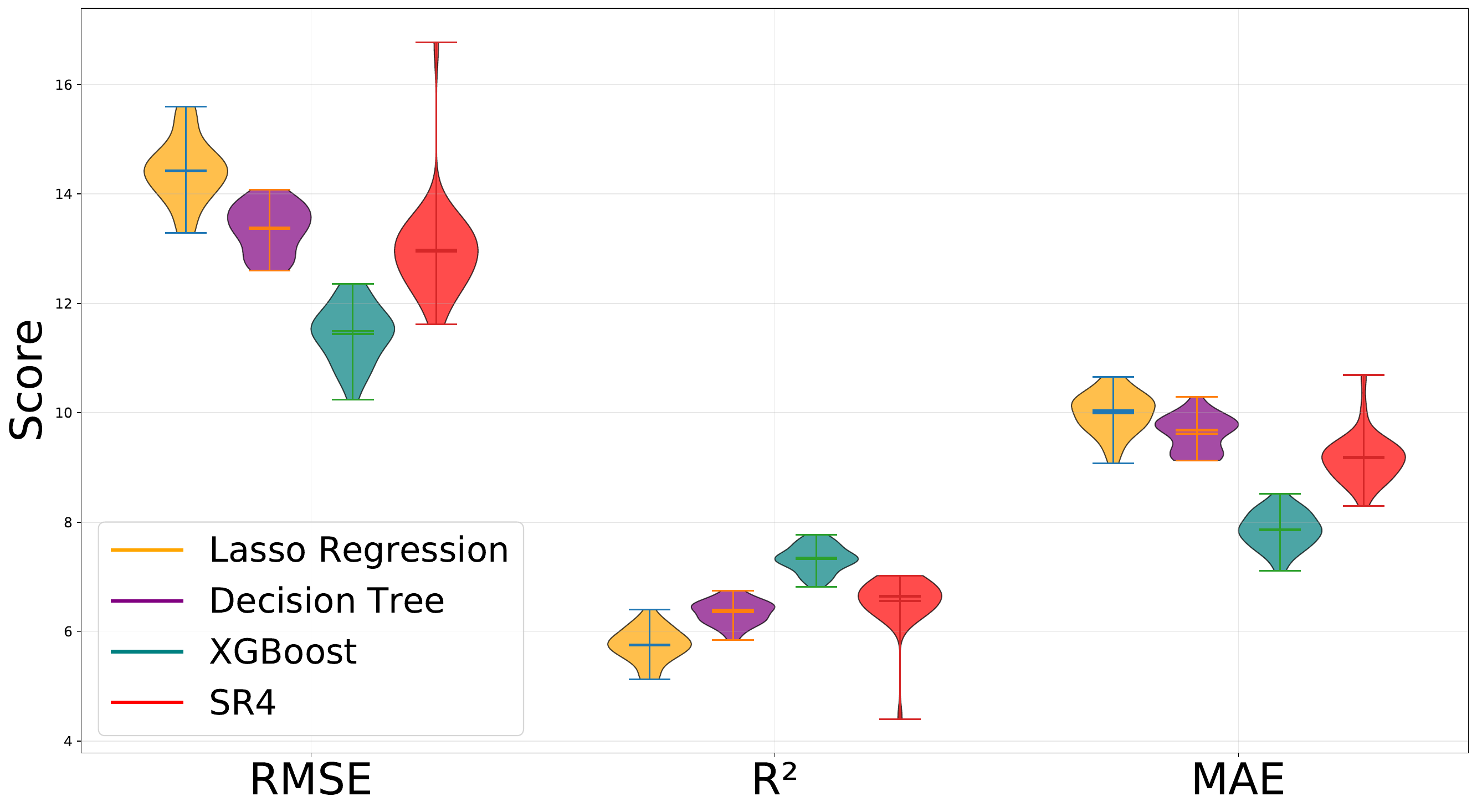}
    \subcaption{Previous}
\end{minipage}

\caption{
Violin plot comparison of model (LASSO regression, Decision Tree, XG boost) performance metrics across multiple datasets for the election regression task that has DEM percentage as a label and does not include party percentage (REP).
}
\label{fig:violin_all_models_dem_wpp_other}
\end{figure}


\begin{table}
\caption{
Average Dice–Sørensen Index $\pm$ standard deviation across 30 trials for the election regression dataset where DEM percentage is the label.
}

\centering
\setlength{\tabcolsep}{3pt}
\renewcommand{\arraystretch}{1.15}

\begin{subtable}{\columnwidth}
\centering
\begin{tabular}{lcccc}
\toprule
\textbf{Dataset}
& \textbf{Random Forest} & \textbf{RuleFit} & \textbf{SR4-Fit} & \textbf{Decision Tree} \\
\midrule
Minimum   & 0.0001$\pm$0.0000 & 0.3333$\pm$0.0000 & \textbf{0.3851$\pm$0.0116} & 0.0027$\pm$0.0043 \\
Standard  & 0.0001$\pm$0.0000 & \textbf{0.7714$\pm$0.0000} & 0.7519$\pm$0.0101 & 0.0020$\pm$0.0030 \\
Expanded  & 0.0001$\pm$0.0000 & 0.7377$\pm$0.0000 & \textbf{0.8337$\pm$0.0102} & 0.0030$\pm$0.0043 \\
Previous  & 0.0001$\pm$0.0000 & \textbf{0.7778$\pm$0.0000} & 0.6863$\pm$0.0146 & 0.0058$\pm$0.0088 \\
\bottomrule
\end{tabular}
\caption{\textbf{DEM as a label, with REP percentage}}
\end{subtable}

\vspace{6pt}

\begin{subtable}{\columnwidth}
\centering
\begin{tabular}{lcccc}
\toprule
\textbf{Dataset}
& \textbf{Random Forest} & \textbf{RuleFit} & \textbf{SR4-Fit} & \textbf{Decision Tree} \\
\midrule
Minimum   & 0.0003$\pm$0.0001 & 0.1032$\pm$0.0014 & \textbf{0.2690$\pm$0.0122} & 0.0001$\pm$0.0000 \\
Standard  & 0.0002$\pm$0.0001 & 0.3034$\pm$0.0031 & \textbf{0.5404$\pm$0.0209} & 0.0029$\pm$0.0045 \\
Expanded  & 0.0001$\pm$0.0001 & \textbf{0.7342$\pm$0.0015} & 0.6638$\pm$0.0118 & 0.3540$\pm$0.2354 \\
Previous  & 0.0020$\pm$0.0005 & 0.7714$\pm$0.0000 & \textbf{0.8051$\pm$0.0077} & 0.0989$\pm$0.1001 \\
\bottomrule
\end{tabular}
\caption{\textbf{DEM as a label, without REP percentage.}}
\end{subtable}

\label{tab:dice_reg_dem_vs_no_rep}
\end{table}


\begin{table}
\caption{
Average number of rules $\pm$ standard deviation across 30 trials for the election regression dataset, where DEM percentage is the label.
}

\centering
\setlength{\tabcolsep}{3pt}
\renewcommand{\arraystretch}{1.15}

\begin{subtable}{\columnwidth}
\centering
\begin{tabular}{lcccc}
\toprule
\textbf{Dataset}
& \textbf{Random Forest} & \textbf{RuleFit} & \textbf{SR4-Fit} & \textbf{Decision Tree} \\
\midrule
Minimum   & 8086.66$\pm$199.12 & 24.00$\pm$0.00 & 20.63$\pm$1.43 & \textbf{9.03$\pm$1.45} \\
Standard  & 8162.07$\pm$233.69 & 35.00$\pm$0.00 & 34.13$\pm$1.03 & \textbf{12.33$\pm$1.03} \\
Expanded  & 7972.60$\pm$224.35 & 61.00$\pm$0.00 & 52.10$\pm$1.09 & \textbf{11.56$\pm$1.69} \\
Previous  & 7856.70$\pm$222.97 & 36.00$\pm$0.00 & 38.83$\pm$1.32 & \textbf{11.10$\pm$1.21} \\
\bottomrule
\end{tabular}
\caption{\textbf{DEM as a label, with REP percentage.}}
\end{subtable}

\vspace{6pt}

\begin{subtable}{\columnwidth}
\centering
\begin{tabular}{lcccc}
\toprule
\textbf{Dataset}
& \textbf{Random Forest} & \textbf{RuleFit} & \textbf{SR4-Fit} & \textbf{Decision Tree} \\
\midrule
Minimum   & 8869.63$\pm$179.67 & 67.87$\pm$1.82 & 26.13$\pm$2.46 & \textbf{1.00$\pm$0.00} \\
Standard  & 9024.40$\pm$202.97 & 85.70$\pm$1.84 & 46.13$\pm$3.35 & \textbf{8.22$\pm$2.15} \\
Expanded  & 8654.03$\pm$138.38 & 59.93$\pm$0.25 & 61.70$\pm$2.31 & \textbf{1.00$\pm$0.00} \\
Previous  & 7946.13$\pm$164.93 & 35.00$\pm$0.00 & 32.30$\pm$0.65 & \textbf{1.53$\pm$0.50} \\
\bottomrule
\end{tabular}
\caption{\textbf{DEM as a label, without REP percentage.}}
\end{subtable}

\label{tab:rules_reg_dem_vs_no_rep}
\end{table}


\begin{table}
\caption{
Average rule complexity $\pm$ standard deviation across 30 trials for the election regression dataset where DEM percentage is the label.
}

\centering
\setlength{\tabcolsep}{3pt}
\renewcommand{\arraystretch}{1.15}

\begin{subtable}{\columnwidth}
\centering
\begin{tabular}{lcccc}
\toprule
\textbf{Dataset}
& \textbf{Random Forest} & \textbf{RuleFit} & \textbf{SR4-Fit} & \textbf{Decision Tree} \\
\midrule
Minimum   & 6.25$\pm$0.03 & 3.00$\pm$0.00 & \textbf{2.84$\pm$0.01} & 4.81$\pm$0.14 \\
Standard  & 6.17$\pm$0.03 & \textbf{1.46$\pm$0.00} & 1.46$\pm$0.01 & 4.80$\pm$0.08 \\
Expanded  & 6.09$\pm$0.03 & 1.78$\pm$0.00 & \textbf{1.30$\pm$0.01} & 4.75$\pm$0.10 \\
Previous  & 6.11$\pm$0.03 & \textbf{1.44$\pm$0.00} & 1.89$\pm$0.05 & 4.73$\pm$0.01 \\
\bottomrule
\end{tabular}
\caption{\textbf{DEM as a label, with REP percentage.}}
\end{subtable}

\vspace{6pt}

\begin{subtable}{\columnwidth}
\centering
\begin{tabular}{lcccc}
\toprule
\textbf{Dataset}
& \textbf{Random Forest} & \textbf{RuleFit} & \textbf{SR4-Fit} & \textbf{Decision Tree} \\
\midrule
Minimum   & 6.36$\pm$0.02 & 5.44$\pm$0.04 & 4.64$\pm$0.13 & \textbf{1.01$\pm$0.00} \\
Standard  & 6.39$\pm$0.02 & 4.44$\pm$0.05 & \textbf{3.28$\pm$0.21} & 4.73$\pm$0.16 \\
Expanded  & 6.34$\pm$0.014 & 1.79$\pm$0.01 & 2.68$\pm$0.12 & \textbf{1.2$\pm$0.61} \\
Previous  & 6.25$\pm$0.03 & 1.46$\pm$0.00 & \textbf{1.39$\pm$0.03} & 2.80$\pm$0.61 \\
\bottomrule
\end{tabular}
\caption{\textbf{DEM as a label, without REP percentage.}}
\end{subtable}

\label{tab:complexity_reg_dem_vs_no_rep}
\end{table}


In the minimum dataset, the line plots in Figure~\ref{fig:lineplot_grid_min_dem} and Figure~\ref{fig:lineplot_grid_min_dem_other} show that when REP percentage is included as a feature, all models achieve relatively low error values, whereas Random Forest  attains slightly lower average error values compared to other models. The distributional comparisons through violin plots in Figure~\ref{fig:violin_all_models_dem} and Figure~\ref{fig:violin_all_models_dem_other} reinforce these findings. When REP percentage is removed, performance declines across all models, with XGBoost  attaining slightly lower average error values compared to other models. In contrast, SR4-Fit maintains comparatively smoother trends despite a modest increase in error, which is further confirmed in Figure~\ref{fig:violin_all_models_dem_wpp} and Figure~\ref{fig:violin_all_models_dem_wpp_other}, where SR4-Fit retains more concentrated distributions. Stability analysis in Table~\ref{tab:dice_reg_dem_vs_no_rep} shows that with both cases, SR4-Fit achieves the highest Dice–Sørensen Index among rule-based models, showing its structural consistency and robustness across different experimental scenarios. Table~\ref{tab:rules_reg_dem_vs_no_rep} shows that in both cases, the Decision Tree has the most compact structure, but the structure is too compact, which may lower robustness and may lower the ability to identify interactions properly. Table~\ref{tab:complexity_reg_dem_vs_no_rep} shows that in the case of including REP percentage as a feature, SR4-Fit has the lower average rule complexity, whereas in the case of removing REP percentage as a feature, the Decision Tree has the overall lower average rule complexity.

In the standard dataset, the line plots in Figure~\ref{fig:lineplot_grid_std_dem} and Figure~\ref{fig:lineplot_grid_std_dem_other} show that when REP percentage is included as a feature, all models achieve lower error values, whereas XGBoost attains slightly lower average error values compared to other models. When the REP percentage is removed, performance declines across all models, with XGBoost still attaining lower average error values compared to other models. Stability analysis in Table~\ref{tab:dice_reg_dem_vs_no_rep} shows that in the case of including REP percentage as a feature, RuleFit achieves the highest Dice–Sørensen Index among rule-based models, whereas when REP percentage is removed, SR4-Fit achieves the highest Dice–Sørensen Index among rule-based models. Table~\ref{tab:rules_reg_dem_vs_no_rep} shows that in both cases, the Decision Tree has the most compact structure, trailed closely by SR4-Fit. Table~\ref{tab:complexity_reg_dem_vs_no_rep} shows that in the case of including REP percentage as a feature, RuleFit has the lower average rule complexity, whereas in the case of removing REP percentage as a feature, the SR4-Fit has the overall lower average rule complexity.

In the expanded dataset, the line plots in Figure~\ref{fig:lineplot_grid_exp_dem} and Figure~\ref{fig:lineplot_grid_exp_dem_other} show that when REP percentage is included as a feature, all models achieve lower error values, whereas XGBoost again attains slightly lower average error values compared to other models. When the REP percentage is removed, performance declines across all models, with XGBoost still attaining lower average error values compared to other models. Stability analysis in Table~\ref{tab:dice_reg_dem_vs_no_rep} shows that in the case of including REP percentage as a feature, SR4-Fit achieves the highest Dice–Sørensen Index among rule-based models, whereas when REP percentage is removed, RuleFit achieves the highest Dice–Sørensen Index among rule-based models. Table~\ref{tab:rules_reg_dem_vs_no_rep} shows that in both cases, the Decision Tree has the most compact structure. Table~\ref{tab:complexity_reg_dem_vs_no_rep} shows that in the case of including REP percentage as a feature, SR4-Fit has the lower average rule complexity, whereas in the case of removing REP percentage as a feature, the Decision Tree has the overall lower average rule complexity.

In the previous dataset, the line plots in Figure~\ref{fig:lineplot_grid_pre_dem} and Figure~\ref{fig:lineplot_grid_pre_dem_other} show that when REP percentage is included as a feature, all models achieve lower error values, whereas XGBoost attains slightly lower average error values compared to other models. When the REP percentage is removed, performance declines across all models, with XGBoost slightly achieving lower average error values compared to other models. Stability analysis in Table~\ref{tab:dice_reg_dem_vs_no_rep} shows that with both cases, SR4-Fit achieves the highest Dice–Sørensen Index among rule-based models. Table~\ref{tab:rules_reg_dem_vs_no_rep} shows that in both cases, the Decision Tree has the most compact structure. Table~\ref{tab:complexity_reg_dem_vs_no_rep} shows that in the case of including REP percentage as a feature, RuleFit has the lower average rule complexity, whereas in the case of removing REP percentage as a feature, the SR4-Fit has the overall lower average rule complexity.
\FloatBarrier

\section{Results of U.S House of Representative Elections Regression (REP)}\label{sup_sec_HouseRR}
\begin{figure}
\centering
\begin{minipage}{0.49\linewidth}
    \centering
    \includegraphics[width=\linewidth]{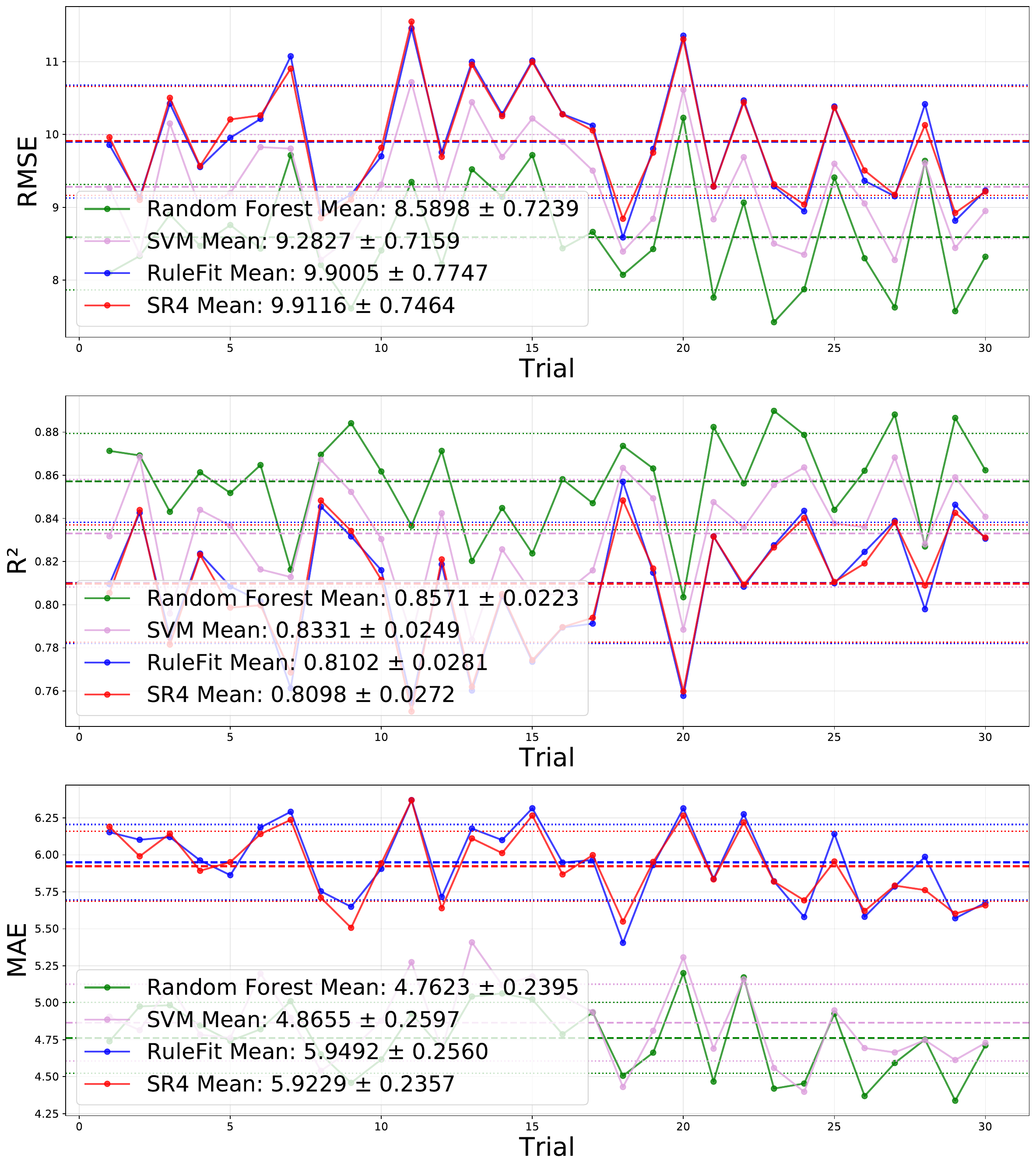}
\end{minipage}\hfill
\begin{minipage}{0.49\linewidth}
    \centering
    \includegraphics[width=\linewidth]{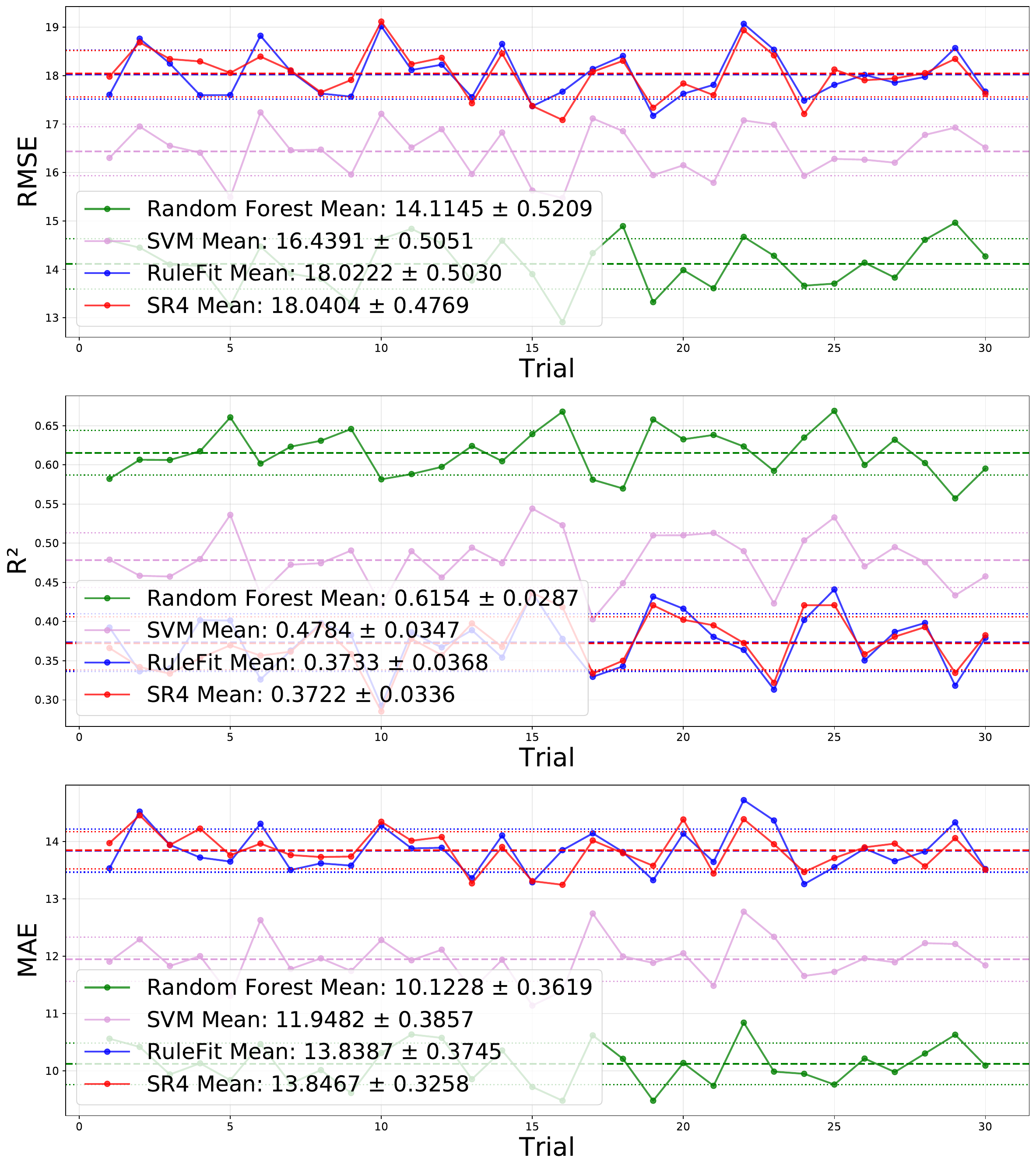}
\end{minipage}

\caption{
Line plot comparison of model performance metrics for minimum data across 30 trials with REP percentage as label. The left panel reports results obtained with party percentage (DEM) included, whereas the right panel shows results with party percentage (DEM) not included.
}
\label{fig:lineplot_grid_min_rep}
\end{figure}

\begin{figure}
\centering
\begin{minipage}{0.49\linewidth}
    \centering
    \includegraphics[width=\linewidth]{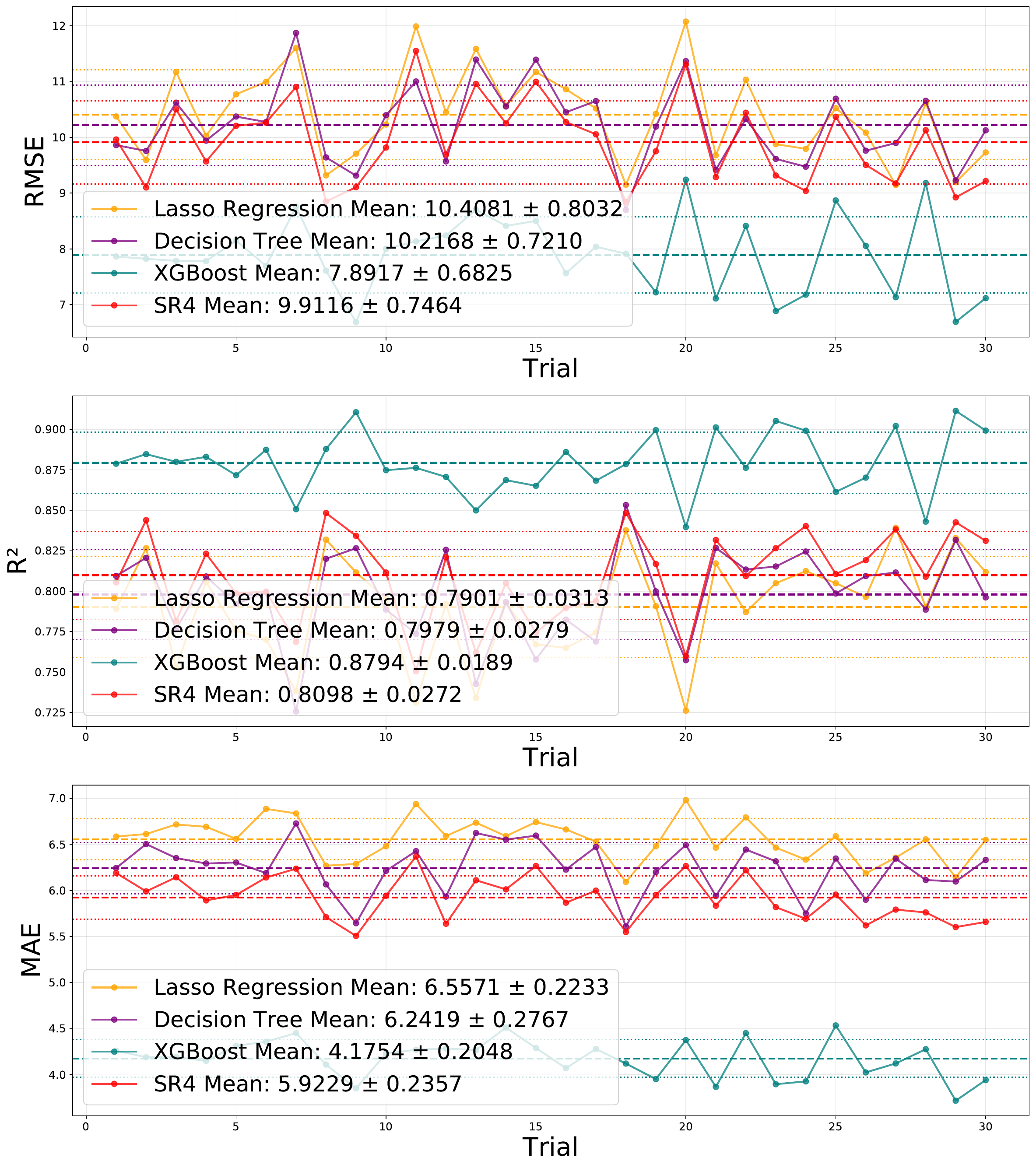}
\end{minipage}\hfill
\begin{minipage}{0.49\linewidth}
    \centering
    \includegraphics[width=\linewidth]{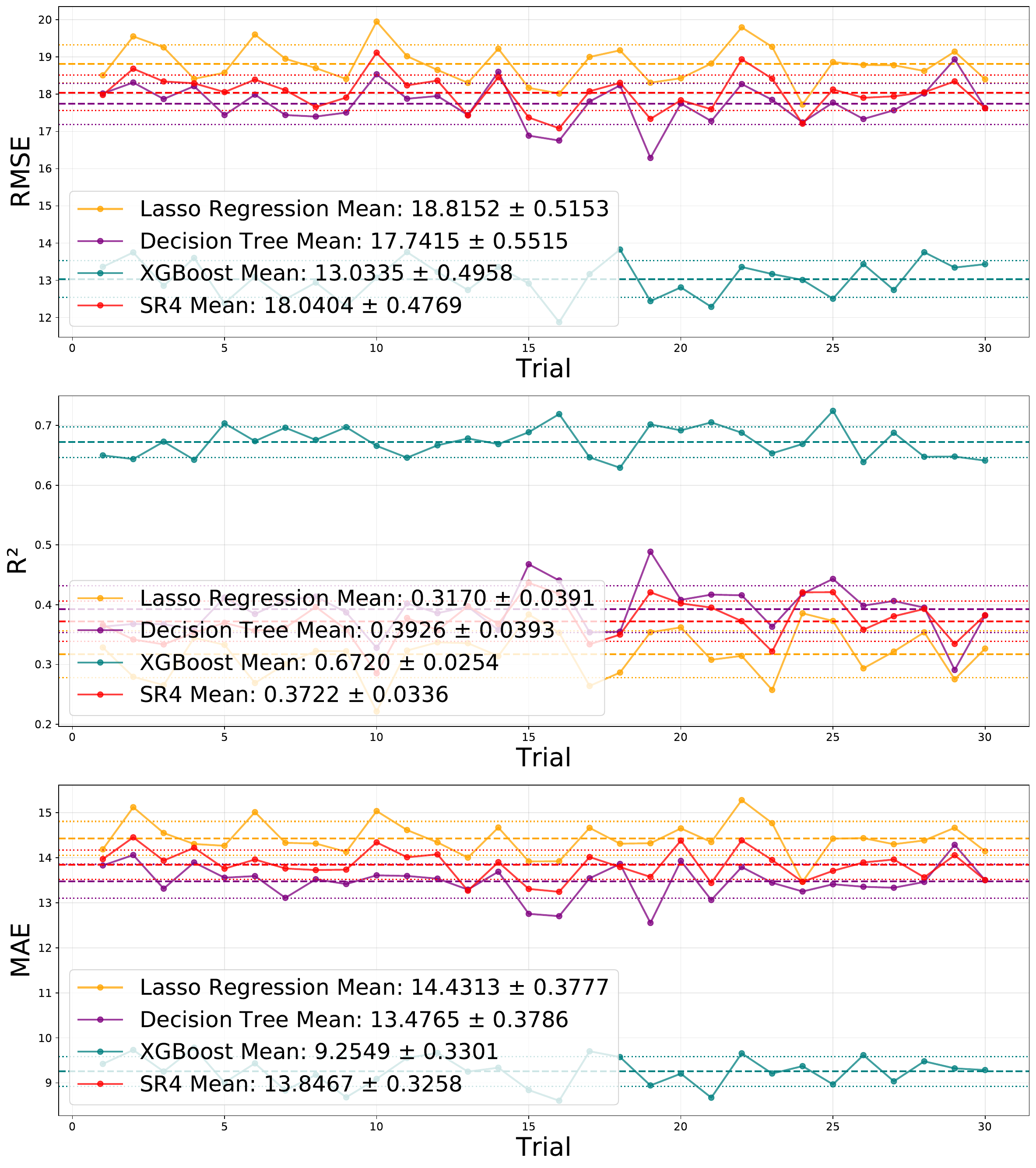}
\end{minipage}

\caption{
Line plot comparison of model (LASSO regression, decision tree, XG boost) performance metrics for minimum data across 30 trials. The left panel reports results obtained with party percentage (DEM) included, whereas the right panel shows results with party percentage (DEM) not included.
}
\label{fig:lineplot_grid_min_rep_other}
\end{figure}

\begin{figure}
\centering
\begin{minipage}{0.49\linewidth}
    \centering
    \includegraphics[width=\linewidth]{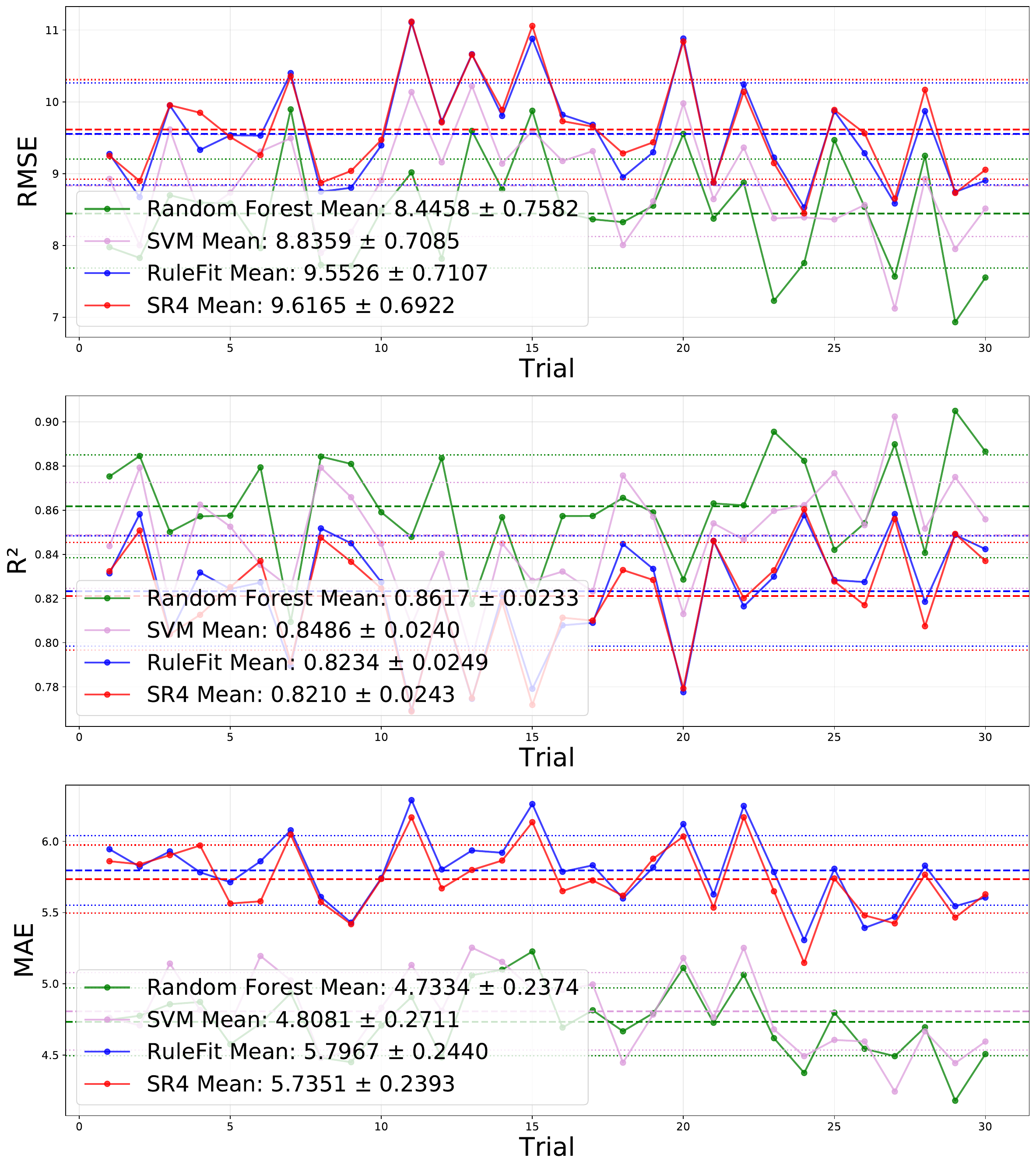}
\end{minipage}\hfill
\begin{minipage}{0.49\linewidth}
    \centering
    \includegraphics[width=\linewidth]{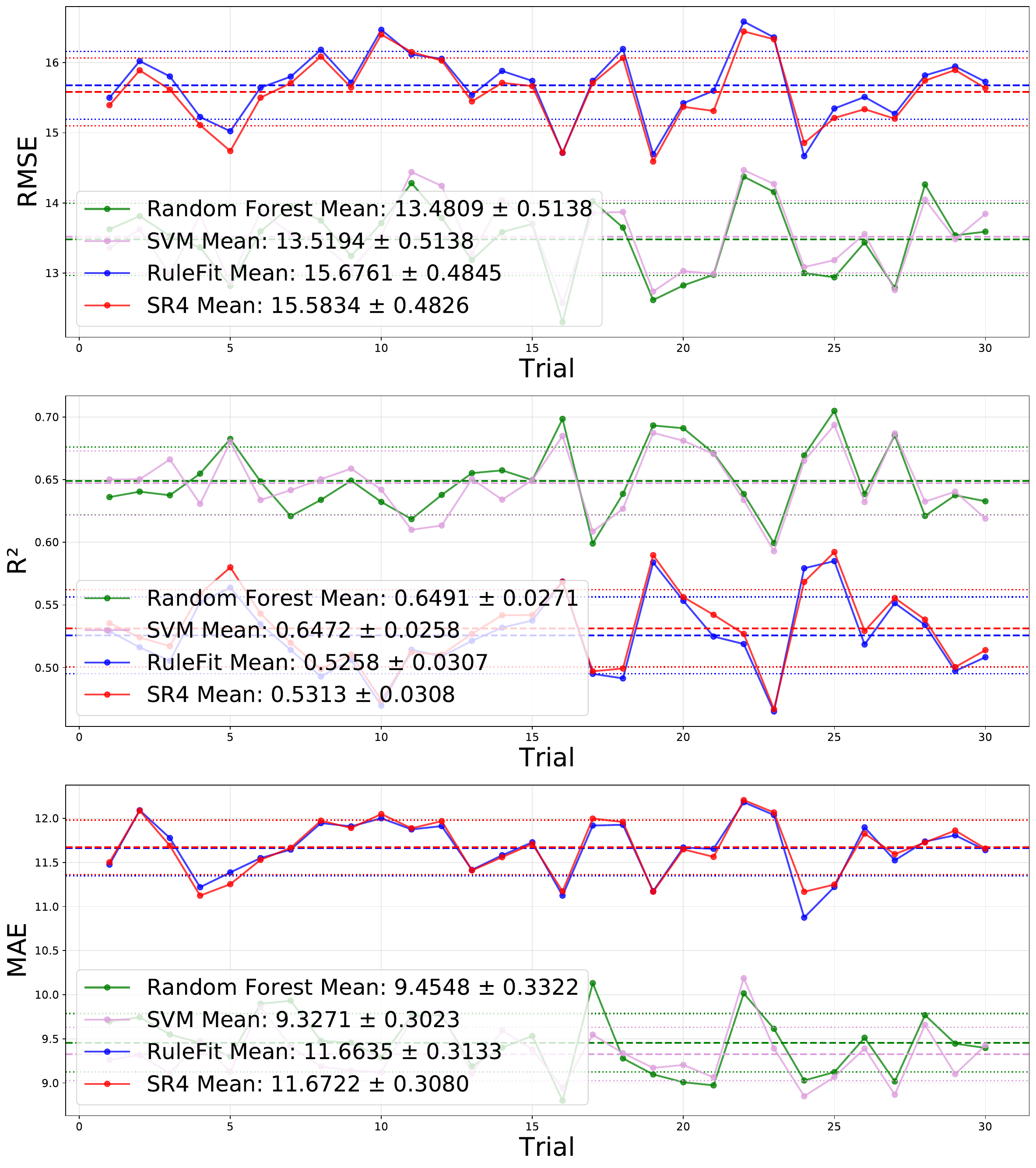}
\end{minipage}

\caption{
Line plot comparison of model performance metrics for standard data across 30 trials with REP percentage as label. The left panel reports results obtained with party percentage (DEM) included, whereas the right panel shows results with party percentage (DEM) not included.
}
\label{fig:lineplot_grid_std_rep}
\end{figure}

\begin{figure}
\centering
\begin{minipage}{0.49\linewidth}
    \centering
    \includegraphics[width=\linewidth]{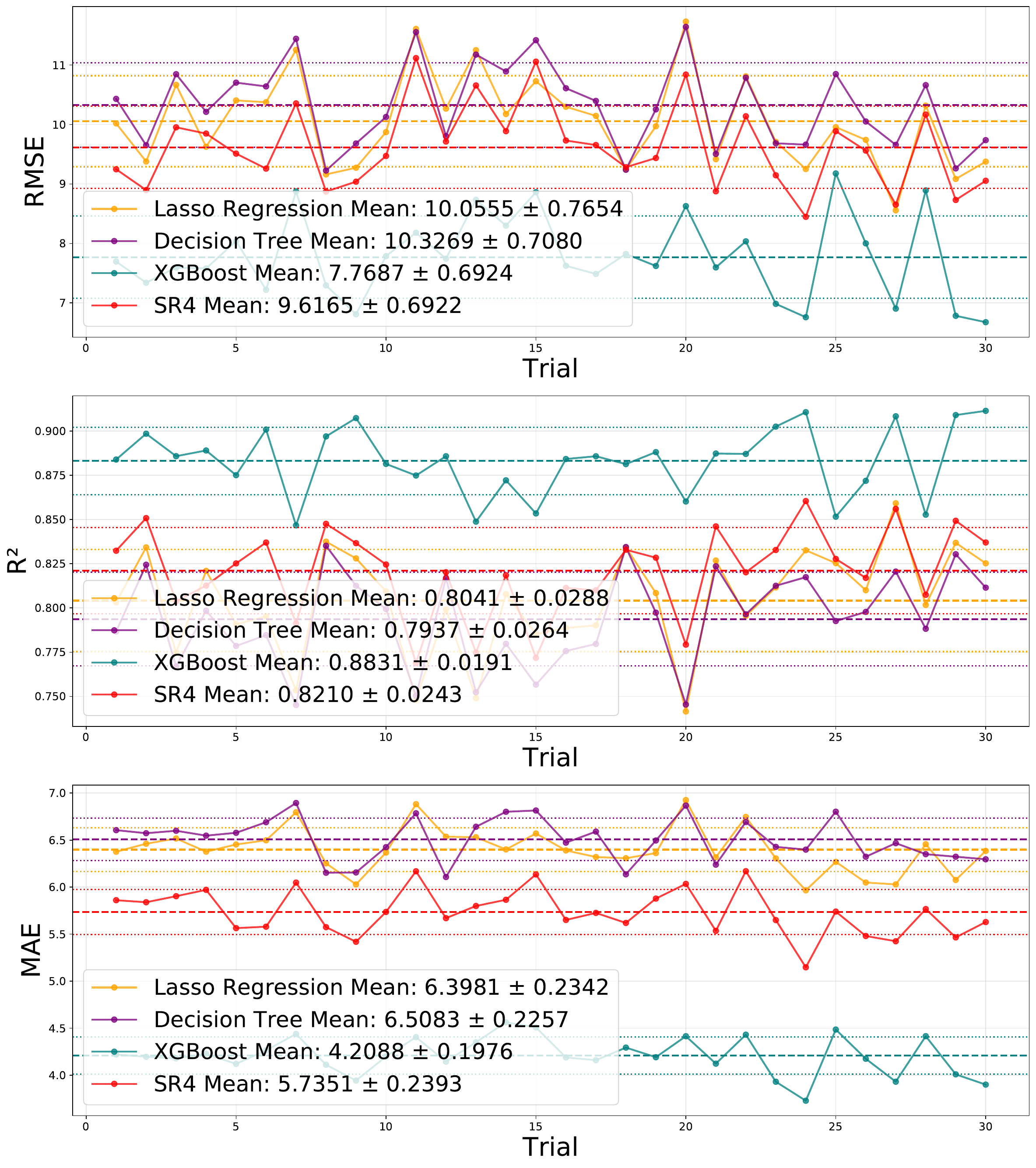}
\end{minipage}\hfill
\begin{minipage}{0.49\linewidth}
    \centering
    \includegraphics[width=\linewidth]{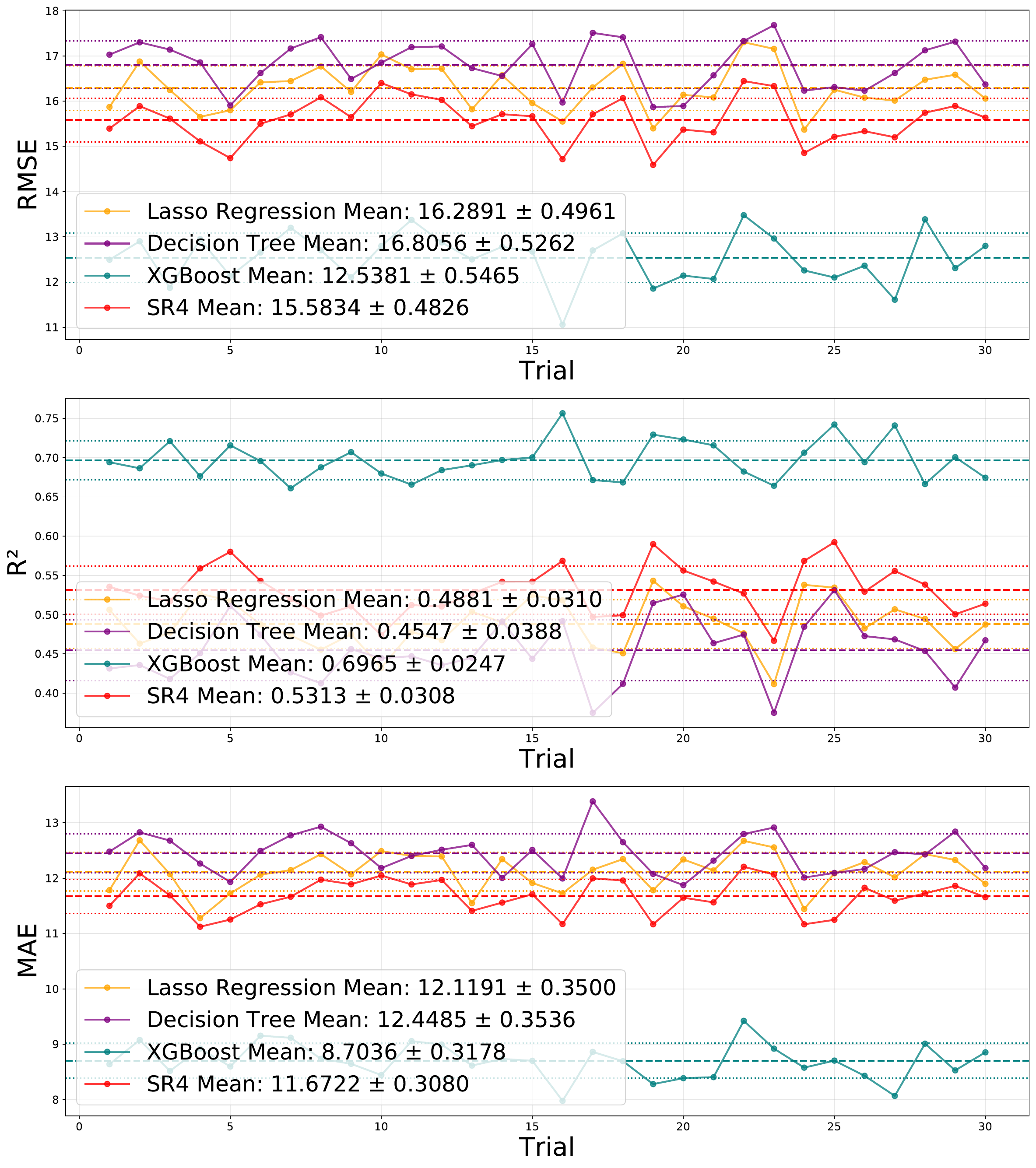}
\end{minipage}

\caption{
Line plot comparison of model (LASSO regression, decision tree, XG boost) performance metrics for standard data across 30 trials. The left panel reports results obtained with party percentage (DEM) included, whereas the right panel shows results with party percentage (DEM) not included.
}
\label{fig:lineplot_grid_std_rep_other}
\end{figure}

\begin{figure}
\centering
\begin{minipage}{0.49\linewidth}
    \centering
    \includegraphics[width=\linewidth]{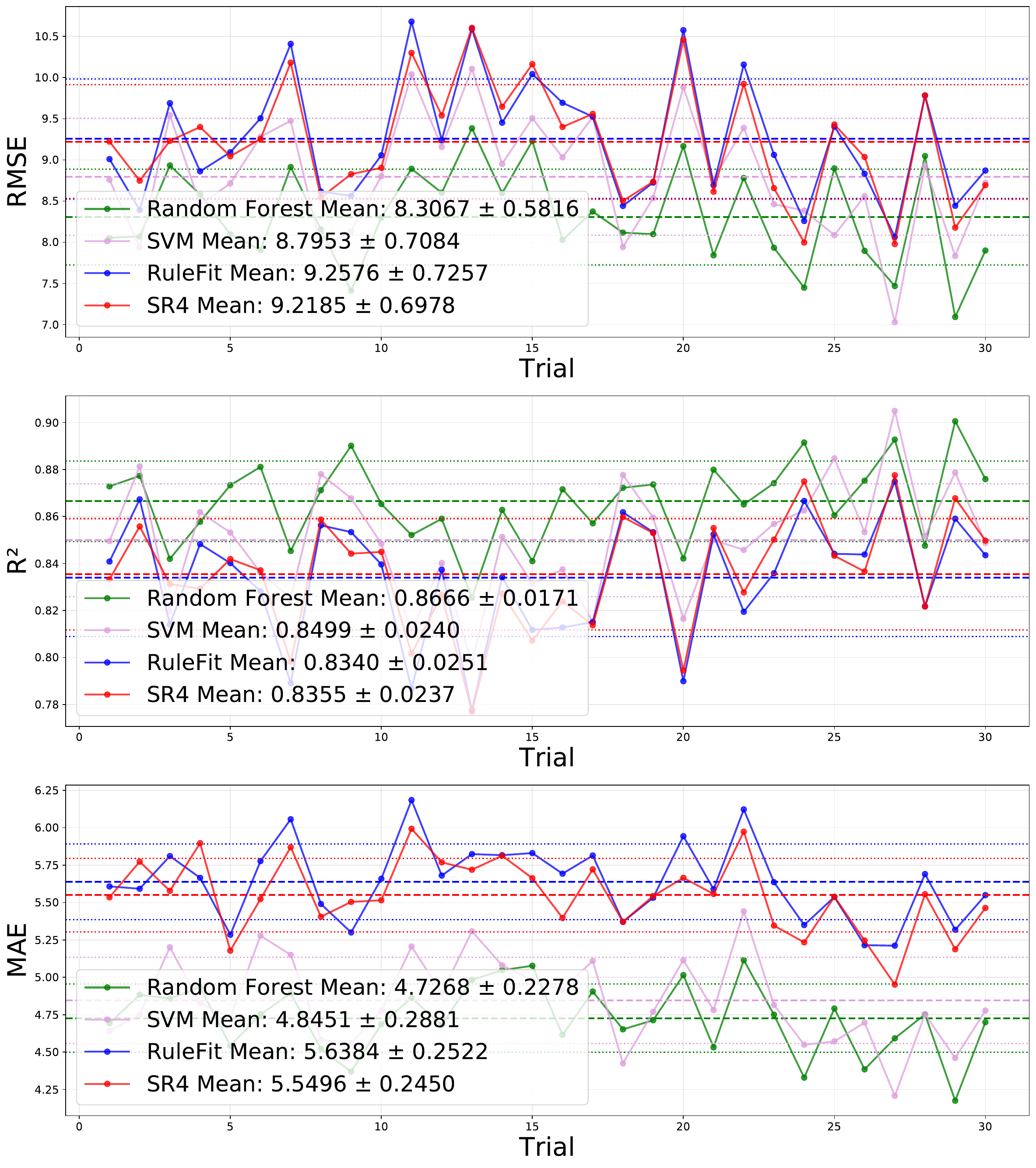}
\end{minipage}\hfill
\begin{minipage}{0.49\linewidth}
    \centering
    \includegraphics[width=\linewidth]{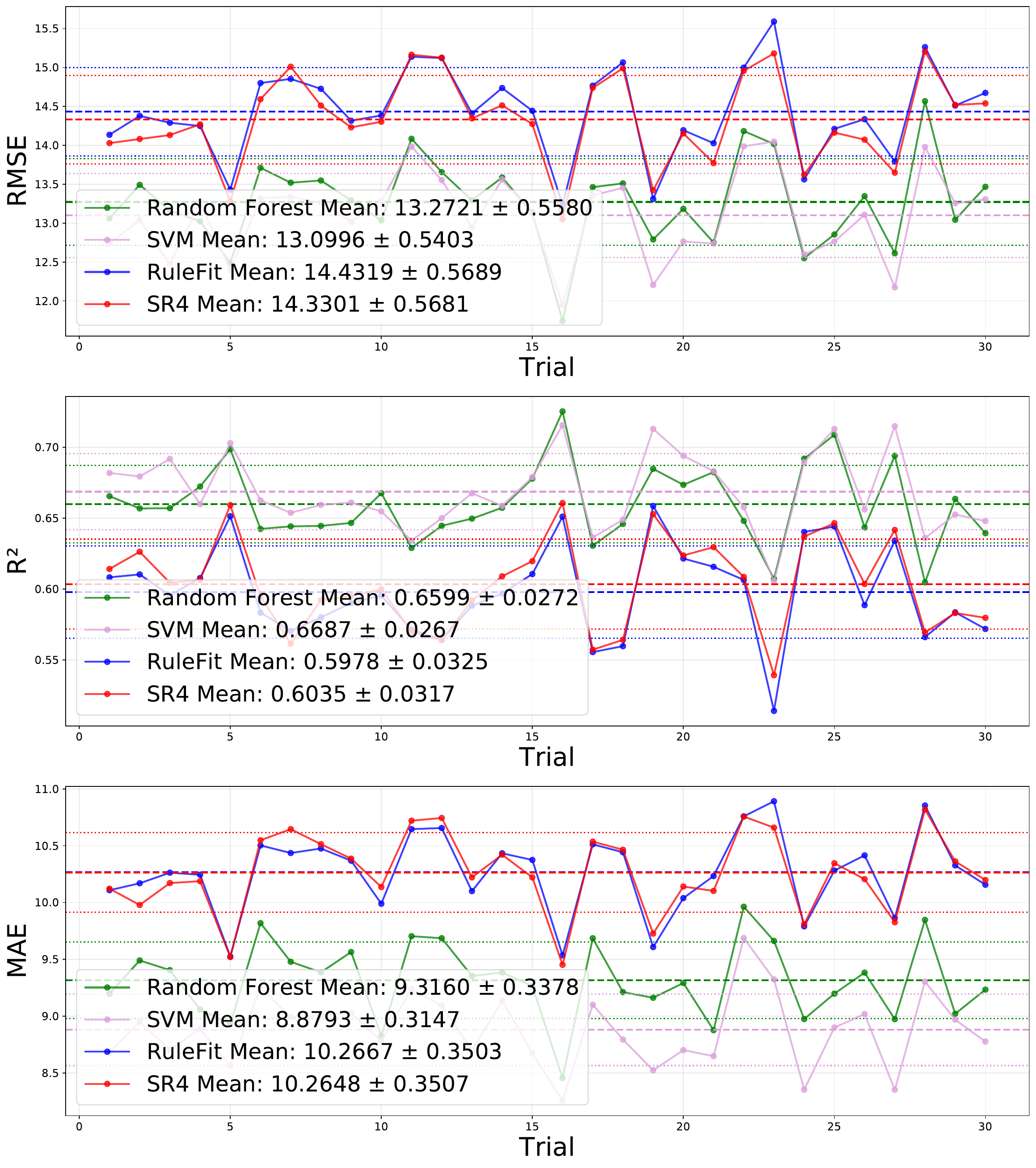}
\end{minipage}

\caption{
Line plot comparison of model performance metrics for expanded data across 30 trials with REP percentage as label. The left panel reports results obtained with party percentage (DEM) included, whereas the right panel shows results with party percentage (DEM) not included.
}
\label{fig:lineplot_grid_exp_rep}
\end{figure}

\begin{figure}
\centering
\begin{minipage}{0.49\linewidth}
    \centering
    \includegraphics[width=\linewidth]{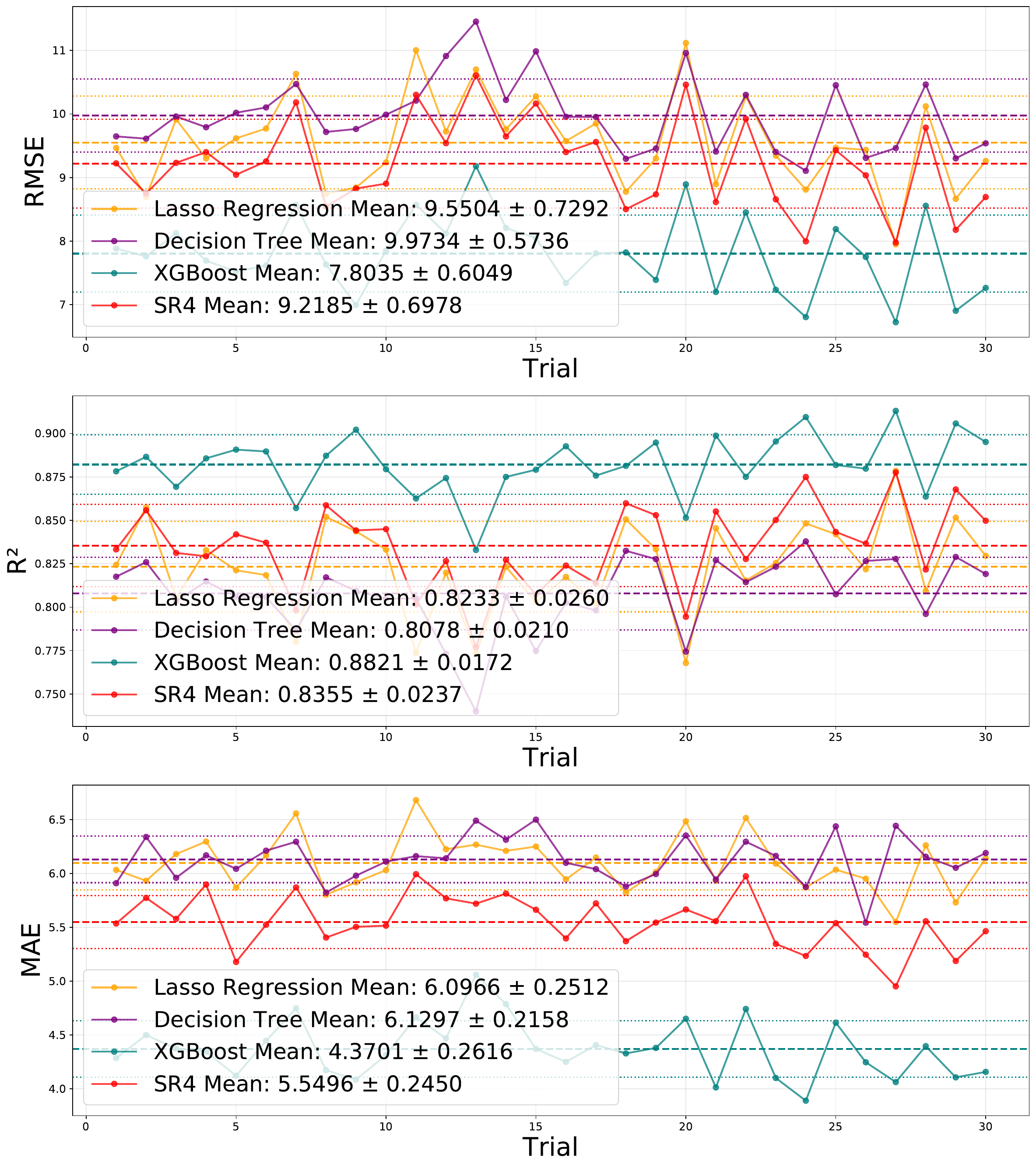}
\end{minipage}\hfill
\begin{minipage}{0.49\linewidth}
    \centering
    \includegraphics[width=\linewidth]{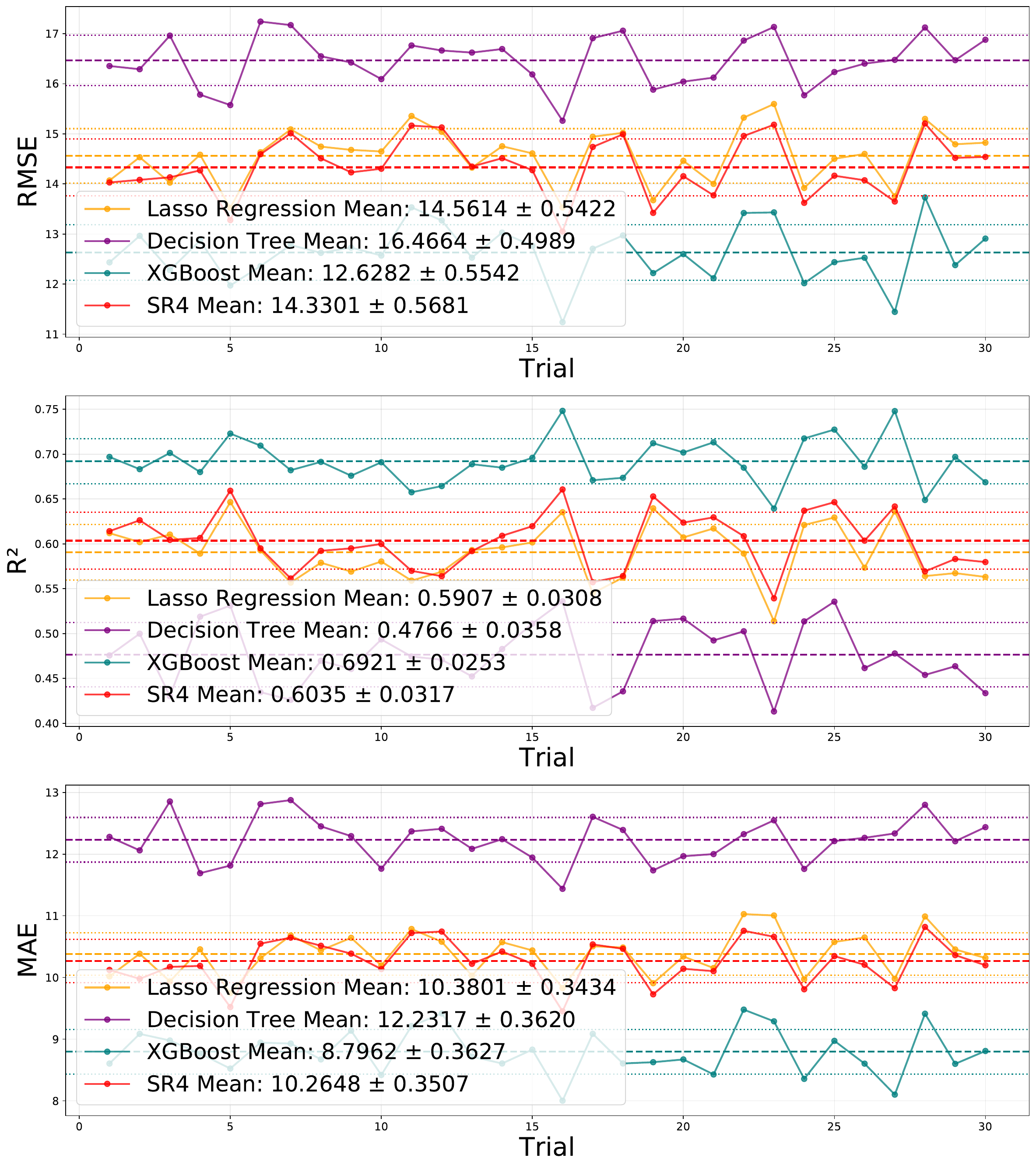}
\end{minipage}

\caption{
Line plot comparison of model (LASSO regression, decision tree, XG boost) performance metrics for expanded data across 30 trials. The left panel reports results obtained with party percentage (DEM) included, whereas the right panel shows results with party percentage (DEM) not included.
}
\label{fig:lineplot_grid_exp_rep_other}
\end{figure}

\begin{figure}
\centering
\begin{minipage}{0.49\linewidth}
    \centering
    \includegraphics[width=\linewidth]{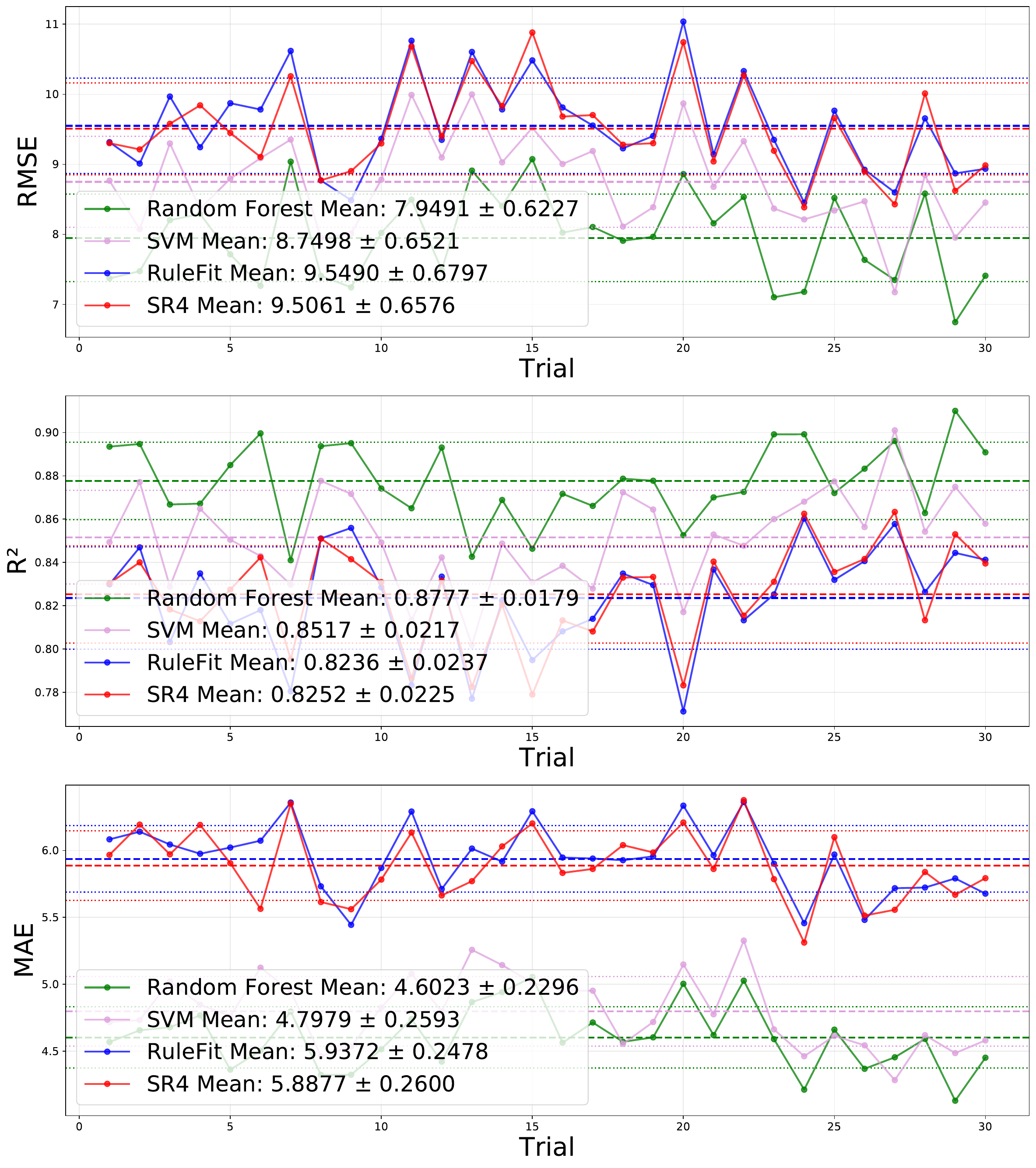}
\end{minipage}\hfill
\begin{minipage}{0.49\linewidth}
    \centering
    \includegraphics[width=\linewidth]{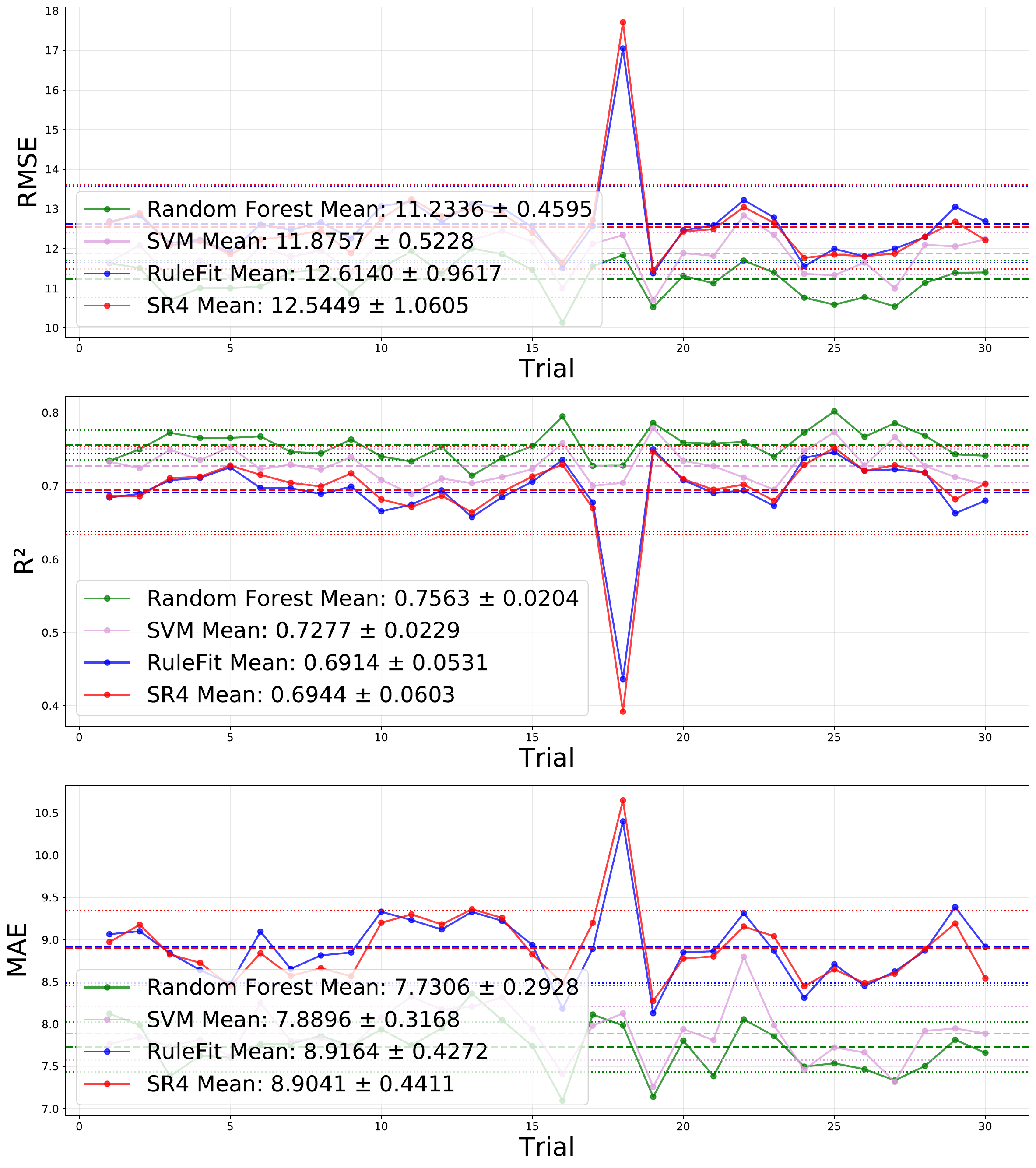}
\end{minipage}

\caption{
Line plot comparison of model performance metrics for previous data across 30 trials with REP percentage as label. The left panel reports results obtained with party percentage (DEM) included, whereas the right panel shows results with party percentage (DEM) not included.
}
\label{fig:lineplot_grid_pre_rep}
\end{figure}

\begin{figure}
\centering
\begin{minipage}{0.49\linewidth}
    \centering
    \includegraphics[width=\linewidth]{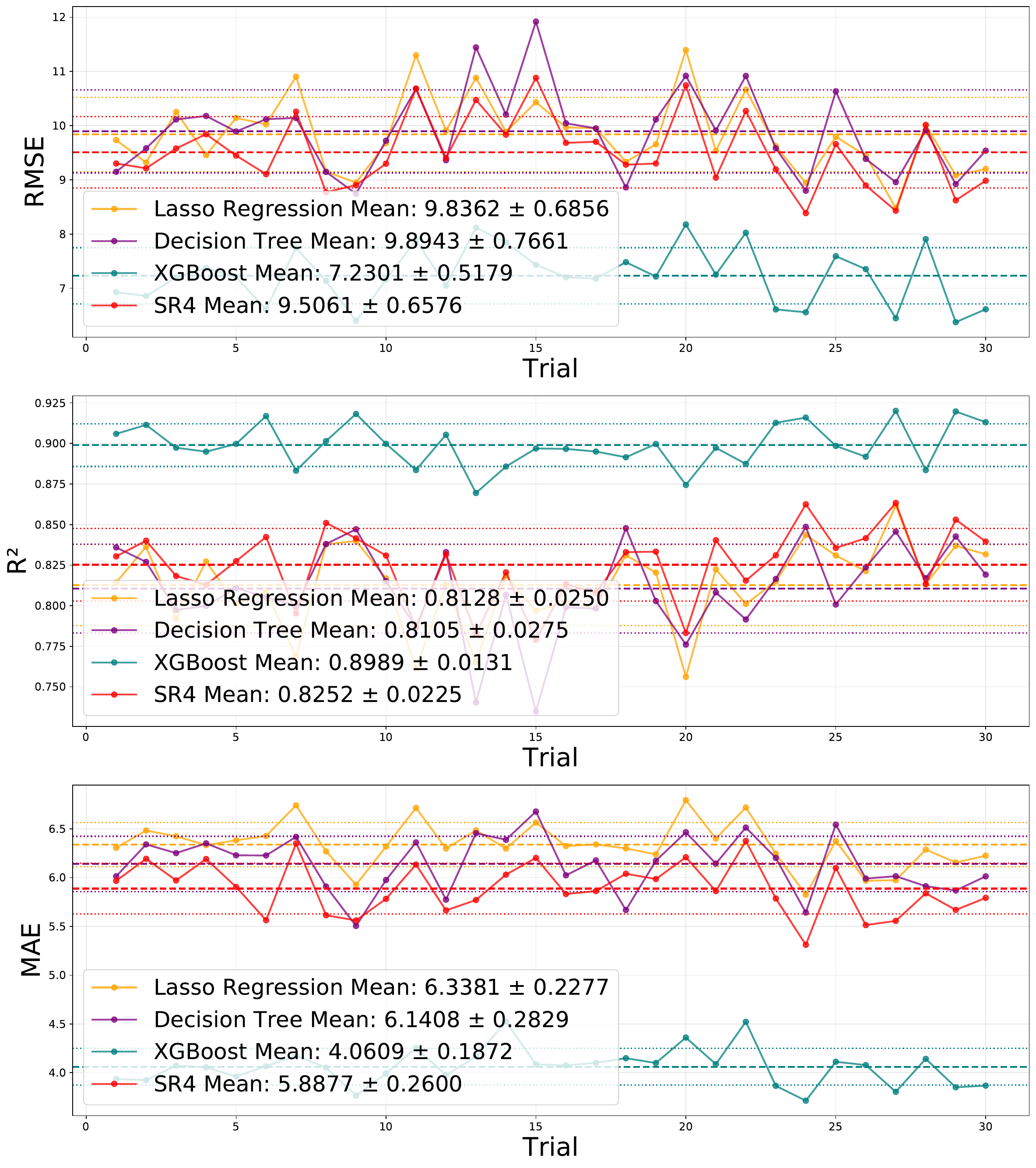}
\end{minipage}\hfill
\begin{minipage}{0.49\linewidth}
    \centering
    \includegraphics[width=\linewidth]{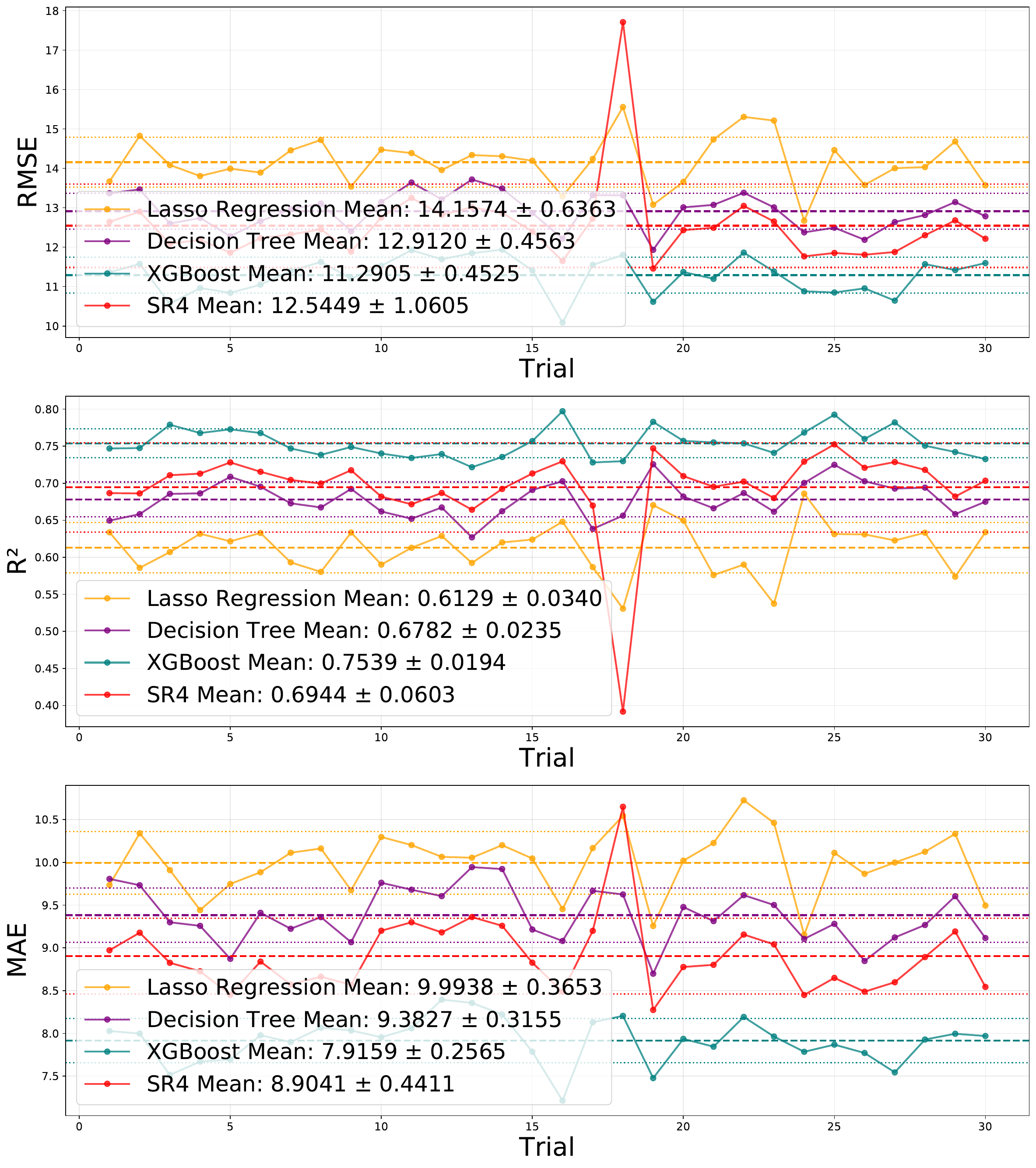}
\end{minipage}

\caption{
Line plot comparison of model (LASSO regression, decision tree, XG boost) performance metrics for previous data across 30 trials. The left panel reports results obtained with party percentage (DEM) included, whereas the right panel shows results with party percentage (DEM) not included.
}
\label{fig:lineplot_grid_pre_rep_other}
\end{figure}


\begin{figure}
\centering
\begin{minipage}{0.48\linewidth}
    \centering
    \includegraphics[width=\linewidth]{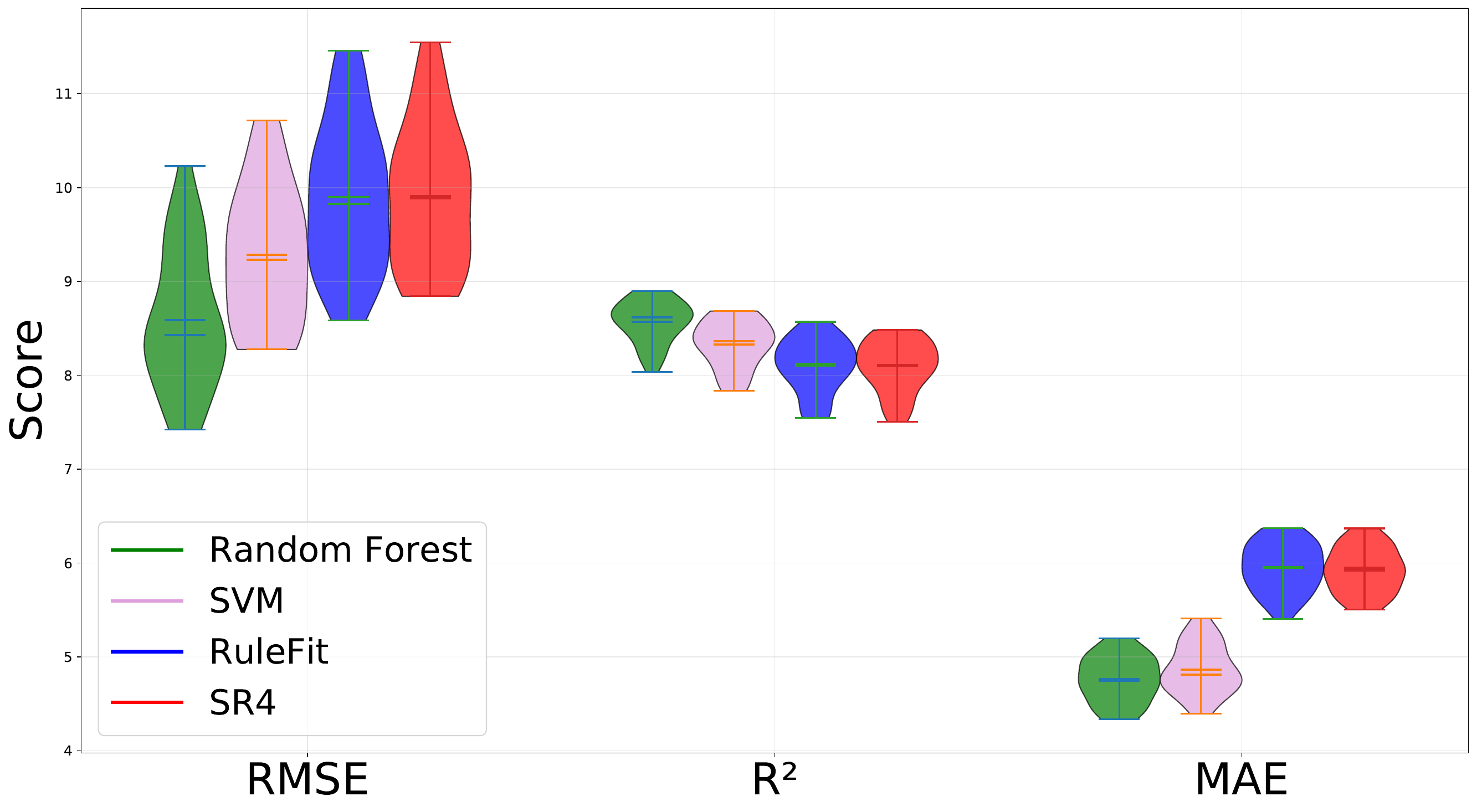}
    \subcaption{Minimum}
\end{minipage}\hfill
\begin{minipage}{0.48\linewidth}
    \centering
    \includegraphics[width=\linewidth]{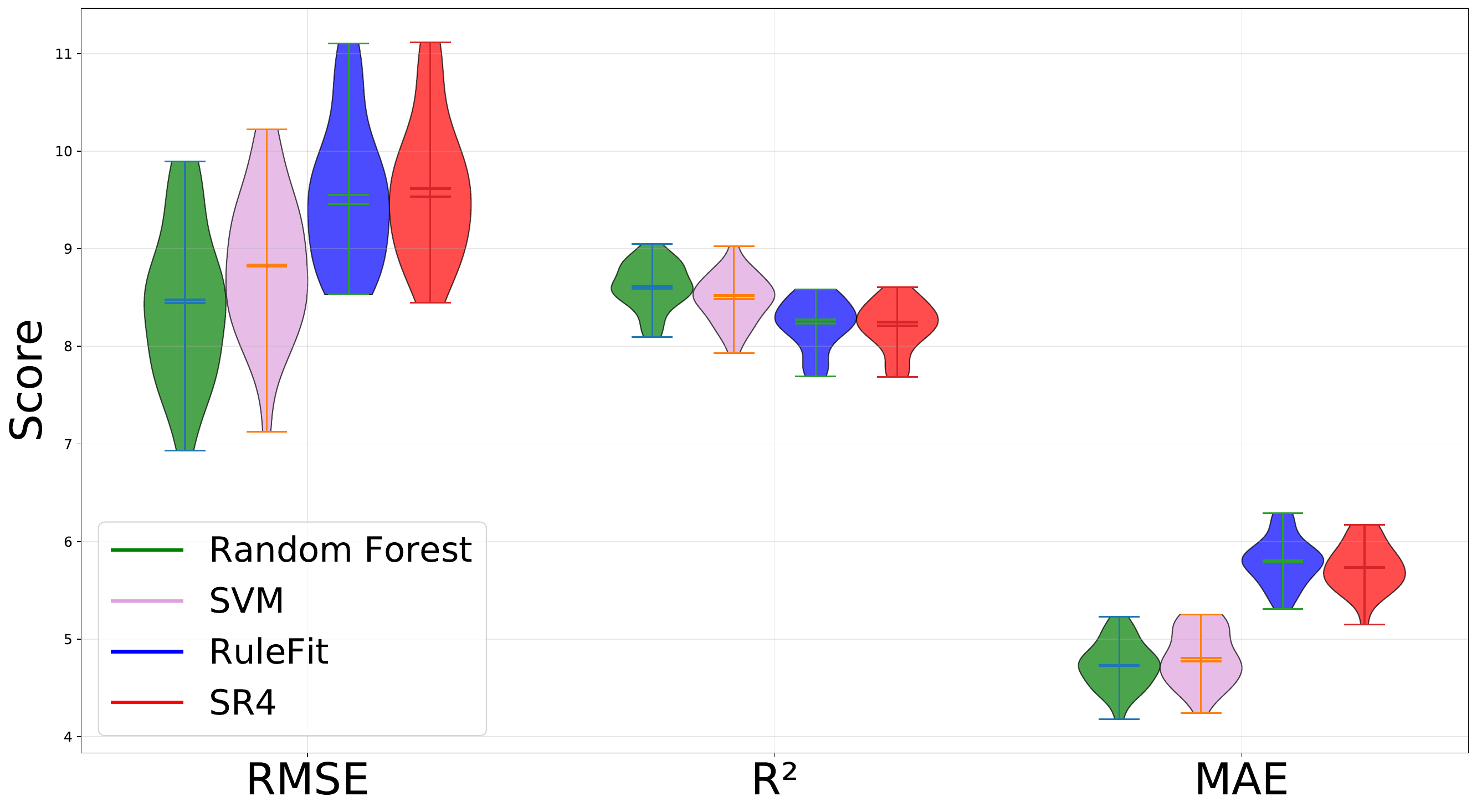}
    \subcaption{Standard}
\end{minipage}

\vspace{0.4cm}

\begin{minipage}{0.48\linewidth}
    \centering
    \includegraphics[width=\linewidth]{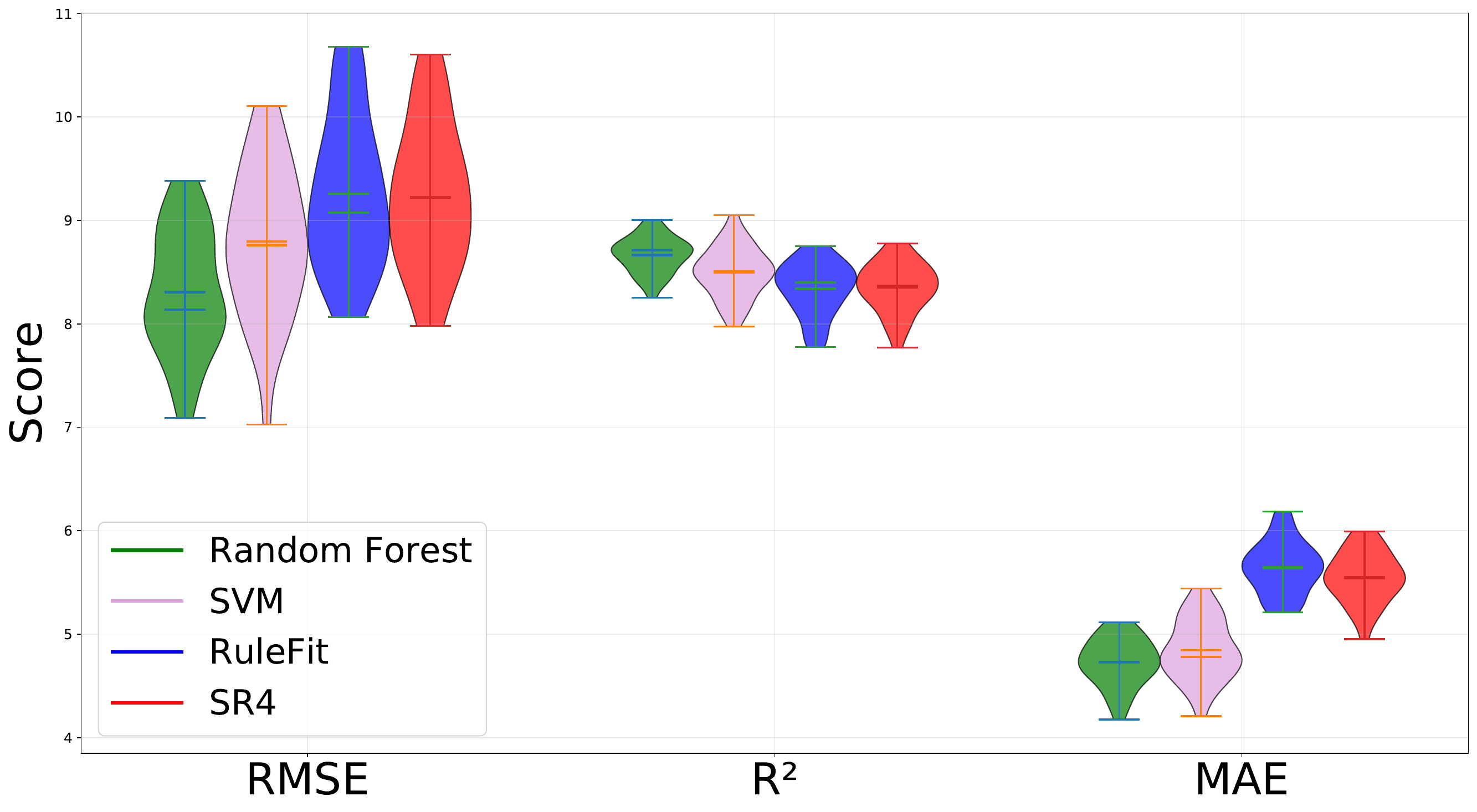}
    \subcaption{Expanded}
\end{minipage}\hfill
\begin{minipage}{0.48\linewidth}
    \centering
    \includegraphics[width=\linewidth]{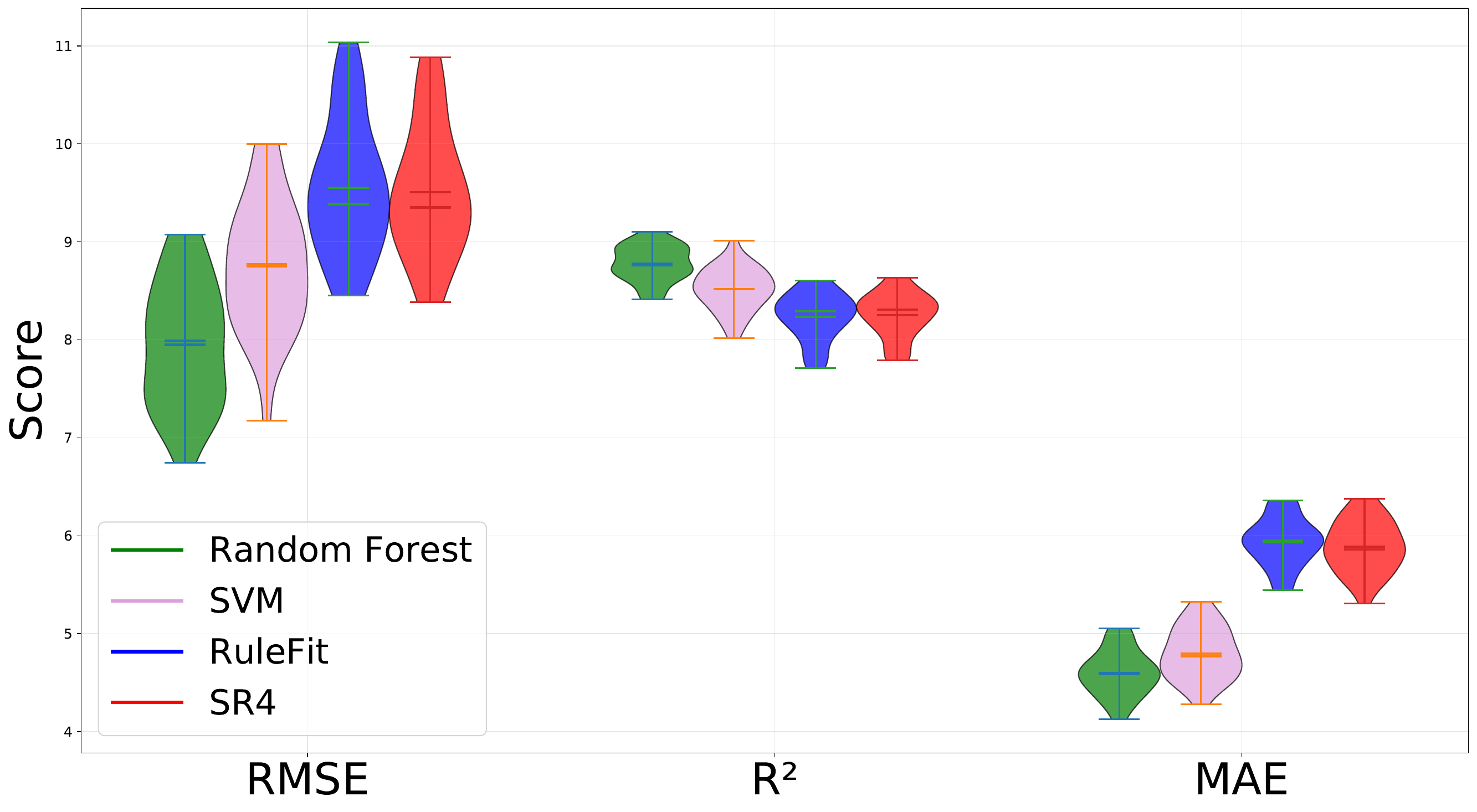}
    \subcaption{Previous}
\end{minipage}

\caption{
Violin plot comparison of model performance metrics across multiple datasets for the election regression task that has REP percentage as a label and includes party percentage (DEM).
}
\label{fig:violin_all_models_rep}
\end{figure}

\begin{figure}
\centering
\begin{minipage}{0.48\linewidth}
    \centering
    \includegraphics[width=\linewidth]{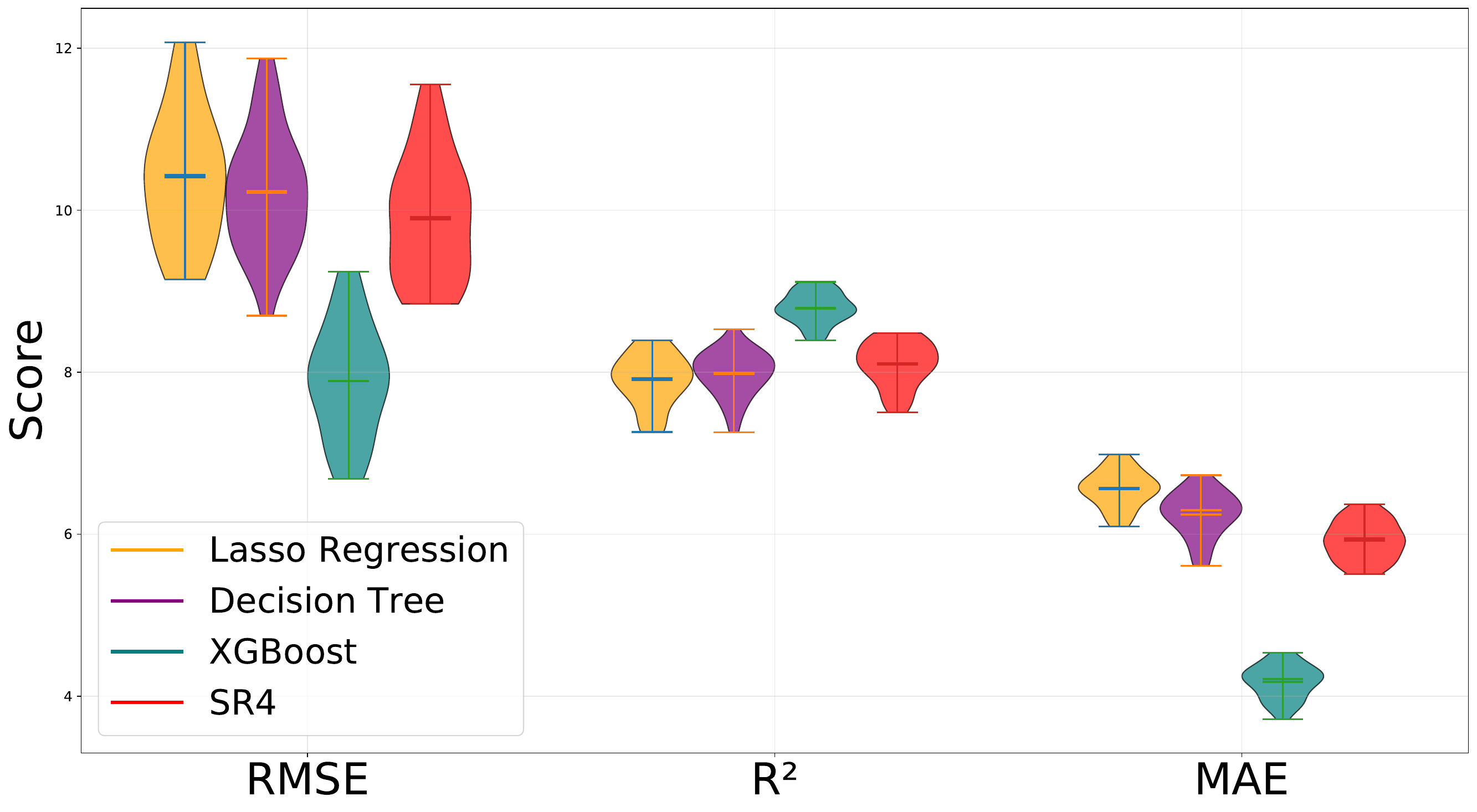}
    \subcaption{Minimum}
\end{minipage}\hfill
\begin{minipage}{0.48\linewidth}
    \centering
    \includegraphics[width=\linewidth]{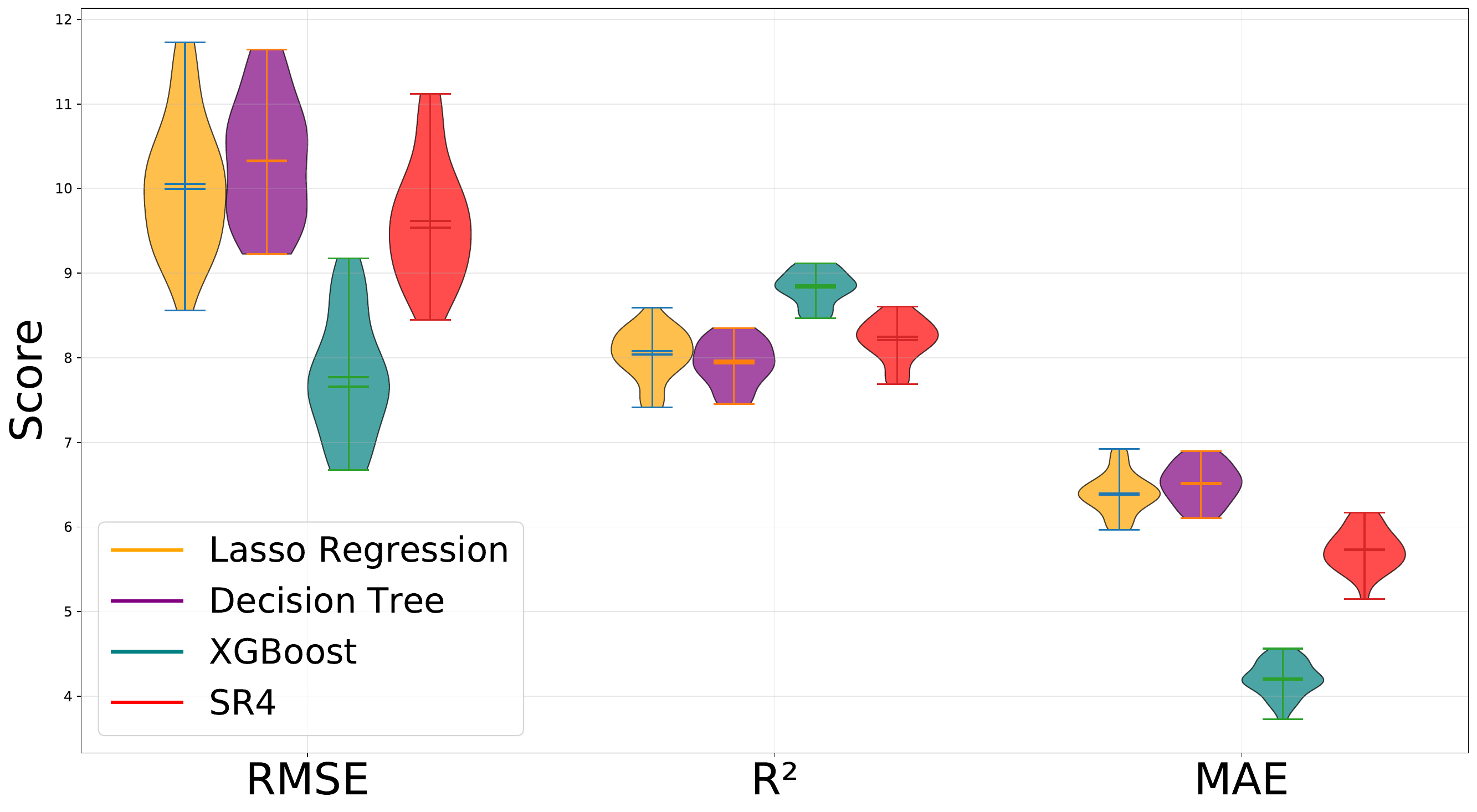}
    \subcaption{Standard}
\end{minipage}

\vspace{0.4cm}

\begin{minipage}{0.48\linewidth}
    \centering
    \includegraphics[width=\linewidth]{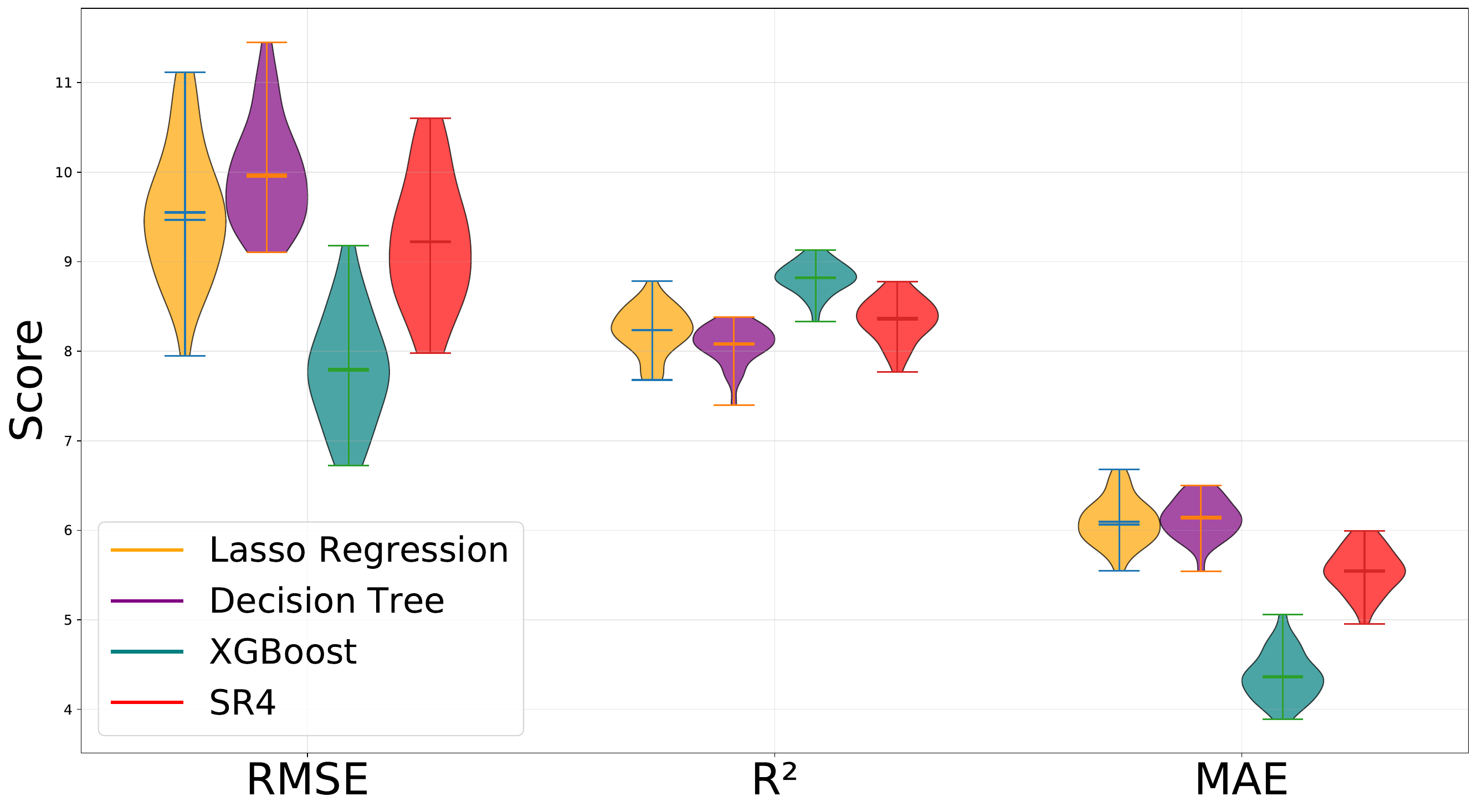}
    \subcaption{Expanded}
\end{minipage}\hfill
\begin{minipage}{0.48\linewidth}
    \centering
    \includegraphics[width=\linewidth]{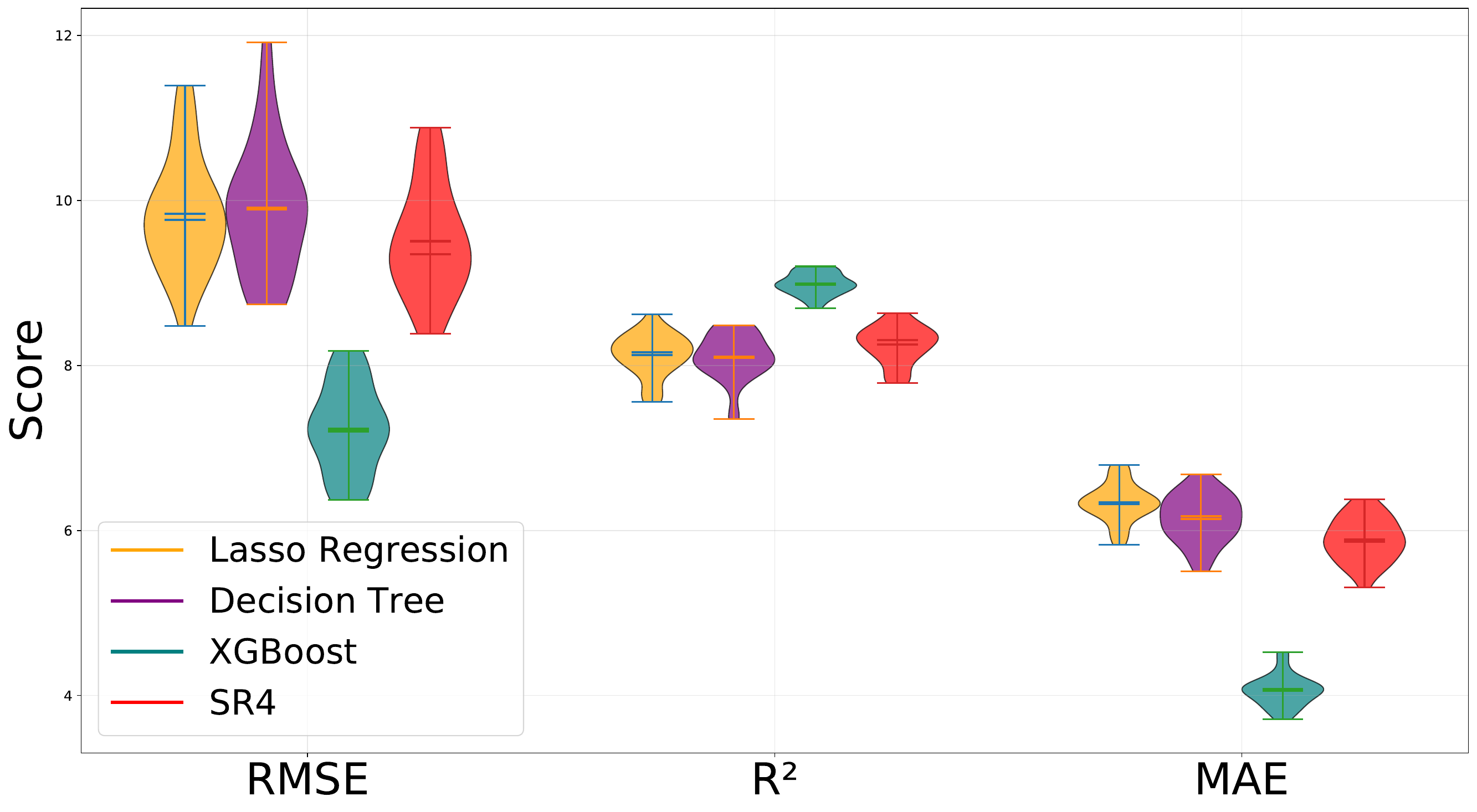}
    \subcaption{Previous}
\end{minipage}

\caption{
Violin plot comparison of model (LASSO regression, decision tree, XG boost) performance metrics across multiple datasets for the election regression task that has REP percentage as a label and includes party percentage (DEM).
}
\label{fig:violin_all_models_rep_other}
\end{figure}


\begin{figure}
\centering
\begin{minipage}{0.48\linewidth}
    \centering
    \includegraphics[width=\linewidth]{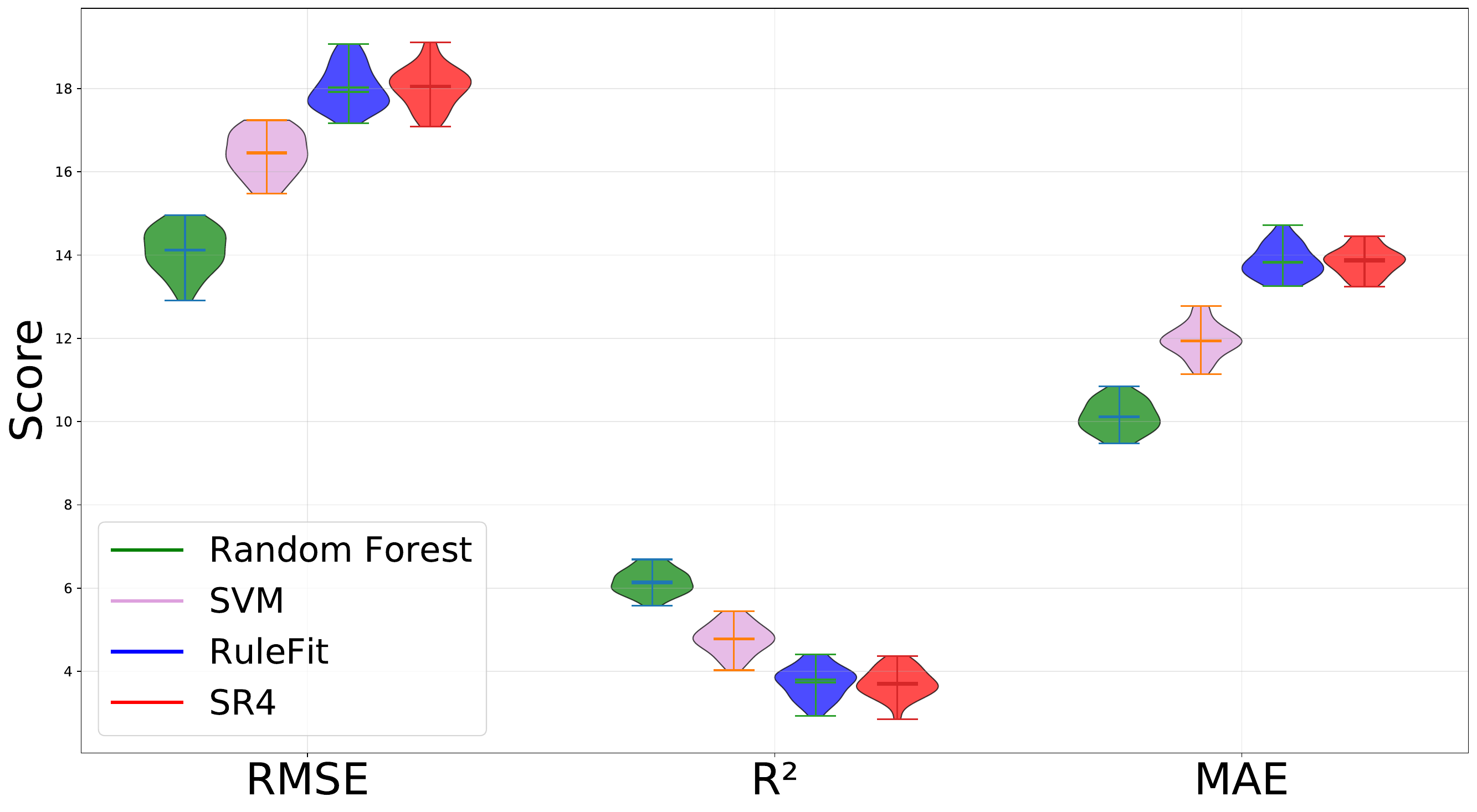}
    \subcaption{Minimum}
\end{minipage}\hfill
\begin{minipage}{0.48\linewidth}
    \centering
    \includegraphics[width=\linewidth]{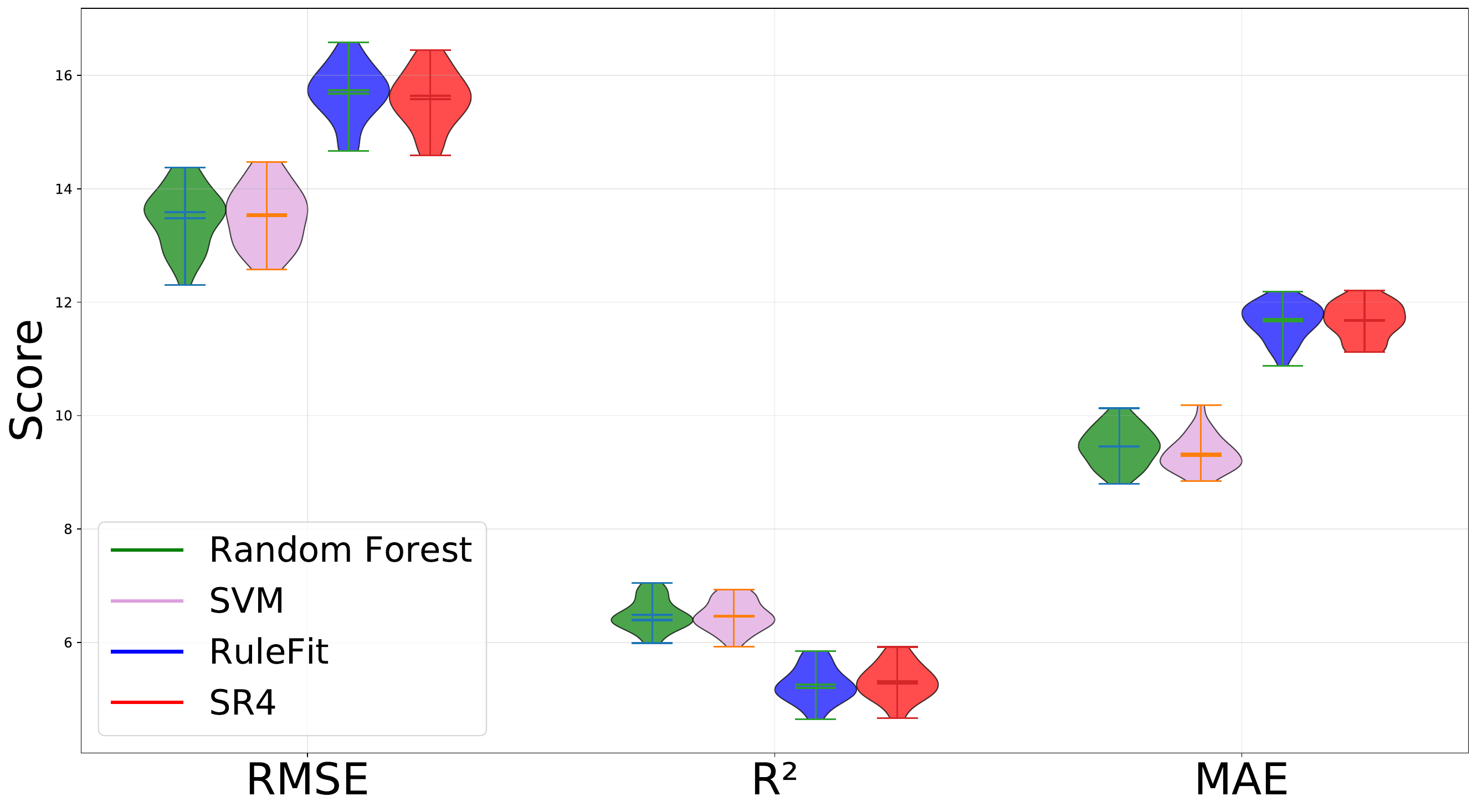}
    \subcaption{Standard}
\end{minipage}

\vspace{0.4cm}

\begin{minipage}{0.48\linewidth}
    \centering
    \includegraphics[width=\linewidth]{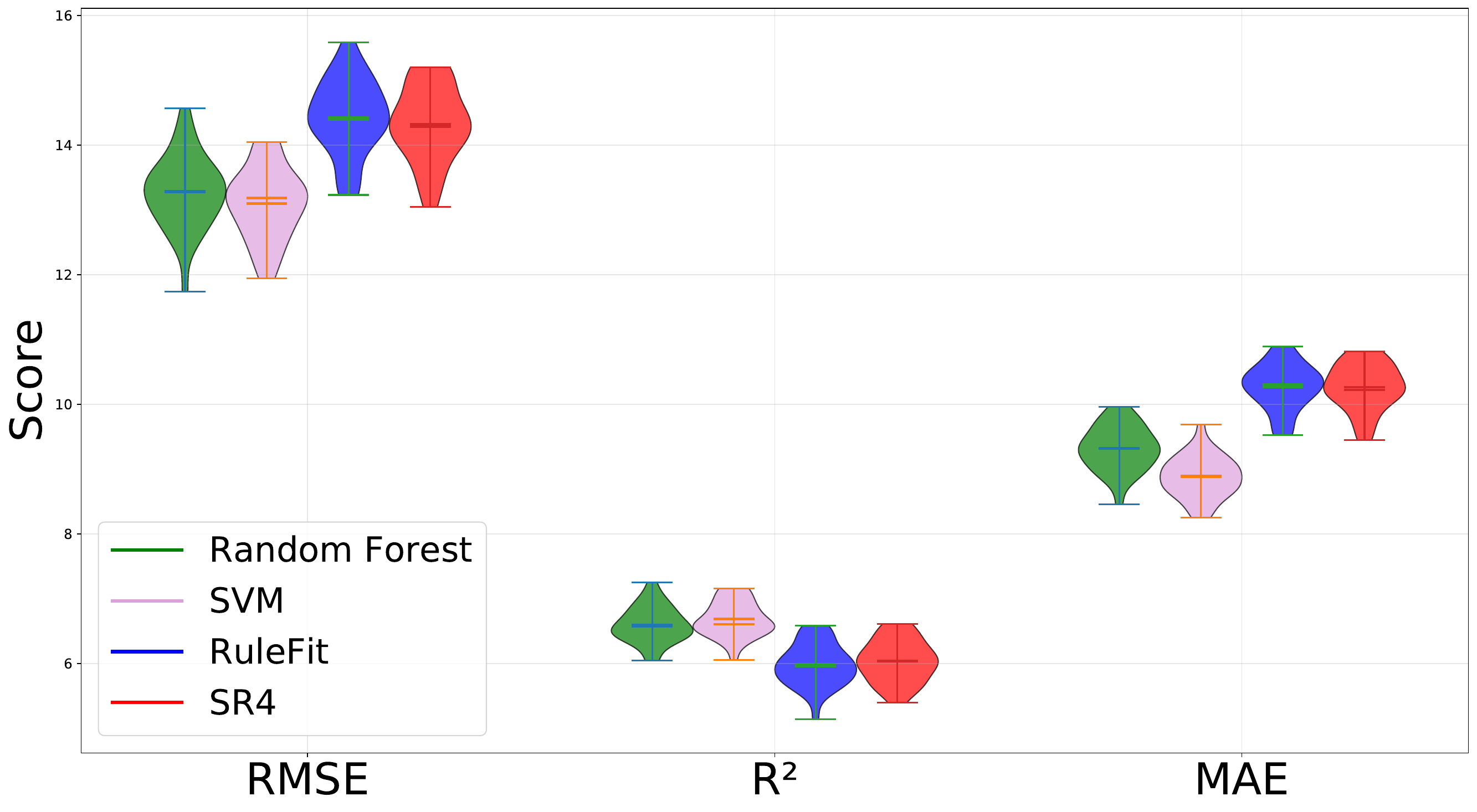}
    \subcaption{Expanded}
\end{minipage}\hfill
\begin{minipage}{0.48\linewidth}
    \centering
    \includegraphics[width=\linewidth]{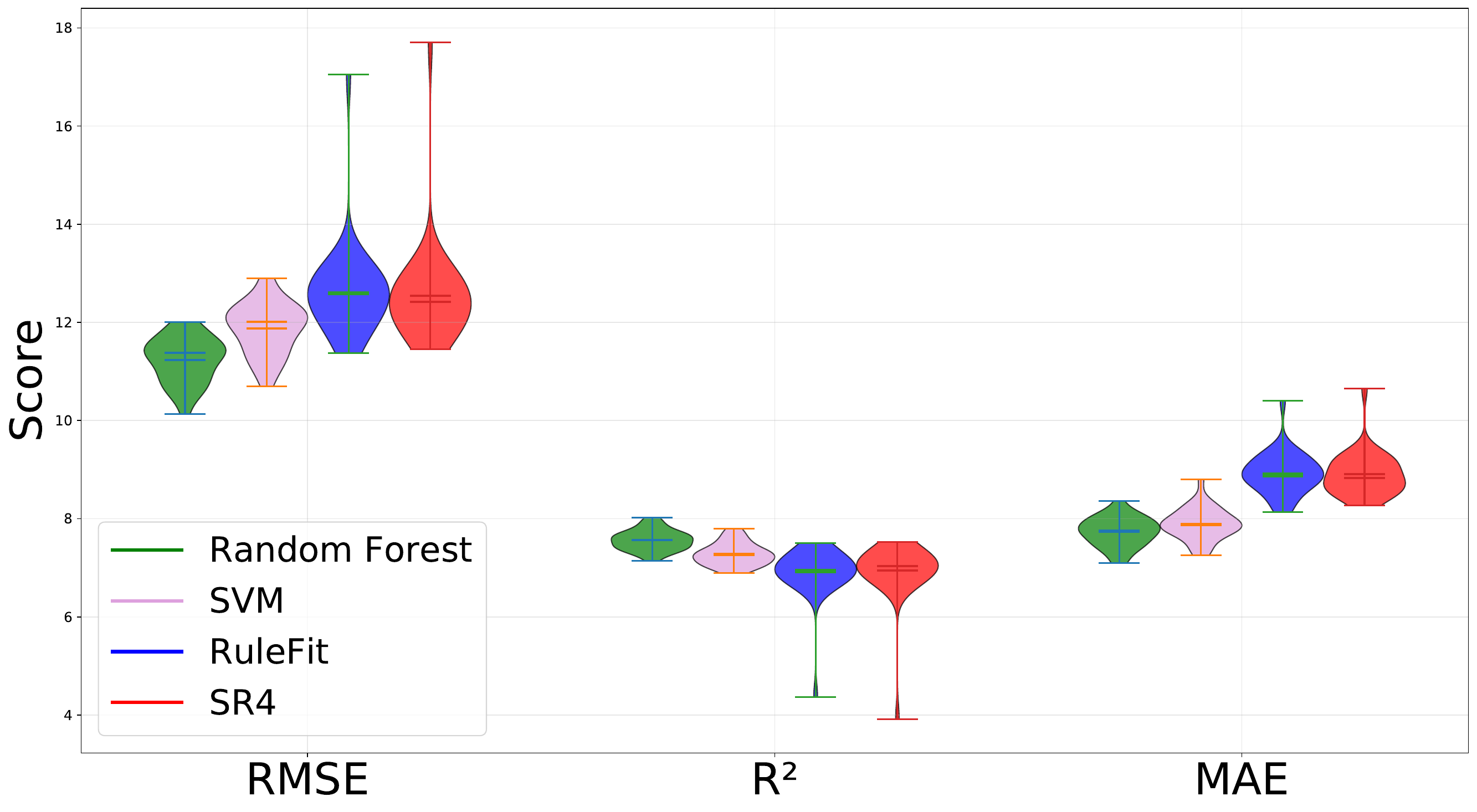}
    \subcaption{Previous}
\end{minipage}

\caption{
Violin plot comparison of model performance metrics across multiple datasets for the election regression task that has REP percentage as a label and does not include party percentage (DEM).
}
\label{fig:violin_all_models_rep_wpp}
\end{figure}

\begin{figure}
\centering
\begin{minipage}{0.48\linewidth}
    \centering
    \includegraphics[width=\linewidth]{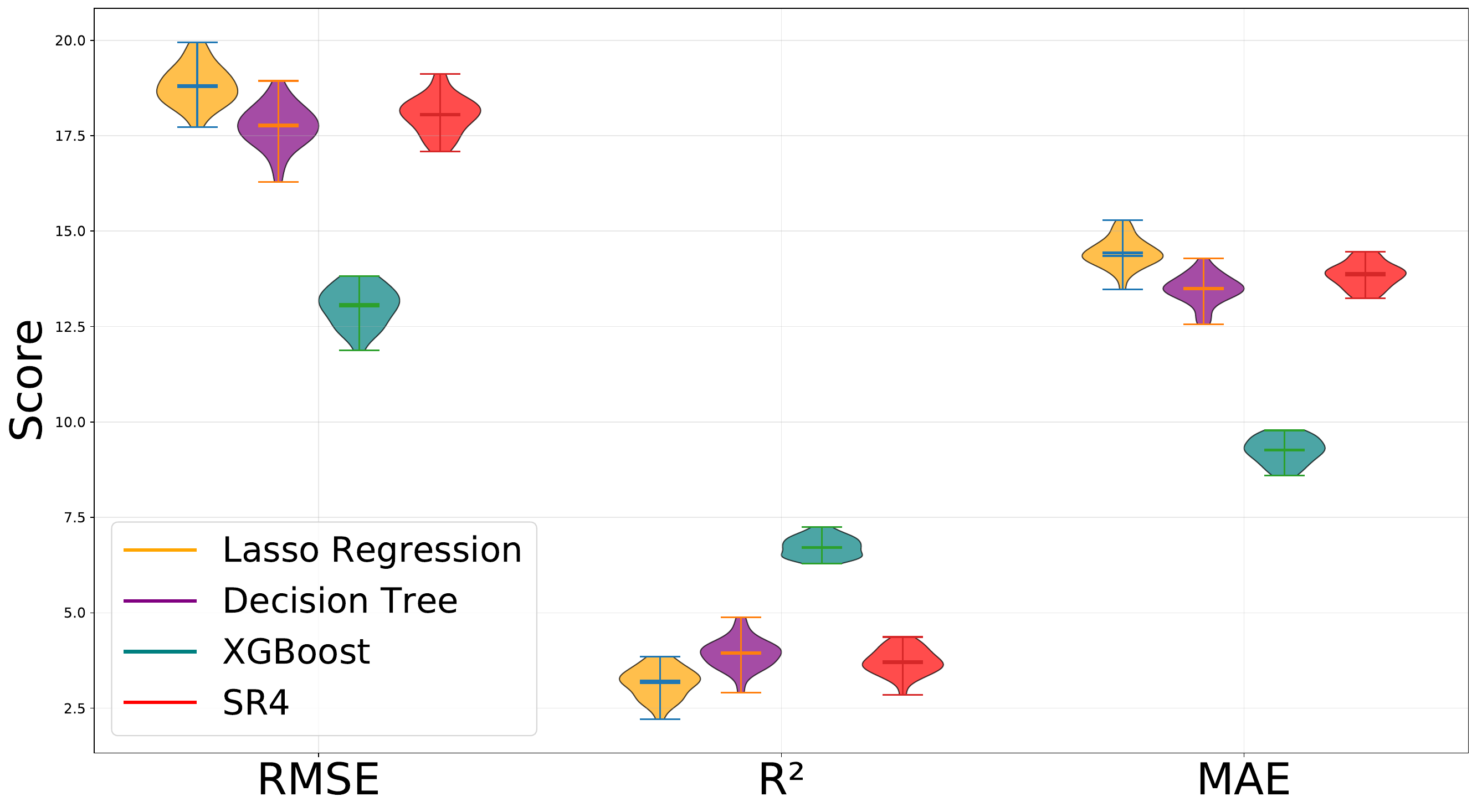}
    \subcaption{Minimum}
\end{minipage}\hfill
\begin{minipage}{0.48\linewidth}
    \centering
    \includegraphics[width=\linewidth]{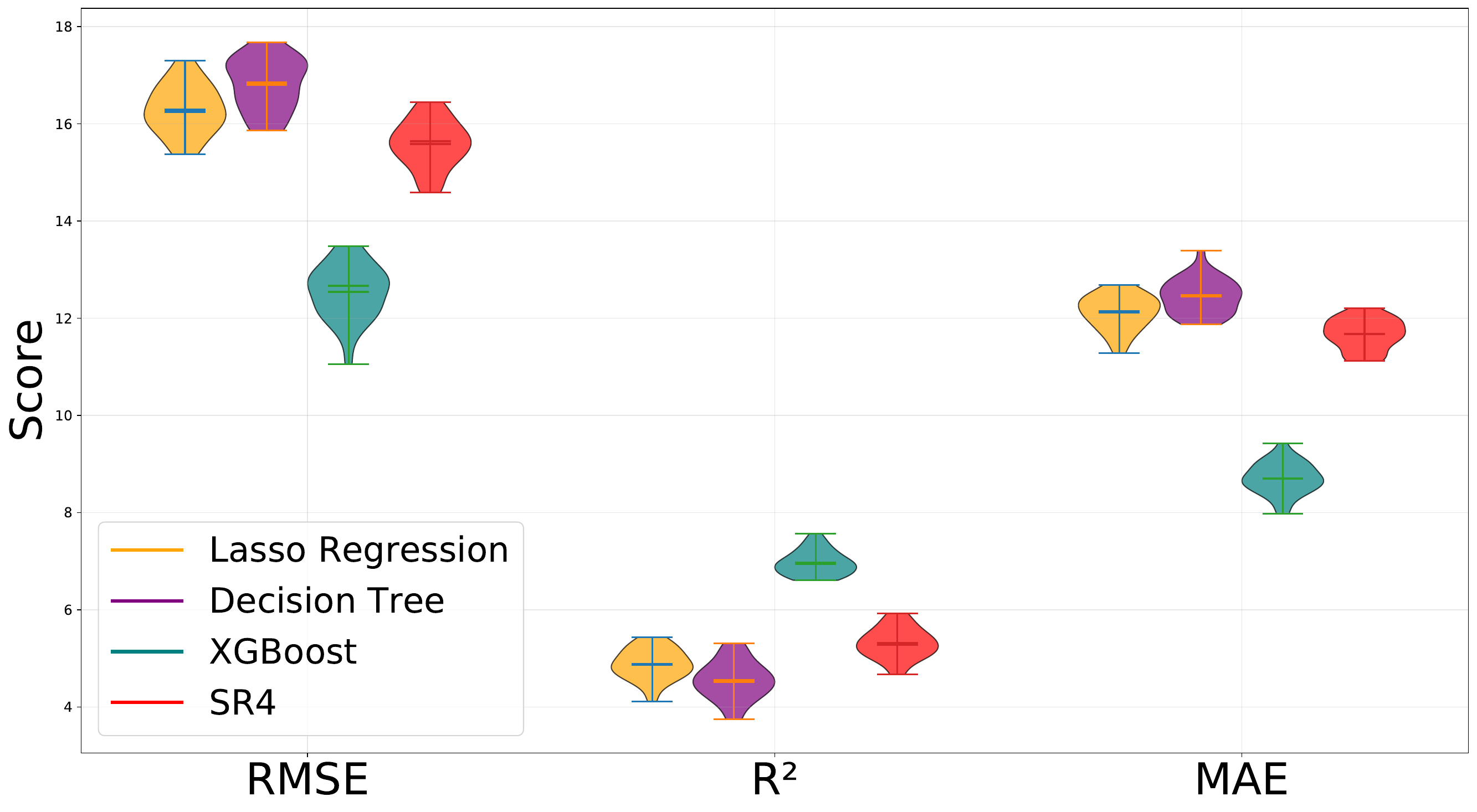}
    \subcaption{Standard}
\end{minipage}

\vspace{0.4cm}

\begin{minipage}{0.48\linewidth}
    \centering
    \includegraphics[width=\linewidth]{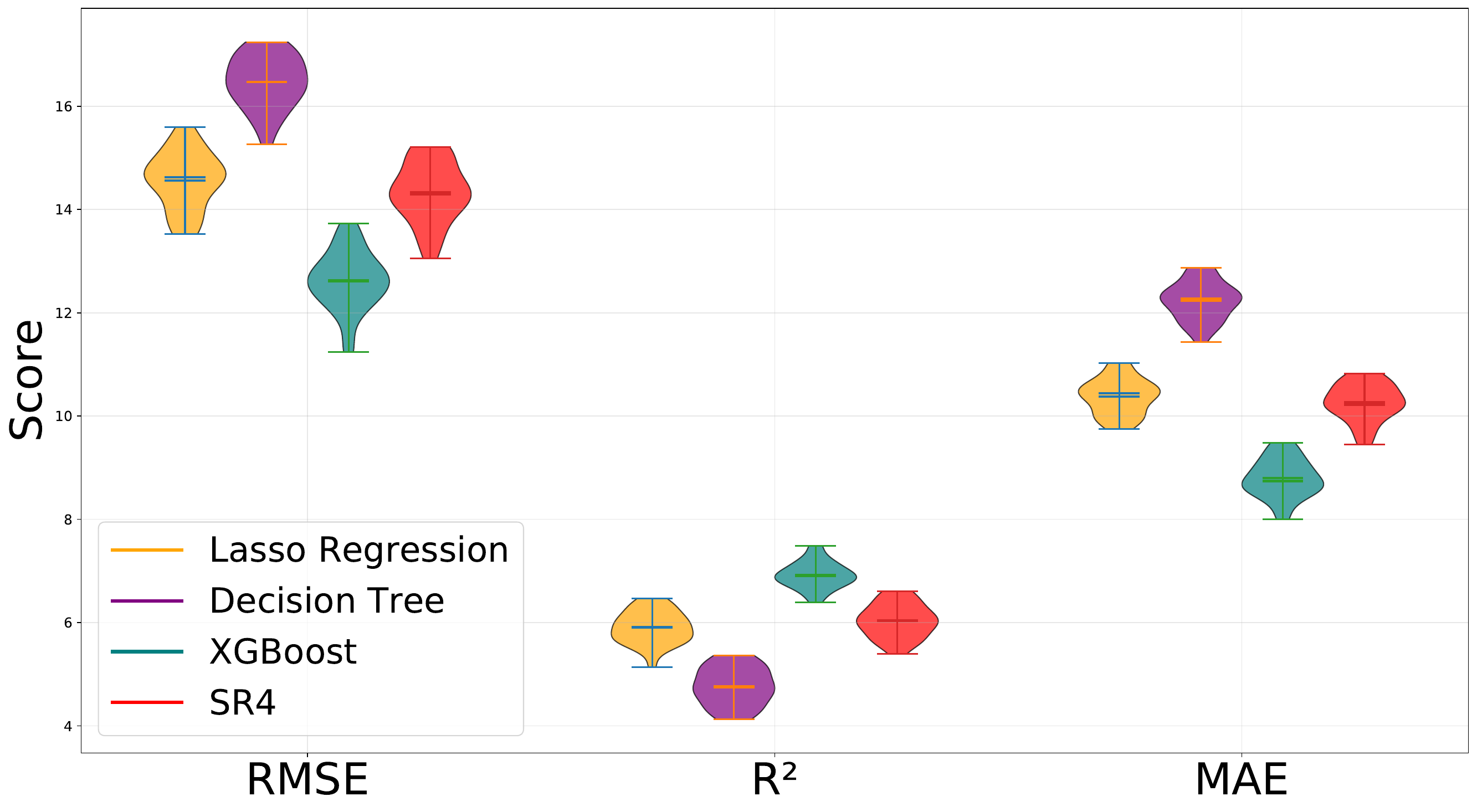}
    \subcaption{Expanded}
\end{minipage}\hfill
\begin{minipage}{0.48\linewidth}
    \centering
    \includegraphics[width=\linewidth]{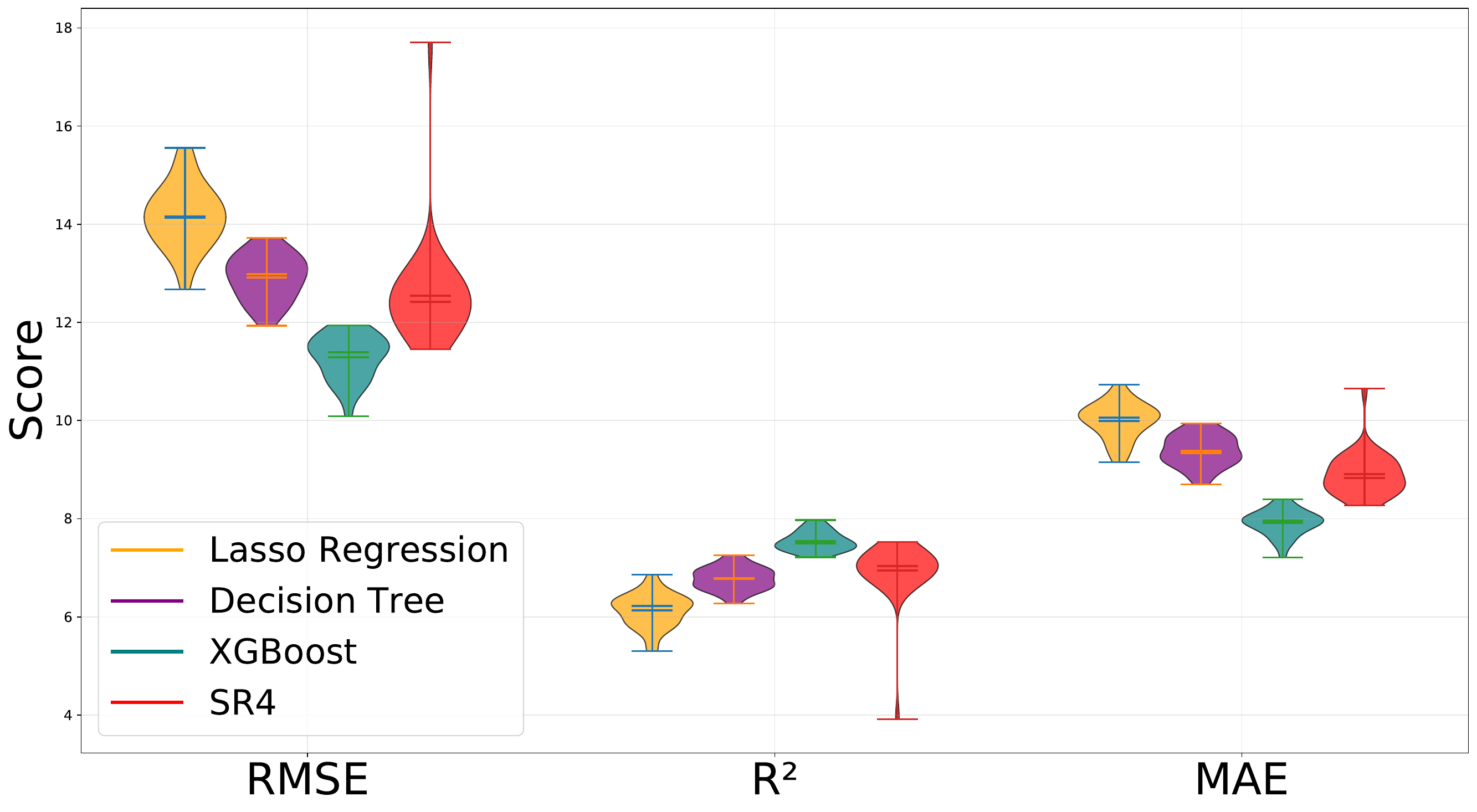}
    \subcaption{Previous}
\end{minipage}

\caption{
Violin plot comparison of model (LASSO regression, decision tree, XG boost) performance metrics across multiple datasets for the election regression task that has REP percentage as a label and does not include party percentage (DEM).
}
\label{fig:violin_all_models_rep_wpp_other}
\end{figure}


\begin{table}
\caption{
Average Dice–Sørensen Index $\pm$ standard deviation across 30 trials for the election regression dataset where REP percentage is the label.
}

\centering
\setlength{\tabcolsep}{3pt}
\renewcommand{\arraystretch}{1.15}

\begin{subtable}{\columnwidth}
\centering
\begin{tabular}{lcccc}
\toprule
\textbf{Dataset}
& \textbf{Random Forest} & \textbf{RuleFit} & \textbf{SR4-Fit} & \textbf{Decision Tree} \\
\midrule
Minimum   & 0.0001$\pm$0.0000 & 0.3333$\pm$0.0000 & \textbf{0.4874$\pm$0.0155} & 0.0092$\pm$0.0155 \\
Standard  & 0.0001$\pm$0.0000 & 0.6279$\pm$0.0000 & \textbf{0.6873$\pm$0.0078} & 0.0092$\pm$0.0155 \\
Expanded  & 0.0001$\pm$0.0000 & 0.7377$\pm$0.0000 & \textbf{0.7806$\pm$0.0075} & 0.0069$\pm$0.0140 \\
Previous  & 0.0001$\pm$0.0000 & \textbf{0.7778$\pm$0.0000} & 0.7132$\pm$0.0068 & 0.0092$\pm$0.0155 \\
\bottomrule
\end{tabular}
\caption{\textbf{REP as a label, with DEM percentage.}}
\end{subtable}

\vspace{6pt}

\begin{subtable}{\columnwidth}
\centering
\begin{tabular}{lcccc}
\toprule
\textbf{Dataset}
& \textbf{Random Forest} & \textbf{RuleFit} & \textbf{SR4-Fit} & \textbf{Decision Tree} \\
\midrule
Minimum   & 0.0004$\pm$0.0001 & \textbf{0.3048$\pm$0.0012} & 0.2666$\pm$0.0121 & 0.0074$\pm$0.0084 \\
Standard  & 0.0001$\pm$0.0000 & \textbf{0.7647$\pm$0.0000} & 0.7584$\pm$0.0021 & 0.0100$\pm$0.0110 \\
Expanded  & 0.0001$\pm$0.0000 & 0.7333$\pm$0.0000 & \textbf{0.7452$\pm$0.0077} & 0.4184$\pm$0.2733 \\
Previous  & 0.0015$\pm$0.0005 & \textbf{0.7714$\pm$0.0000} & 0.6814$\pm$0.0078 & 0.1356$\pm$0.1127 \\
\bottomrule
\end{tabular}
\caption{\textbf{REP as a label, without DEM percentage.}}
\end{subtable}

\label{tab:dice_reg_rep_vs_no_dem}
\end{table}


\begin{table}
\caption{
Average number of rules $\pm$ standard deviation across 30 trials for the election regression dataset, where REP percentage is the label.
}

\centering
\setlength{\tabcolsep}{3pt}
\renewcommand{\arraystretch}{1.15}

\begin{subtable}{\columnwidth}
\centering
\begin{tabular}{lcccc}
\toprule
\textbf{Dataset}
& \textbf{Random Forest} & \textbf{RuleFit} & \textbf{SR4-Fit} & \textbf{Decision Tree} \\
\midrule
Minimum   & 8161.43$\pm$200.06 & 24.00$\pm$0.00 & 15.86$\pm$0.34 & \textbf{1.00$\pm$0.00} \\
Standard  & 8225.93$\pm$225.83 & 43.00$\pm$0.00 & 39.00$\pm$0.52 & \textbf{1.00$\pm$0.00} \\
Expanded  & 8457.63$\pm$232.08 & 61.00$\pm$0.00 & 56.96$\pm$1.12 & \textbf{1.00$\pm$0.00} \\
Previous  & 8205.06$\pm$199.47 & 36.00$\pm$0.00 & 39.26$\pm$0.78 & \textbf{1.00$\pm$0.00} \\
\bottomrule
\end{tabular}
\caption{\textbf{REP as a label, with DEM percentage.}}
\end{subtable}

\vspace{6pt}

\begin{subtable}{\columnwidth}
\centering
\begin{tabular}{lcccc}
\toprule
\textbf{Dataset}
& \textbf{Random Forest} & \textbf{RuleFit} & \textbf{SR4-Fit} & \textbf{Decision Tree} \\
\midrule
Minimum   & 8797.16$\pm$179.66 & 22.96$\pm$0.18 & 26.36$\pm$2.41 & \textbf{10.50$\pm$2.28} \\
Standard  & 9513.73$\pm$133.31 & 34.00$\pm$0.00 & 32.96$\pm$0.18 & \textbf{10.13$\pm$1.45} \\
Expanded  & 9256.36$\pm$162.12 & 60.00$\pm$0.00 & 55.03$\pm$1.15 & \textbf{1.00$\pm$0.00} \\
Previous  & 8044.26$\pm$152.79 & 35.00$\pm$0.00 & 38.16$\pm$0.91 & \textbf{1.60$\pm$0.49} \\
\bottomrule
\end{tabular}
\caption{\textbf{REP as a label, without DEM percentage.}}
\end{subtable}

\label{tab:rules_reg_rep_vs_no_dem}
\end{table}


\begin{table}
\caption{
Average rule complexity $\pm$ standard deviation across 30 trials for the election regression dataset, where REP percentage is the label.
}

\centering
\setlength{\tabcolsep}{3pt}
\renewcommand{\arraystretch}{1.15}

\begin{subtable}{\columnwidth}
\centering
\begin{tabular}{lcccc}
\toprule
\textbf{Dataset}
& \textbf{Random Forest} & \textbf{RuleFit} & \textbf{SR4-Fit} & \textbf{Decision Tree} \\
\midrule
Minimum   & 6.23$\pm$0.03 & 3.00$\pm$0.00 & 2.01$\pm$0.02 & \textbf{1.00$\pm$0.00} \\
Standard  & 6.17$\pm$0.03 & 2.11$\pm$0.00 & 1.93$\pm$0.03 & \textbf{1.00$\pm$0.00} \\
Expanded  & 6.18$\pm$0.03 & 1.78$\pm$0.00 & 1.64$\pm$0.03 & \textbf{1.20$\pm$0.61} \\
Previous  & 6.16$\pm$0.03 & 1.44$\pm$0.00 & 1.86$\pm$0.04 & \textbf{1.00$\pm$0.00} \\
\bottomrule
\end{tabular}
\caption{\textbf{REP as a label, with DEM percentage.}}
\end{subtable}

\vspace{6pt}

\begin{subtable}{\columnwidth}
\centering
\begin{tabular}{lcccc}
\toprule
\textbf{Dataset}
& \textbf{Random Forest} & \textbf{RuleFit} & \textbf{SR4-Fit} & \textbf{Decision Tree} \\
\midrule
Minimum   & 6.32$\pm$0.02 & \textbf{3.08$\pm$0.02} & 4.66$\pm$0.13 & 4.59$\pm$0.17 \\
Standard  & 6.42$\pm$0.01 & \textbf{1.47$\pm$0.00} & 1.48$\pm$0.01 & 4.73$\pm$0.13 \\
Expanded  & 6.39$\pm$0.03 & 1.80$\pm$0.00 & 1.76$\pm$0.05 & \textbf{1.20$\pm$0.61} \\
Previous  & 6.30$\pm$0.02 & \textbf{1.45$\pm$0.00} & 1.95$\pm$0.05 & 2.86$\pm$0.50 \\
\bottomrule
\end{tabular}
\caption{\textbf{REP as a label, without DEM percentage.}}
\end{subtable}

\label{tab:complexity_reg_rep_vs_no_dem}
\end{table}


In the minimum dataset, the line plots in Figure~\ref{fig:lineplot_grid_min_rep} and Figure~\ref{fig:lineplot_grid_min_rep_other} show that when DEM percentage is included as a feature, all models achieve relatively low error values, whereas XGBoost attains slightly lower average error values compared to other models. The distributional comparisons through violin plots in Figure~\ref{fig:violin_all_models_rep} and Figure~\ref{fig:violin_all_models_rep_other} reinforce these findings. When the DEM percentage is removed, performance declines across all models, with XGBoost still achieving lower average error values compared to other models, which is further confirmed in Figure~\ref{fig:violin_all_models_rep_wpp} and Figure~\ref{fig:violin_all_models_rep_wpp_other}. Stability analysis in Table~\ref{tab:dice_reg_rep_vs_no_dem} shows that in the case of including DEM percentage as a feature, SR4-Fit achieves the highest Dice–Sørensen Index among rule-based models, whereas when DEM percentage is removed, RuleFit achieves the highest Dice–Sørensen Index among rule-based models. Table~\ref{tab:rules_reg_rep_vs_no_dem} shows that in both cases, the decision tree has the most compact structure, but the structure is too compact, which may lower robustness and may lower the ability to identify interactions properly. Table~\ref{tab:complexity_reg_rep_vs_no_dem} shows that in the case of including DEM percentage as a feature, the decision tree has the lower average rule complexity, whereas in the case of removing DEM percentage as a feature, the RuleFit has the overall lower average rule complexity.

In the standard dataset, the line plots in Figure~\ref{fig:lineplot_grid_std_rep} and Figure~\ref{fig:lineplot_grid_std_rep_other} show that when DEM percentage is included as a feature, all models achieve lower error values, whereas XGBoost attains slightly lower average error values compared to other models. When the DEM percentage is removed, performance declines across all models, with XGBoost still attaining lower average error values compared to other models. Stability analysis in Table~\ref{tab:dice_reg_rep_vs_no_dem} shows that in the case of including DEM percentage as a feature, SR4-Fit achieves the highest Dice–Sørensen Index among rule-based models, whereas when DEM percentage is removed, RuleFit achieves the highest Dice–Sørensen Index among rule-based models. Table~\ref{tab:rules_reg_rep_vs_no_dem} shows that in both cases, the decision tree has the most compact structure. Table~\ref{tab:complexity_reg_rep_vs_no_dem} shows that in the case of including DEM percentage as a feature, the decision tree has the lower average rule complexity, whereas in the case of removing DEM percentage as a feature, the RuleFit has the overall lower average rule complexity.

In the expanded dataset, the line plots in Figure~\ref{fig:lineplot_grid_exp_rep} and Figure~\ref{fig:lineplot_grid_exp_rep_other} show that when DEM percentage is included as a feature, all models achieve lower error values, whereas XGBoost again attains slightly lower average error values compared to other models. When the DEM percentage is removed, performance declines across all models, with XGBoost still attaining lower average error values compared to other models. Stability analysis in Table~\ref{tab:dice_reg_rep_vs_no_dem} shows that with both cases, SR4-Fit achieves the highest Dice–Sørensen Index among rule-based models. Table~\ref{tab:rules_reg_rep_vs_no_dem} shows that in both cases, the decision tree has the most compact structure. Table~\ref{tab:complexity_reg_dem_vs_no_rep} shows that in both cases, the decision tree has lower average rule complexity compared to all models.

In the previous dataset, the line plots in Figure~\ref{fig:lineplot_grid_pre_rep} and Figure~\ref{fig:lineplot_grid_pre_rep_other} show that when DEM percentage is included as a feature, all models achieve lower error values, whereas XGBoost attains lower average error values compared to other models. When the DEM percentage is removed, performance declines across all models, with random forest slightly achieving lower average error values compared to other models. Stability analysis in Table~\ref{tab:dice_reg_rep_vs_no_dem} shows that with both cases, RuleFit achieves the highest Dice–Sørensen Index among rule-based models. Table~\ref{tab:rules_reg_rep_vs_no_dem} shows that in both cases, the decision tree has the most compact structure. Table~\ref{tab:complexity_reg_rep_vs_no_dem} shows that in the case of including DEM percentage as a feature, the decision tree has the lower average rule complexity, whereas in the case of removing DEM percentage as a feature, the RuleFit has the overall lower average rule complexity.

\FloatBarrier

\section{Results on Standard Public Datasets Classification}\label{sup_sec_BenchC}
\begin{figure}
\centering
\begin{minipage}{0.49\linewidth}
    \centering
    \includegraphics[width=\linewidth]{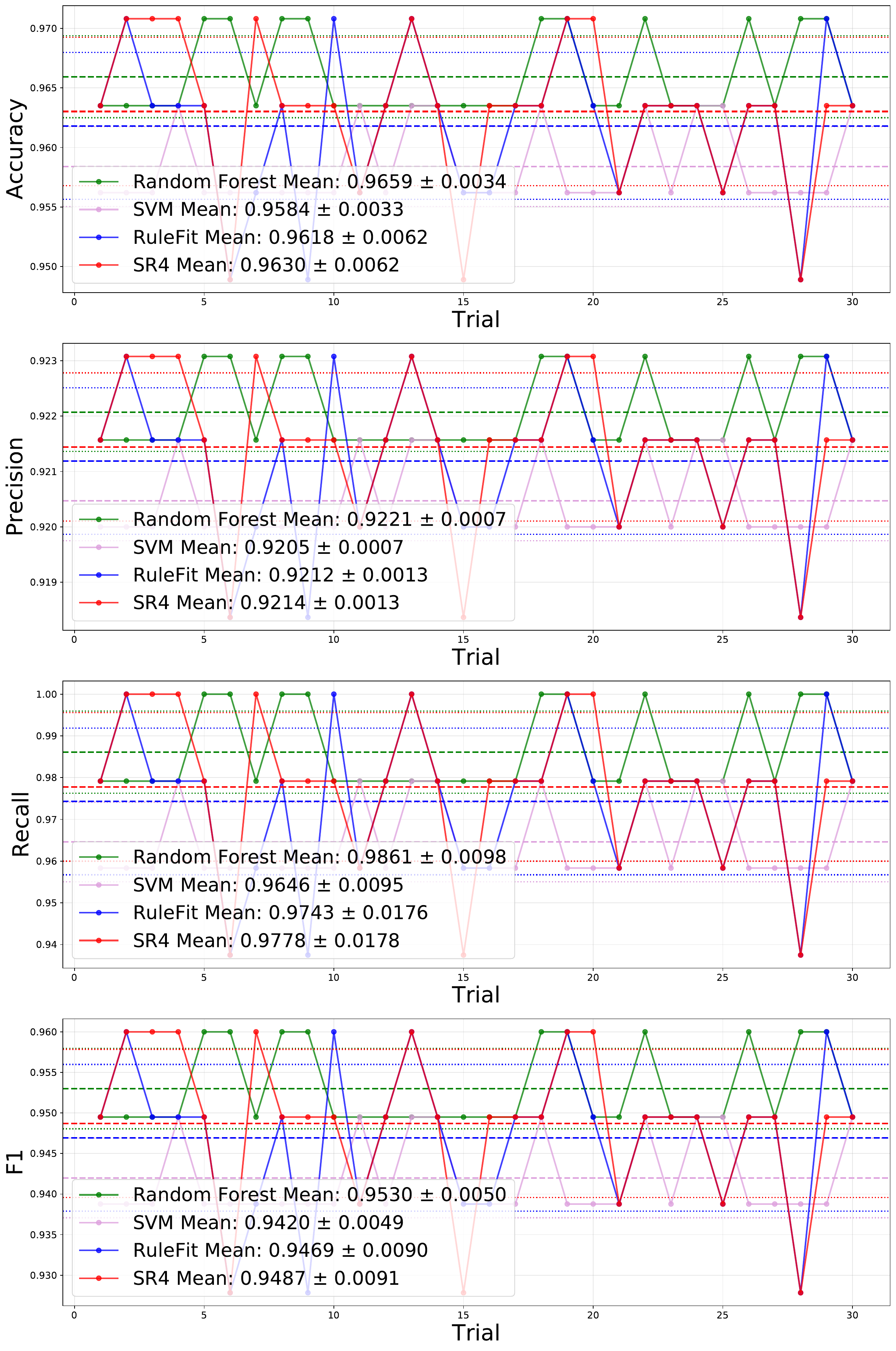}
\end{minipage}\hfill
\begin{minipage}{0.49\linewidth}
    \centering
    \includegraphics[width=\linewidth]{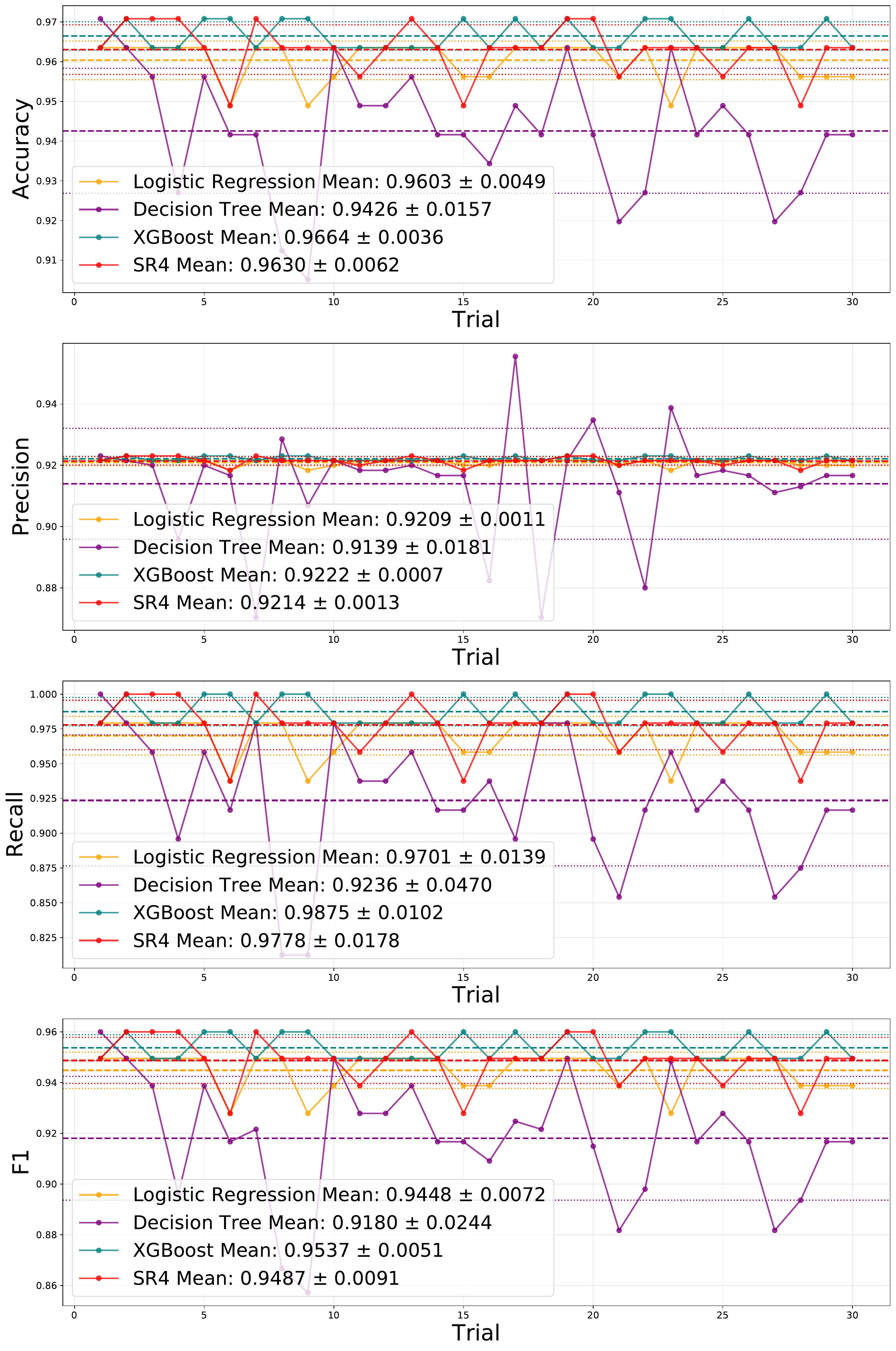}
\end{minipage}

\caption{
Line plot comparison of model performance metrics for breast cancer data across 30 trials.
The left panel reports results obtained for models—random forest, SVM, rulefit, and SR4-fit ,
whereas the right panel shows results with models logistic regression, decision tree, and XGboost.
}
\label{fig:lineplot_breast}
\end{figure}


\begin{figure}
\centering
\begin{minipage}{0.49\linewidth}
    \centering
    \includegraphics[width=\linewidth]{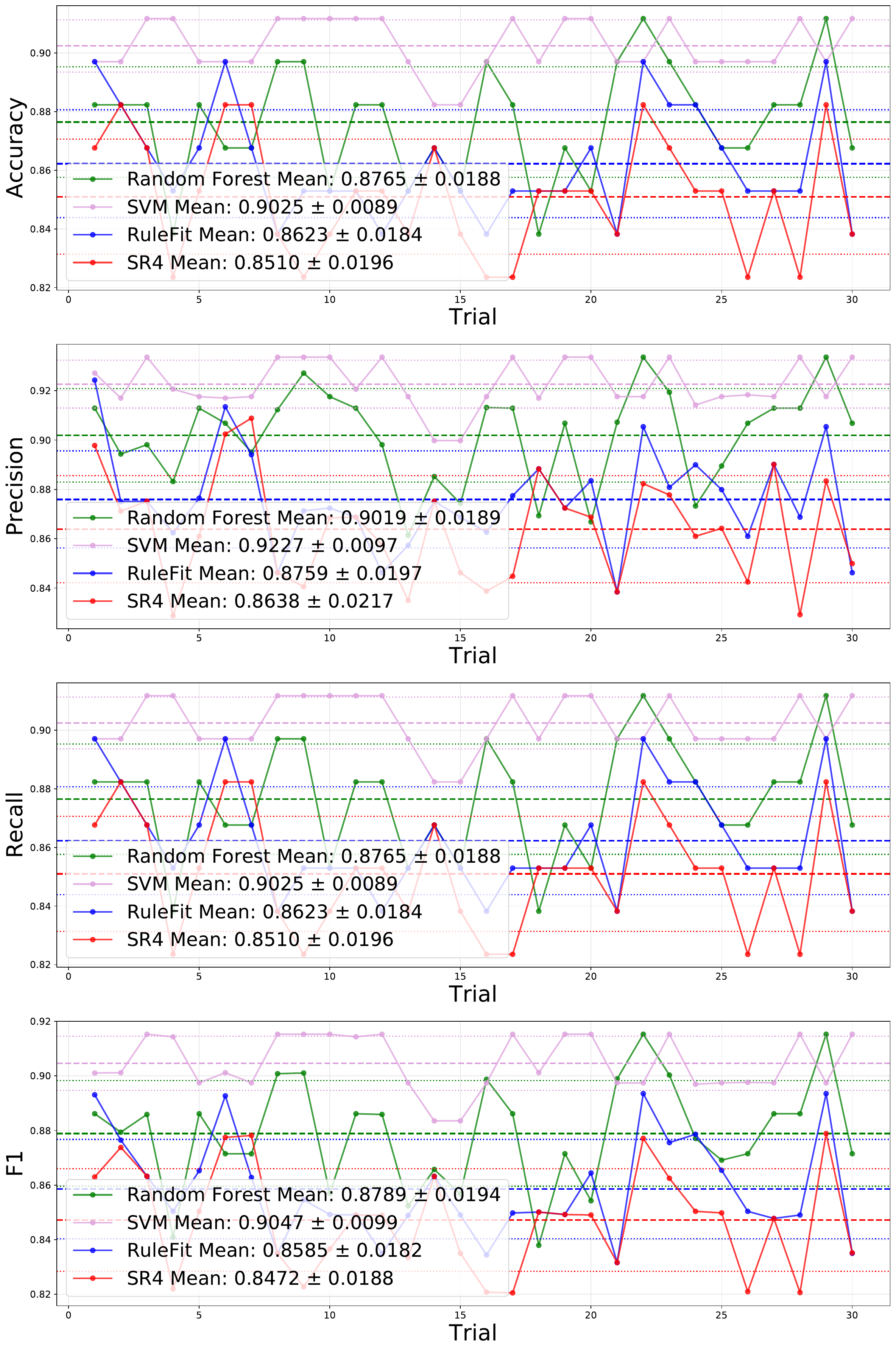}
\end{minipage}\hfill
\begin{minipage}{0.49\linewidth}
    \centering
    \includegraphics[width=\linewidth]{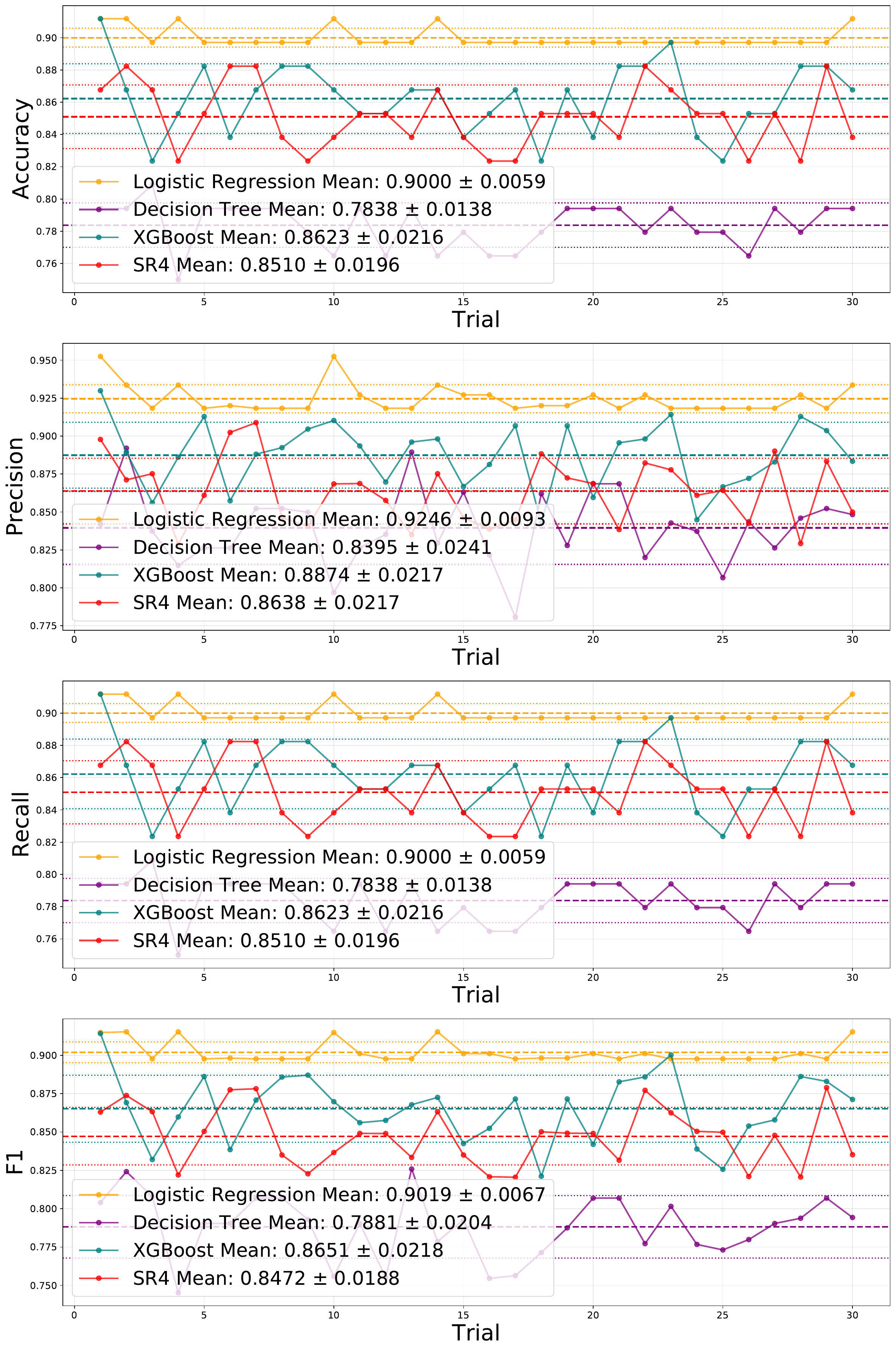}
\end{minipage}

\caption{
Line plot comparison of model performance metrics for E. coli data across 30 trials.
The left panel reports results obtained for models—random forest, SVM, rulefit, and SR4-fit ,
whereas the right panel shows results with models logistic regression, decision tree, and XGboost.
}
\label{fig:lineplot_ecoli}
\end{figure}


\begin{figure}
\centering
\begin{minipage}{0.49\linewidth}
    \centering
    \includegraphics[width=\linewidth]{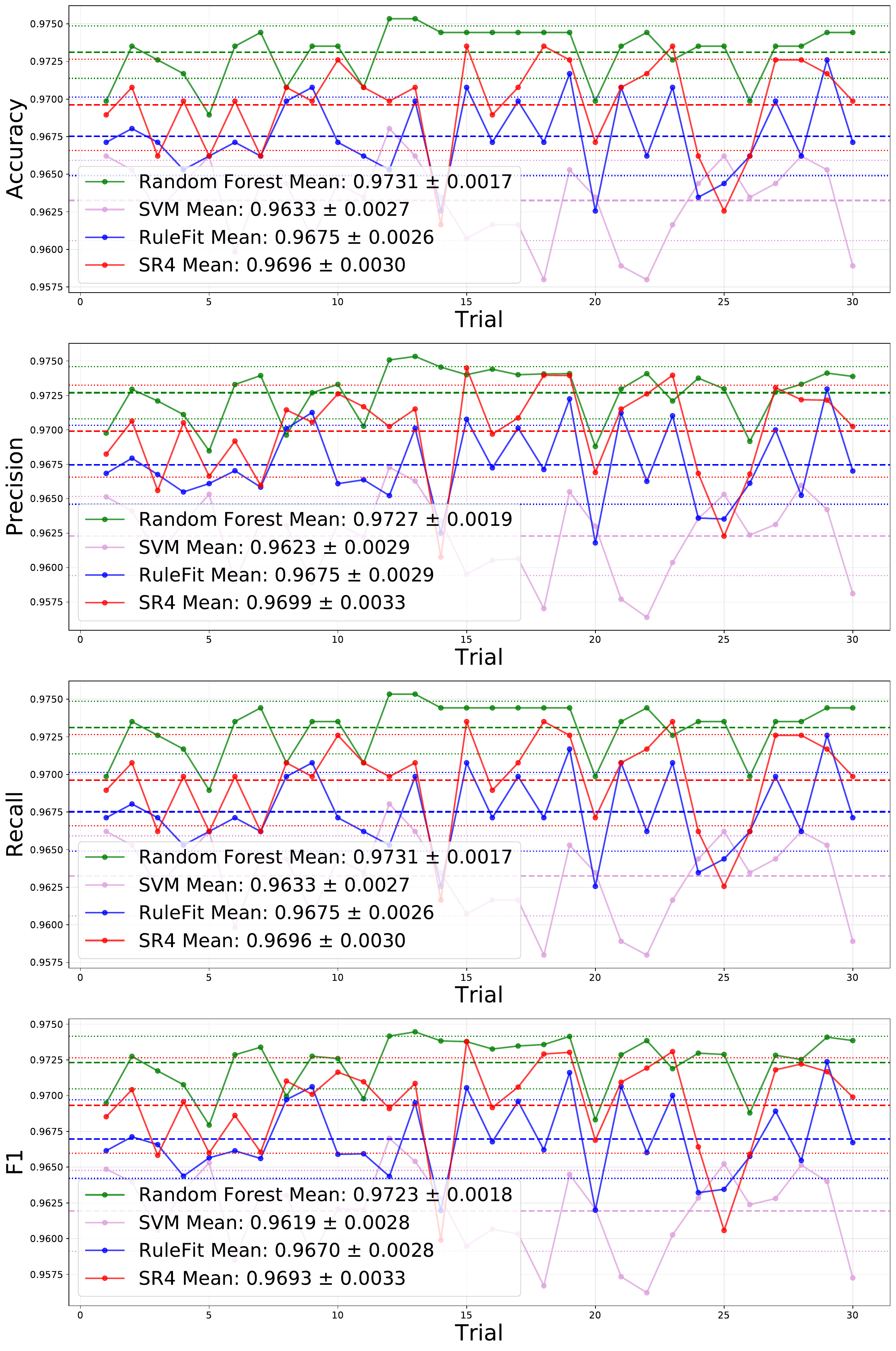}
\end{minipage}\hfill
\begin{minipage}{0.49\linewidth}
    \centering
    \includegraphics[width=\linewidth]{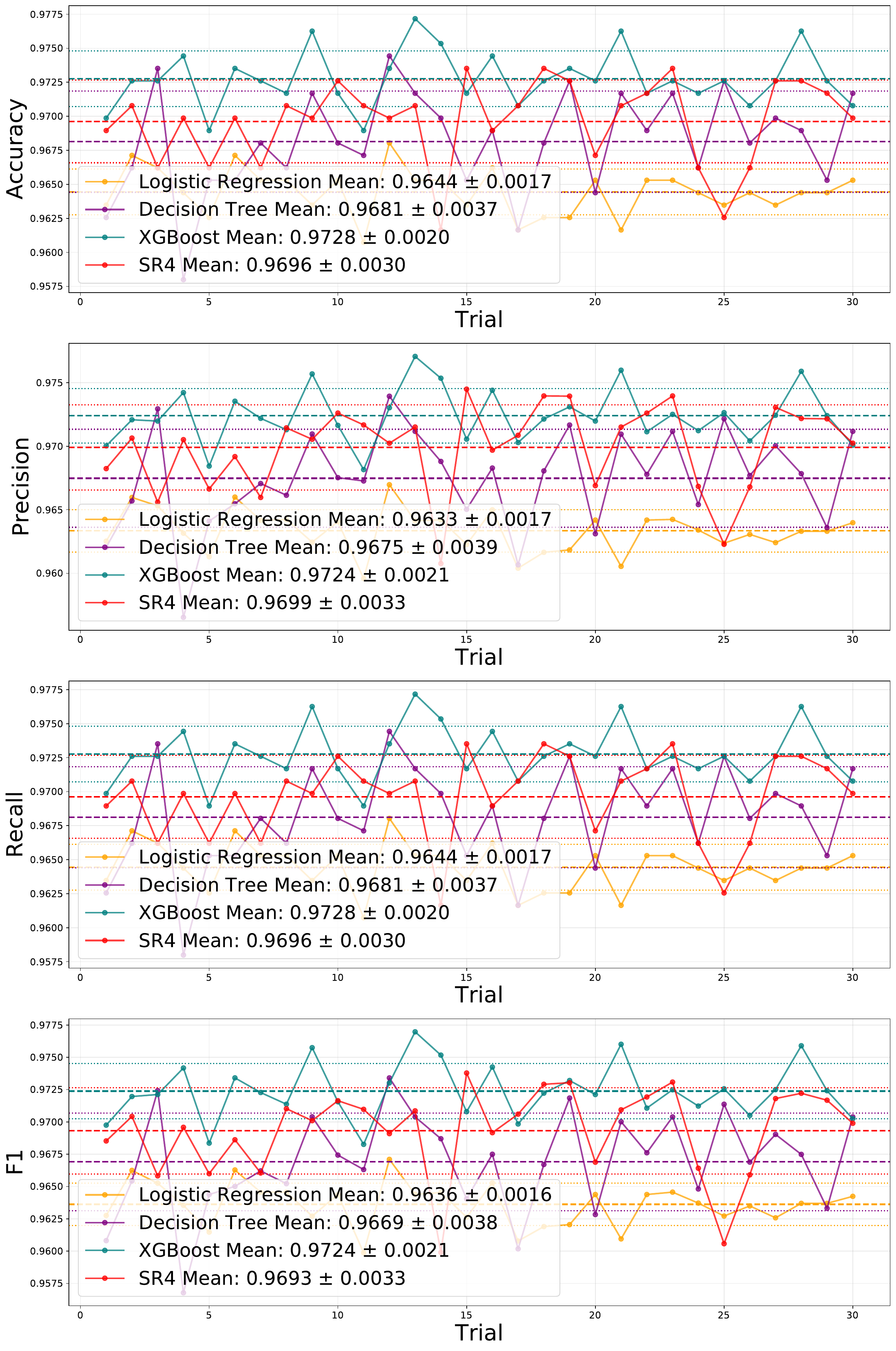}
\end{minipage}

\caption{
Line plot comparison of model performance metrics for page blocks data across 30 trials.
The left panel reports results obtained for models—random forest, SVM, rulefit, and SR4-fit ,
whereas the right panel shows results with models logistic regression, decision tree, and XGboost.
}
\label{fig:lineplot_page}
\end{figure}


\begin{figure}
\centering
\begin{minipage}{0.49\linewidth}
    \centering
    \includegraphics[width=\linewidth]{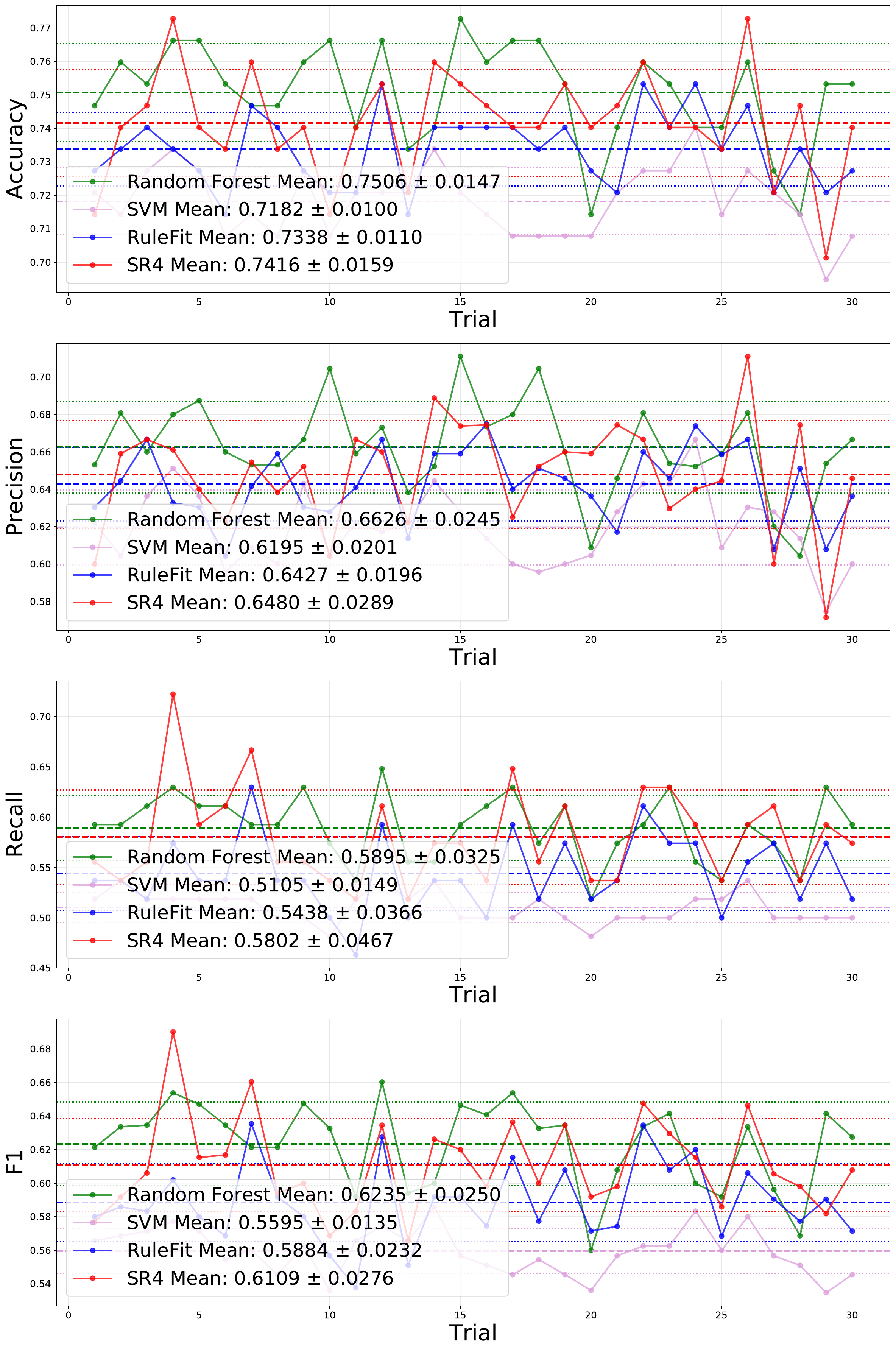}
\end{minipage}\hfill
\begin{minipage}{0.49\linewidth}
    \centering
    \includegraphics[width=\linewidth]{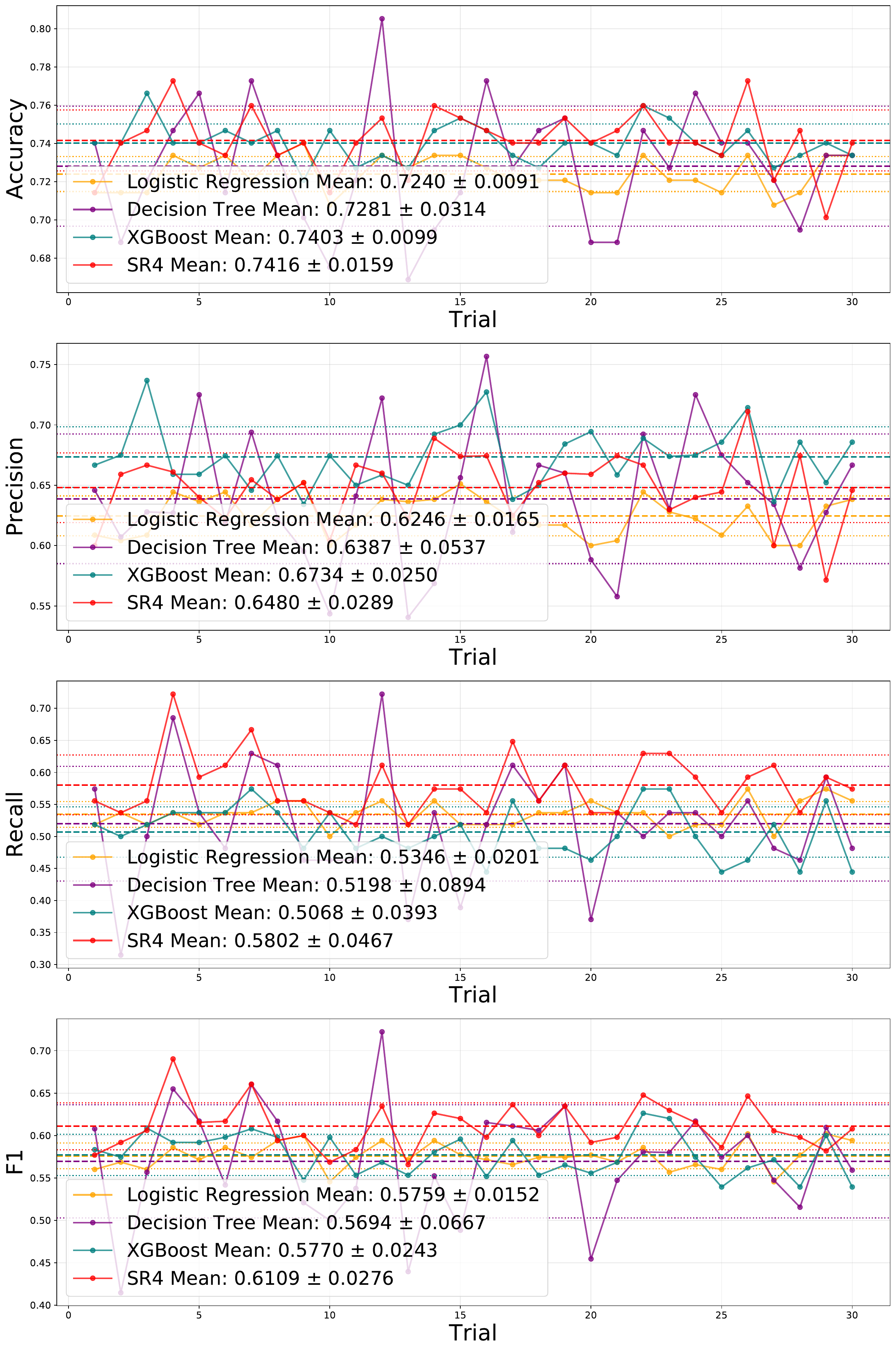}
\end{minipage}

\caption{
Line plot comparison of model performance metrics for Pima Indians data across 30 trials. The left panel reports results obtained for models—random forest, SVM, rulefit, and SR4-fit ,
whereas the right panel shows results with models logistic regression, decision tree, and XGboost.
}
\label{fig:lineplot_pima}
\end{figure}


\begin{figure}
\centering
\begin{minipage}{0.49\linewidth}
    \centering
    \includegraphics[width=\linewidth]{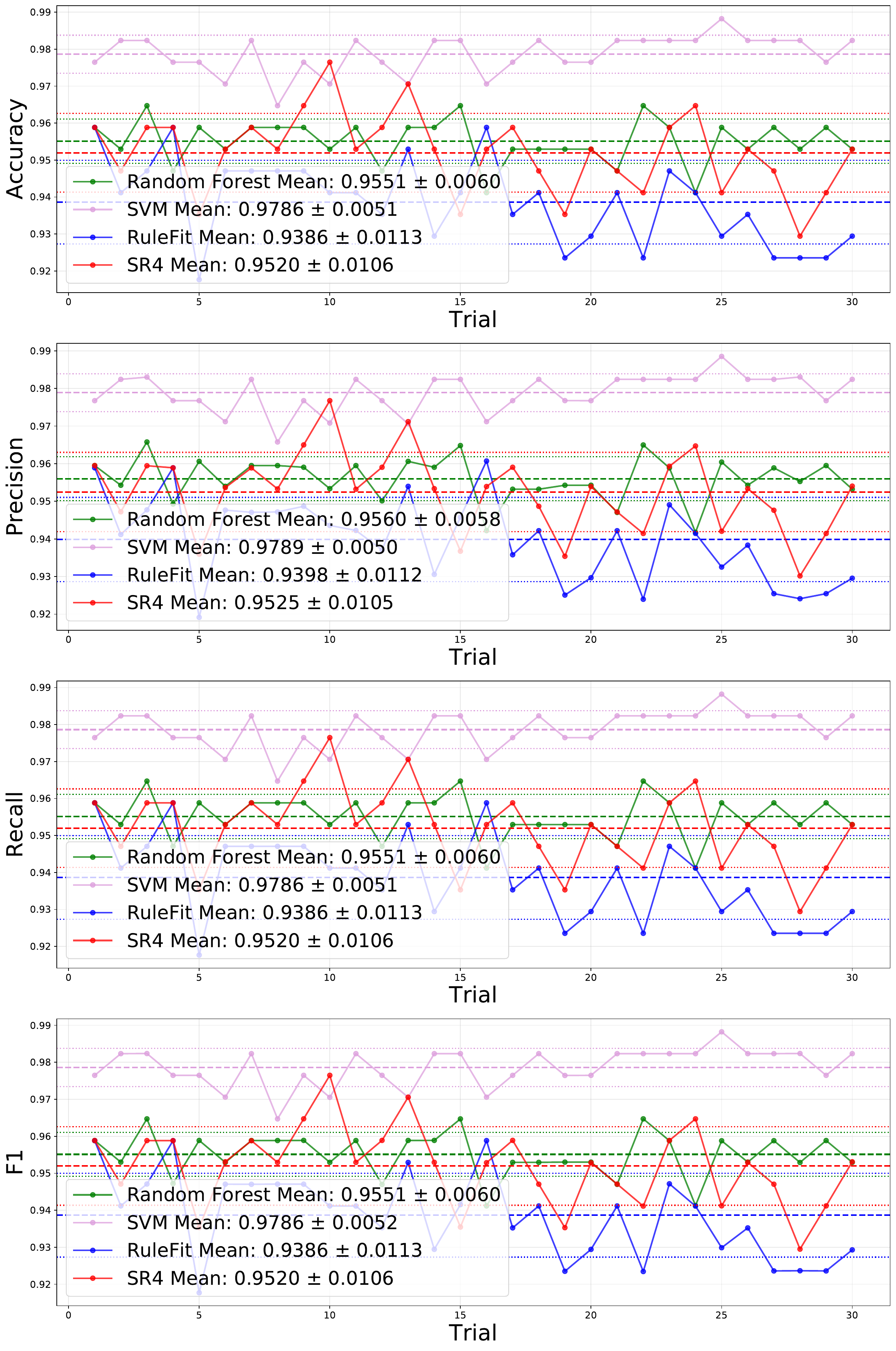}
\end{minipage}\hfill
\begin{minipage}{0.49\linewidth}
    \centering
    \includegraphics[width=\linewidth]{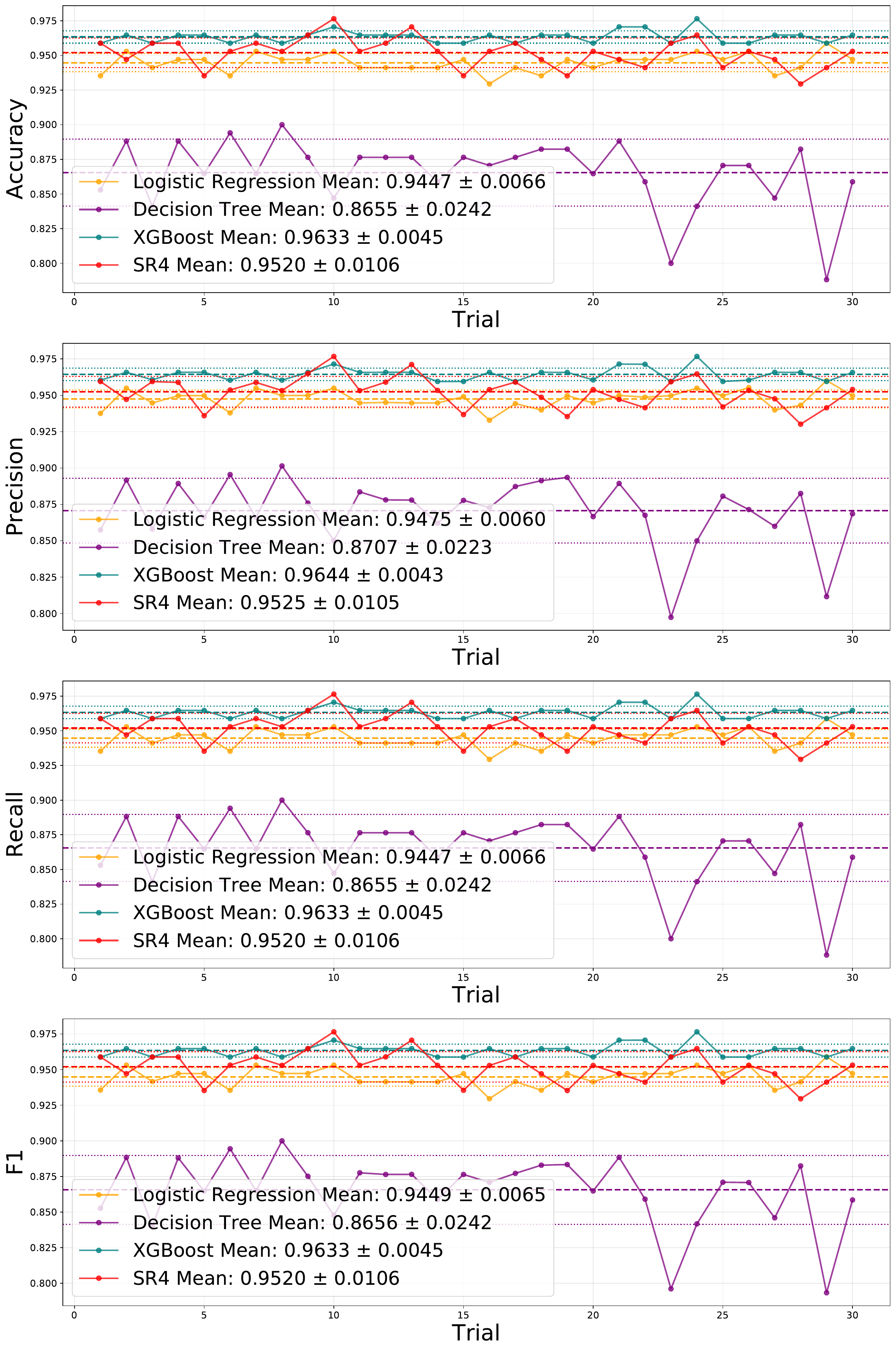}
\end{minipage}

\caption{
Line plot comparison of model performance metrics for vehicle data across 30 trials. The left panel reports results obtained for models—random forest, SVM, rulefit, and SR4-fit ,
whereas the right panel shows results with models logistic regression, decision tree, and XGboost.
}
\label{fig:lineplot_vehicle}
\end{figure}


\begin{figure}
\centering
\begin{minipage}{0.49\linewidth}
    \centering
    \includegraphics[width=\linewidth]{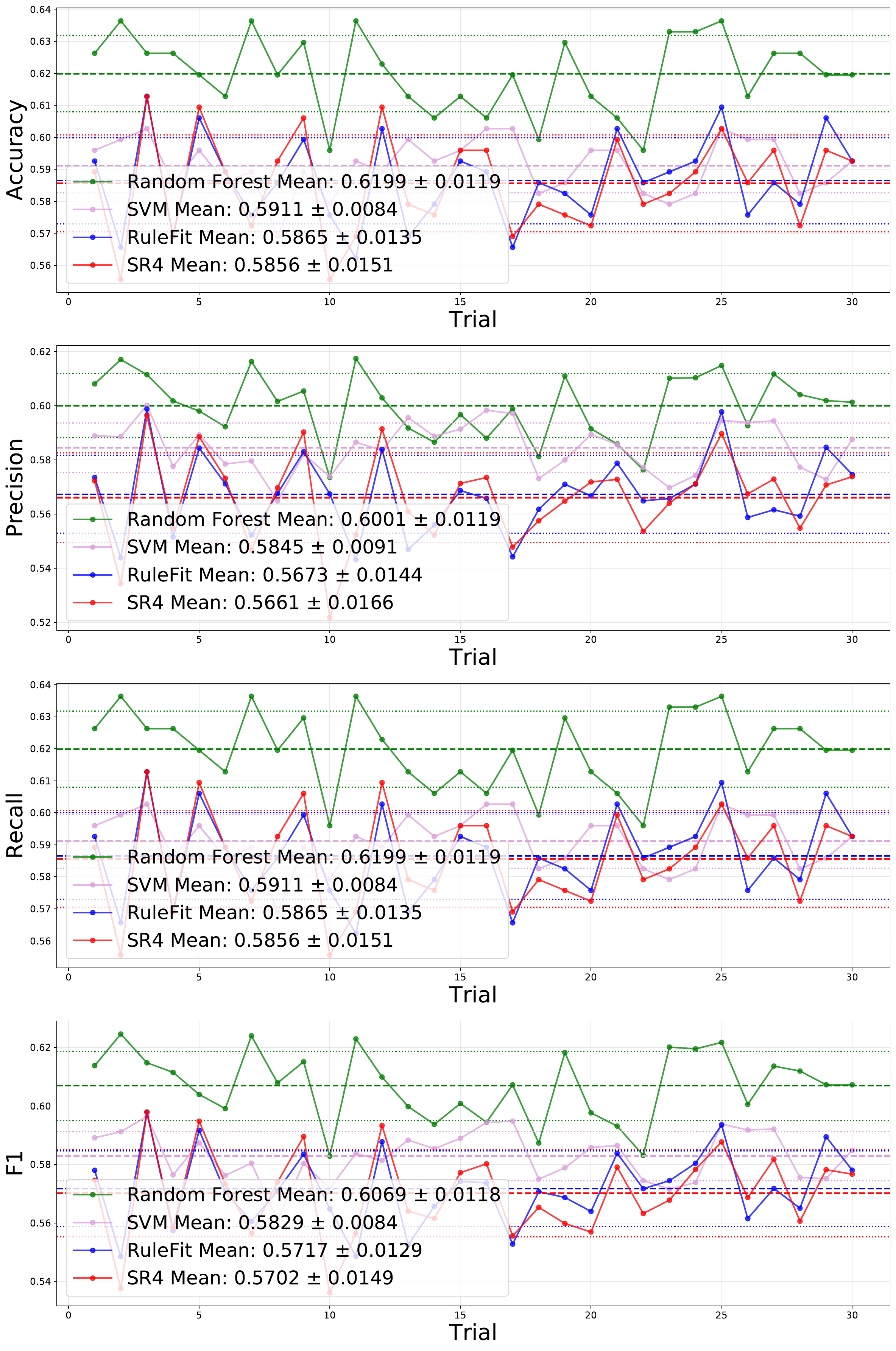}
\end{minipage}\hfill
\begin{minipage}{0.49\linewidth}
    \centering
    \includegraphics[width=\linewidth]{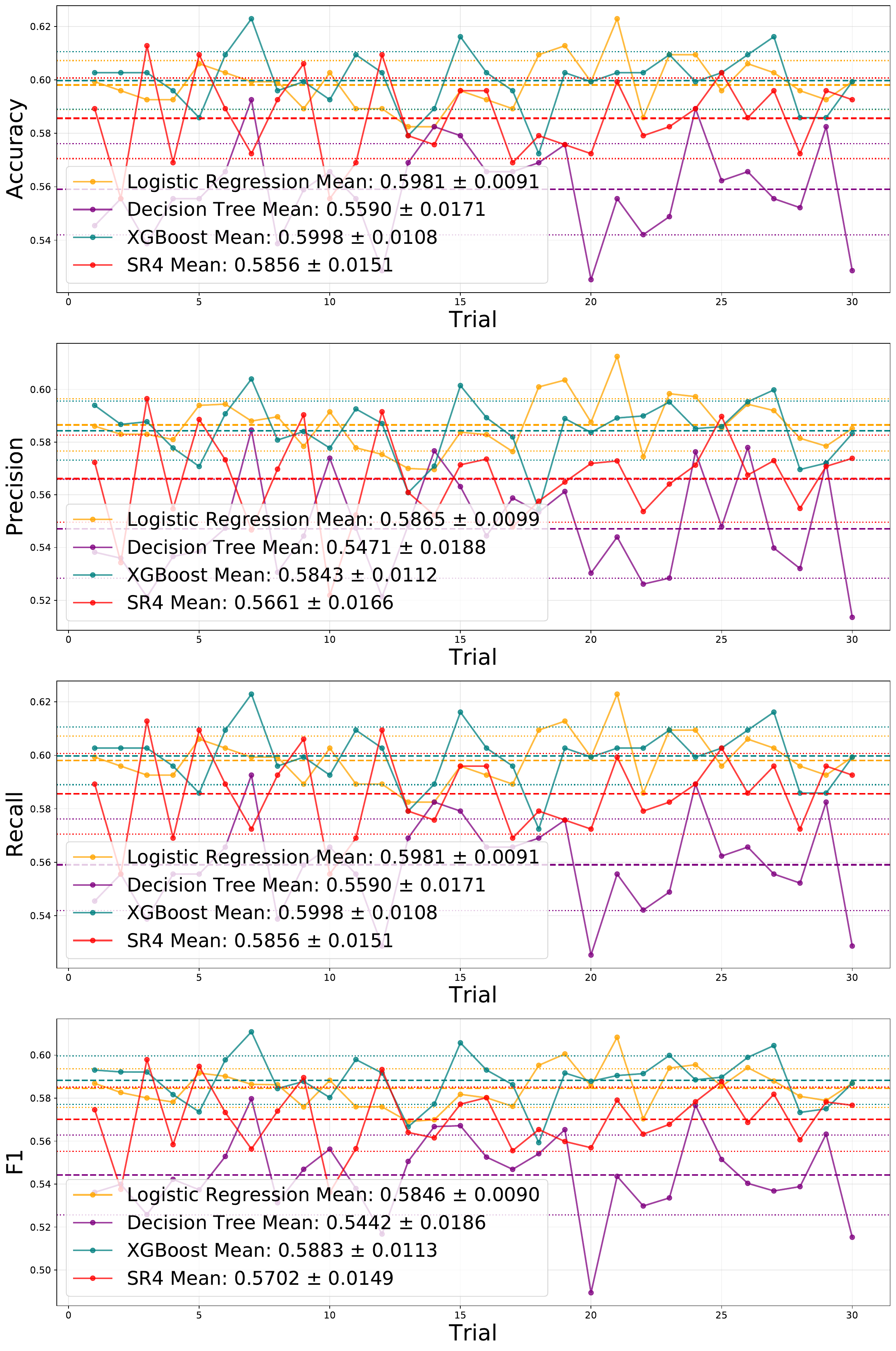}
\end{minipage}

\caption{
Line plot comparison of model performance metrics for yeast data across 30 trials. The left panel reports results obtained for models—random forest, SVM, rulefit, and SR4-fit ,
whereas the right panel shows results with models logistic regression, decision tree, and XGboost.
}
\label{fig:lineplot_yeast}
\end{figure}


\begin{figure}
\centering
\scriptsize

\begin{minipage}{0.48\linewidth}
    \centering
    \includegraphics[width=\linewidth]{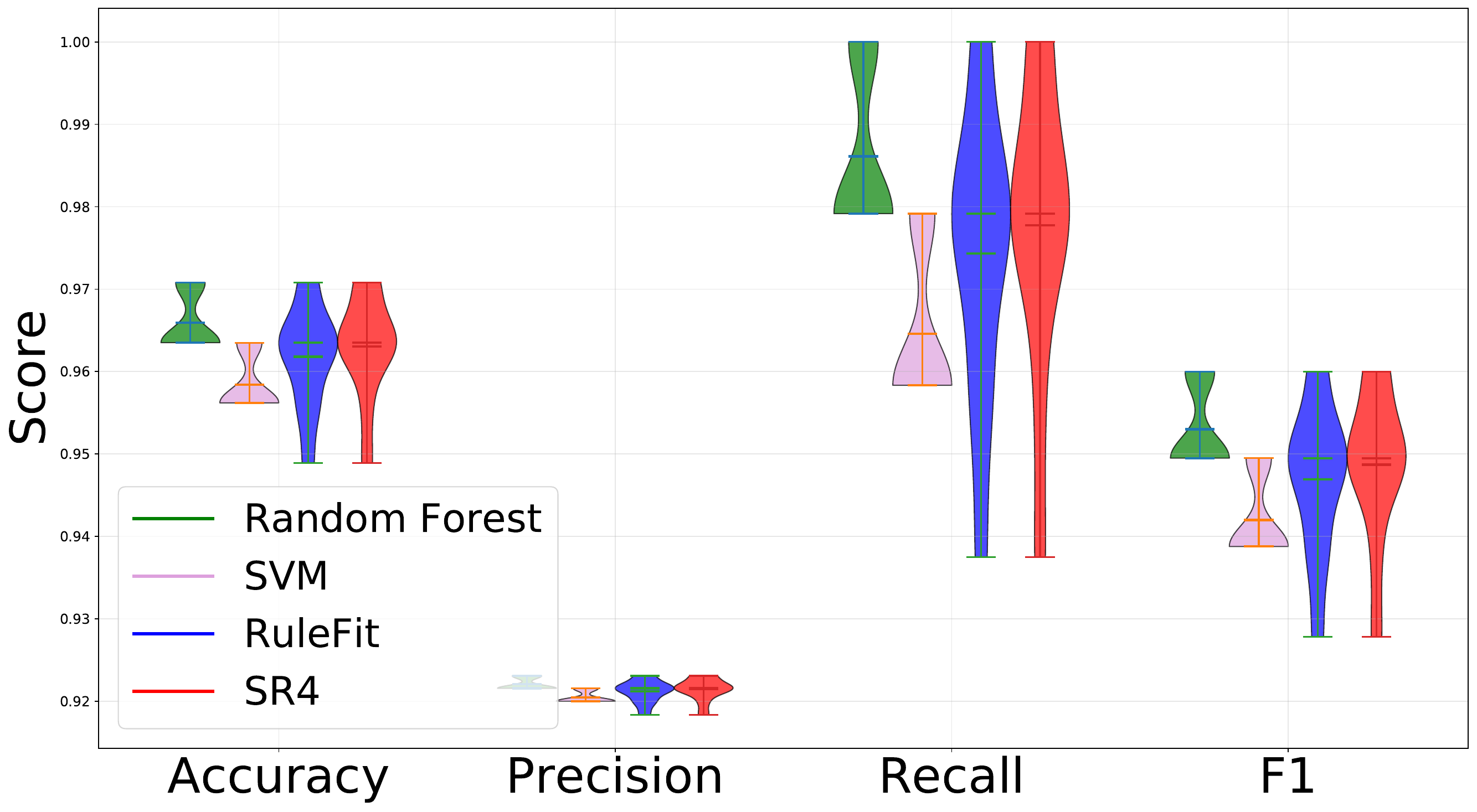}
    \subcaption{Breast Cancer}
\end{minipage}\hfill
\begin{minipage}{0.48\linewidth}
    \centering
    \includegraphics[width=\linewidth]{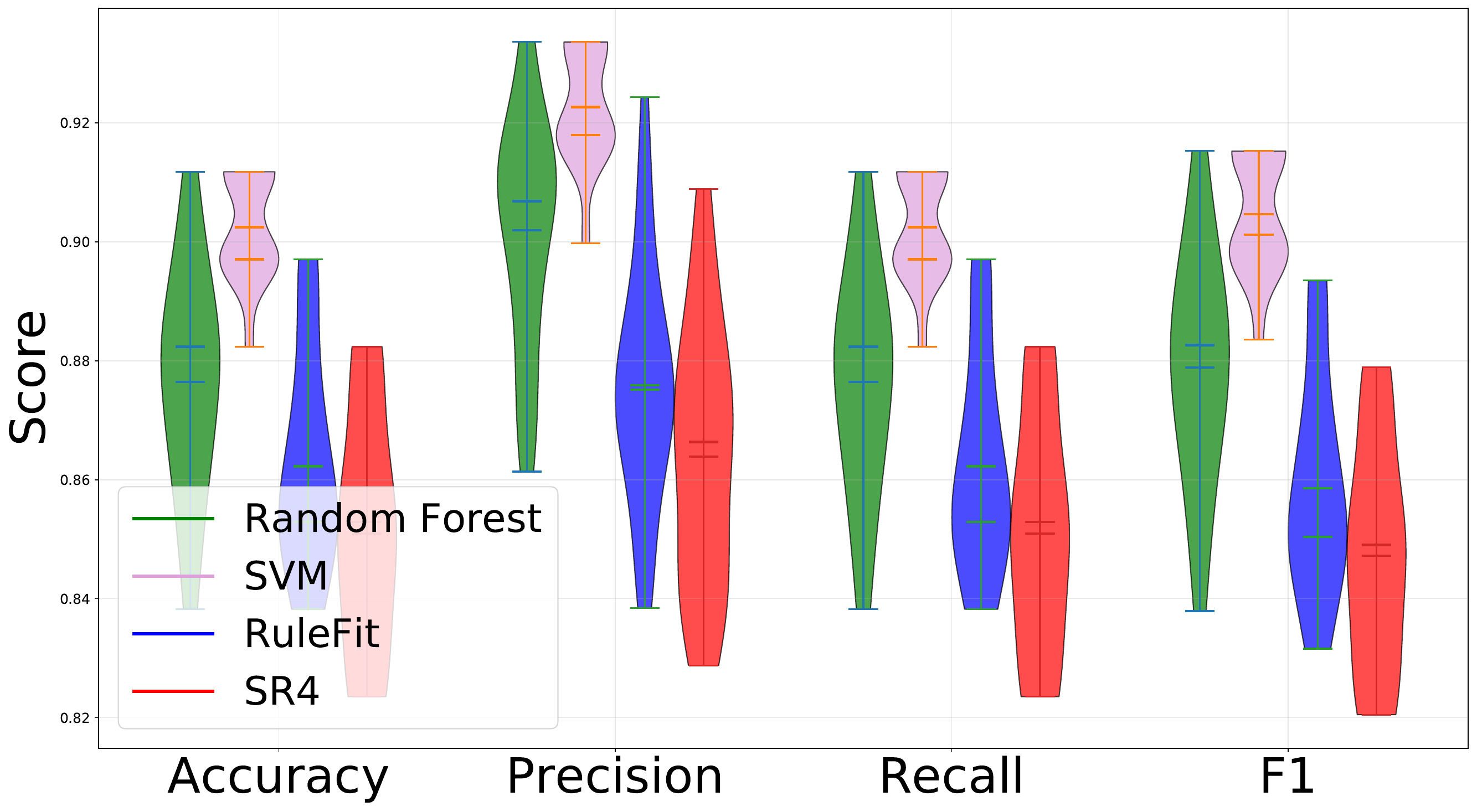}
    \subcaption{E. Coli}
\end{minipage}

\vspace{0.4cm}

\begin{minipage}{0.48\linewidth}
    \centering
    \includegraphics[width=\linewidth]{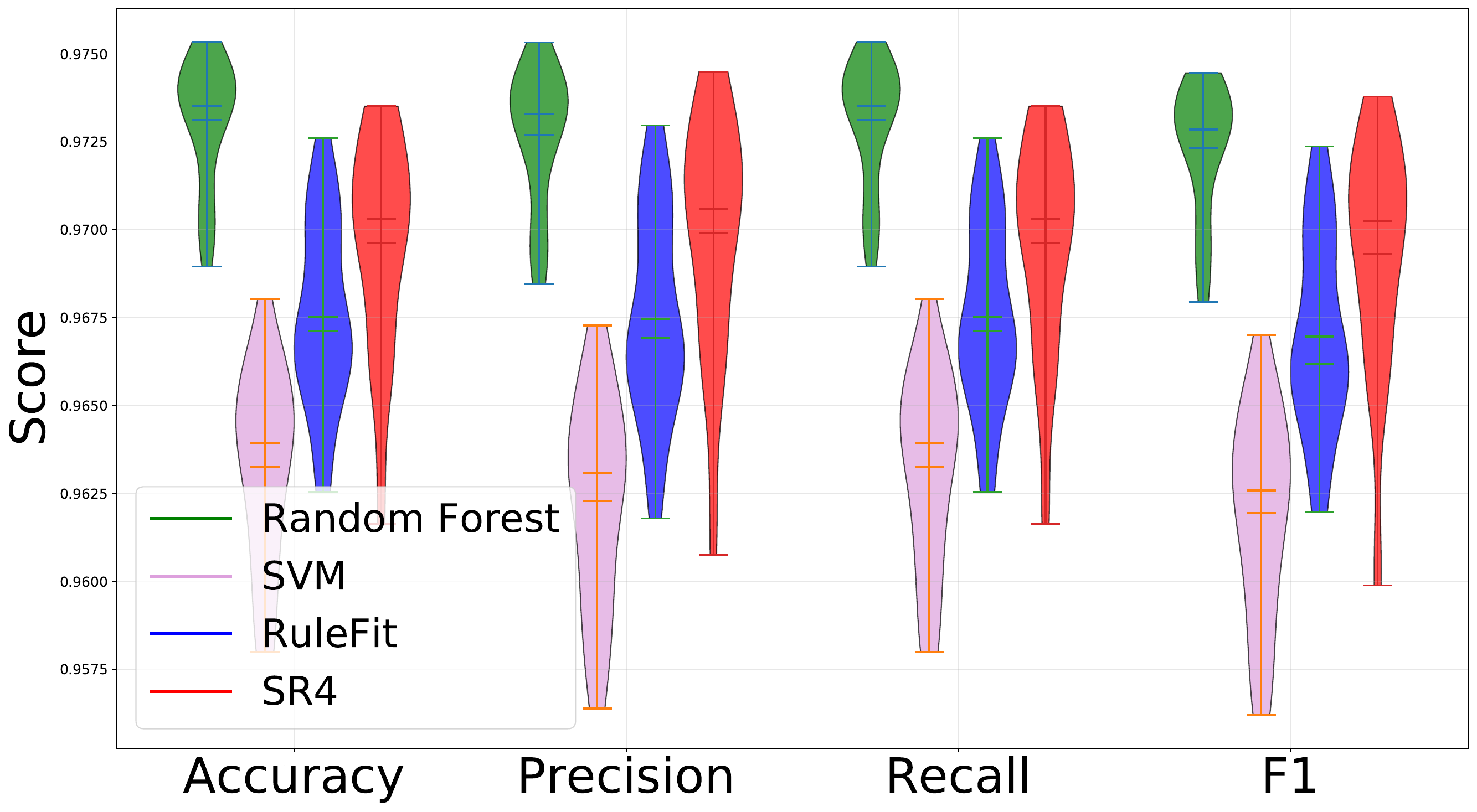}
    \subcaption{Page blocks}
\end{minipage}\hfill
\begin{minipage}{0.48\linewidth}
    \centering
    \includegraphics[width=\linewidth]{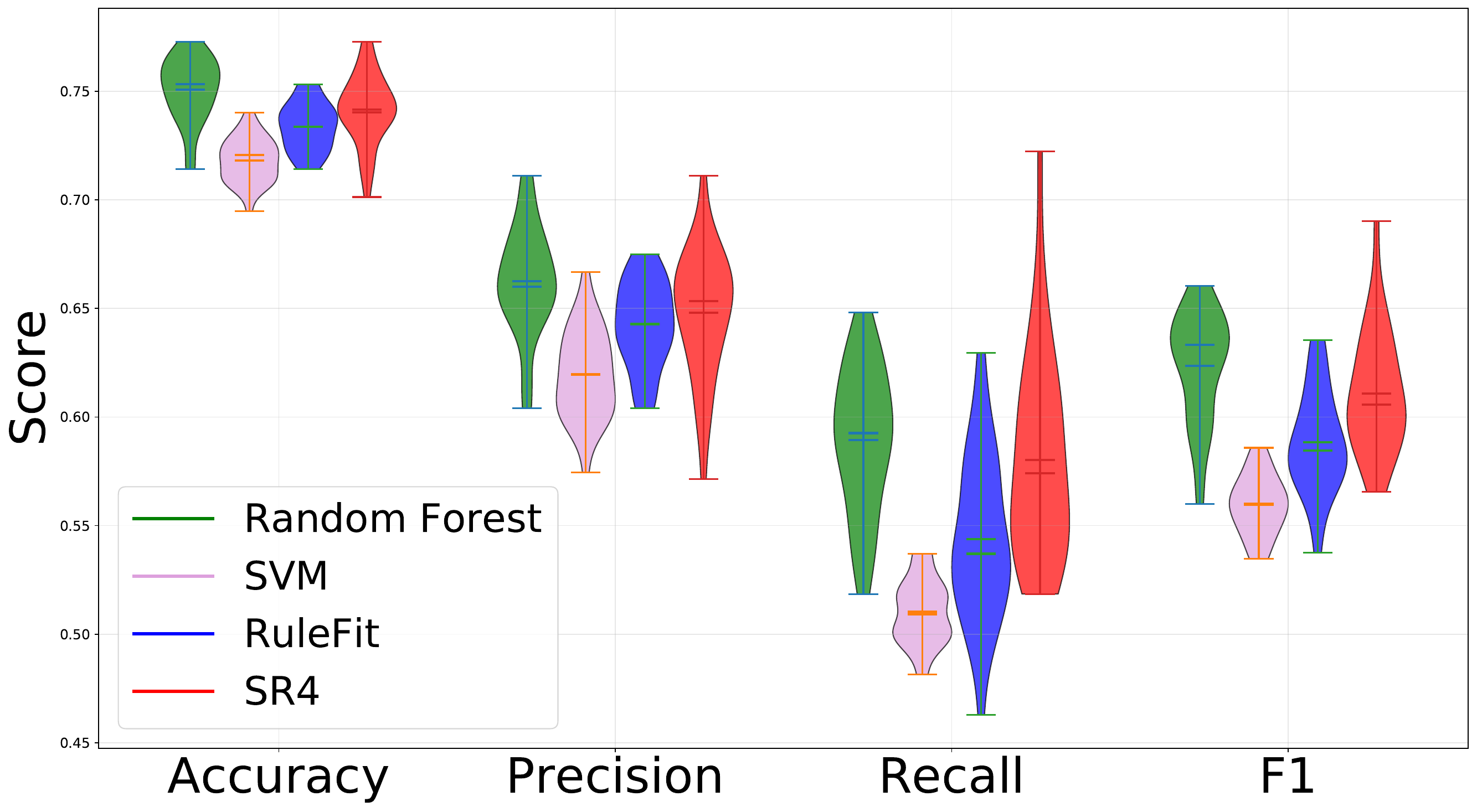}
    \subcaption{Pima Indians}
\end{minipage}

\vspace{0.4cm}

\begin{minipage}{0.48\linewidth}
    \centering
    \includegraphics[width=\linewidth]{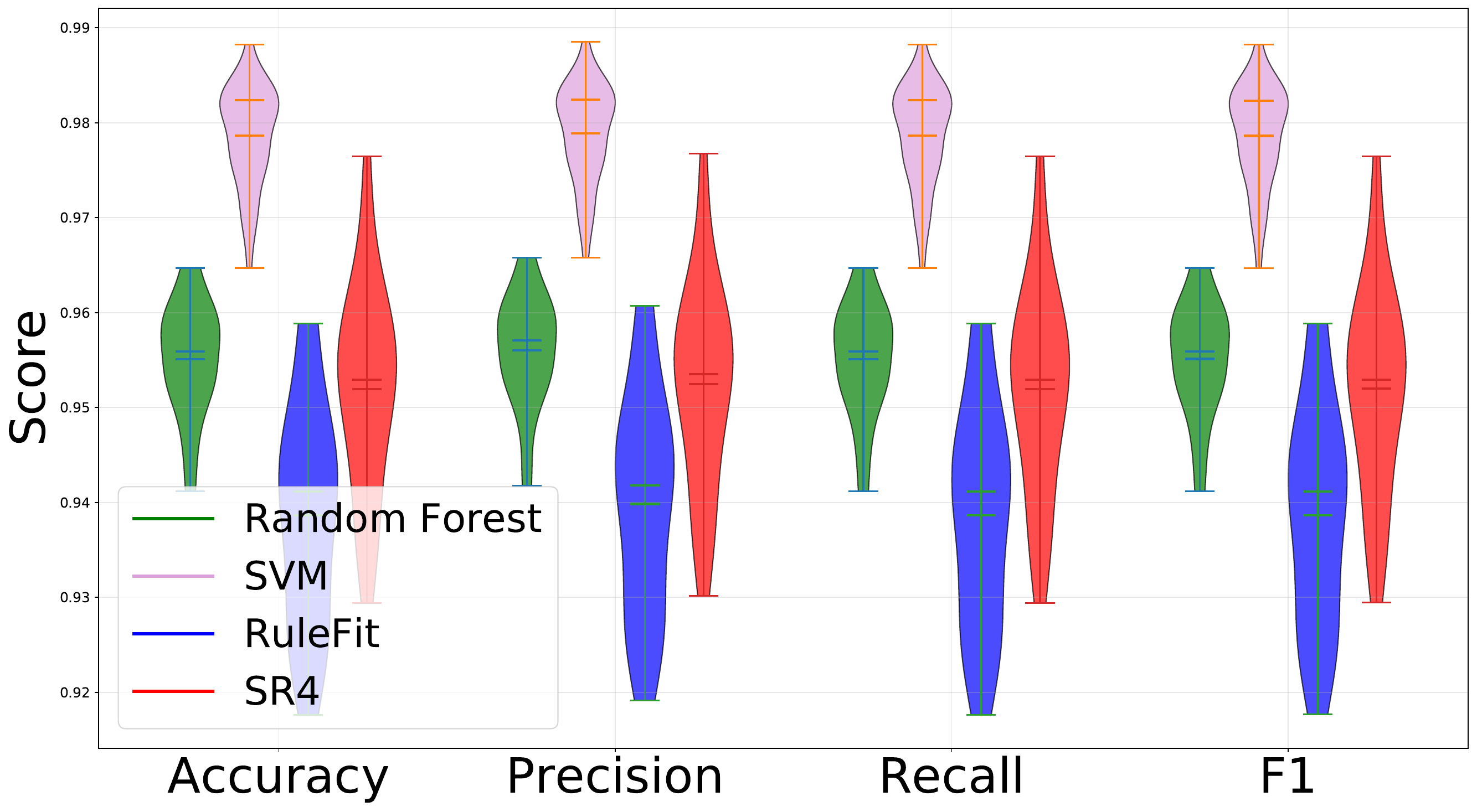}
    \subcaption{Vehicle}
\end{minipage}\hfill
\begin{minipage}{0.48\linewidth}
    \centering
    \includegraphics[width=\linewidth]{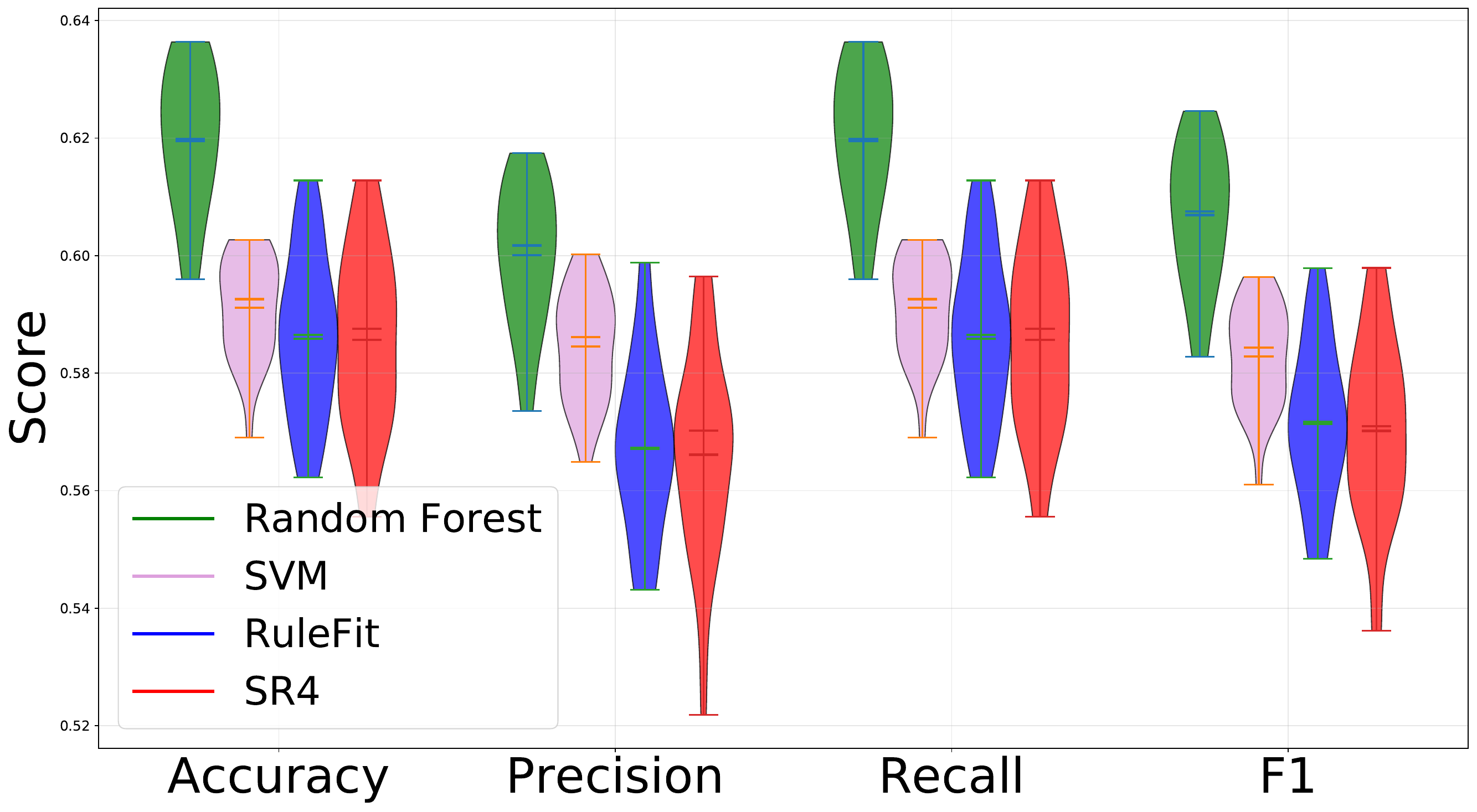}
    \subcaption{Yeast}
\end{minipage}

\caption{
Violin plot comparison of model (random forest, SVM, RuleFit, and SR4-fit) performance across multiple standard benchmark classification datasets for different prediction metrics.
}
\label{fig:violin_all_models_class}
\end{figure}

\begin{figure}
\centering
\scriptsize

\begin{minipage}{0.48\linewidth}
    \centering
    \includegraphics[width=\linewidth]{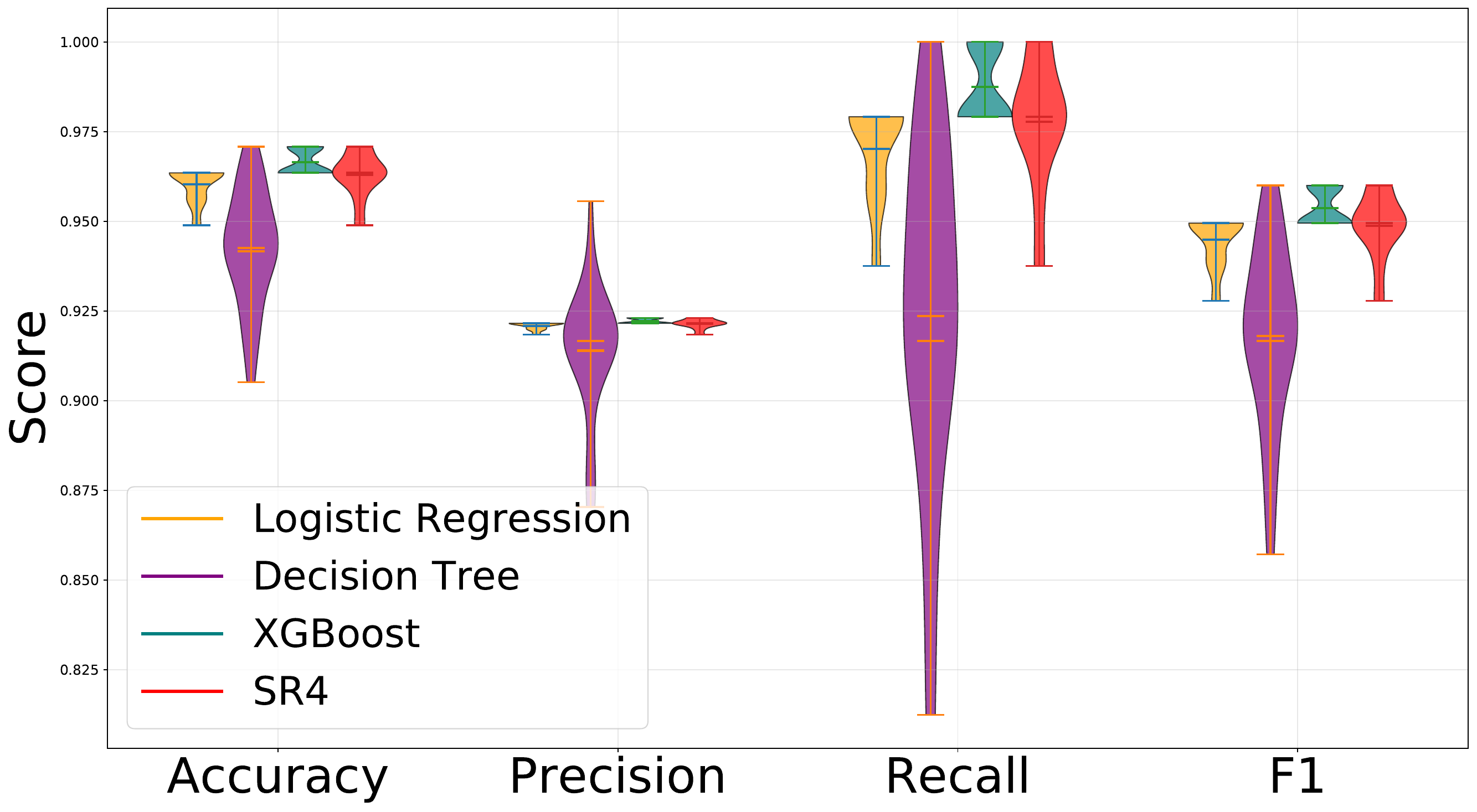}
    \subcaption{Breast Cancer}
\end{minipage}\hfill
\begin{minipage}{0.48\linewidth}
    \centering
    \includegraphics[width=\linewidth]{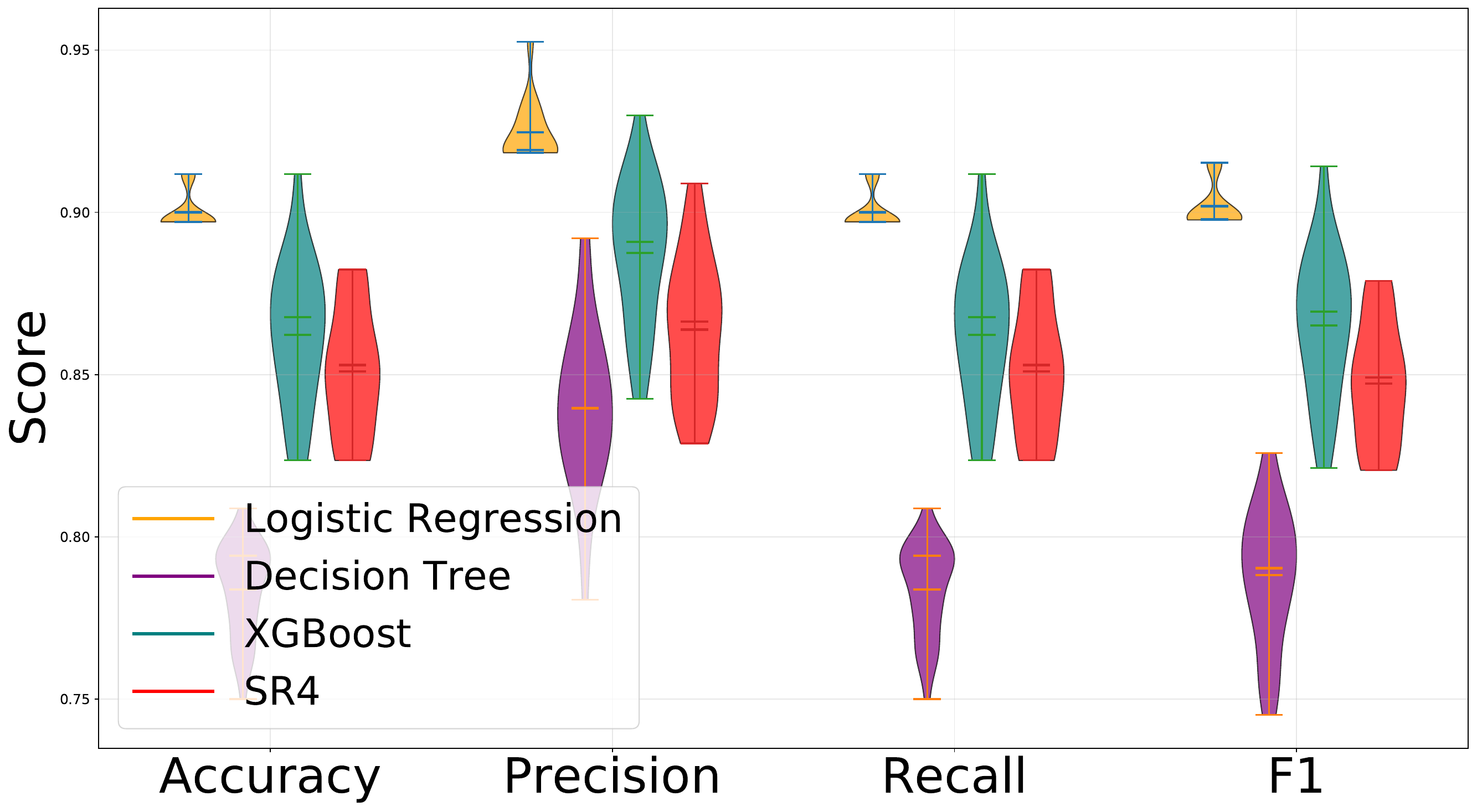}
    \subcaption{E. coli}
\end{minipage}

\vspace{0.4cm}

\begin{minipage}{0.48\linewidth}
    \centering
    \includegraphics[width=\linewidth]{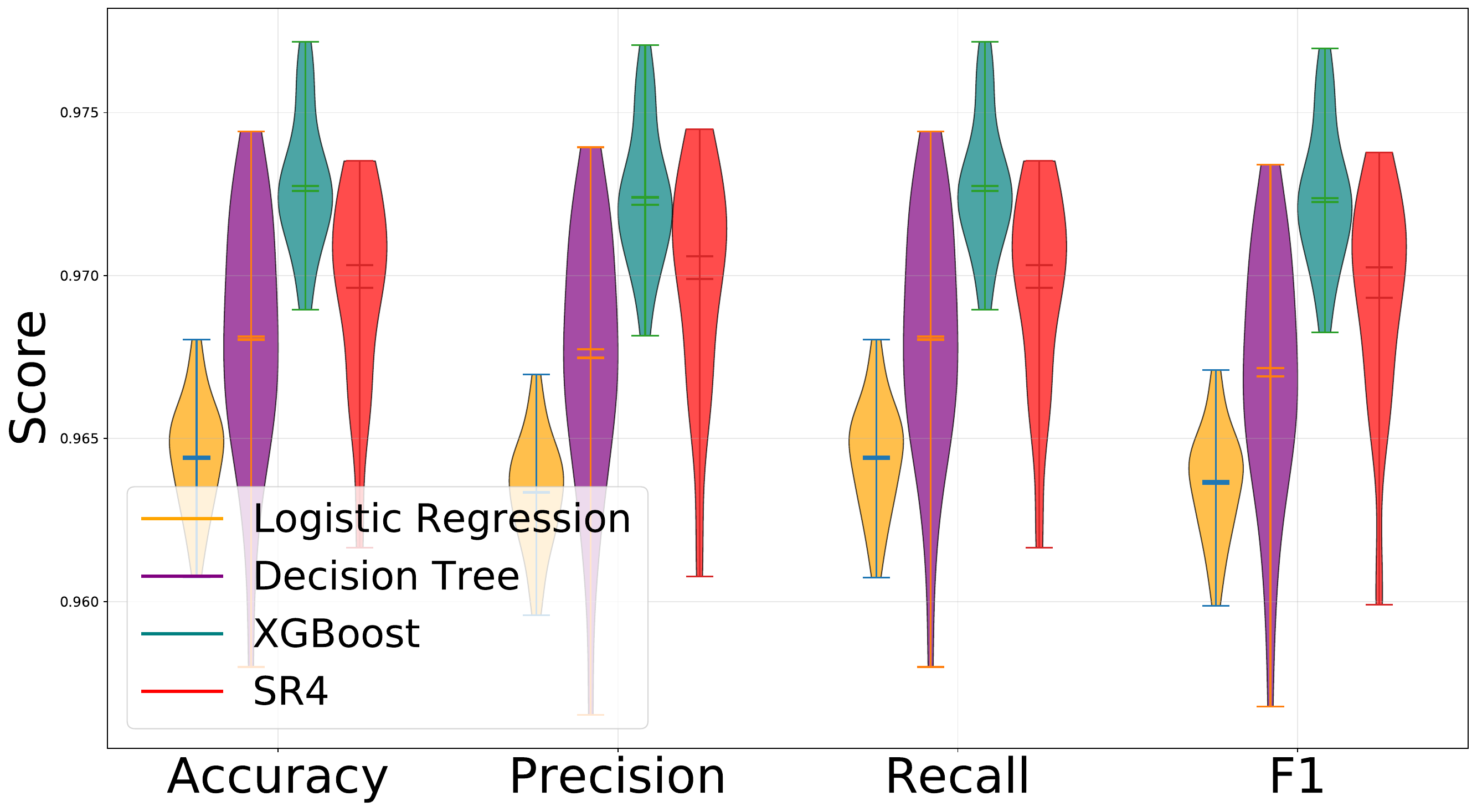}
    \subcaption{Page blocks}
\end{minipage}\hfill
\begin{minipage}{0.48\linewidth}
    \centering
    \includegraphics[width=\linewidth]{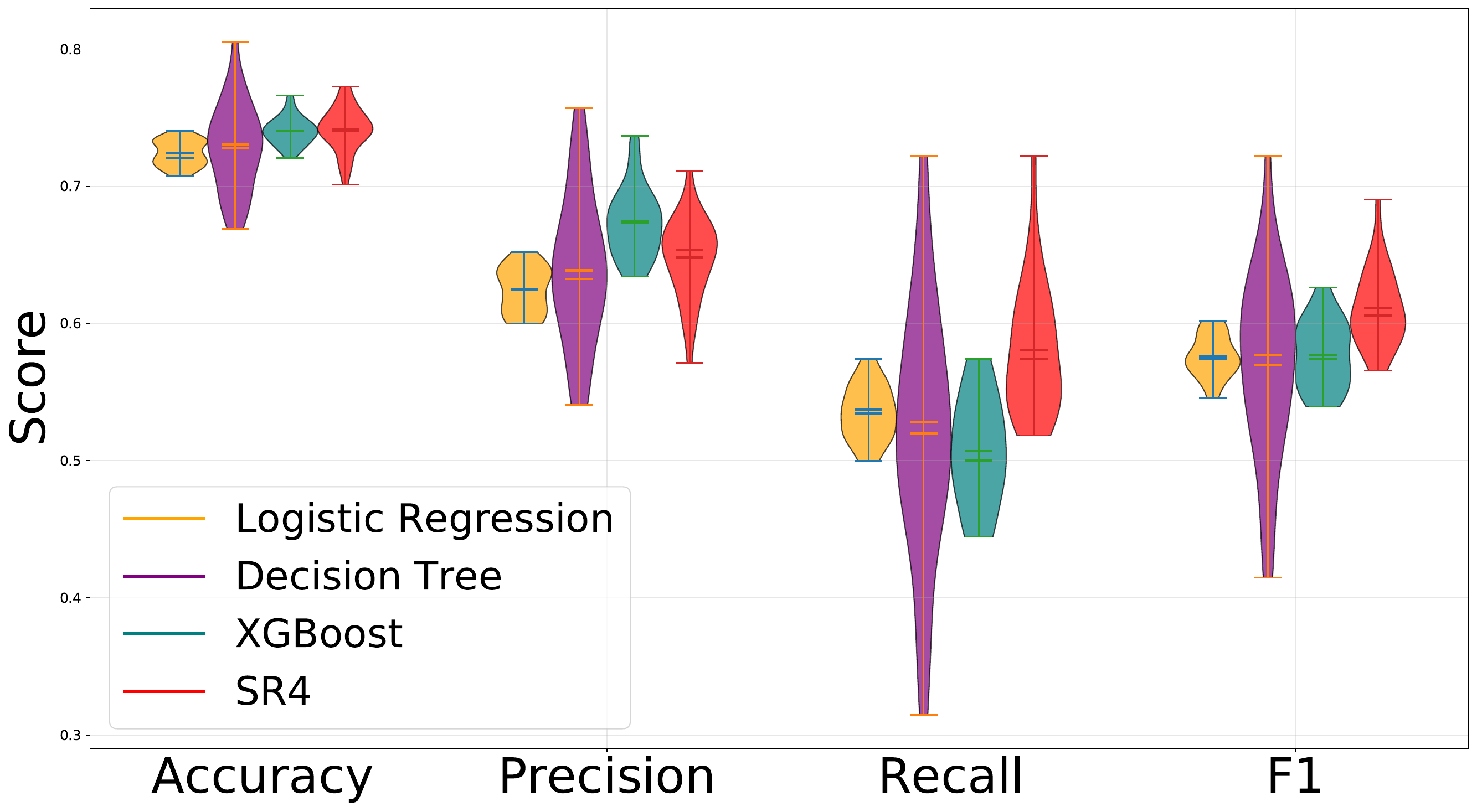}
    \subcaption{Pima Indians}
\end{minipage}

\vspace{0.4cm}

\begin{minipage}{0.48\linewidth}
    \centering
    \includegraphics[width=\linewidth]{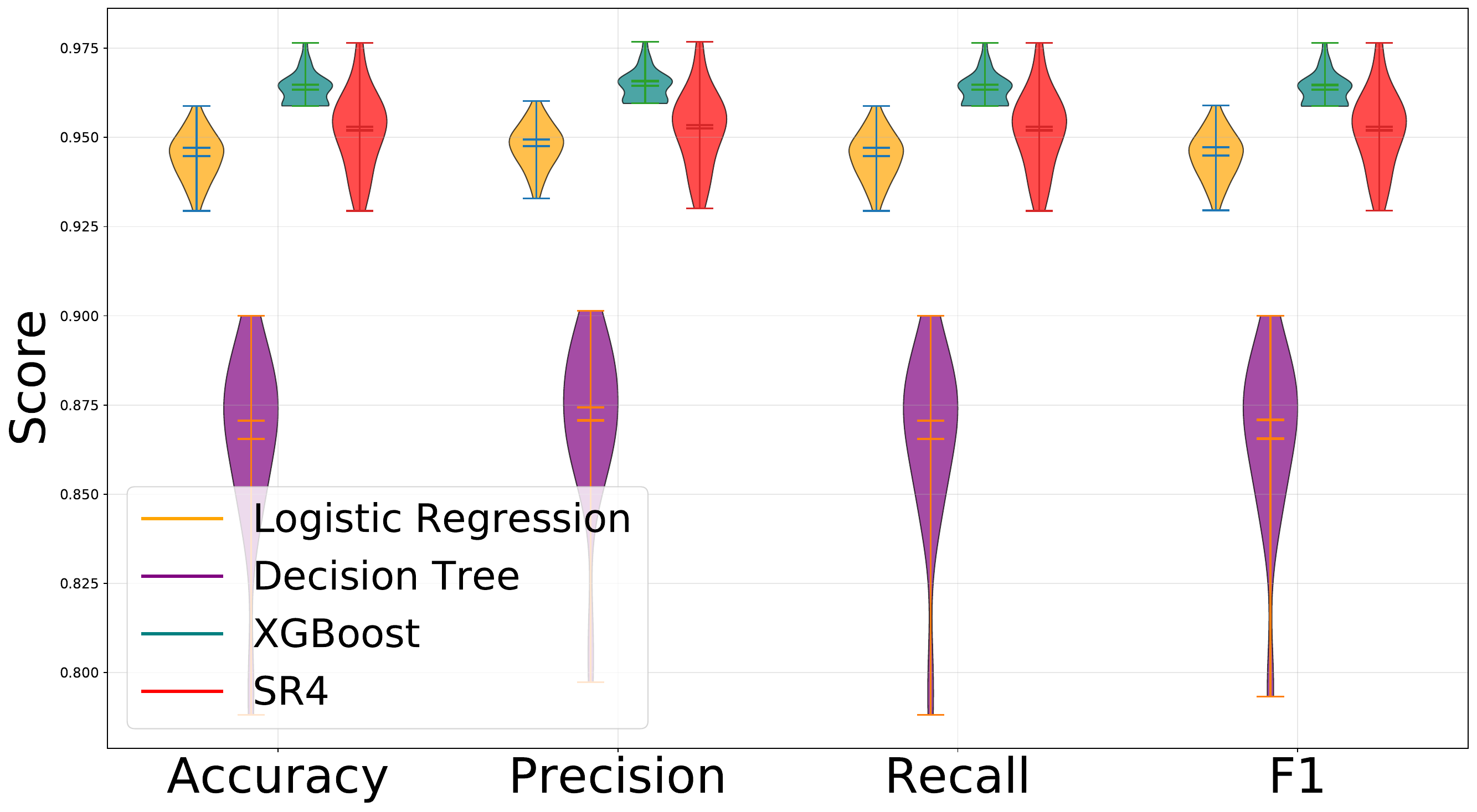}
    \subcaption{Vehicle}
\end{minipage}\hfill
\begin{minipage}{0.48\linewidth}
    \centering
    \includegraphics[width=\linewidth]{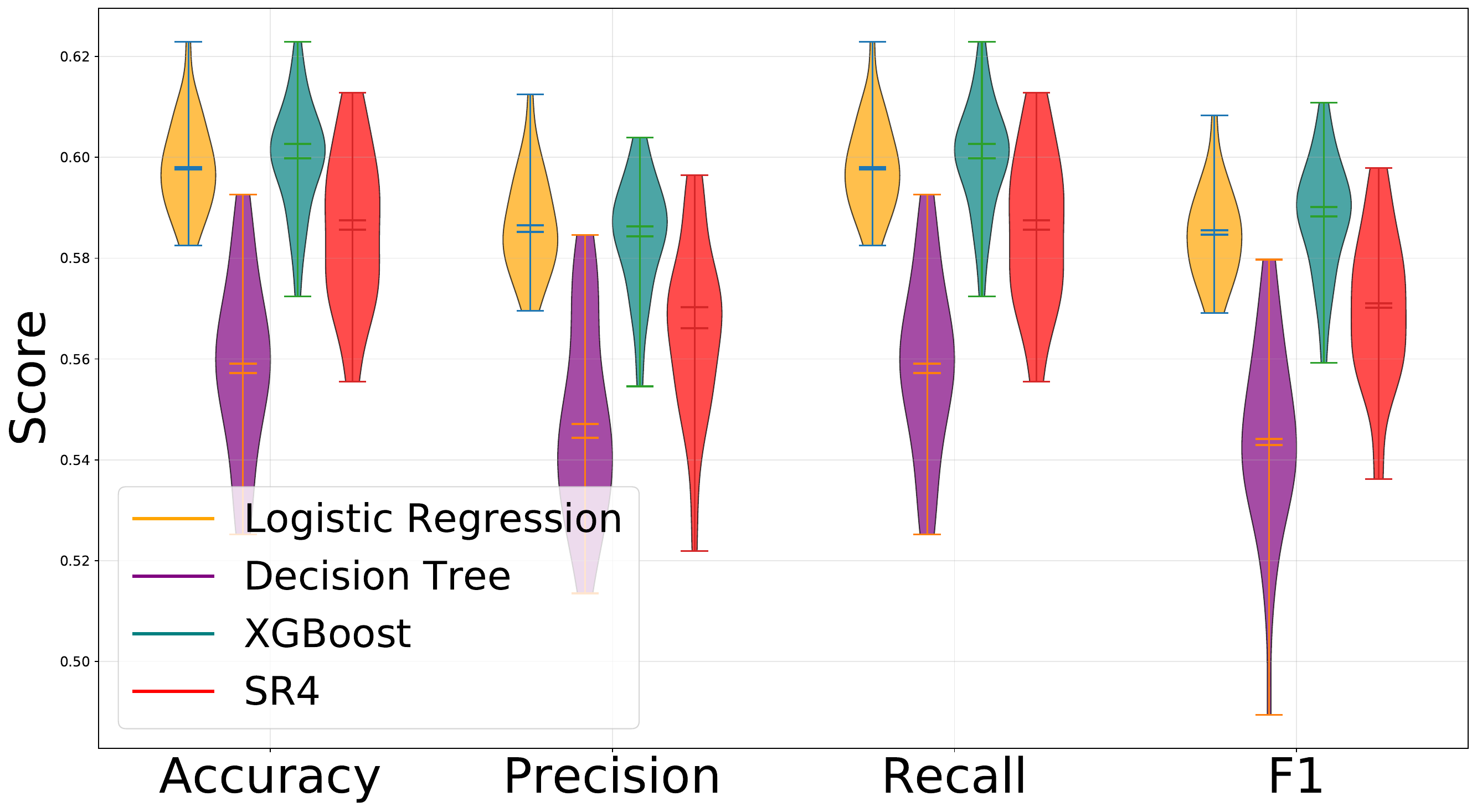}
    \subcaption{Yeast}
\end{minipage}

\caption{
Violin plot comparison of model (logistic regression, decision tree, and XGBoost) performance across multiple standard benchmark classification datasets for different prediction metrics.
}
\label{fig:violin_all_models_other_class}
\end{figure}


\begin{table}
\caption{
Average Dice–Sørensen Index $\pm$ standard deviation across 30 trials for public classification datasets. Bold indicates the best-performing method per dataset.
}

\centering
\setlength{\tabcolsep}{3pt}
\renewcommand{\arraystretch}{1.15}

\begin{tabular}{lcccc}
\toprule
\textbf{Dataset}
& \textbf{Random Forest} & \textbf{RuleFit} & \textbf{SR4-Fit} & \textbf{Decision Tree} \\
\midrule
Breast Cancer
& 0.3307$\pm$0.0186 & \textbf{0.4449$\pm$0.0199} & \textbf{0.4449$\pm$0.0199} & 0.0558$\pm$0.0540 \\
E. coli
& 0.1160$\pm$0.0150 & \textbf{0.1762$\pm$0.0140} & \textbf{0.1762$\pm$0.0140} & 0.0414$\pm$0.0347 \\
Page Blocks
& 0.1090$\pm$0.0155 & \textbf{0.3232$\pm$0.0294} & \textbf{0.3232$\pm$0.0294} & 0.0135$\pm$0.0090 \\
Pima Indians
& 0.0454$\pm$0.0078 & \textbf{0.5153$\pm$0.0115} & \textbf{0.5153$\pm$0.0115} & 0.0021$\pm$0.0032 \\
Vehicle
& 0.1993$\pm$0.0146 & \textbf{0.5231$\pm$0.0272} & \textbf{0.5231$\pm$0.0272} & 0.0593$\pm$0.0360 \\
Yeast
& 0.1182$\pm$0.0146 & \textbf{0.1807$\pm$0.0182} & \textbf{0.1807$\pm$0.0182} & 0.0121$\pm$0.0098 \\
\bottomrule
\end{tabular}

\label{tab:stability_class}
\end{table}


\begin{table}
\caption{
Average number of rules $\pm$ standard deviation across 30 trials for public classification datasets. Lower values indicate more compact models.
}

\centering
\setlength{\tabcolsep}{3pt}
\renewcommand{\arraystretch}{1.15}

\begin{tabular}{lcccc}
\toprule
\textbf{Dataset}
& \textbf{Random Forest} & \textbf{RuleFit} & \textbf{SR4-Fit} & \textbf{Decision Tree} \\
\midrule
Breast Cancer
& 104.43$\pm$11.58 & 21.13$\pm$1.11 & 21.13$\pm$1.11 & \textbf{9.10$\pm$0.66} \\
E. coli
& 117.17$\pm$13.48 & 57.00$\pm$0.00 & 57.00$\pm$0.00 & \textbf{11.03$\pm$0.67} \\
Page Blocks
& 50.27$\pm$8.77 & 43.10$\pm$2.15 & 43.10$\pm$2.15 & \textbf{18.23$\pm$0.89} \\
Pima Indians
& 37.47$\pm$8.19 & \textbf{15.67$\pm$0.48} & \textbf{15.67$\pm$0.48} & 20.77$\pm$1.96 \\
Vehicle
& 147.37$\pm$11.09 & 40.07$\pm$0.78 & 40.07$\pm$0.78 & \textbf{18.57$\pm$1.33} \\
Yeast
& 57.00$\pm$8.87 & 57.00$\pm$0.00 & 57.00$\pm$0.00 & \textbf{24.07$\pm$1.41} \\
\bottomrule
\end{tabular}

\label{tab:avg_rules_class}
\end{table}


\begin{table}
\caption{
Average rule complexity (number of conditions per rule) $\pm$ standard deviation across 30 trials for public classification datasets. Bold indicates less complex rules.
}

\centering
\setlength{\tabcolsep}{3pt}
\renewcommand{\arraystretch}{1.15}

\begin{tabular}{lcccc}
\toprule
\textbf{Dataset}
& \textbf{Random Forest} & \textbf{RuleFit} & \textbf{SR4-Fit} & \textbf{Decision Tree} \\
\midrule
Breast Cancer
& \textbf{1.99$\pm$0.01} & 2.97$\pm$0.12 & 2.97$\pm$0.12 & 3.42$\pm$0.15 \\
E. coli
& \textbf{1.97$\pm$0.01} & 3.55$\pm$0.01 & 3.55$\pm$0.01 & 3.60$\pm$0.08 \\
Page Blocks
& \textbf{1.97$\pm$0.01} & 2.47$\pm$0.04 & 2.47$\pm$0.04 & 4.41$\pm$0.05 \\
Pima Indians
& \textbf{1.96$\pm$0.01} & 2.28$\pm$0.07 & 2.28$\pm$0.07 & 4.61$\pm$0.08 \\
Vehicle
& \textbf{1.99$\pm$0.01} & 2.05$\pm$0.04 & 2.05$\pm$0.04 & 4.52$\pm$0.09 \\
Yeast
& \textbf{1.96$\pm$0.02} & 2.66$\pm$0.03 & 2.66$\pm$0.03 & 4.69$\pm$0.06 \\
\bottomrule
\end{tabular}

\label{tab:avg_complexity_class}
\end{table}


Across all six classification datasets and 30 trials, SR4-Fit demonstrates a strong balance of predictive performance, rule stability, and low variance, making it the most robust and interpretable method overall. Although individual models occasionally outperform it on accuracy, SR4-Fit repeatedly delivers the most consistent and stable behavior across trials, as reflected in the line plots and violin plots.

In the Breast Cancer dataset, XGBoost has the highest performance with respect to prediction metrics, but SR4-Fit trails very closely behind XGBoost, as seen in Figure~\ref{fig:lineplot_breast}. The violin plots for breast cancer in Figure~\ref{fig:violin_all_models_class} and Figure~\ref{fig:violin_all_models_other_class} reinforce these findings. Despite this small difference in accuracy, SR4-Fit significantly outperforms in stability, achieving the highest Dice–Sørensen Index, along with RuleFit, as seen in Table~\ref{tab:stability_class}. Table~\ref{tab:avg_rules_class} shows that the decision tree has a compact structure, whereas with respect to average rule complexity in Table~\ref{tab:avg_complexity_class} random forest has lower average rule complexity compared to all rule-based models.

In the E. coli dataset, SVM has the highest performance with respect to prediction metrics, as seen in Figure~\ref{fig:lineplot_ecoli}. The violin plots for ecoli in Figure~\ref{fig:violin_all_models_class} and Figure~\ref{fig:violin_all_models_other_class} reinforce these findings.  SR4-Fit and RuleFit significantly outperform in stability, achieving the highest Dice–Sørensen Index, as seen in Table~\ref{tab:stability_class}. Table~\ref{tab:avg_rules_class} shows that the decision tree has a compact structure, whereas with respect to average rule complexity in Table~\ref{tab:avg_complexity_class} random forest has lower average rule complexity compared to all rule-based models.

On the Page blocks dataset, random forest has the highest performance with respect to prediction metrics, but SR4-Fit trails very closely behind random forest, as seen in Figure~\ref{fig:lineplot_page}. The violin plots for page blocks in Figure~\ref{fig:violin_all_models_class} and Figure~\ref{fig:violin_all_models_other_class} show the distribution of the prediction metrics across trials.  SR4-Fit and RuleFit continue to significantly outperform in stability, achieving the highest Dice–Sørensen Index, as seen in Table~\ref{tab:stability_class}. Table~\ref{tab:avg_rules_class} shows that the decision tree has a compact structure, whereas with respect to average rule complexity in Table~\ref{tab:avg_complexity_class} random forest has lower average rule complexity compared to all rule-based models.

On the Pima Indians dataset, random forest continues to have the highest performance with respect to prediction metrics, but SR4-Fit trails very closely behind random forest, as seen in Figure~\ref{fig:lineplot_pima}. The violin plots for Pima Indians in Figure~\ref{fig:violin_all_models_class} and Figure~\ref{fig:violin_all_models_other_class} show the distribution of the prediction metrics across trials.  SR4-Fit and RuleFit continue to significantly outperform in stability, achieving the highest Dice–Sørensen Index, as seen in Table~\ref{tab:stability_class}. Table~\ref{tab:avg_rules_class} shows that the SR4-Fit and RuleFit have a compact structure, whereas, with respect to average rule complexity in Table~\ref{tab:avg_complexity_class} random forests continue to have lower average rule complexity compared to all rule-based models.

In the vehicle dataset, SVM has the highest performance with respect to prediction metrics, but SR4-Fit trails very closely behind SVM, as seen in Figure~\ref{fig:lineplot_vehicle}. The violin plots for the vehicle in Figure~\ref{fig:violin_all_models_class} and Figure~\ref{fig:violin_all_models_other_class} show the distribution of the prediction metrics across trials.  SR4-Fit and RuleFit continue to significantly outperform in stability, achieving the highest Dice–Sørensen Index, as seen in Table~\ref{tab:stability_class}. Table~\ref{tab:avg_rules_class} shows that the decision tree has a compact structure, whereas, with respect to average rule complexity in Table~\ref{tab:avg_complexity_class} random forests continue to have lower average rule complexity compared to all rule-based models.

Finally, in the vehicle dataset, random forest has the highest performance with respect to prediction metrics, as seen in Figure~\ref{fig:lineplot_yeast}. The violin plots for the yeast in Figure~\ref{fig:violin_all_models_class} and Figure~\ref{fig:violin_all_models_other_class} show the distribution of the prediction metrics across trials.  SR4-Fit and RuleFit continue to significantly outperform in stability, achieving the highest Dice–Sørensen Index, as seen in Table~\ref{tab:stability_class}. Table~\ref{tab:avg_rules_class} shows that the decision tree has a compact structure, whereas, with respect to average rule complexity in Table~\ref{tab:avg_complexity_class} random forests continue to have lower average rule complexity compared to all rule-based models.

\FloatBarrier

\section{Results on Standard Public Datasets Regression}\label{sup_sec_BenchR}
\begin{figure}
\centering
\begin{minipage}{0.49\linewidth}
    \centering
    \includegraphics[width=\linewidth]{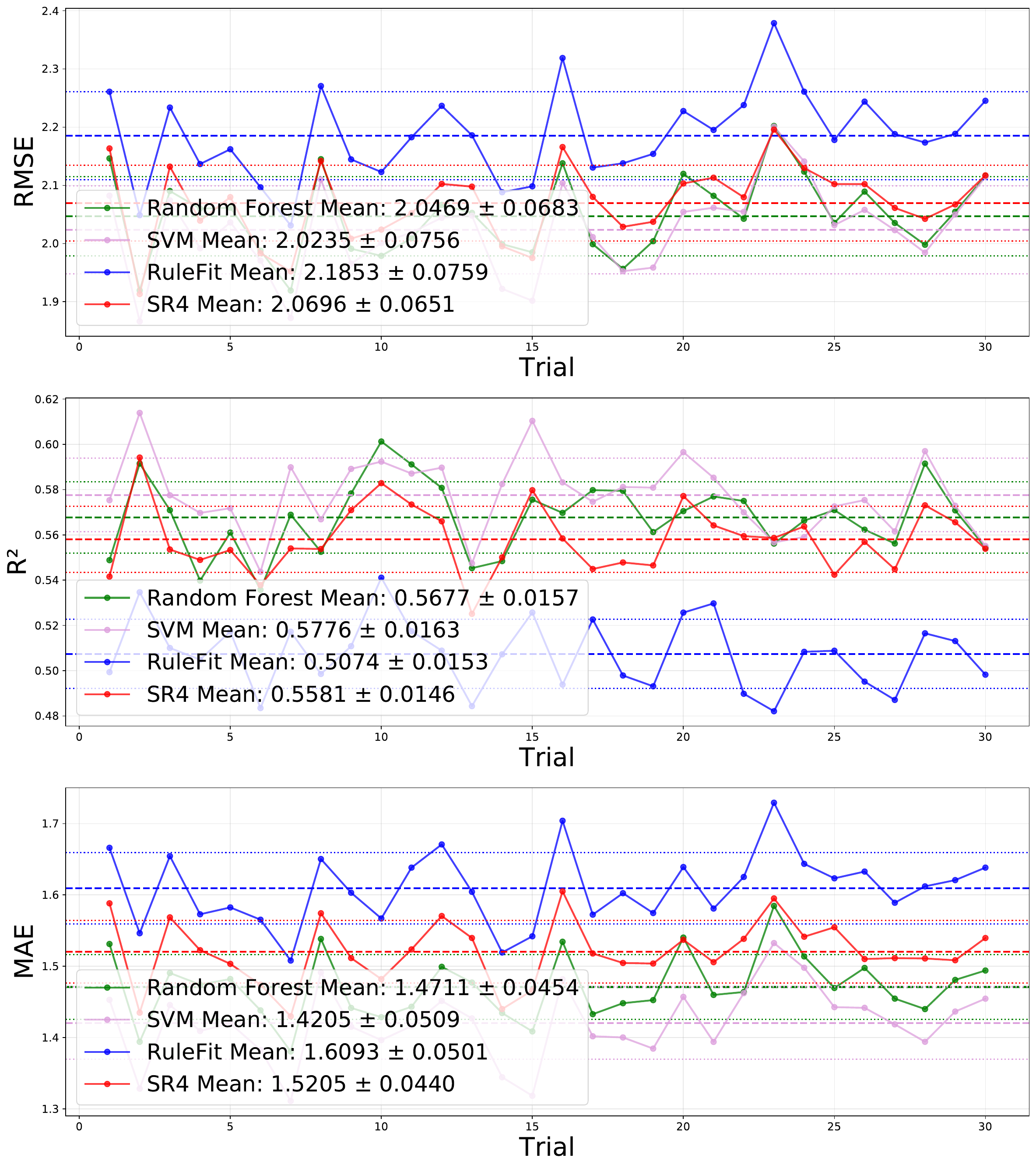}
\end{minipage}\hfill
\begin{minipage}{0.49\linewidth}
    \centering
    \includegraphics[width=\linewidth]{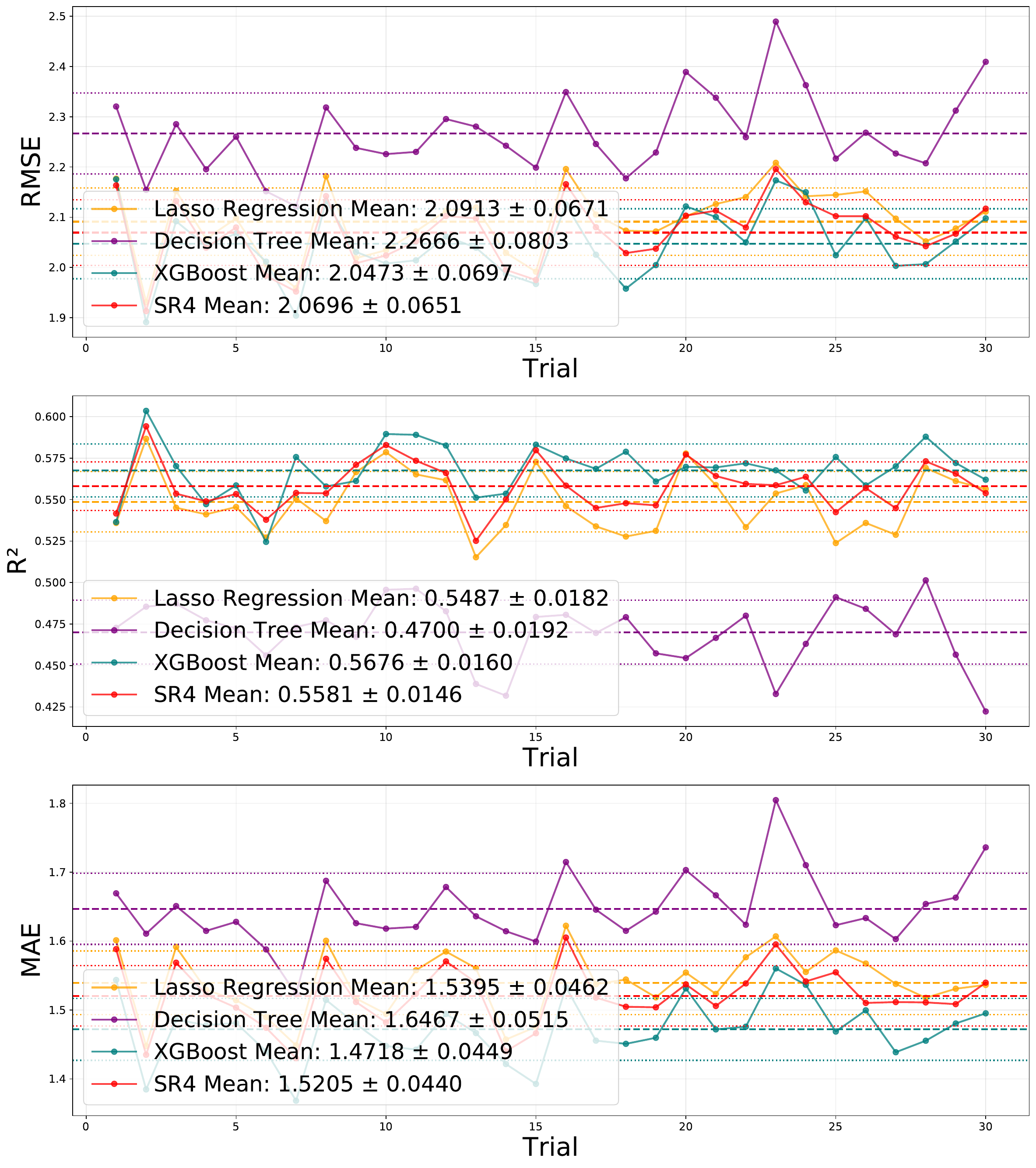}
\end{minipage}

\caption{
Line plot comparison of model performance metrics for abalone data across 30 trials.
The left panel reports results obtained for models—random forest, SVM, RuleFit, and SR4-fit ,
whereas the right panel shows results with models LASSO regression, decision tree, and XGBoost.
}
\label{fig:lineplot_abalone}
\end{figure}


\begin{figure}
\centering
\begin{minipage}{0.49\linewidth}
    \centering
    \includegraphics[width=\linewidth]{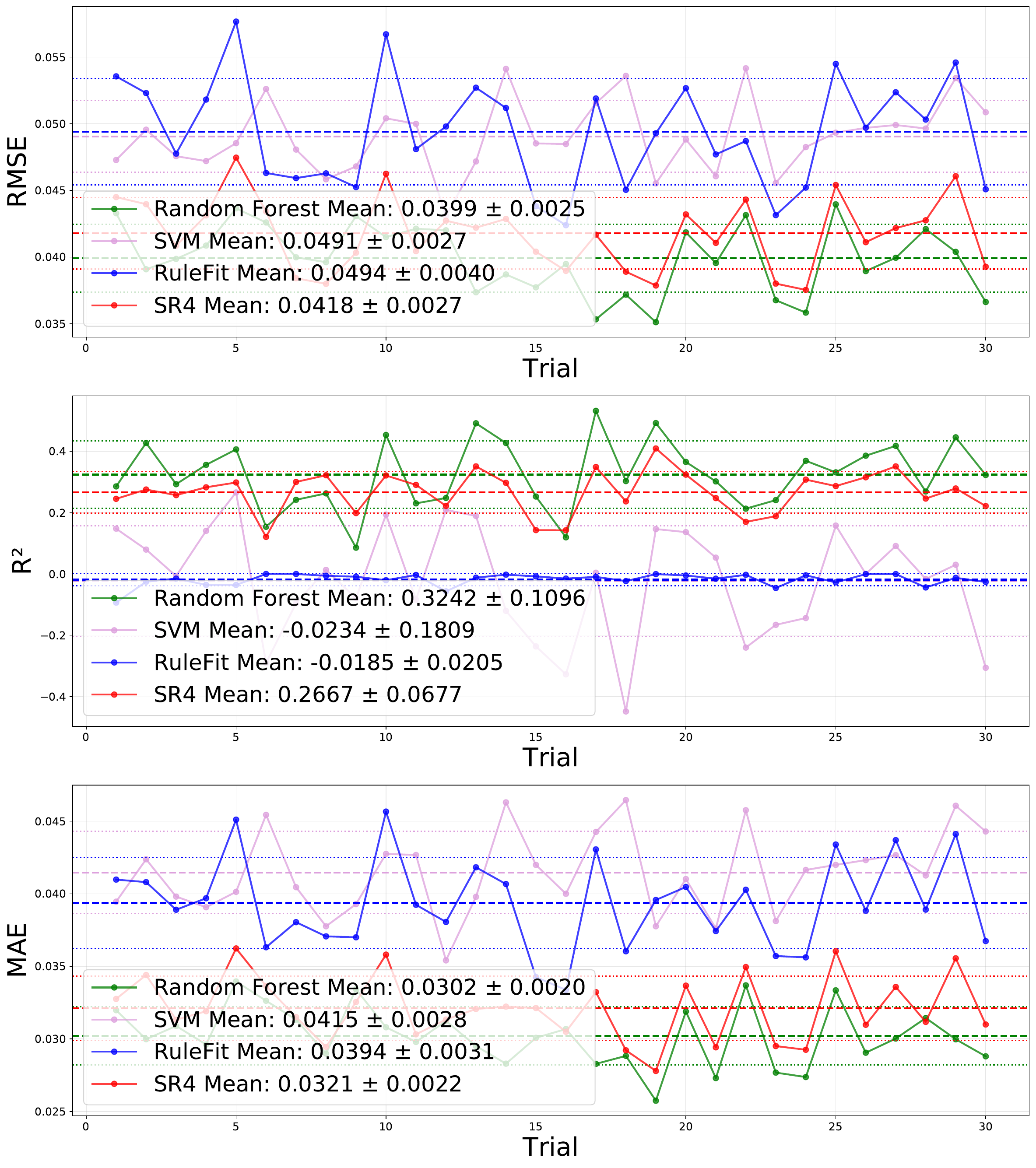}
\end{minipage}\hfill
\begin{minipage}{0.49\linewidth}
    \centering
    \includegraphics[width=\linewidth]{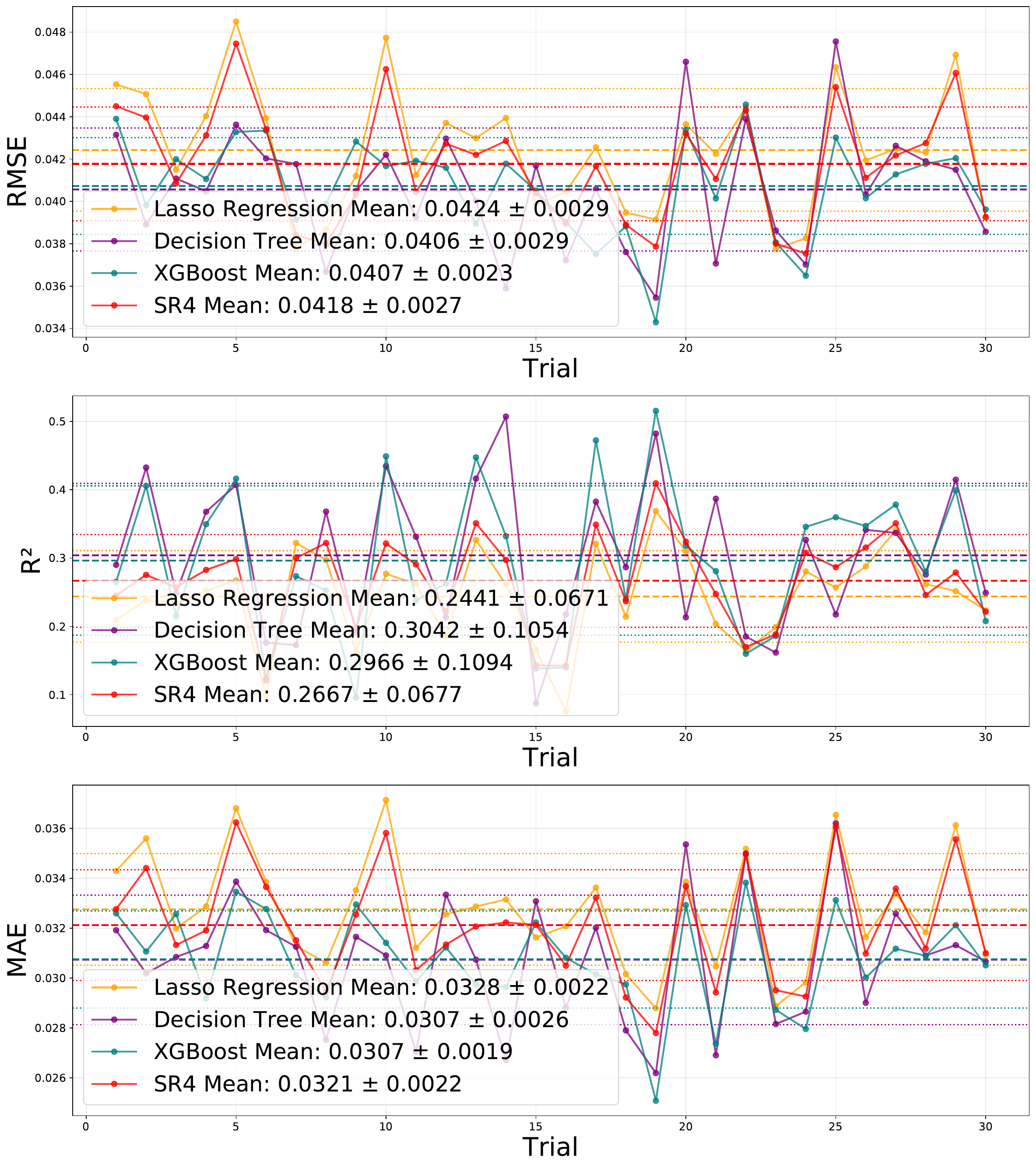}
\end{minipage}

\caption{
Line plot comparison of model performance metrics for bone data across 30 trials.
The left panel reports results obtained for models—random forest, SVM, RuleFit, and SR4-fit,
whereas the right panel shows results with models LASSO regression, decision tree, and XGBoost.
}
\label{fig:lineplot_bone}
\end{figure}


\begin{figure}
\centering
\begin{minipage}{0.49\linewidth}
    \centering
    \includegraphics[width=\linewidth]{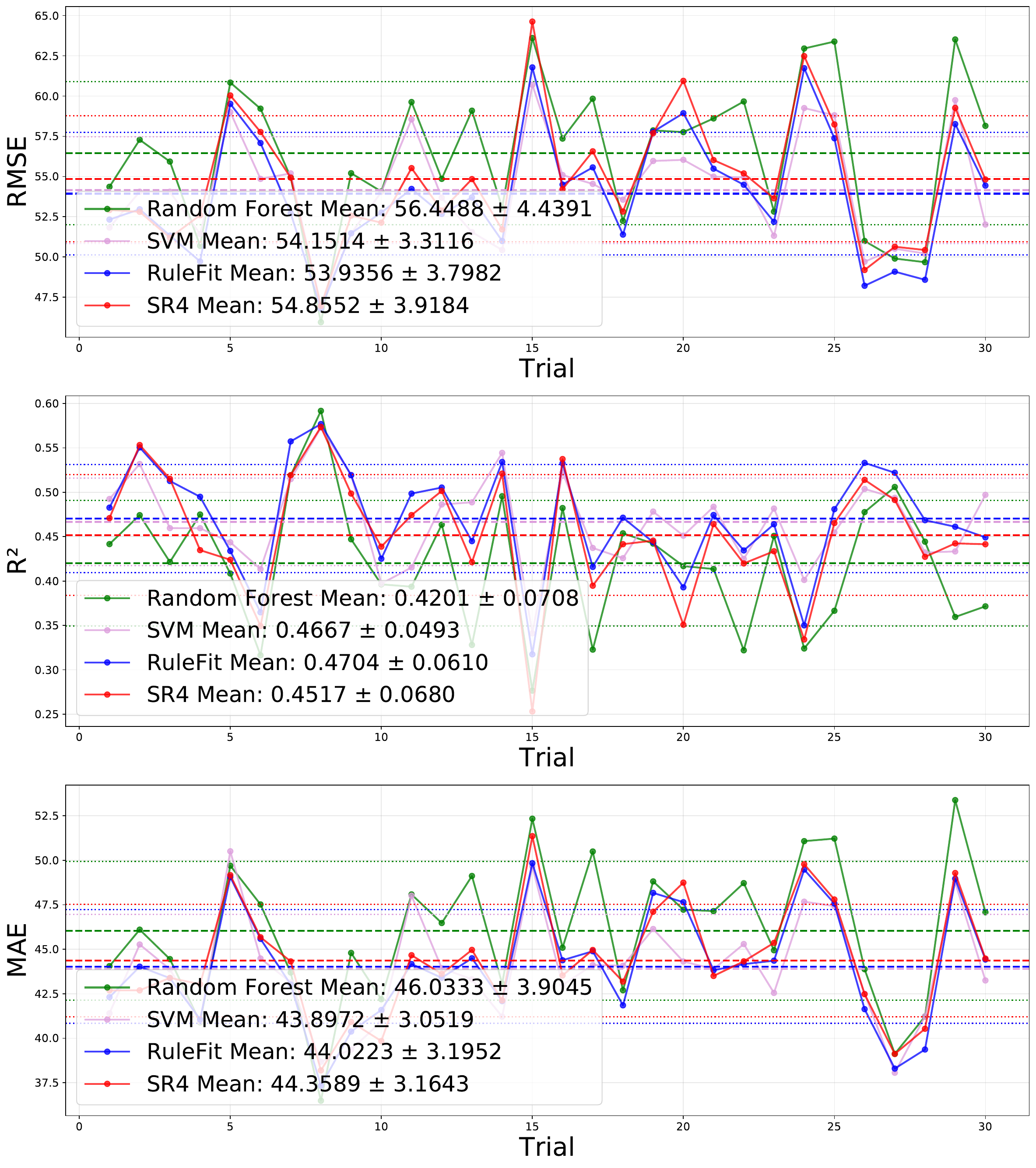}
\end{minipage}\hfill
\begin{minipage}{0.49\linewidth}
    \centering
    \includegraphics[width=\linewidth]{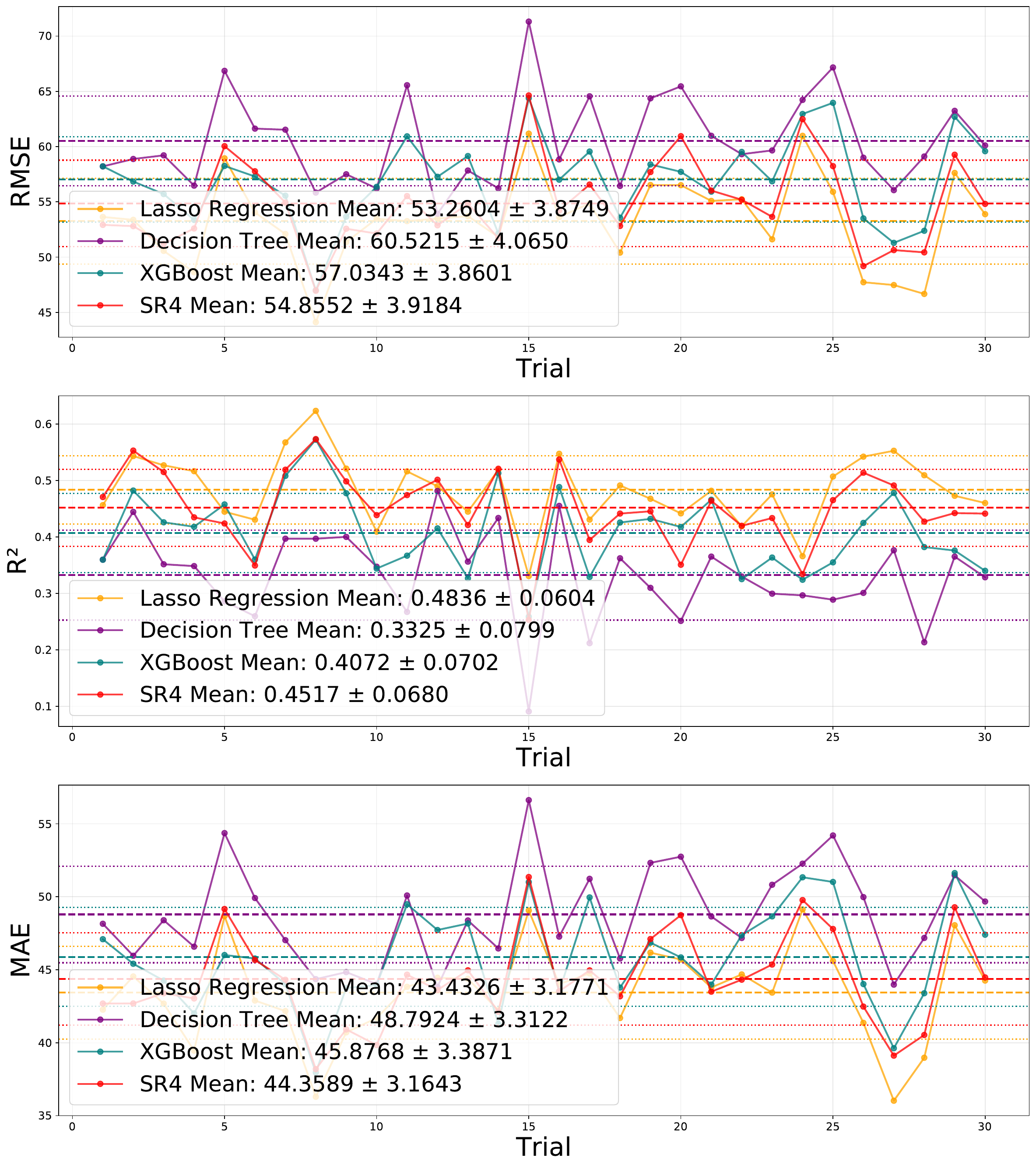}
\end{minipage}

\caption{
Line plot comparison of model performance metrics for diabetes data across 30 trials.
The left panel reports results obtained for models—random forest, SVM, RuleFit, and SR4-fit ,
whereas the right panel shows results with models LASSO regression, decision tree, and XGBoost.
}
\label{fig:lineplot_diabetes}
\end{figure}


\begin{figure}
\centering
\begin{minipage}{0.49\linewidth}
    \centering
    \includegraphics[width=\linewidth]{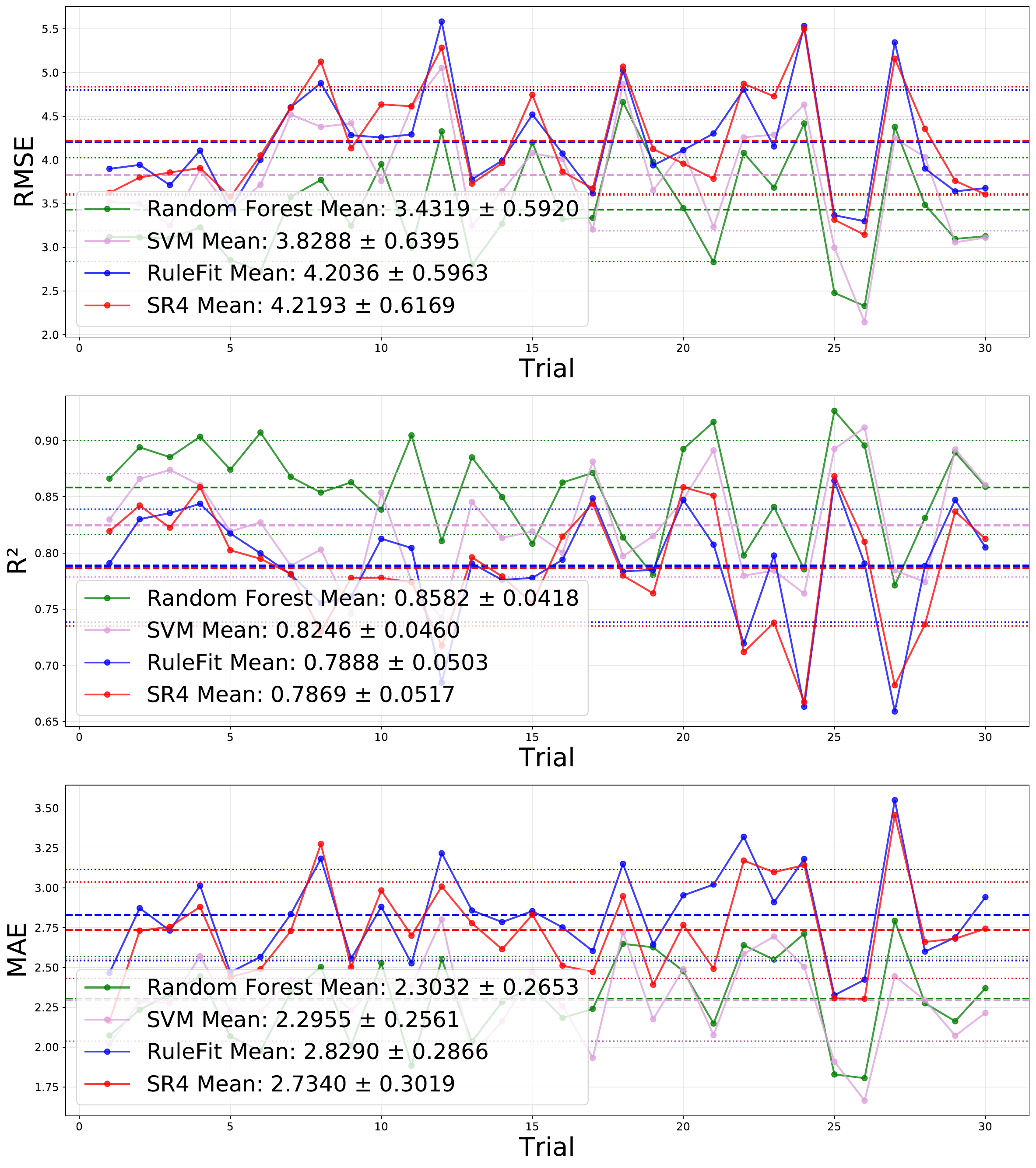}
\end{minipage}\hfill
\begin{minipage}{0.49\linewidth}
    \centering
    \includegraphics[width=\linewidth]{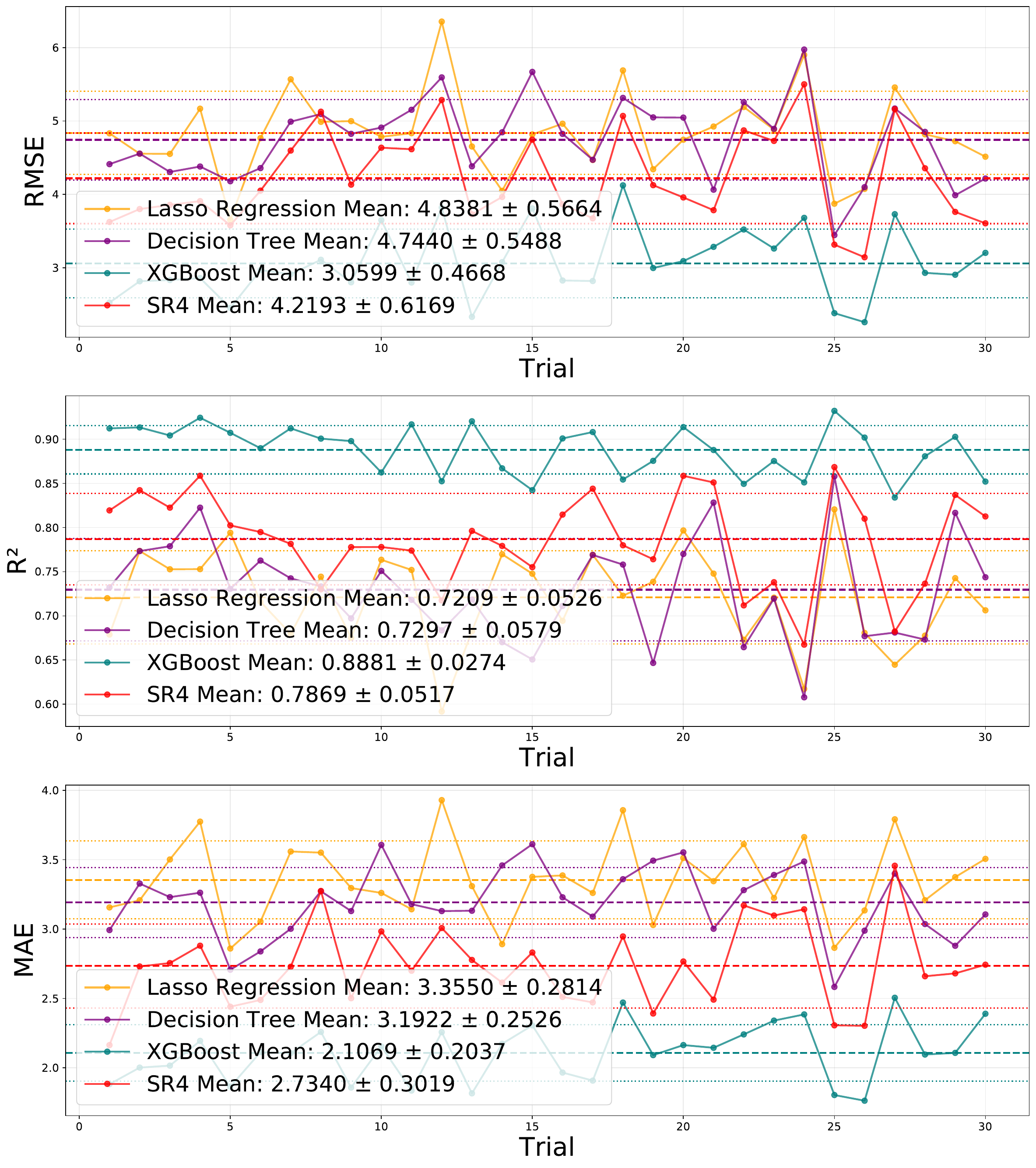}
\end{minipage}

\caption{
Line plot comparison of model performance metrics for housing data across 30 trials.
The left panel reports results obtained for models—random forest, SVM, RuleFit, and SR4-fit ,
whereas the right panel shows results with models LASSO regression, decision tree, and XGBoost.
}
\label{fig:lineplot_housing}
\end{figure}


\begin{figure}
\centering
\begin{minipage}{0.49\linewidth}
    \centering
    \includegraphics[width=\linewidth]{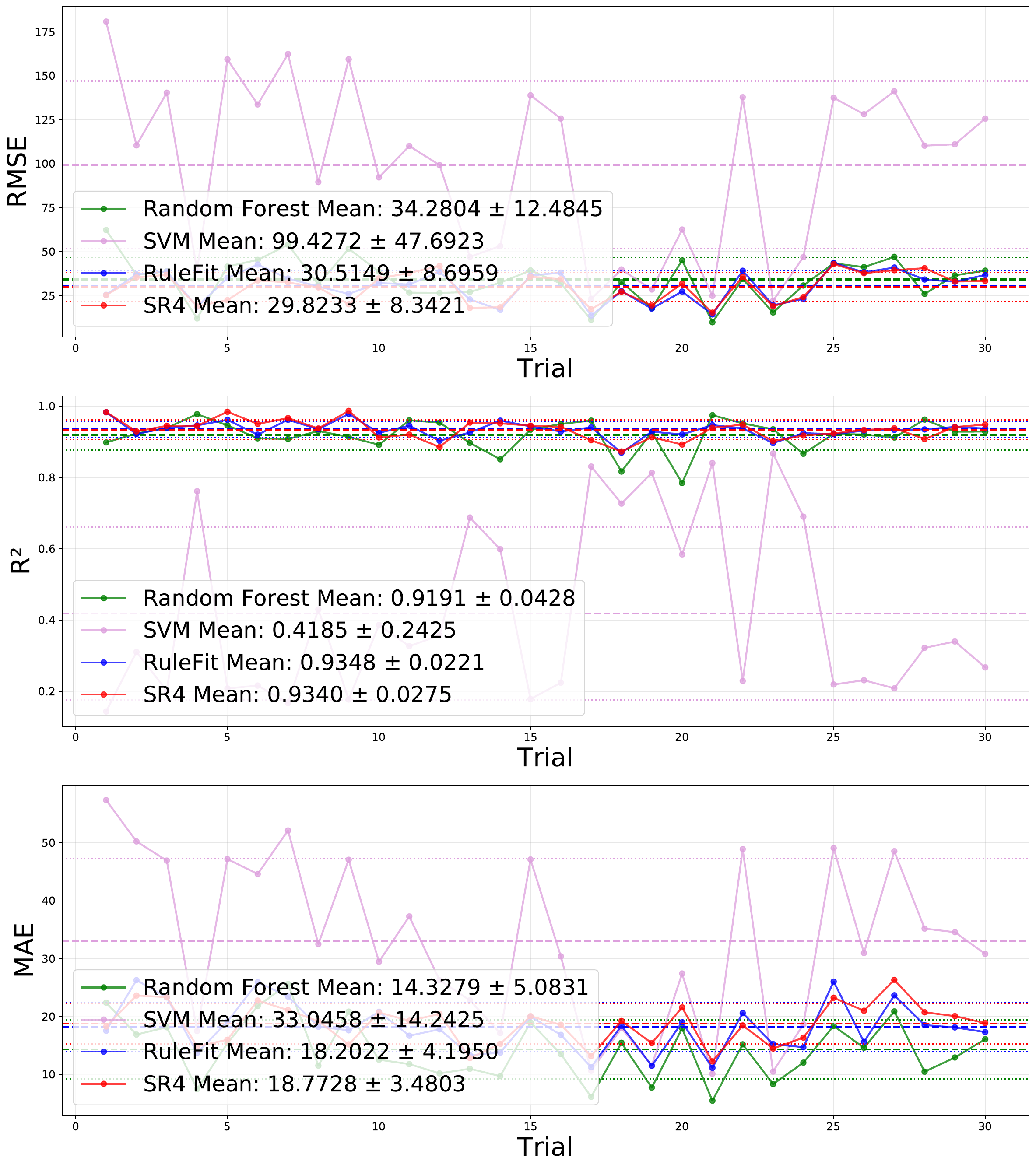}
\end{minipage}\hfill
\begin{minipage}{0.49\linewidth}
    \centering
    \includegraphics[width=\linewidth]{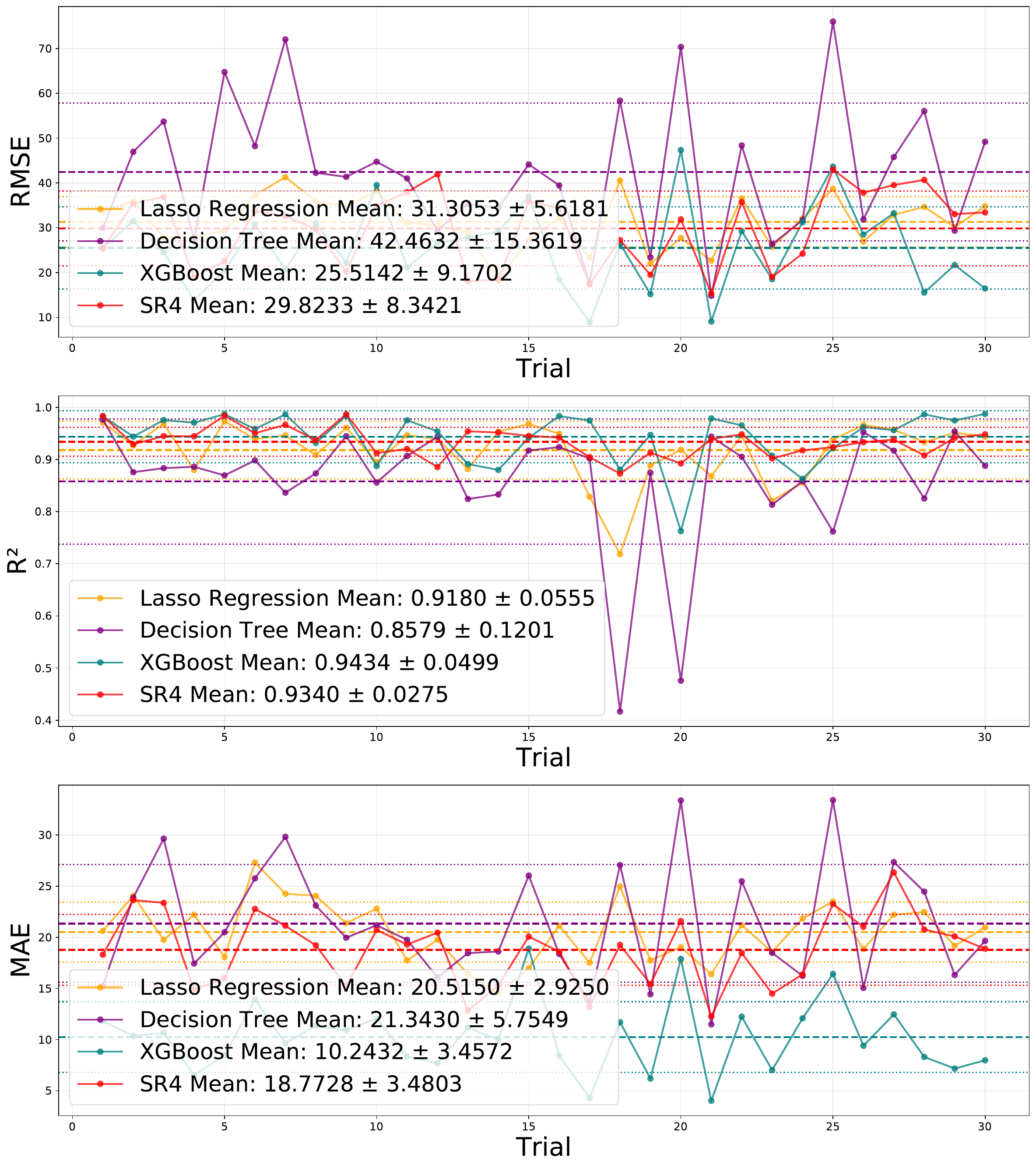}
\end{minipage}

\caption{
Line plot comparison of model performance metrics for machine data across 30 trials.
The left panel reports results obtained for models—random forest, SVM, RuleFit, and SR4-fit ,
whereas the right panel shows results with models LASSO regression, decision tree, and XGBoost.
}
\label{fig:lineplot_machine}
\end{figure}


\begin{figure}
\centering
\begin{minipage}{0.49\linewidth}
    \centering
    \includegraphics[width=\linewidth]{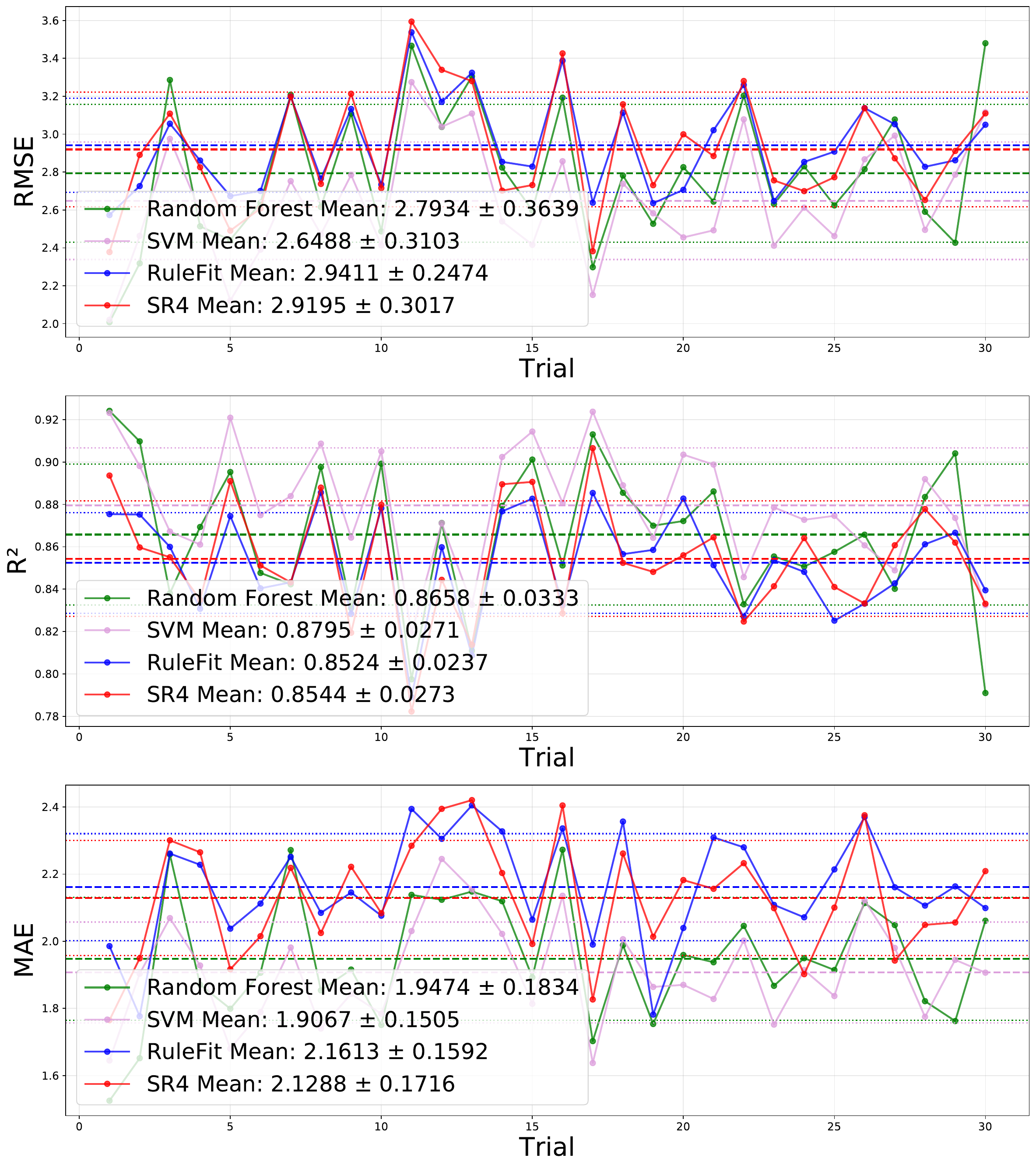}
\end{minipage}\hfill
\begin{minipage}{0.49\linewidth}
    \centering
    \includegraphics[width=\linewidth]{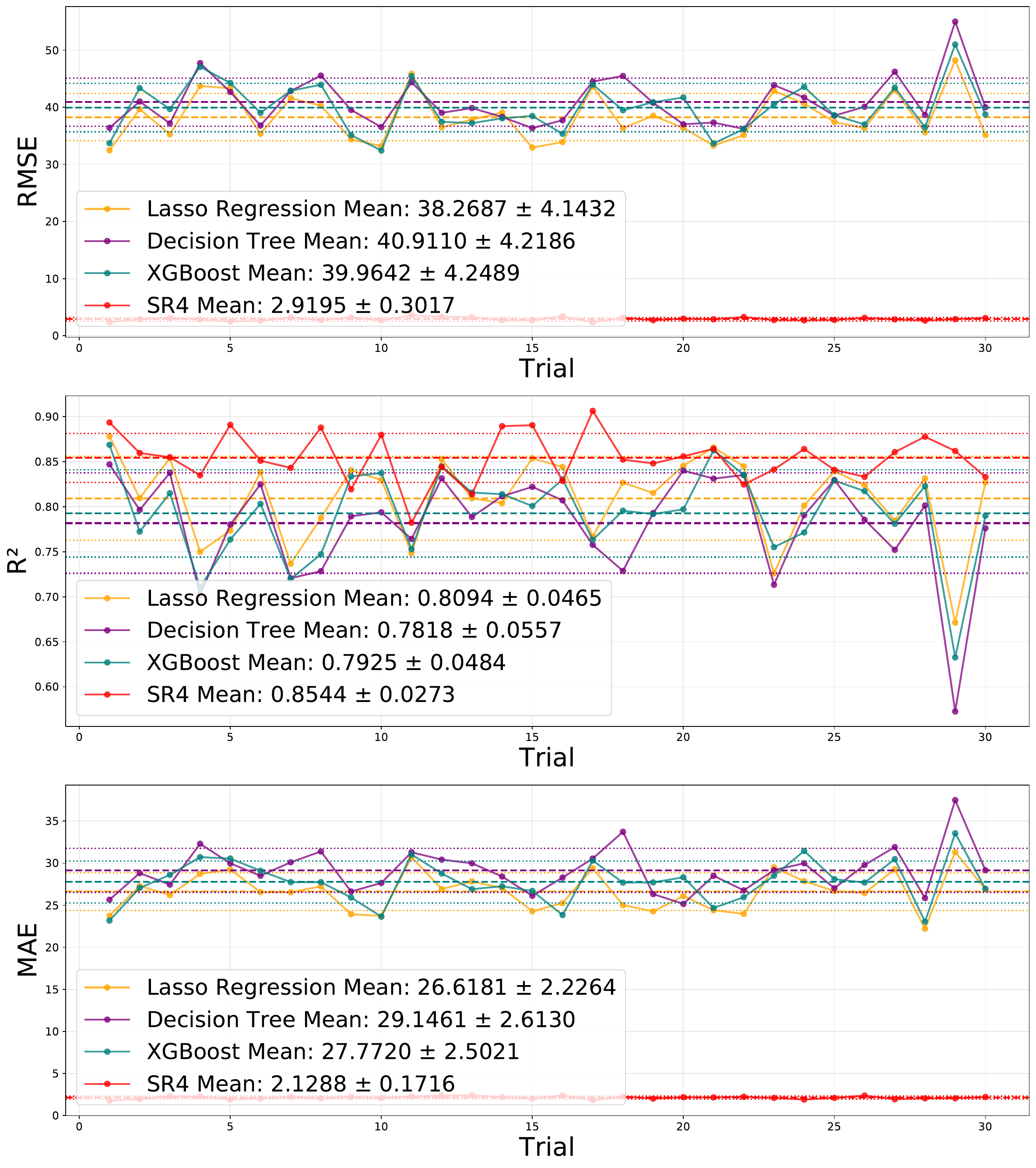}
\end{minipage}

\caption{
Line plot comparison of model performance metrics for MPG data across 30 trials.
The left panel reports results obtained for models—random forest, SVM, RuleFit, and SR4-fit ,
whereas the right panel shows results with models LASSO regression, decision tree, and XGBoost.
}
\label{fig:lineplot_mpg}
\end{figure}


\begin{figure}
\centering
\begin{minipage}{0.49\linewidth}
    \centering
    \includegraphics[width=\linewidth]{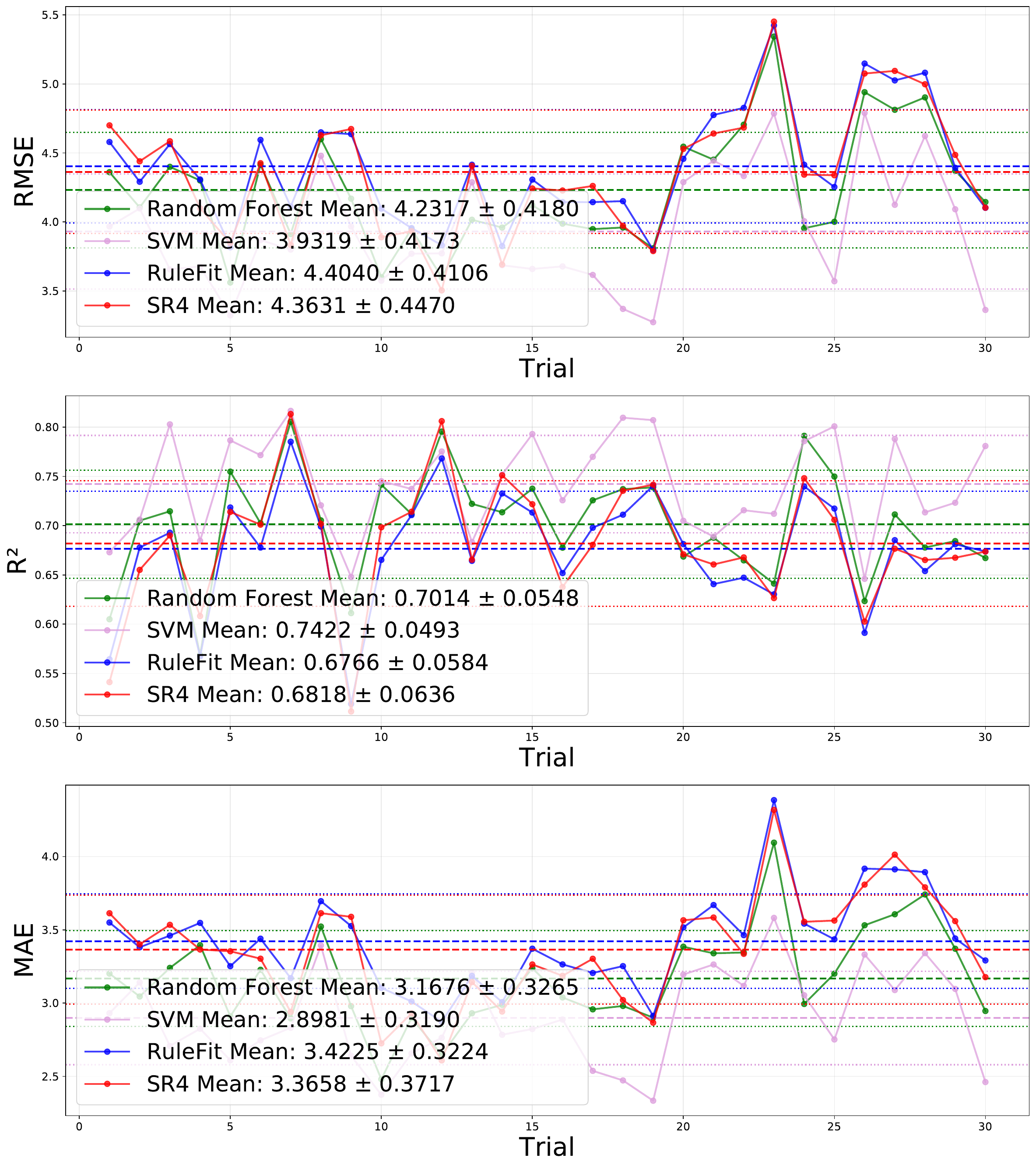}
\end{minipage}\hfill
\begin{minipage}{0.49\linewidth}
    \centering
    \includegraphics[width=\linewidth]{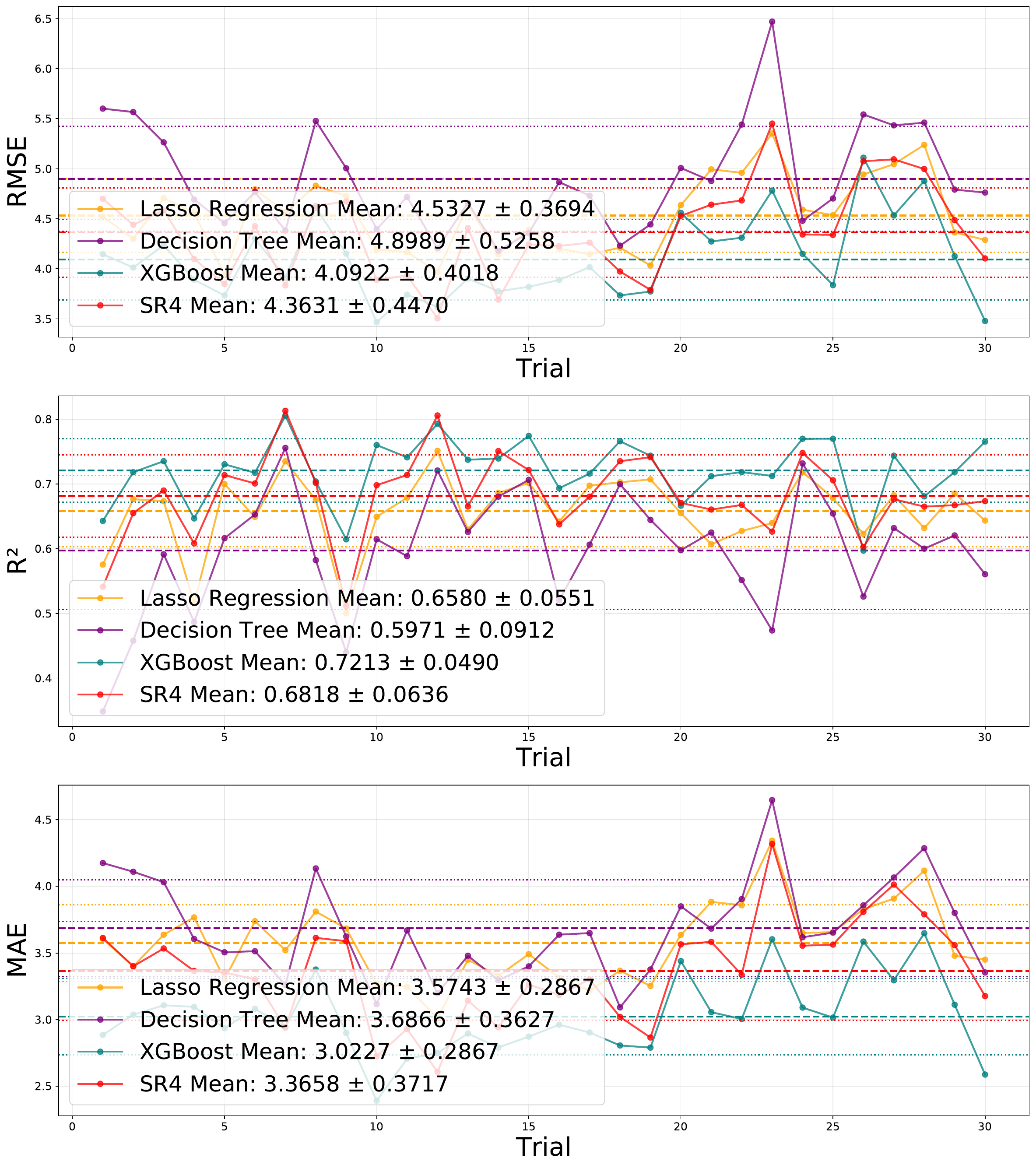}
\end{minipage}

\caption{
Line plot comparison of model performance metrics for ozone data across 30 trials.
The left panel reports results obtained for models—random forest, SVM, RuleFit, and SR4-fit ,
whereas the right panel shows results with models LASSO regression, decision tree, and XGBoost.
}
\label{fig:lineplot_ozone}
\end{figure}


\begin{figure}
\centering
\begin{minipage}{0.49\linewidth}
    \centering
    \includegraphics[width=\linewidth]{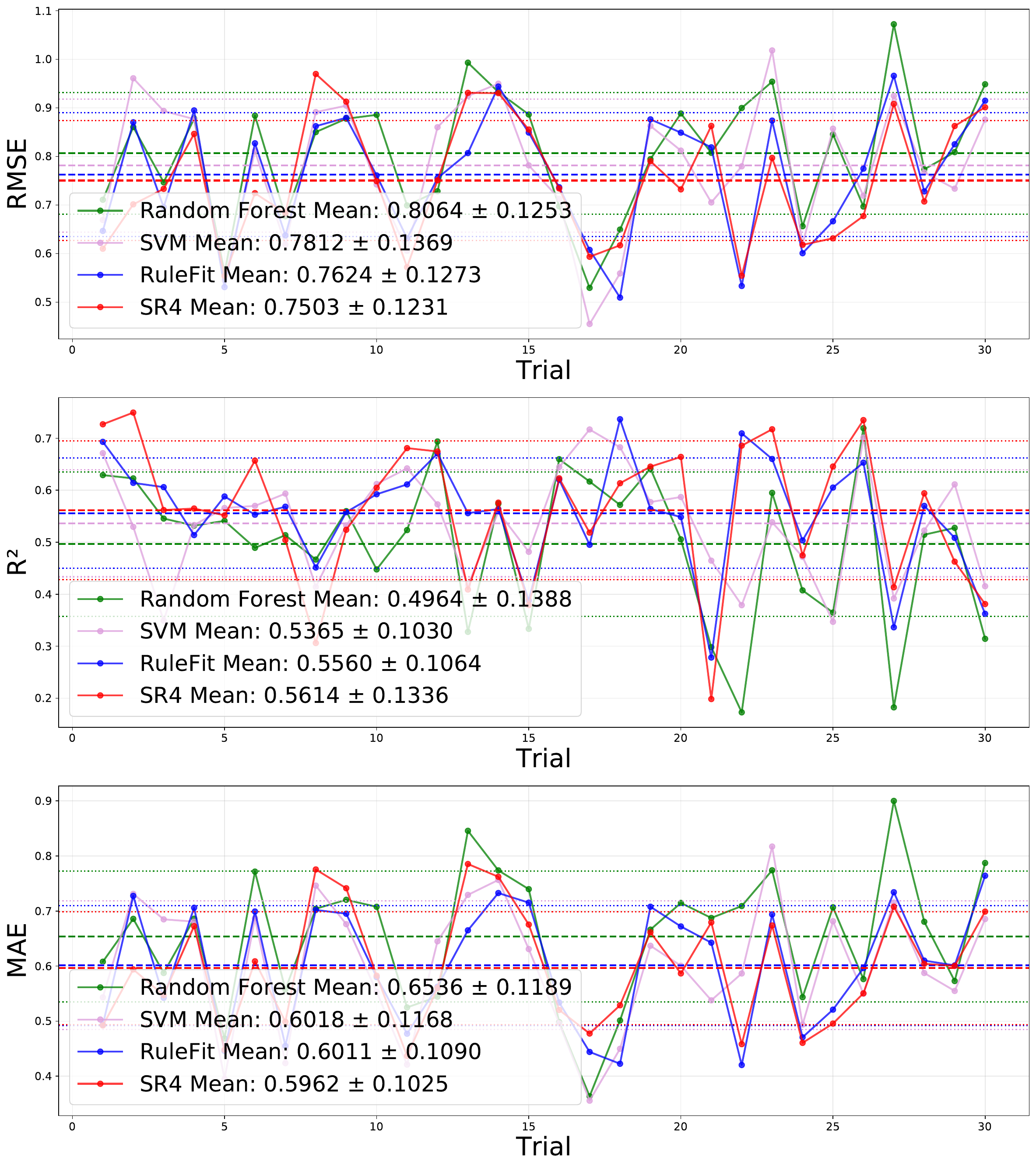}
\end{minipage}\hfill
\begin{minipage}{0.49\linewidth}
    \centering
    \includegraphics[width=\linewidth]{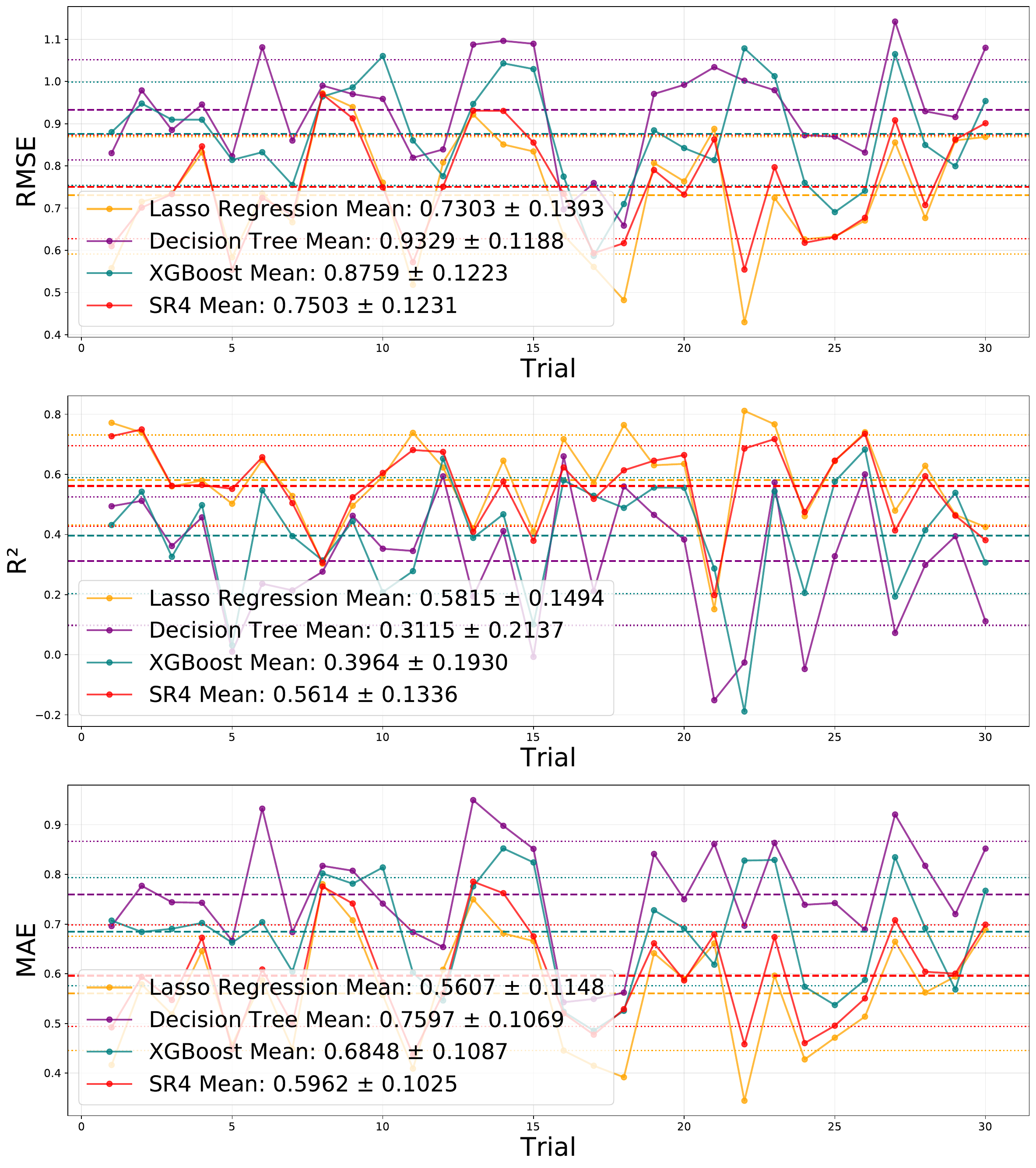}
\end{minipage}

\caption{
Line plot comparison of model performance metrics for prostate data across 30 trials.
The left panel reports results obtained for models—random forest, SVM, RuleFit, and SR4-fit ,
whereas the right panel shows results with models LASSO regression, decision tree, and XGBoost.
}
\label{fig:lineplot_prostate}
\end{figure}

\begin{figure}
\centering
\scriptsize

\begin{minipage}{0.48\linewidth}
    \centering
    \includegraphics[width=\linewidth]{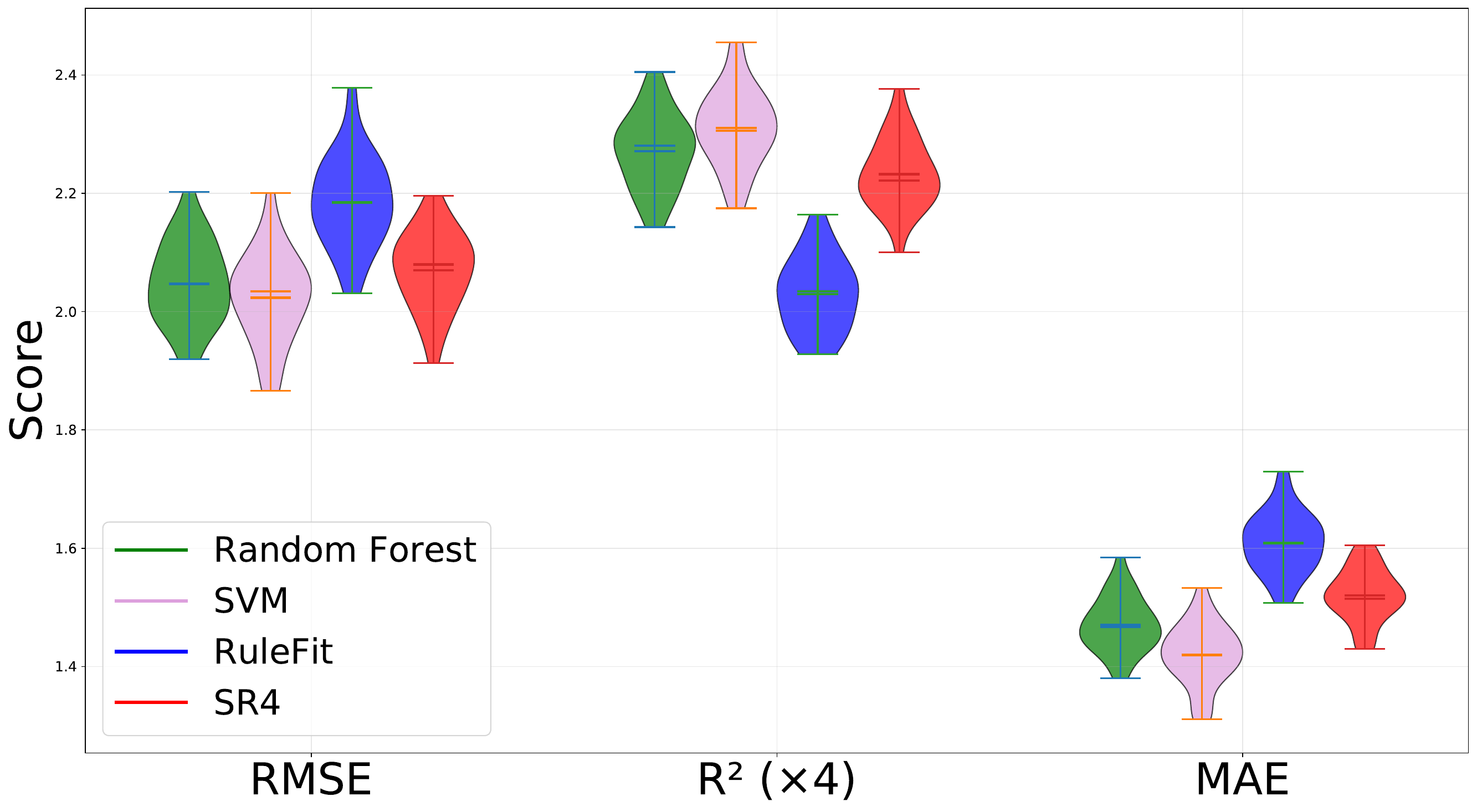}
    \subcaption{Abalone}
\end{minipage}\hfill
\begin{minipage}{0.48\linewidth}
    \centering
    \includegraphics[width=\linewidth]{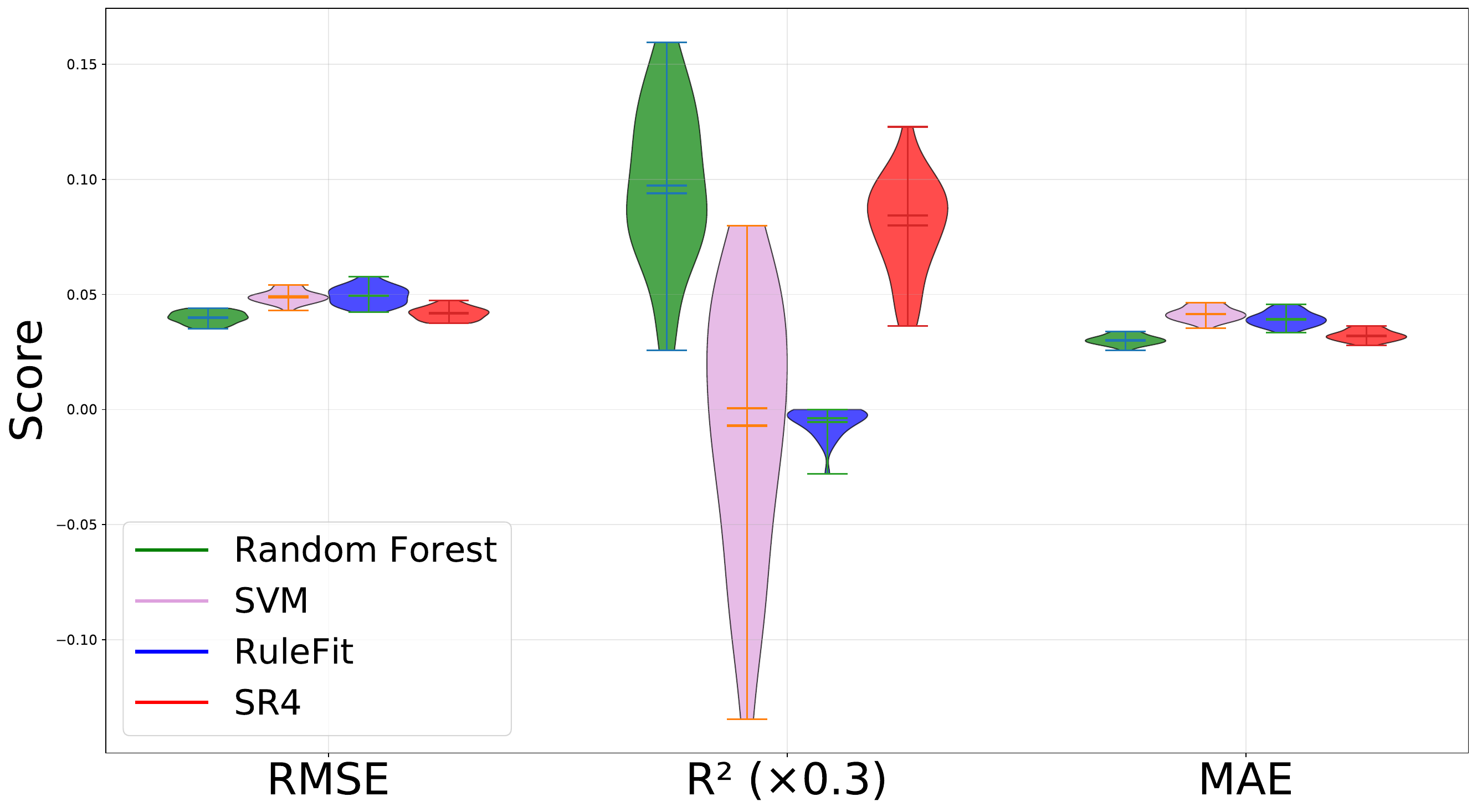}
    \subcaption{Bone}
\end{minipage}

\vspace{0.4cm}

\begin{minipage}{0.48\linewidth}
    \centering
    \includegraphics[width=\linewidth]{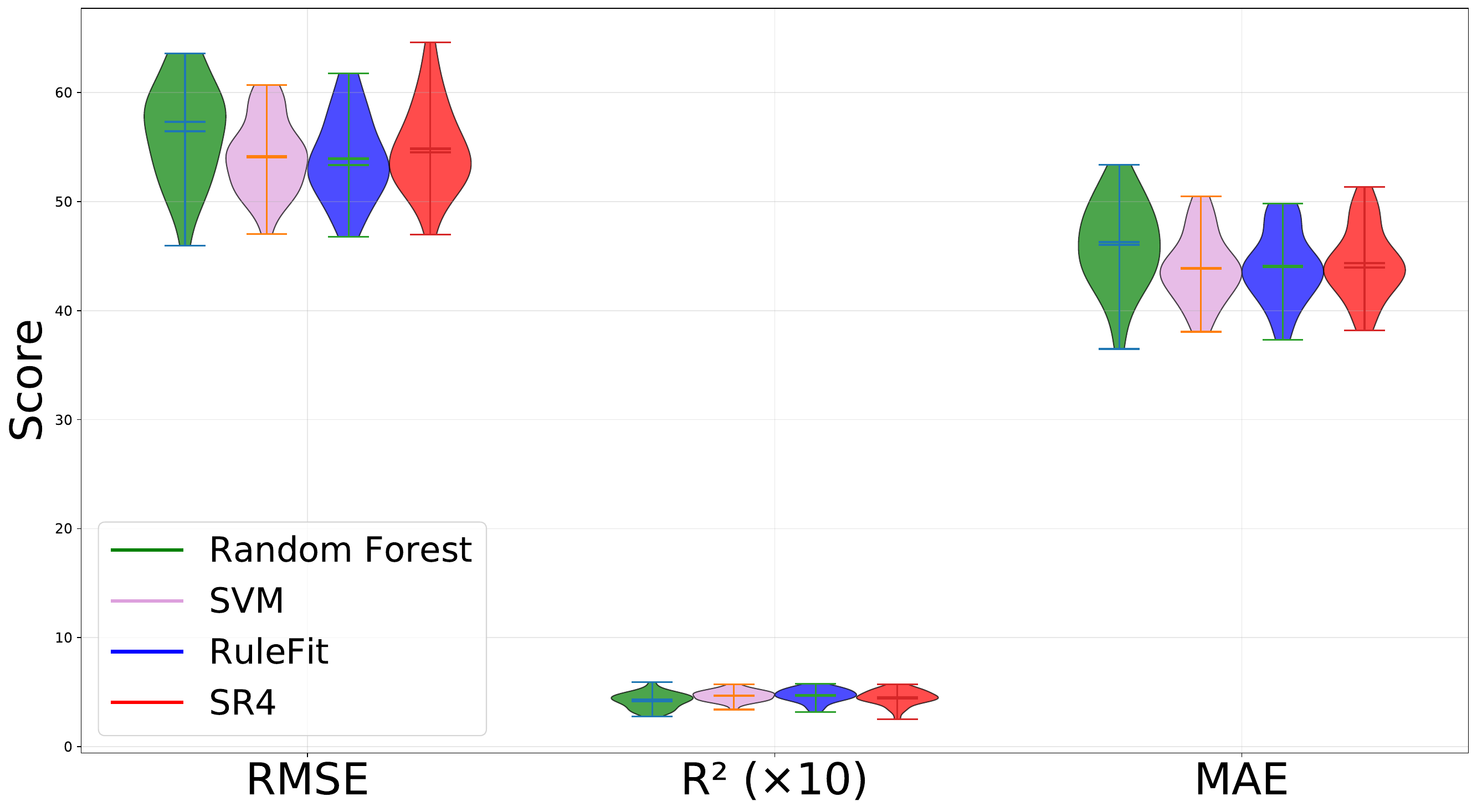}
    \subcaption{Diabetes}
\end{minipage}\hfill
\begin{minipage}{0.48\linewidth}
    \centering
    \includegraphics[width=\linewidth]{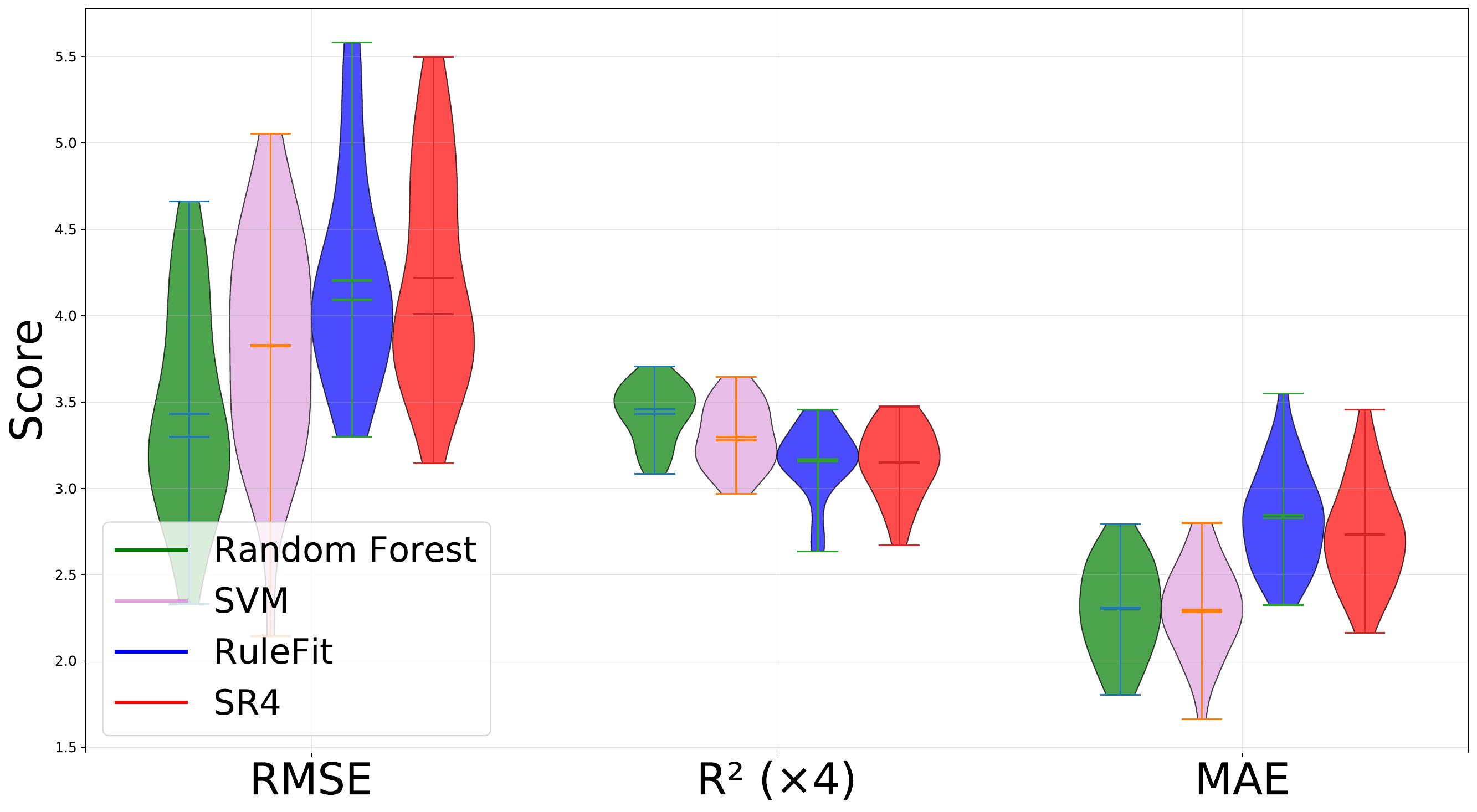}
    \subcaption{Housing}
\end{minipage}

\vspace{0.4cm}

\begin{minipage}{0.48\linewidth}
    \centering
    \includegraphics[width=\linewidth]{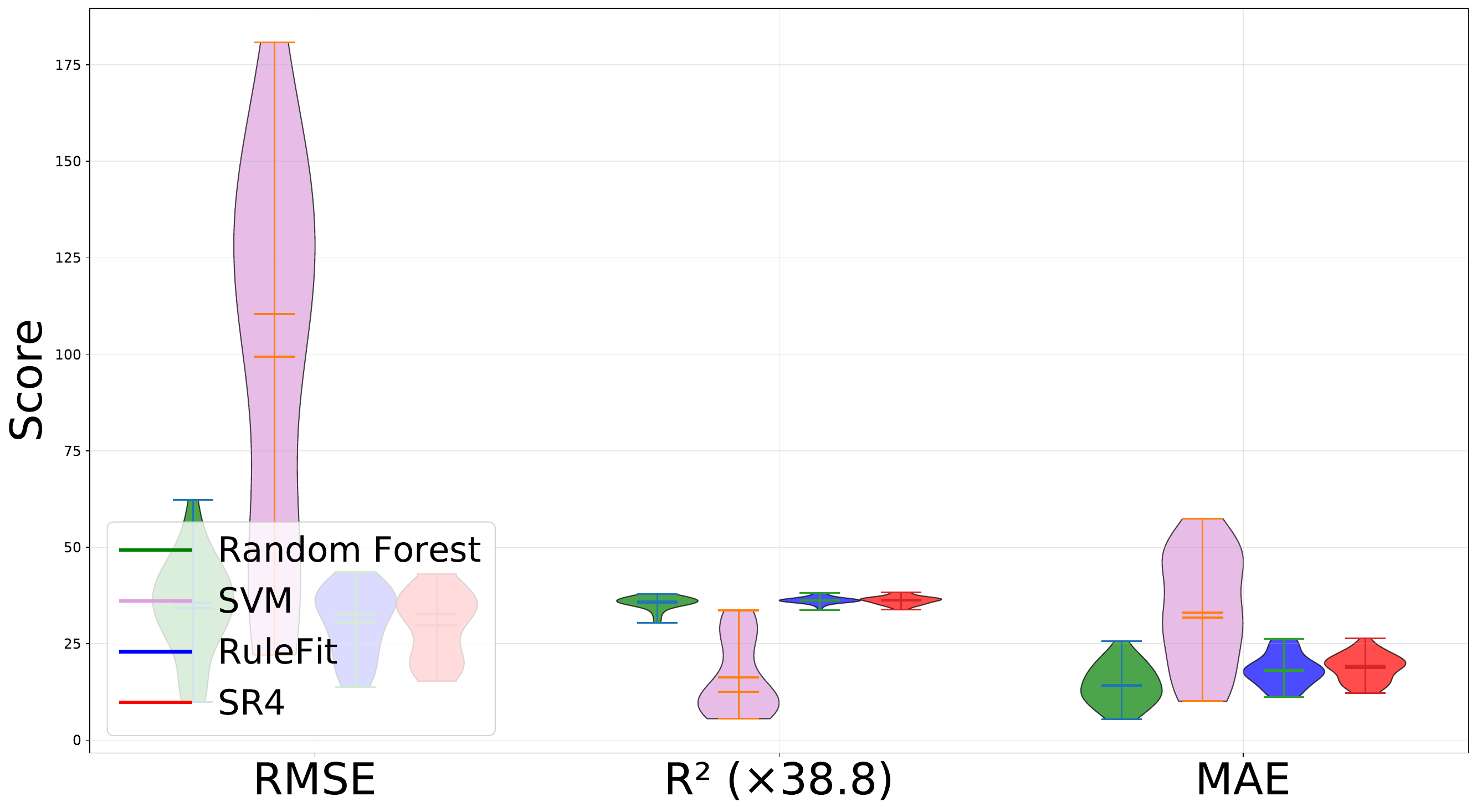}
    \subcaption{Machine}
\end{minipage}\hfill
\begin{minipage}{0.48\linewidth}
    \centering
    \includegraphics[width=\linewidth]{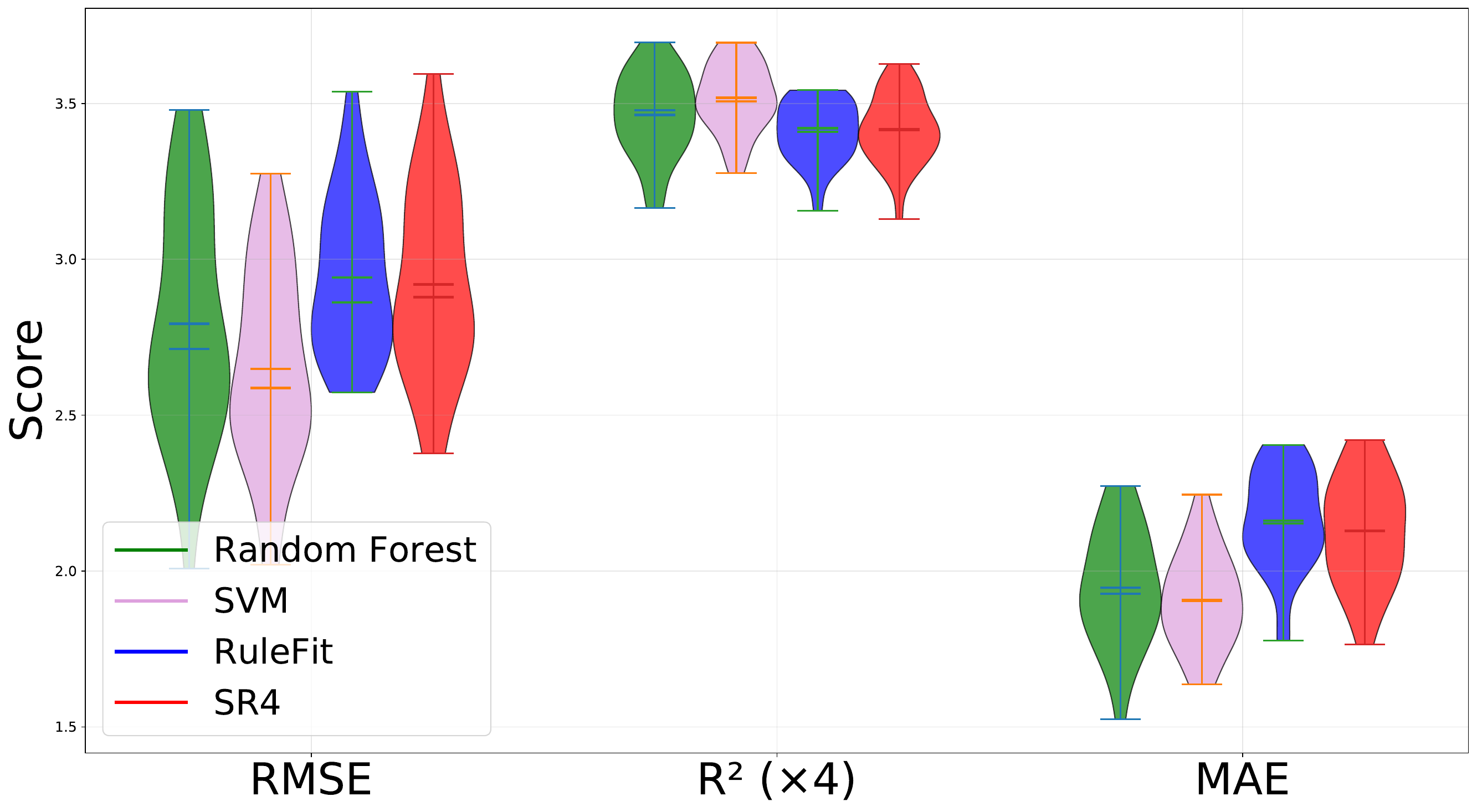}
    \subcaption{MPG}
\end{minipage}

\vspace{0.4cm}

\begin{minipage}{0.48\linewidth}
    \centering
    \includegraphics[width=\linewidth]{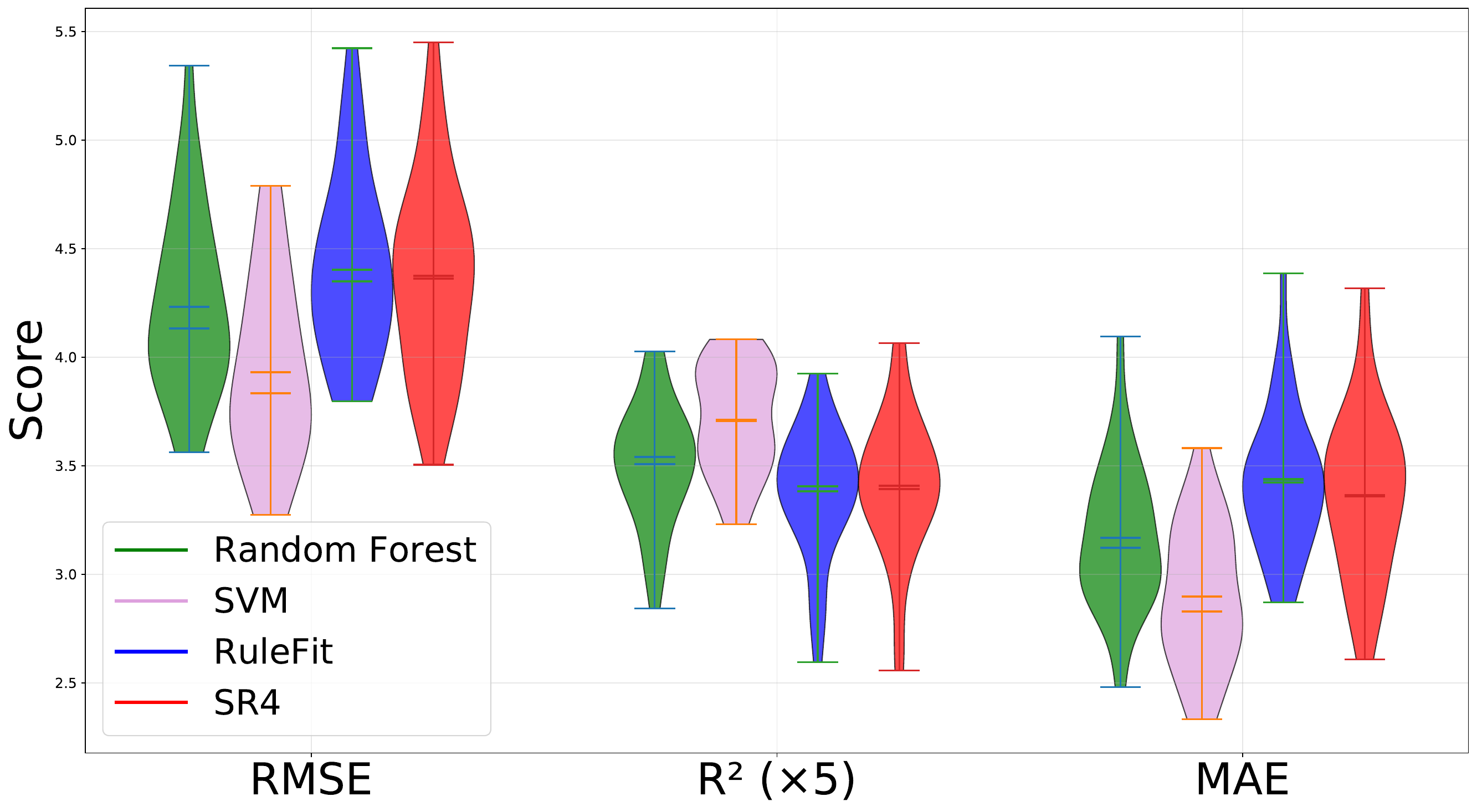}
    \subcaption{Ozone}
\end{minipage}\hfill
\begin{minipage}{0.48\linewidth}
    \centering
    \includegraphics[width=\linewidth]{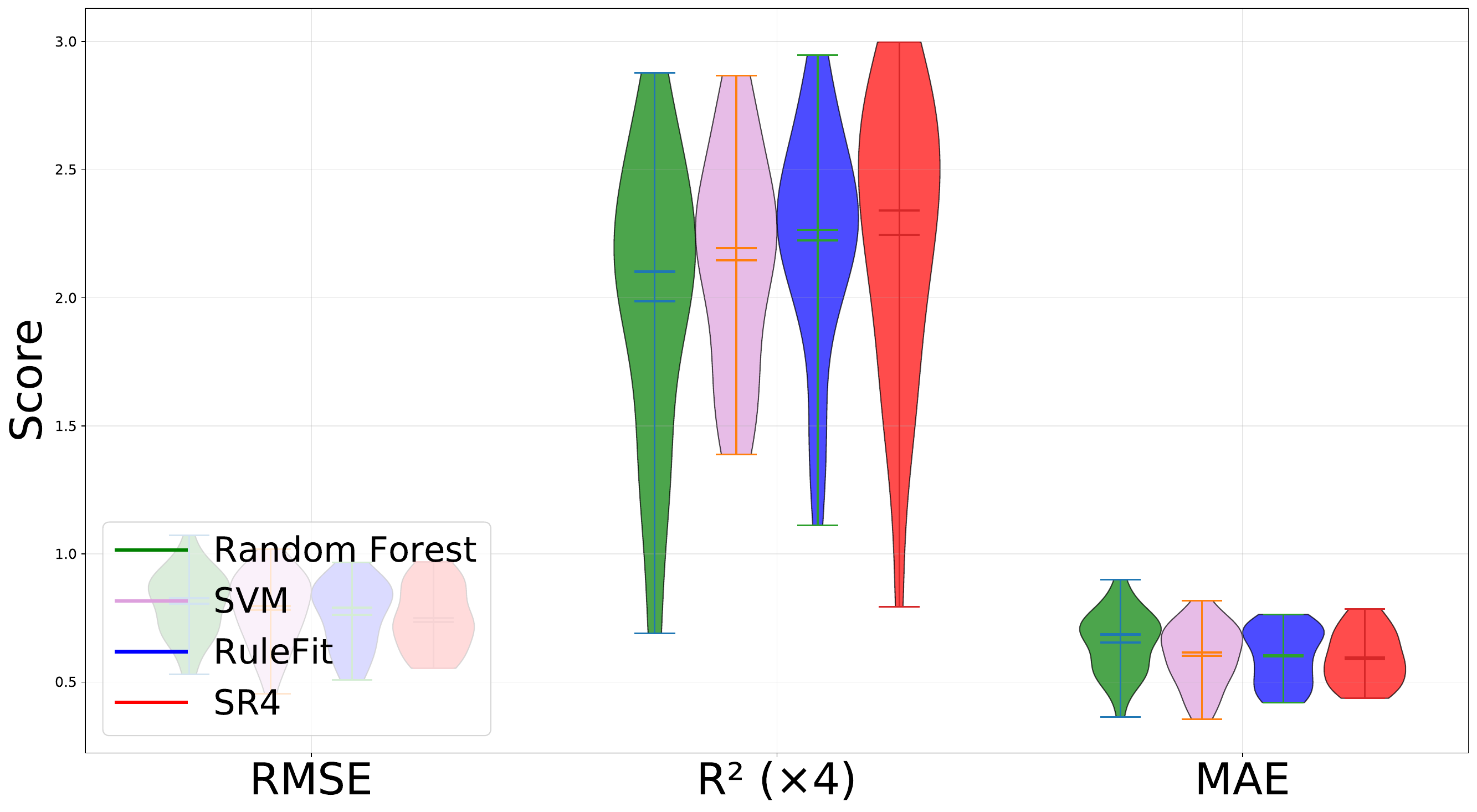}
    \subcaption{Prostate}
\end{minipage}

\caption{
Violin plot comparison of models (random forest, SVM, RuleFit, and SR4-fit) performance across multiple public regression datasets for different prediction metrics.
}
\label{fig:violin_all_models_reg}
\end{figure}

\begin{figure}[t]
\centering
\scriptsize

\begin{minipage}{0.48\linewidth}
    \centering
    \includegraphics[width=\linewidth]{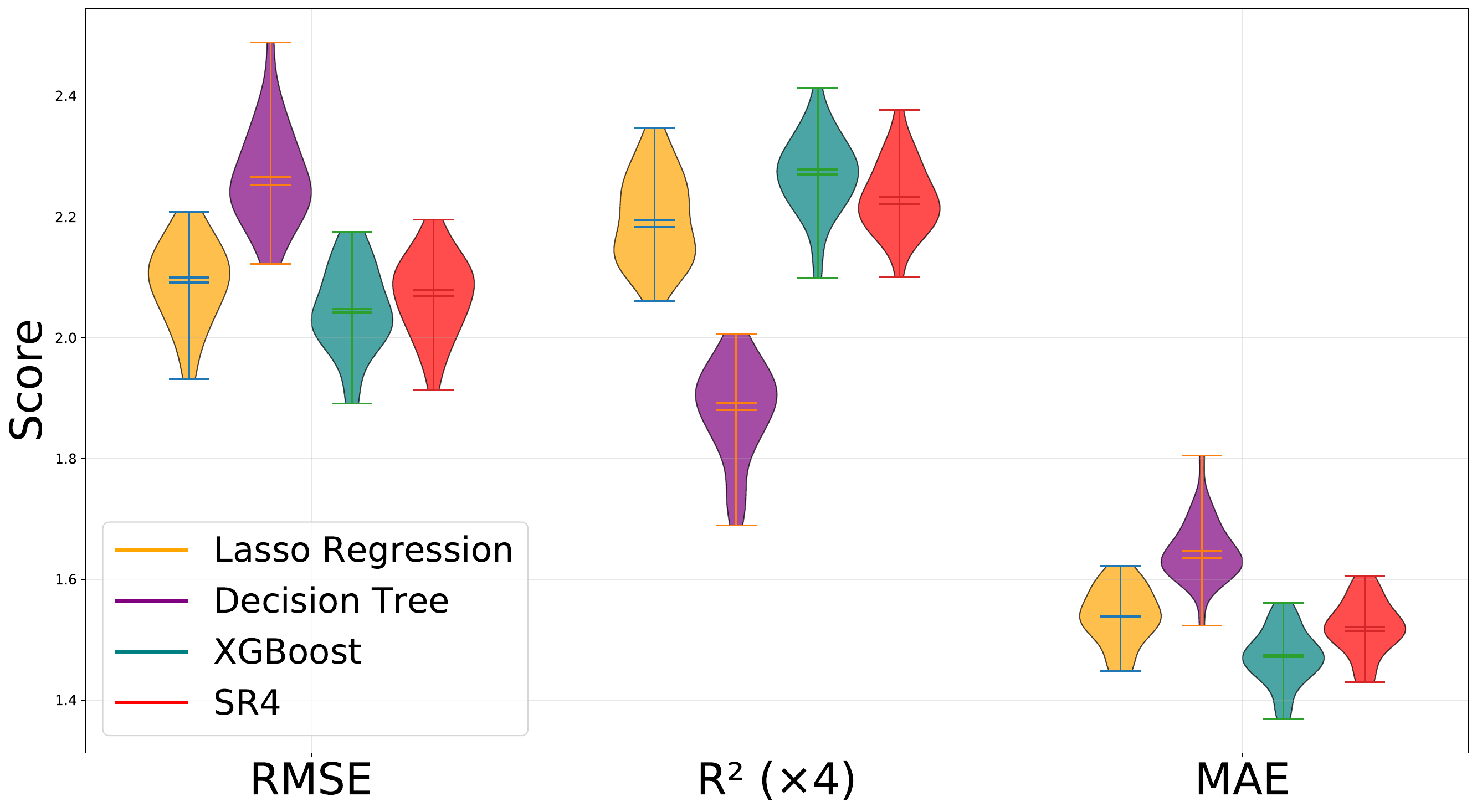}
    \subcaption{Abalone}
\end{minipage}\hfill
\begin{minipage}{0.48\linewidth}
    \centering
    \includegraphics[width=\linewidth]{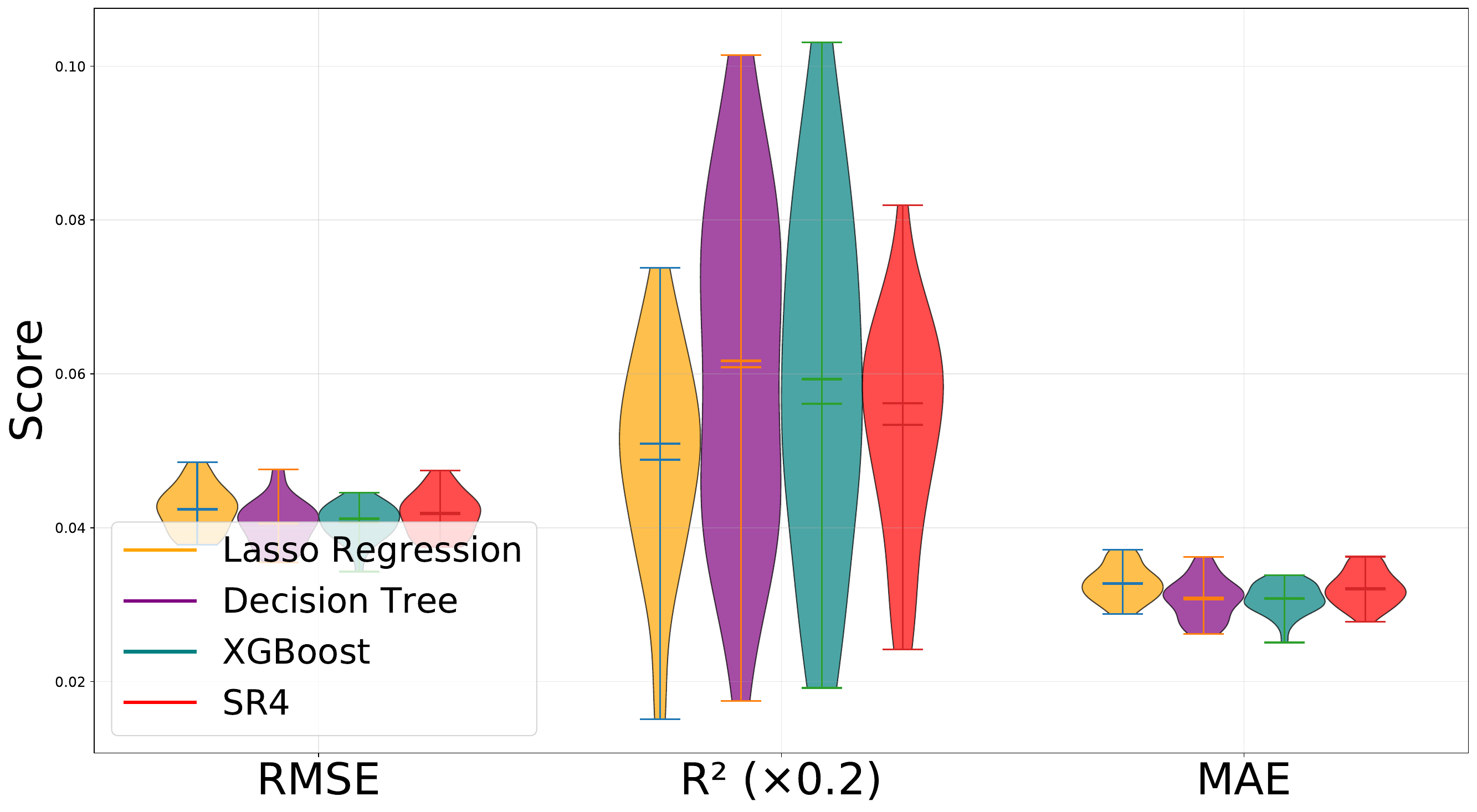}
    \subcaption{Bone}
\end{minipage}

\vspace{0.4cm}

\begin{minipage}{0.48\linewidth}
    \centering
    \includegraphics[width=\linewidth]{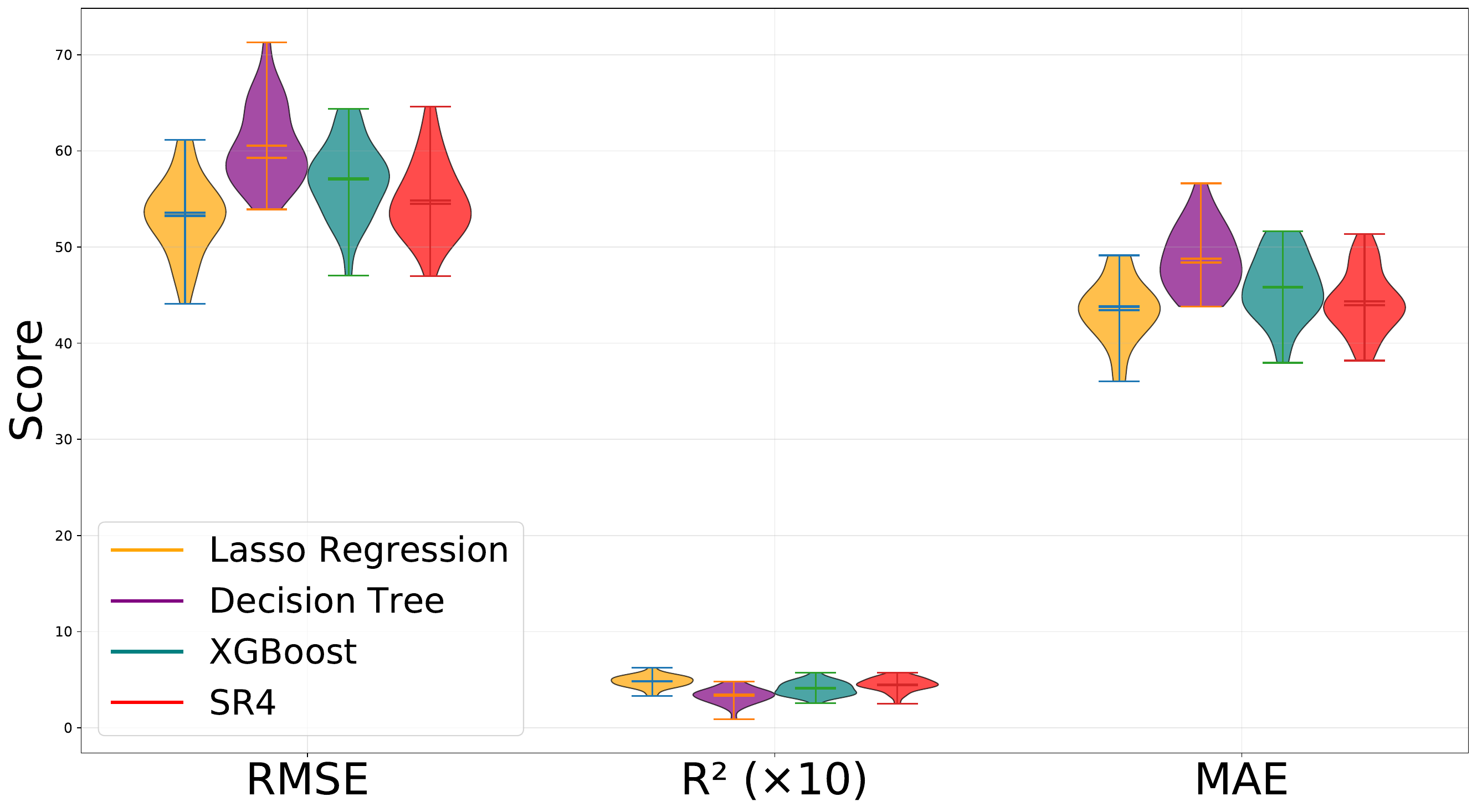}
    \subcaption{Diabetes}
\end{minipage}\hfill
\begin{minipage}{0.48\linewidth}
    \centering
    \includegraphics[width=\linewidth]{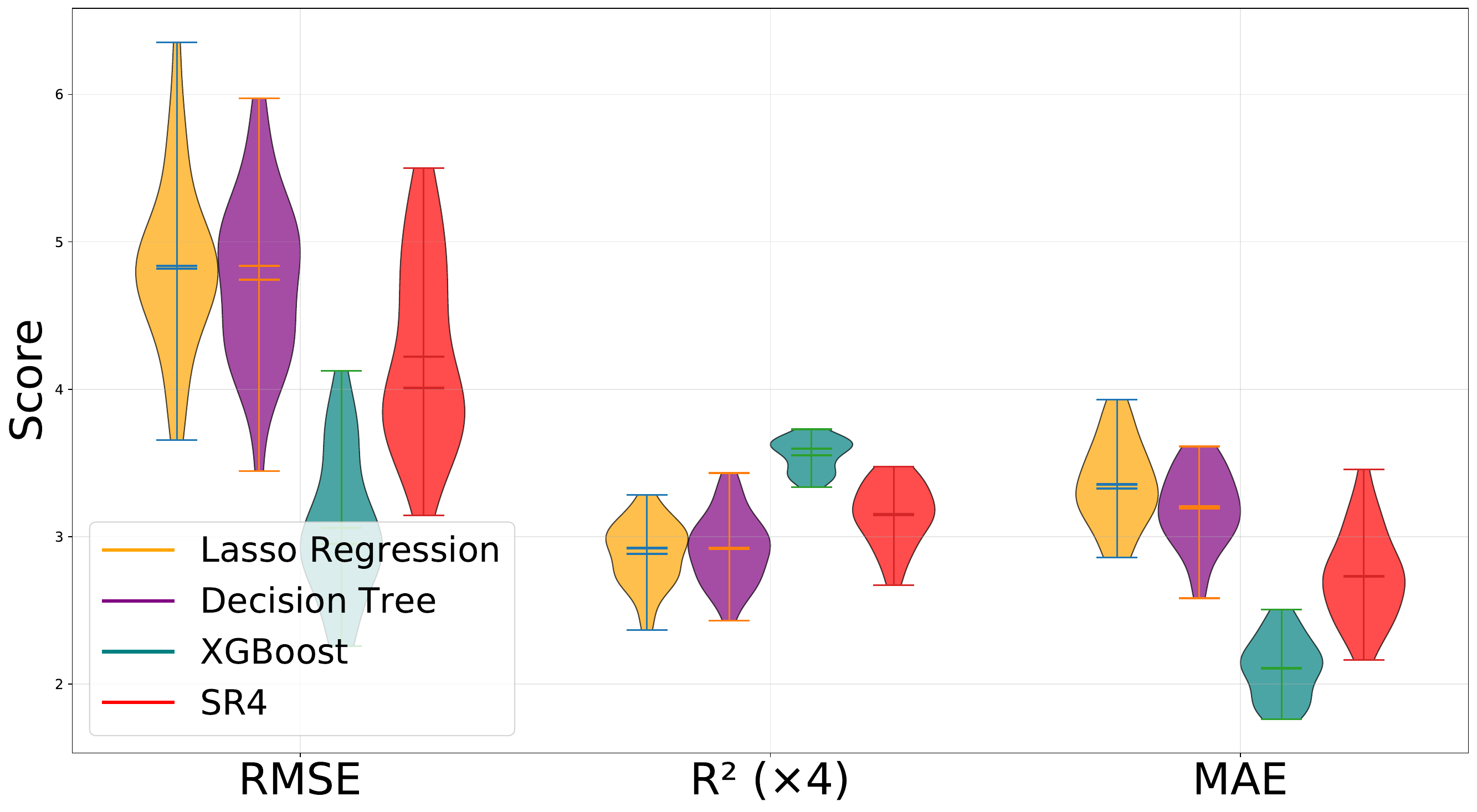}
    \subcaption{Housing}
\end{minipage}

\vspace{0.4cm}

\begin{minipage}{0.48\linewidth}
    \centering
    \includegraphics[width=\linewidth]{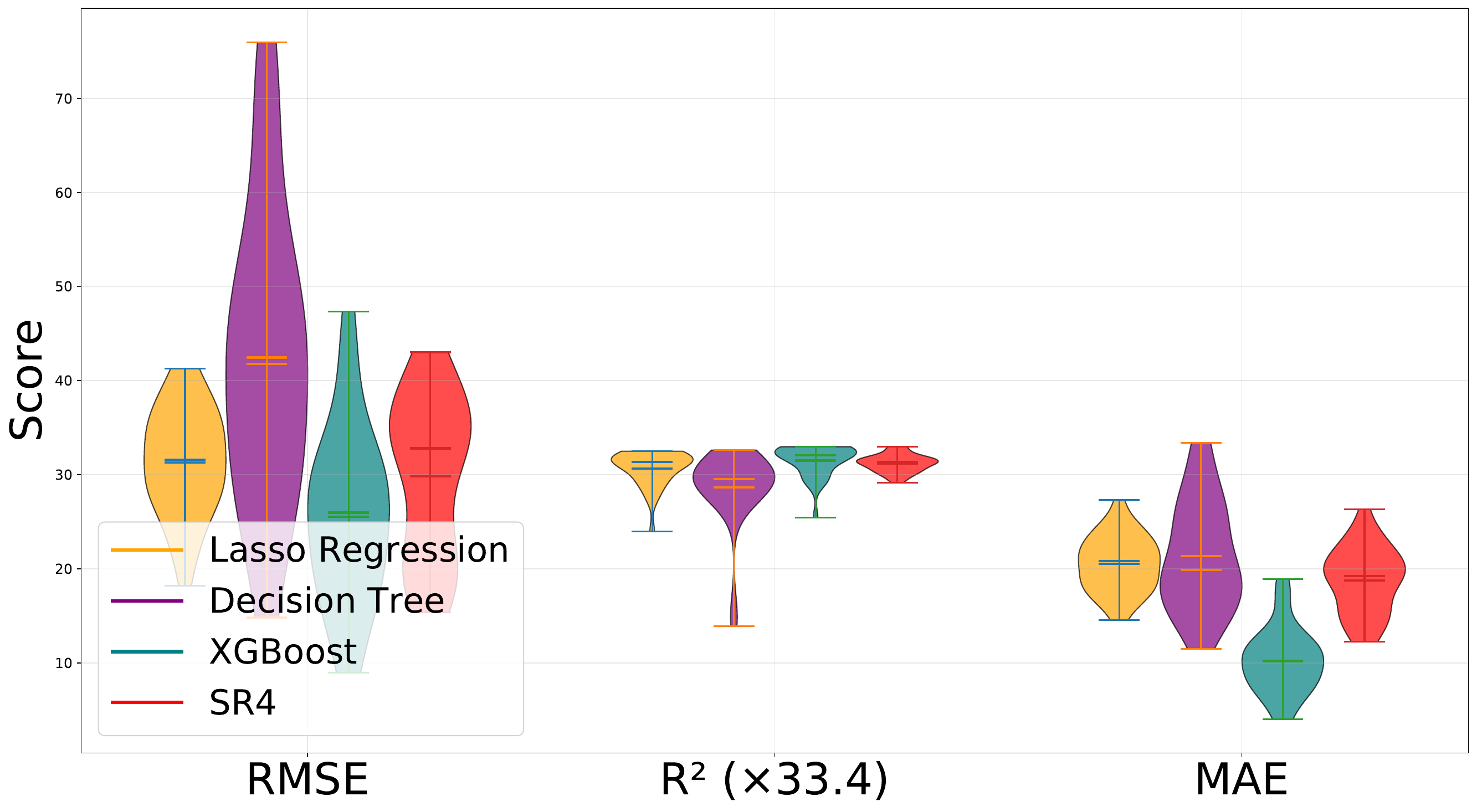}
    \subcaption{Machine}
\end{minipage}\hfill
\begin{minipage}{0.48\linewidth}
    \centering
    \includegraphics[width=\linewidth]{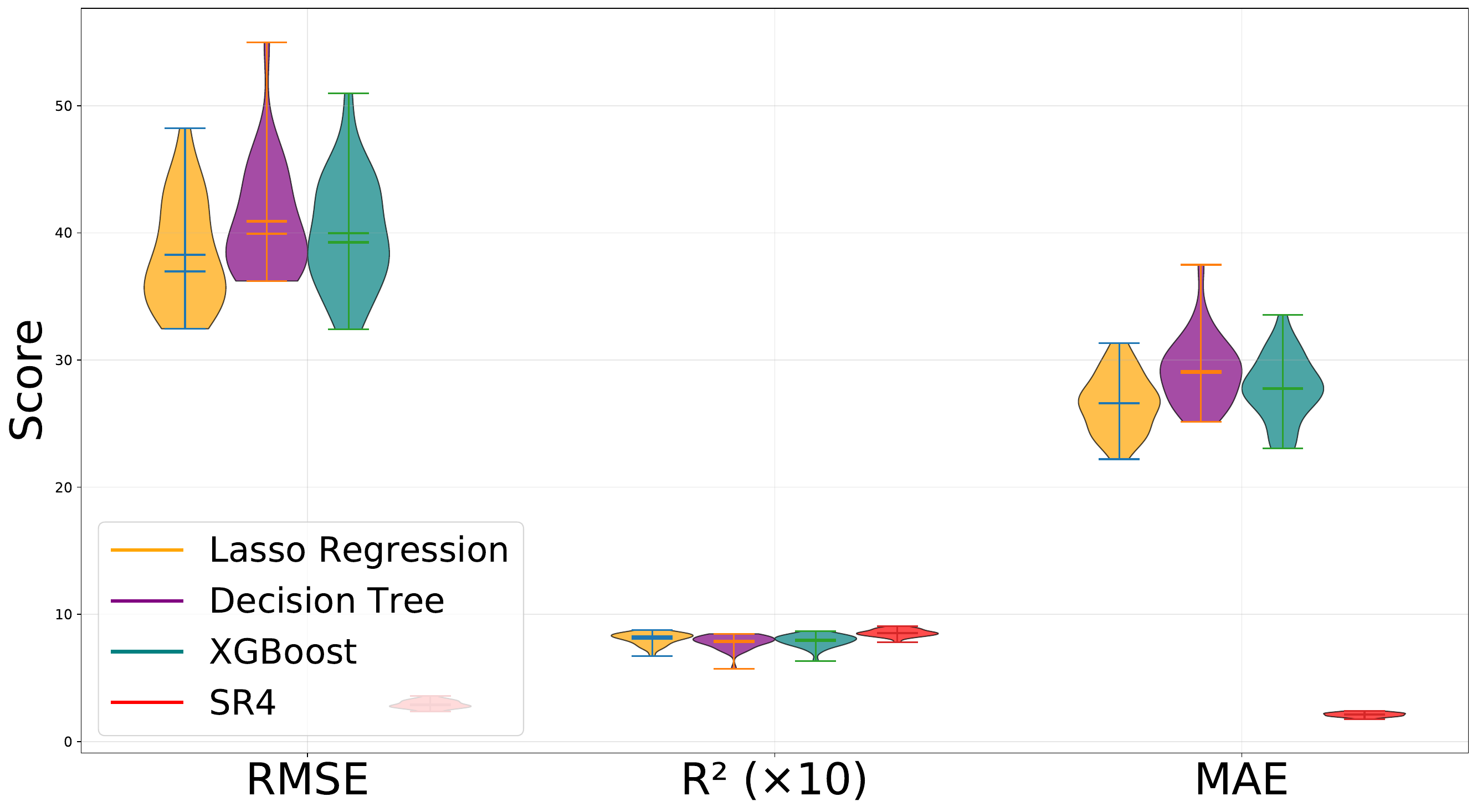}
    \subcaption{MPG}
\end{minipage}

\vspace{0.4cm}

\begin{minipage}{0.48\linewidth}
    \centering
    \includegraphics[width=\linewidth]{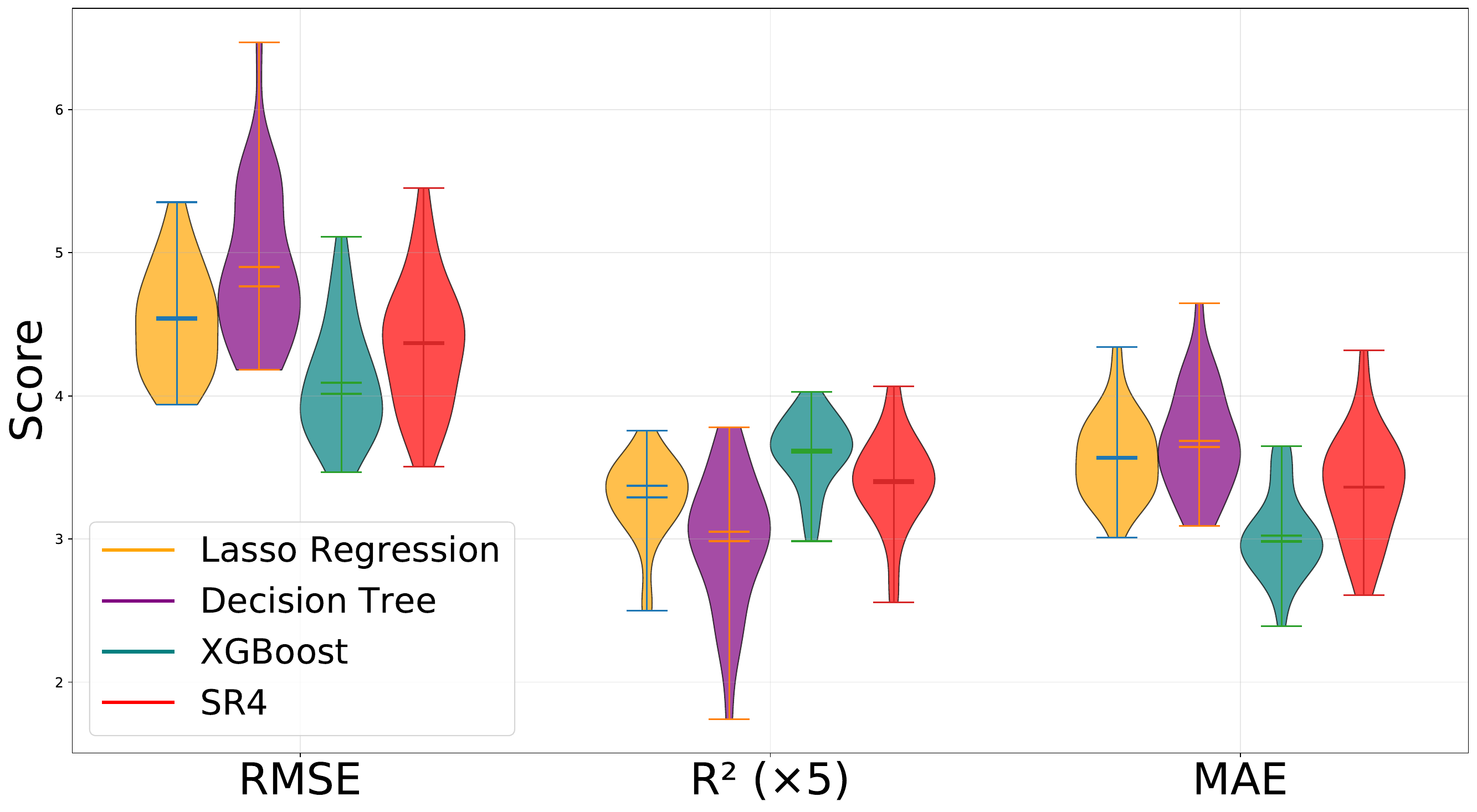}
    \subcaption{Ozone}
\end{minipage}\hfill
\begin{minipage}{0.48\linewidth}
    \centering
    \includegraphics[width=\linewidth]{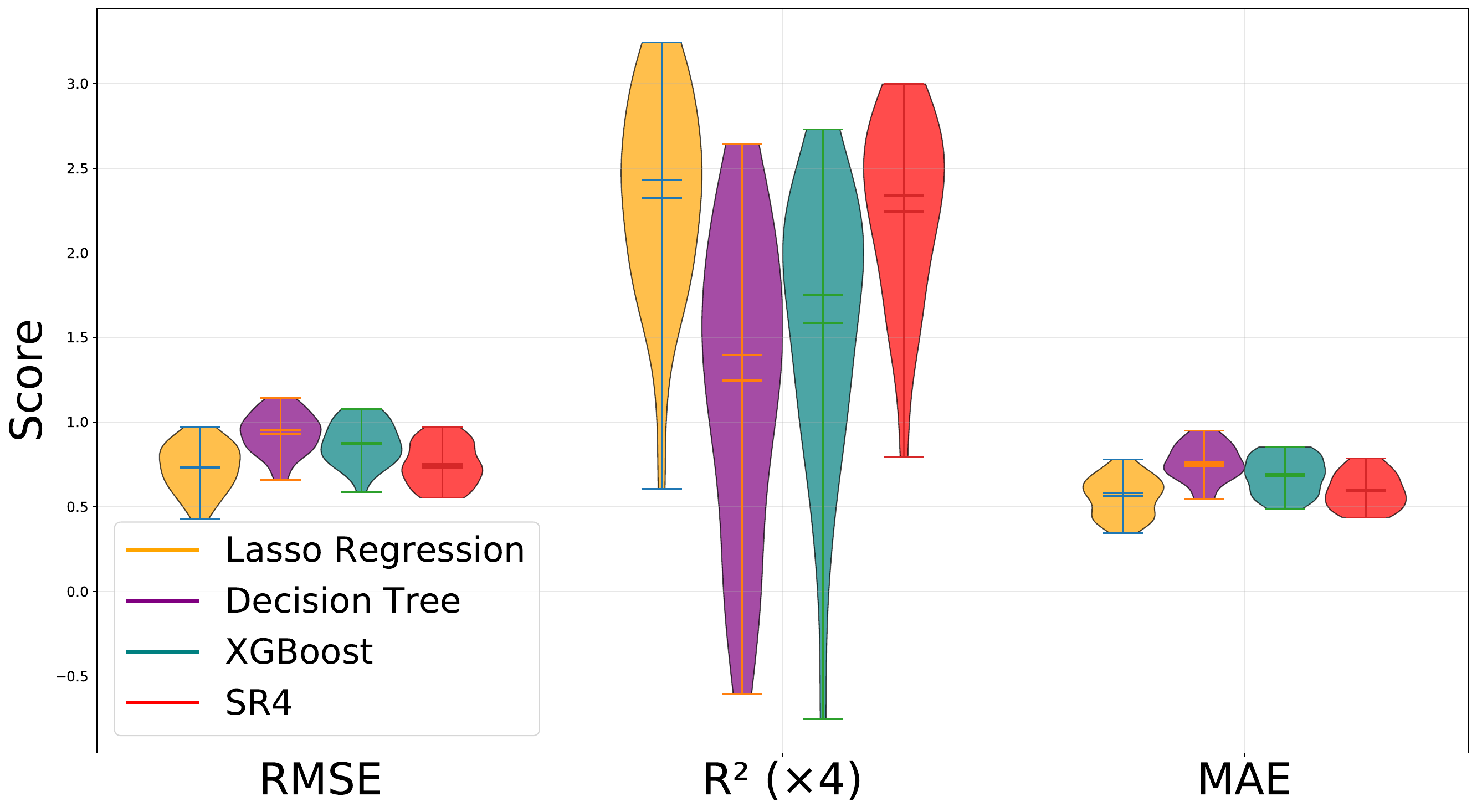}
    \subcaption{Prostate}
\end{minipage}

\caption{
Violin plot comparison of models (LASSO regression, decision tree, and XGBoost) performance across multiple public regression datasets for different prediction metrics.
}
\label{fig:violin_all_models_reg_other}
\end{figure}


\begin{table}
\caption{
Average Dice–Sørensen Index $\pm$ standard deviation across 30 trials for public regression datasets. Bold indicates the best-performing method per dataset.
}

\centering
\setlength{\tabcolsep}{3pt}
\renewcommand{\arraystretch}{1.15}

\begin{tabular}{lcccc}
\toprule
\textbf{Dataset}
& \textbf{Random Forest} & \textbf{RuleFit} & \textbf{SR4-Fit} & \textbf{Decision Tree} \\
\midrule
Abalone   & 0.0001$\pm$0.0000 & 0.5000$\pm$0.0000 & \textbf{0.5011$\pm$0.0028} & 0.2506$\pm$0.1684 \\
Bone      & 0.0012$\pm$0.0003 & 0.2727$\pm$0.0000 & \textbf{0.3710$\pm$0.0326} & 0.0230$\pm$0.0228 \\
Diabetes  & 0.0001$\pm$0.0001 & 0.5556$\pm$0.0000 & \textbf{0.5848$\pm$0.0156} & 0.1425$\pm$0.0970 \\
Housing   & 0.0005$\pm$0.0002 & \textbf{0.6190$\pm$0.0000} & 0.5880$\pm$0.0143 & 0.0428$\pm$0.0668 \\
Machine   & 0.0090$\pm$0.0020 & 0.4138$\pm$0.0141 & \textbf{0.6146$\pm$0.0167} & 0.0169$\pm$0.0205 \\
MPG       & 0.0005$\pm$0.0001 & 0.5000$\pm$0.0000 & 0.3743$\pm$0.0095 & \textbf{1.0000$\pm$0.0000} \\
Ozone     & 0.0008$\pm$0.0002 & 0.4286$\pm$0.0000 & \textbf{0.4649$\pm$0.0138} & 0.0511$\pm$0.0506 \\
Prostate  & 0.0034$\pm$0.0009 & 0.5032$\pm$0.0048 & \textbf{0.5574$\pm$0.0427} & 0.0084$\pm$0.0130 \\
\bottomrule
\end{tabular}

\label{tab:regression_stability}
\end{table}


\begin{table}
\caption{
Average number of rules $\pm$ standard deviation across 30 trials for public regression datasets. Lower values indicate more compact models.
}

\centering
\setlength{\tabcolsep}{3pt}
\renewcommand{\arraystretch}{1.15}

\begin{tabular}{lcccc}
\toprule
\textbf{Dataset}
& \textbf{Random Forest} & \textbf{RuleFit} & \textbf{SR4-Fit} & \textbf{Decision Tree} \\
\midrule
Abalone   & 8666.53$\pm$101.33 & 16.00$\pm$0.00 & 15.96$\pm$0.18 & \textbf{1.00$\pm$0.00} \\
Bone      & 8151.33$\pm$298.36 & 11.00$\pm$0.00 & 2.73$\pm$0.44 & \textbf{1.03$\pm$0.18} \\
Diabetes  & 7823.36$\pm$209.19 & 18.00$\pm$0.00 & 17.00$\pm$0.69 & \textbf{1.00$\pm$0.00} \\
Housing   & 7244.06$\pm$142.93 & 21.00$\pm$0.00 & 21.90$\pm$0.95 & \textbf{3.06$\pm$1.08} \\
Machine   & 4166.13$\pm$207.00 & 21.80$\pm$1.5177 & 14.66$\pm$0.84 & \textbf{2.66$\pm$0.80} \\
MPG       & 9090.80$\pm$205.65 & 16.00$\pm$0.00 & 21.40$\pm$1.13 & \textbf{1.00$\pm$0.00} \\
Ozone     & 7968.26$\pm$144.58 & 28.00$\pm$0.00 & 25.70$\pm$1.23 & \textbf{2.20$\pm$0.48} \\
Prostate  & 3220.96$\pm$75.75  & 15.90$\pm$0.30 & 12.33$\pm$1.24 & \textbf{3.33$\pm$0.88} \\
\bottomrule
\end{tabular}

\label{tab:regression_rules}
\end{table}

\begin{table}
\caption{
Average rule complexity $\pm$ standard deviation across 30 trials for public regression datasets. Bold indicates less complex rules
}

\centering
\setlength{\tabcolsep}{3pt}
\renewcommand{\arraystretch}{1.15}

\begin{tabular}{lcccc}
\toprule
\textbf{Dataset}
& \textbf{Random Forest} & \textbf{RuleFit} & \textbf{SR4-Fit} & \textbf{Decision Tree} \\
\midrule
Abalone   & 6.54$\pm$0.02 & 2.00$\pm$0.00 & 1.99$\pm$0.01 & \textbf{1.00$\pm$0.00} \\
Bone      & 5.95$\pm$0.03 & 2.45$\pm$0.00 & 2.24$\pm$0.15 & \textbf{1.73$\pm$0.98} \\
Diabetes  & 5.88$\pm$0.02 & 1.88$\pm$0.00 & 1.82$\pm$0.05 & \textbf{1.00$\pm$0.00} \\
Housing   & 5.89$\pm$0.02 & \textbf{1.76$\pm$0.00} & 2.22$\pm$0.07 & 3.00$\pm$0.00 \\
Machine   & 5.62$\pm$0.05 & 2.61$\pm$0.14 & \textbf{1.75$\pm$0.07} & 2.72$\pm$0.58 \\
MPG       & 5.98$\pm$0.02 & 2.00$\pm$0.00 & 2.87$\pm$0.06 & \textbf{1.00$\pm$0.00} \\
Ozone     & 5.86$\pm$0.01 & 2.71$\pm$0.00 & \textbf{2.59$\pm$0.07} & 2.98$\pm$0.09 \\
Prostate  & 4.75$\pm$0.05 & 1.98$\pm$0.04 & \textbf{1.79$\pm$0.14} & 2.39$\pm$0.31 \\
\bottomrule
\end{tabular}

\label{tab:regression_complexity}
\end{table}


Across all eight regression datasets and 30 trials, SR4-Fit demonstrates a strong balance of predictive performance, rule stability, and low variance, making it the most robust and interpretable method overall. Although individual models occasionally outperform it on prediction metrics, SR4-Fit repeatedly delivers the most consistent and stable behavior across trials.

In the abalone dataset, SVM has the lowest error values, but SR4-Fit trails very closely behind SVM, as seen in Figure~\ref{fig:lineplot_abalone}. The violin plots for abalone in Figure~\ref{fig:violin_all_models_reg} and Figure~\ref{fig:violin_all_models_reg_other} reinforce these findings. SR4-Fit significantly outperforms in stability, achieving the highest Dice–Sorensen Index, as seen in Table~\ref{tab:regression_stability}. Table~\ref{tab:regression_rules} shows that the decision tree has a compact structure, whereas, with respect to average rule complexity in Table~\ref{tab:regression_complexity} decision tree has lower average rule complexity compared to all rule-based models.

In the bone dataset, random forest has the lowest error values, but SR4-Fit trails very closely behind random forest, as seen in Figure~\ref{fig:lineplot_bone}. The violin plots for bone in Figure~\ref{fig:violin_all_models_reg} and Figure~\ref{fig:violin_all_models_reg_other} reinforce these findings. SR4-Fit significantly outperforms in stability, achieving the highest Dice–Sorensen Index, as seen in Table~\ref{tab:regression_stability}. Table~\ref{tab:regression_rules} shows that the decision tree has a compact structure, whereas, with respect to average rule complexity in Table~\ref{tab:regression_complexity} decision tree has lower average rule complexity compared to all rule-based models.

In the diabetes dataset, linear regression has the lowest error values, but SR4-Fit and RuleFit trail very closely behind, as seen in Figure~\ref{fig:lineplot_diabetes}. The violin plots for diabetes in Figure~\ref{fig:violin_all_models_reg} and Figure~\ref{fig:violin_all_models_reg_other} reinforce these findings. SR4-Fit significantly outperforms in stability, achieving the highest Dice–Sorensen Index, as seen in Table~\ref{tab:regression_stability}. Table~\ref{tab:regression_rules} shows that the decision tree has a compact structure, whereas, with respect to average rule complexity in Table~\ref{tab:regression_complexity} decision tree has lower average rule complexity compared to all rule-based models.

In the housing dataset, XGBoost has the lowest error values, but SR4-Fit trails very closely behind XGBoost, as seen in Figure~\ref{fig:lineplot_housing}. The violin plots for housing in Figure~\ref{fig:violin_all_models_reg} and Figure~\ref{fig:violin_all_models_reg_other} reinforce these findings. RuleFit outperforms in stability, achieving the highest Dice–Sorensen Index, as seen in Table~\ref{tab:regression_stability}. Table~\ref{tab:regression_rules} shows that the decision tree has a compact structure, whereas, with respect to average rule complexity in Table~\ref{tab:regression_complexity} RuleFit has lower average rule complexity compared to all rule-based models.

In the machine dataset, XGBoost has the lowest error values, but SR4-Fit trails very closely behind XGBoost, as seen in Figure~\ref{fig:lineplot_machine}. The violin plots for the machine in Figure~\ref{fig:violin_all_models_reg} and Figure~\ref{fig:violin_all_models_reg_other} reinforce these findings. SR4-Fit outperforms in stability, achieving the highest Dice–Sorensen Index, as seen in Table~\ref{tab:regression_stability}. Table~\ref{tab:regression_rules} shows that the decision tree has a compact structure, whereas, with respect to average rule complexity in Table~\ref{tab:regression_complexity} SR4-Fit has lower average rule complexity compared to all rule-based models.

In the MPG dataset, SVM has the lowest error values, but SR4-Fit and RuleFit trail very closely behind, as seen in Figure~\ref{fig:lineplot_mpg}. The violin plots for MPG in Figure~\ref{fig:violin_all_models_reg} and Figure~\ref{fig:violin_all_models_reg_other} reinforce these findings. The decision tree outperforms in stability, achieving the highest Dice–Sorensen Index, as seen in Table~\ref{tab:regression_stability}. Table~\ref{tab:regression_rules} shows that the decision tree has a compact structure, whereas, with respect to average rule complexity in Table~\ref{tab:regression_complexity} decision tree has lower average rule complexity compared to all rule-based models.

In the ozone dataset, SVM has the lowest error values, but SR4-Fit trails very closely behind, as seen in Figure~\ref{fig:lineplot_ozone}. The violin plots for the ozone in Figure~\ref{fig:violin_all_models_reg} and Figure~\ref{fig:violin_all_models_reg_other} reinforce these findings. SR4-Fit outperforms in stability, achieving the highest Dice–Sorensen Index, as seen in Table~\ref{tab:regression_stability}. Table~\ref{tab:regression_rules} shows that the decision tree has a compact structure, whereas, with respect to average rule complexity in Table~\ref{tab:regression_complexity} SR4-Fit has lower average rule complexity compared to all rule-based models.

In the prostate dataset, SVM has the lowest error values, but SR4-Fit trails very closely behind, as seen in Figure~\ref{fig:lineplot_prostate}. The violin plots for the ozone in Figure~\ref{fig:violin_all_models_reg} and Figure~\ref{fig:violin_all_models_reg_other} reinforce these findings. SR4-Fit outperforms in stability, achieving the highest Dice–Sorensen Index, as seen in Table~\ref{tab:regression_stability}. Table~\ref{tab:regression_rules} shows that the decision tree has a compact structure, whereas, with respect to average rule complexity in Table~\ref{tab:regression_complexity} SR4-Fit has lower average rule complexity compared to all rule-based models.

\FloatBarrier

\clearpage
\addtocounter{page}{-1}

\end{document}